\RequirePackage{fix-cm} 
\documentclass[ twoside,openright,titlepage,numbers=noenddot,headinclude,
                footinclude=true,cleardoublepage=empty,abstractoff, 
                BCOR=10mm,paper=a4,fontsize=11pt,
                ngerman,american,%
                ]{scrbook}

\PassOptionsToPackage{utf8}{inputenc}
	\usepackage{inputenc}

\PassOptionsToPackage{eulerchapternumbers,listings,
					 pdfspacing,
					 beramono,eulermath,parts}{classicthesis}                                        

\newcommand{\myTitle}{Extracting Temporal and Causal Relations between Events\xspace}

\newcommand{\myName}{Paramita Mirza\xspace}

\newcommand{\myFaculty}{Put data here\xspace}

\newcommand{\myUni}{Put data here\xspace}

\newcommand{\myTime}{April 2016\xspace}
\newcommand{\myVersion}{version 0.1\xspace}

\newcommand{\role}[2]{$[_{\mathrm{#1}}$~#2$]$}

\newcommand{\event}[1]{\textit{\textbf{#1}}}
\newcommand{\eventattr}[2]{\textit{\textbf{#1}}\textsubscript{~\textsc{#2}}}
\newcommand{\timex}[1]{\textit{#1}}
\newcommand{\timexattr}[2]{\textit{#1}\textsubscript{~\textsc{#2}}}
\newcommand{\signal}[1]{\ul{#1}}
\newcommand{\signalattr}[2]{\ul{#1}\textsubscript{~\textsc{#2}}}
\newcommand{\wordattr}[2]{\textit{\textbf{#1}}\textsubscript{~\textsc{#2}}}
\newcommand{\entity}[1]{\textit{[\textbf{#1}]}}

\newcounter{dummy} 
\providecommand{\mLyX}{L\kern-.1667em\lower.25em\hbox{Y}\kern-.125emX\@}

	\usepackage{babel}                  

\usepackage{csquotes}
\PassOptionsToPackage{%
	backend=bibtex8,bibencoding=ascii,%
	language=auto,%
    style=authoryear-comp, 
    bibstyle=authoryear,dashed=false, 
    sorting=nyt, 
    maxbibnames=10, 
    maxcitenames=2, 
    natbib=true 
}{biblatex}
    \usepackage{biblatex}
    
\DeclareBibliographyCategory{publications}
\newcommand{\pubcite}[1]{%
  \addtocategory{publications}{#1}%
  \defbibcheck{key#1}{
    \iffieldequalstr{entrykey}{#1}
      {}
      {\skipentry}}%
  \printbibliography[heading=none,check=key#1]%
}

\PassOptionsToPackage{fleqn}{amsmath}       
    \usepackage{amsmath}

\PassOptionsToPackage{T1}{fontenc} 
    \usepackage{fontenc}     
\usepackage{textcomp} 
\usepackage{scrhack} 
\usepackage{xspace} 
\usepackage{mparhack} 
\usepackage{fixltx2e} 
\PassOptionsToPackage{printonlyused,smaller}{acronym} 
    \usepackage{acronym} 

\usepackage{afterpage}
\usepackage{longtable}
\usepackage{graphicx,color,multirow,multicol,enumerate}
\usepackage{minitoc}
\usepackage{soul}

\usepackage{tabularx} 
\usepackage{caption}
\usepackage{listings} 
    \usepackage{hyperref}  
\PassOptionsToPackage{pdftex}{graphicx}
    \usepackage{graphicx}

\hypersetup{%
    colorlinks=true, linktocpage=true, pdfstartpage=3, pdfstartview=FitV,%
    breaklinks=true, pdfpagemode=UseNone, pageanchor=true, pdfpagemode=UseOutlines,%
    plainpages=false, bookmarksnumbered, bookmarksopen=true, bookmarksopenlevel=1,%
    hypertexnames=true, pdfhighlight=/O,
    urlcolor=webbrown, linkcolor=RoyalBlue, citecolor=webgreen, 
    pdftitle={\myTitle},%
    pdfauthor={\textcopyright\ \myName, \myUni, \myFaculty},%
    pdfsubject={},%
    pdfkeywords={},%
    pdfcreator={pdfLaTeX},%
    pdfproducer={LaTeX with hyperref and classicthesis}%
}   

\makeatletter
\@ifpackageloaded{babel}%
    {%
       \addto\extrasamerican{%
                }%
       \addto\extrasngerman{%
                }%
    }{\relax}
\makeatother

\usepackage{classicthesis} 

\areaset[current]{\dimexpr\textwidth+\marginparwidth+\marginparsep}{761pt}
\usepackage{arydshln}
\usepackage{adjustbox}
\usepackage{footnote}
\usepackage{enumitem}

\usepackage{qtree}
\usepackage{caption}
\usepackage{subcaption}
\usepackage{tikz}
\usepackage{tikz-dependency}
\usetikzlibrary{automata,arrows,shapes.geometric}
\usepackage{algorithmic}
\usepackage{verbatim}

\numberwithin{figure}{chapter}
\numberwithin{table}{chapter}
\numberwithin{equation}{chapter}

\begin{document}
\frenchspacing
\raggedbottom
\selectlanguage{american} 
\pagenumbering{roman}
\pagestyle{plain}

\setcounter{minitocdepth}{2} 
\dominitoc[n]
\setul{1pt}{.4pt}
\begin{titlepage}

\begin{center}
	\vspace*{1 cm} 
	\textbf{\Large PhD Dissertation}\\\ \hrulefill\\\
	
	\begin{figure}[h!]\centerline{\includegraphics[width=0.7\textwidth]{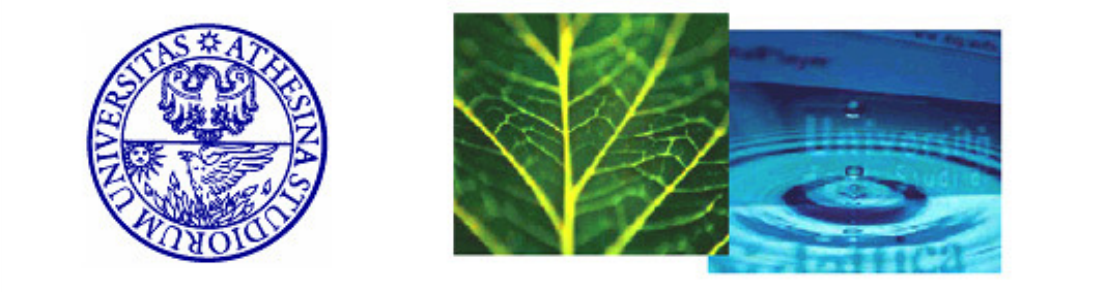}}\end{figure}
	\textbf{\Large{International Doctorate School in Information and\\Communication Technologies}}\\
	\vspace{0.4 cm}\LARGE{DISI - University of Trento}\\
	
	\vspace{0.8 cm}\huge{\textsc{\myTitle}}\\\vspace{0.8 cm}

	\begin{tabular}{l}\LARGE{\myName}\\\end{tabular}

	\begin{flushleft}\begin{tabular}{l}
		\Large{Advisor:}\\
		\Large{Dr. Sara Tonelli}\\
		\Large{Universit\`{a} degli Studi di Trento}\\
	\end{tabular}\end{flushleft}
	
	\vspace{4cm}\hrulefill\\\large{\myTime}
\end{center}

\end{titlepage}   
\thispagestyle{empty}

\hfill

\vfill

\noindent\myName: \textit{\myTitle} 
\textcopyright\ \myTime

%
%
%
%
%

\cleardoublepage
\thispagestyle{empty}
\refstepcounter{dummy}
\pdfbookmark[1]{Dedication}{Dedication}

\vspace*{3cm}

\begin{center}
    \emph{Post hoc ergo propter hoc.\\
    Post `doc' ergo propter `doc'.}   
\end{center}

\medskip

\cleardoublepage
\pdfbookmark[1]{Abstract}{Abstract}
\begingroup
\let\clearpage\relax
\let\cleardoublepage\relax
\let\cleardoublepage\relax

\chapter*{Abstract}

Structured information resulting from temporal information processing is crucial for a variety of natural language processing tasks, for instance to generate timeline summarization of events from news documents, or to answer temporal/causal-related questions about some events. In this thesis we present a framework for an integrated temporal and causal relation extraction system. We first develop a robust extraction component for each type of relations, i.e. temporal order and causality. We then combine the two extraction components into an integrated relation extraction system, CATENA---CAusal and Temporal relation Extraction from NAtural language texts---, by utilizing the presumption about event precedence in causality, that causing events must happened \texttt{BEFORE} resulting events. Several resources and techniques to improve our relation extraction systems are also discussed, including word embeddings and training data expansion. Finally, we report our adaptation efforts of temporal information processing for languages other than English, namely Italian and Indonesian.

\vspace{20pt}
\noindent\textbf{Keywords}\\
{[}Temporal information extraction, Temporal ordering of events, Event causality{]}

\vfill


\endgroup			

\cleardoublepage
\pdfbookmark[1]{Publications}{publications}
\chapter*{Publications}



\nocite{*}

\pubcite{mirza-tonelli:2016:ACL}
\pubcite{Mirza2016}
\pubcite{mirza-minard:2015:SemEval}
\pubcite{mirza-minard:2014:Evalita}
\pubcite{mirza-tonelli:2014:Coling}
\pubcite{mirza:2014:P14-3}
\pubcite{mirza-tonelli:2014:EACL}
\pubcite{mirza-EtAl:2014:CAtoCL}

\cleardoublepage
\pdfbookmark[1]{Acknowledgments}{acknowledgments}

\begin{flushright}{\slshape    
    Research is what I'm doing \\
    when I don't know what I'm doing.} \\ \medskip
    --- Wernher von Braun
\end{flushright}

\bigskip

\begingroup
\let\clearpage\relax
\let\cleardoublepage\relax
\let\cleardoublepage\relax
\chapter*{Acknowledgments}

First and foremost, I would like to express my sincere gratitude to my advisor, Sara Tonelli, for the continuous support during my PhD journey, for her patience, motivation, and immense knowledge. I could not have imagined having a better advisor and mentor. Thank you for helping me find out what I should do, when I don't know what I'm doing.

I would like to thank the members of my thesis examination committee: Alessandro Lenci, German Rigau and Steven Bethard, for their insightful comments, feedback and suggestions to widen my research taking into account various perspectives.

I thank the NewsReader Project\footnote{This work has been supported by the European Union's 7th Framework Programme via the NewsReader Project (ICT-
316404) \url{http://www.newsreader-project.eu/}.} and the people involved in it: Anne-Lyse Minard, Luciano Serafini, Manuela Speranza and Rachele Sprugnoli, for fruitful research collaborations. I also thank Rosella Gennari and Pierpaolo Vittorini for the `remote' collaboration; Ilija Ilievski, Min-Yen Kan and Hwee Tou Ng for the enlightening ideas and discussions in Singapore.

Thanks to \textit{``Algorithm for Temporal Information Processing of Text and their Applications''} by Oleksandr Kolomiyets, and \textit{``Finding Event, Temporal and Causal Structure in Text: A Machine Learning Approach''} by Steven Bethard, which served as \textit{``The Hitchhiker's Guide to the Thesis-writing Galaxy''} for this particular topic of PhD research.

Thanks to my fellow officemates---FBK Crew---for the stimulating and fun discussions during our lunches and coffee-breaks for the last three and a half years. Thanks to friends in Trento and Bolzano for their supports and cheers.

Thanks to my family---my mum, dad and my sister---for your endless support all these years; my dearest cats, for your warmth and purrs. Thanks to Simon, for everything.

\bigskip

\textit{Special thanks to `trees' for providing countless papers, for printing research papers and thesis drafts.}

\endgroup

\pagestyle{scrheadings}
\cleardoublepage
\refstepcounter{dummy}
\pdfbookmark[1]{\contentsname}{tableofcontents}
\setcounter{tocdepth}{2} 
\setcounter{secnumdepth}{3} 
\manualmark
\markboth{\spacedlowsmallcaps{\contentsname}}{\spacedlowsmallcaps{\contentsname}}
\tableofcontents 
\automark[section]{chapter}
\renewcommand{\chaptermark}[1]{\markboth{\spacedlowsmallcaps{#1}}{\spacedlowsmallcaps{#1}}}
\renewcommand{\sectionmark}[1]{\markright{\thesection\enspace\spacedlowsmallcaps{#1}}}
\clearpage

\begingroup 
    \let\clearpage\relax
    \let\cleardoublepage\relax
    \let\cleardoublepage\relax
    \refstepcounter{dummy}
    \pdfbookmark[1]{\listfigurename}{lof}
    \listoffigures

    \vspace{8ex}

    \refstepcounter{dummy}
    \pdfbookmark[1]{\listtablename}{lot}
    \listoftables
        
    \vspace{8ex}
    
%
       
\endgroup

\cleardoublepage\pagenumbering{arabic}
\cleardoublepage
\chapter{Introduction}\label{ch:introduction}


When the Greek government missed its 1.6 billion euro payment to the IMF as its bailout expired on 30 June 2015, people started to look for information such as, \textit{What is going on? Why did it happen and what will happen next?} While trying to relate current events to past events, news readers may ask themselves more questions, such as \textit{When did the crisis start? How did Greece get to this point?} A compact summary that represents the development of a story over time, be it over the course of one day, several months or even years, would be very beneficial for providing information that the readers need.  

\afterpage{
\begin{figure}
\centering
\includegraphics[scale=0.8]{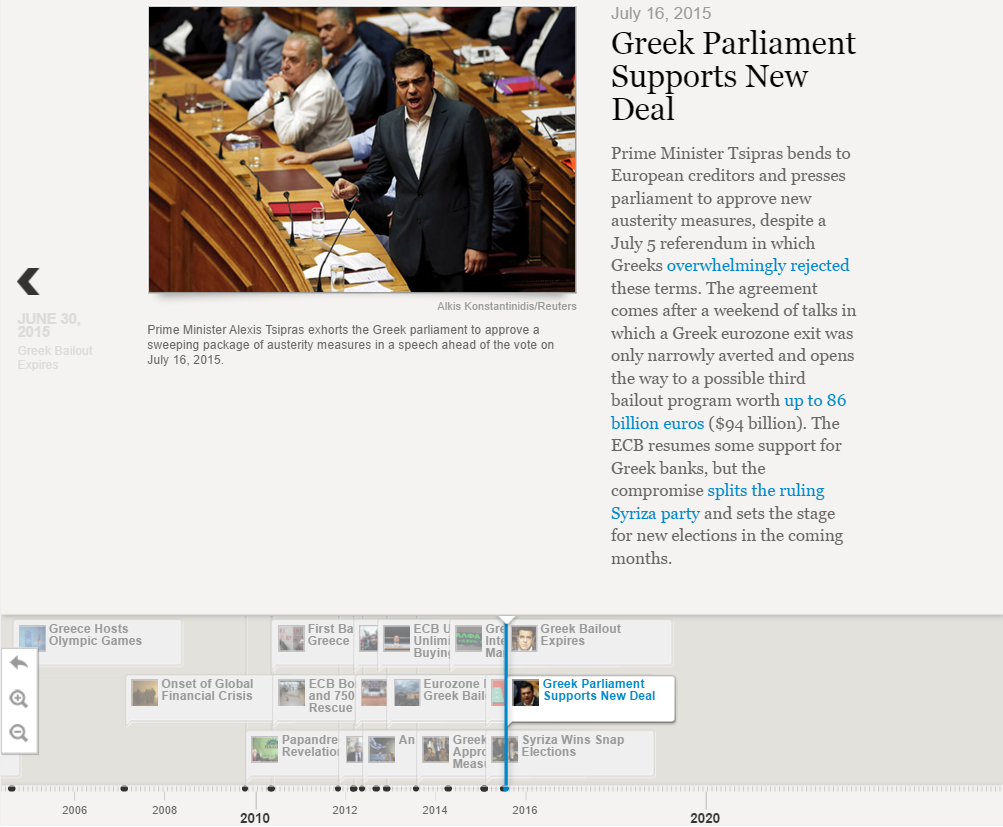}
\caption[The LOF caption]{Greece's debt crisis in a timeline\footnotemark\label{fig:timeline}}
\end{figure}
\footnotetext{Source: \url{http://www.cfr.org/greece/timeline-greeces-debt-crisis/p36451}}
}

Timeline summarization has become a widely adopted, natural way to present news stories in a compact manner. An example of a timeline is illustrated in Figure~\ref{fig:timeline}. News agencies often manually construct and maintain timelines for major events, but constructing such visual summaries requires a considerable amount of human effort and does not scale well, especially given enormous and expanding news data in the web, with millions of documents added daily. This is where \textit{information extraction} comes into play. Information extraction (IE) is part of natural language processing (NLP), and aims to automatically extract information from unstructured text into predefined structures.

Newspaper articles are often used to describe \textit{events} occurring in a certain time, and specify the \textit{temporal order} of these events. Consider, for example, the following excerpt from a news article in Figure~\ref{fig:timeline} published on July 16, 2015:

\begin{quotation}
\textit{Prime Minister Tsipras bends to European creditors and presses parliament to approve new austerity measures, despite a July 5 referendum in which Greeks overwhelmingly rejected these terms. The agreement comes after a weekend of talks in which a Greek eurozone exit was only narrowly averted and opens the way to a possible third bailout program worth up to 86 billion euros (\$94 billion). The ECB resumes some support for Greek banks, but the compromise splits the ruling Syriza party and sets the stage for new elections in the coming months.}
\end{quotation}

Human readers can easily comprehend that: \textit{there was a referendum on July 5}, \textit{there were talks that last for a weekend}, \textit{there was an agreement that comes after the talks} and \textit{there will be new elections in the coming months}; and order these facts in chronological order. They can also infer that \textit{the agreement} and \textit{the compromise} refer to the same entity, and that \textit{there will be new elections} because \textit{the ruling Syriza party is split}. This kind of text comprehension---building structured information about events and their temporal-causal relations---is an ultimate goal of \textit{temporal information processing}, in which the main task is extracting temporal information from texts.

Structured information resulting from temporal information processing is in fact crucial for a variety of natural language tasks, particularly summarization and question answering. In summarization tasks, given a large set of texts, a system is required to generate a much smaller text still containing all important contents of the original. For texts describing events, knowing which events are important and linking them in a temporal-causal structure would allow an automatic generation of a timeline-style summary. In question answering, a system is asked questions in natural language and expected to return the answers by looking for the appropriate information in a large set of documents. For example, having the temporal-causal structure about Greece's debt crisis would allow questions such as \textit{When did the talks resulting in the agreement take place?} or \textit{What is the reason for new elections in the coming months?} to be answered. To answer the first question it is necessary to infer that the talks happened during the weekend (most probably) before the news article is published. Meanwhile, answering the second question requires knowledge of the causing event, which is the \textit{splitting of the Syriza party}.

Furthermore, domain-specific structured temporal-causal information, e.g., about events involving a specific company extracted from financial news, or about chains of symptoms and diagnosis extracted from clinical reports, could be exploited in decision making support systems.

Building a system for extracting from text such temporal-causal information, specifically the temporal and causal relations between events found in the text, is the main focus of this thesis. Temporal and causal relations are closely related, as by common sense, a cause must precede its effect. In our research, we aim to exploit this presumption to improve the performance of our integrated temporal and causal relation extraction system. 




There are several annotation frameworks for modelling temporal information, i.e. temporal entities and relations, in a text. TimeML \parencite{pustejovsky2003}, being one of the prominent ones, is a language specification for \textit{events} and \textit{temporal expressions}, which was developed in the context of the TERQAS workshop\footnote{\texttt{http://www.timeml.org/site/terqas/index.html}}. An \textit{event} is defined as something that happens/occurs or a state that holds true, which can be expressed by a verb (e.g. \emph{killed}, \emph{acquire}), a noun (e.g. \emph{earthquake}, \emph{merger}), an adjective (e.g. \emph{injured}, \emph{retired}), as well as a nominalization either from verbs or adjectives (e.g. \emph{investigation}, \emph{bankruptcy}). 

The distinctive feature of TimeML is the separation of the representation of temporal entities, i.e. events and temporal expressions, from the anchoring or ordering dependencies. Instead of treating a temporal expression as an event argument, TimeML introduces \textit{temporal link} annotations to establish dependencies (temporal relations) between events and temporal expressions. Moreover, in TimeML, all types of events are annotated because every event takes part in the temporal network.\footnote{Except for \textit{generics} as in ``\textbf{Use} of corporate jets for political \textbf{travel} is legal.'' \parencite{timeml2006}} These are the main reason why we adopted the definitions of temporal entities and temporal relations from TimeML for this research.

As an illustration, consider our previous news excerpt, now annotated with temporal entities according to TimeML definitions:
\begin{quotation}
\textit{Prime Minister Tsipras \entity{bends} to European creditors and \entity{presses} parliament to \entity{approve} new austerity measures, despite a \entity{July 5} \entity{referendum} in which Greeks overwhelmingly \entity{rejected} these terms. The \entity{agreement} \entity{comes} after \entity{a weekend} of \entity{talks} in which a Greek eurozone \entity{exit} was only narrowly \entity{averted} and \entity{opens} the way to a possible third bailout \entity{program} worth up to 86 billion euros (\$94 billion). The ECB \entity{resumes} some \entity{support} for Greek banks, but the \entity{compromise} \entity{splits} the ruling Syriza party and \entity{sets} the stage for new \entity{elections} in \entity{the coming months}.}
\end{quotation}
Given such annotated texts, a relation extraction system should be able to identify, for example: \texttt{IS\_INCLUDED} (\textit{referendum}, \textit{July 5}), \texttt{DURING} (\textit{talks}, \textit{a weekend}), \texttt{AFTER} (\textit{agreement}, \textit{talks}), \texttt{IS\_INCLUDED} (\textit{elections}, \textit{the coming months}) and \texttt{CAUSE} (\textit{splits}, \textit{elections}).

\section{Motivations and Goals}
\label{sec:motiv-and-goals}

\paragraph{Temporal Relations}

\paragraph{} TimeML is the annotation framework used in a series of evaluation campaigns for temporal information processing called TempEval \parencite{verhagen-EtAl:2007:SemEval-2007,verhagen-EtAl:2010:SemEval,uzzaman-EtAl:2013:SemEval-2013}, in which the ultimate goal is the automatic identification of temporal expressions, events and temporal relations within a text. In TempEval, the temporal information processing task is divided into several sub-problems. Given a text, the extraction task basically includes: (i) identifying temporal entities mentioned in the text and (ii) identifying the temporal relations between them. In this research, we take the best performing systems in TempEval as our baseline.


The best performing extraction system for complete temporal information extraction achieves 30.98\% F1-score. According to the results reported in TempEval, the main limiting factor seems to be the low performance of temporal relation extraction systems (36.26\% F1-score). This is the main reason why we focus our research on temporal relation extraction. Meanwhile, the extraction systems for temporal entities already achieve quite good results (>80\% F1-scores). Therefore, and to limit the scope of our thesis, we assume that the annotation of temporal entities is already given.



In our attempt to improve the performance of the extraction system for temporal relations, we explore several research directions, which will be explained in the following paragraphs.

\paragraph{Causal Relations}
\begin{quotation}
\textit{A cause should always precede its effect. --- Anonymous}
\end{quotation}

The first research direction for improving the performance of temporal relation extraction is related to the connection between temporal and causal relations, based on the assumption that there is a temporal constraint in causality regarding event precedence. We aimed to investigate whether extracting causal relations between events can benefit temporal relation extraction. Apart from the efforts to improve the temporal relation extraction system, the recognition of causality between events is also crucial to reconstruct a causal chain of events in a story. This could be exploited, for example, in question answering systems, decision making support systems and for predicting future events given a chain of events. Having an integrated extraction system for both temporal and causal relations is one of the goals of this research. 

Unfortunately, unlike for temporal relations, there was no corpus available for building (and evaluating) an automatic extraction system for event causality, specifically the one that provides comprehensive account of how causality can be expressed in a text without limiting the effort to specific connectives. This motivated us to build annotation guidelines for explicit causality in text, and to annotate the TimeBank corpus, in which gold annotated events and temporal relations were already present. The resulting causality corpus, which we called \textit{Causal-TimeBank}, enabled the
adaptation of existing temporal processing systems to the extraction of causal information, and made it easier for us to investigate the relation between temporal and causal information.

\paragraph{Word Embeddings}
\begin{quotation}
\textit{You shall know a word by the company it keeps. --- \textcite{firth57synopsis}}
\end{quotation}

Word embeddings and deep learning techniques are gaining momentum in the NLP research, as they are seen as powerful tools to solve several NLP tasks, such as language modelling, relation extraction and sentiment analysis. Word embedding is a way to capture the semantics of a word via a low-dimensional vector, based on the distribution of other words around this word.

In this research, we explored the effect of using lexical semantic information about event words, based on word embeddings, on temporal relation extraction between events. For example, whether the word embeddings can capture that \textit{attack} often happens \texttt{BEFORE} \textit{injured}. 

\paragraph{Training Data Expansion}
\begin{quotation}
\textit{We don’t have better algorithms. We just have more data. --- Google’s Research Director Peter Norvig}
\end{quotation}

The scarcity of annotated data is often an issue in building extraction systems with supervised learning approach. One widely known approach to gain more training examples is semi-supervised learning, as for some NLP tasks it was shown that unlabelled data, when used in conjunction with a small amount of labelled data, can produce considerable improvement in learning accuracy.

We investigated two approaches to expand the training data for temporal and causal relation extraction, namely (i) \textit{temporal reasoning on demand} for temporal relation type classification and (ii) \textit{self-training}, a wrapper method for \textit{semi-supervised learning}, for causal relation extraction. 

\paragraph{}Finally,

\begin{quotation}
\textit{To have another language is to possess a second soul. --- Charlemagne}
\end{quotation}

Research on temporal information processing has been gaining a lot of attention from the NLP community, but most research efforts have focused only on English. In this research we explore the adaptation of our temporal information processing system for two languages other than English, i.e. Italian and Indonesian.

\section{Contributions}
\label{sec:contributions}

The following contributions are presented in this thesis:
\begin{itemize}[itemsep=1pt]
\item A hybrid approach for building an improved temporal relation extraction system, partly inspired by the sieve-based architecture of CAEVO \parencite{chambers-etal:2014:TACL}. Our approach is arguably more  efficient than CAEVO, because (i) the temporal closure inference over extracted temporal relations is run only once and (ii) we use less classifiers in general.
\item Annotation guidelines for annotating explicit causality between events, strongly inspired by TimeML. Compared with existing attempts for annotating causality in text, we aim to provide a more comprehensive account of how causality can be expressed in a text, without limiting the effort to specific connectives.
\item An event causality corpus, Causal-TimeBank, is made available to the research community, to support evaluations or developments of supervised learning systems for extracting causal relations between events.
\item A hybrid approach for building an improved causal relation extraction system, making use of the constructed event causality corpus.
\item An integrated extraction system for temporal and causal relations, which exploits the assumption about event precedence when two events are connected by causality.
\item Preliminary results on how word embeddings can be exploited for temporal relation extraction.
\item An investigation into the impact of training data expansion for temporal and causal relation extraction.
\item A summary of our adaptation efforts of temporal information processing for Italian and Indonesian languages. 
\end{itemize}

\section{Structure of the Thesis}
\label{sec:structure}

This thesis is organized as follows. In Chapter~\ref{ch:background}, we provide background information about natural language processing and information extraction, and discuss approaches widely used for information extraction tasks. Chapter~\ref{ch:auto-event-extraction} introduces the task of temporal information processing that comprises the TimeML annotation standard, annotated corpora and related evaluation campaigns. We also give a brief overview of state-of-the-art methods for extracting temporal information from text.

Chapter~\ref{ch:temp-rel-type} focuses on our hybrid approach for building an improved temporal relation extraction system. In Chapter~\ref{ch:annotating-causality} we present annotation guidelines for explicit causality between events. We also provide some statistics from the resulting causality-annotated corpus, Causal-TimeBank, on the behaviour of causal cues in a text. Chapter~\ref{ch:caus-rel-recognition} provides details on the hybrid approach for extracting causal relations between events from a text. In Chapter~\ref{ch:integrated-system} we describe our approach for building an integrated system for both temporal and causal relations, making use of the assumption about the temporal constraint of causality. 

Chapter~\ref{ch:deep-learning} provides preliminary results on the effects of using word embeddings for extracting temporal relations between events. Chapter~\ref{ch:training-data-expansion} discusses the impacts of our training data expansion approaches for temporal relation type classification and causal relation extraction. In Chapter~\ref{ch:multilinguality} we address the multilinguality issue, by providing a review of our adaptation efforts of the temporal information processing task for Italian and Indonesian. 

Finally, Chapter~\ref{ch:conclusion} discusses the lesson learned from this research work, and possible fruitful directions for future research.

\clearpage
\chapter{Background}
\label{ch:background}
\minitoc

In this chapter, we provide background information about natural language processing and information extraction, as well as methods and techniques widely used in the experiments described in this thesis.

\section{Natural Language Processing}


\subsection{Morphological Analysis}

Morphological analysis refers to the identification, analysis and description of the structure and formation of a given languages's \textit{morphemes} and other linguistic units, such as stems, affixes, part-of-speech, intonations and stresses, or implied context. A \textit{morpheme} is defined as the smallest meaningful unit of a language. Consider a word like \textit{unhappiness} containing three morphemes, each carrying a certain amount of meaning: \textit{un} means ``not'', \textit{ness} means ``being in a state or condition'' and \textit{happy}. \textit{Happy} is a \textit{free morpheme}, and considered as a \textit{root}, because it can appear on its own. \textit{Bound morphemes}, typically \textit{affixes}, have to be attached to a free morpheme, thus, we cannot have sentences in English such as ``Jason feels very un ness today''. Morphological analysis is a very important step for natural language processing, especially when dealing with morphologically complex languages. 


\paragraph{Stemming and Lemmatization} A \textit{stem} may be a root (e.g. \textit{run}) or a word with derivational morphemes (e.g. the derived verbs \textit{standard-ize}). For instance, the root of \textit{destabilized} is \textit{stabil-} (i.e. a form of \textit{stable} that does not occur alone), and the stem is \textit{de-stabil-ize}, which includes the derivational affixes \textit{de-} and \textit{-ize} but not the inflectional past tense suffix \textit{-(e)d}. In other words, a stem is a part of a word that inflectional affixes attach to. A \textit{lemma} refers to a dictionary form of a word. A typical example of this are the words see, \textit{sees}, \textit{seeing} and \textit{saw}, which all have the same \textit{see}-lemma.

\paragraph{Part-of-Speech Tagging} In natural language, words are divided into two broad categories: \textit{open} and \textit{closed} classes. Open classes do not have a fixed word membership, and encompass nouns, verbs, adjectives and adverbs. Closed classes, contrastingly, have a relatively fixed word membership. They include function words, such as articles, prepositions, auxiliary verbs and pronouns, which have a high occurrence frequency in linguistic expressions. Part-of-speech (PoS) tagging is  the problem of assigning each word in a sentence the part of speech that it assumes in that sentence, according to their different lexical categories (noun, verb, adjective, adverb, preposition, pronoun, etc.).

A PoS tagset specifies the set of PoS categories being distinguished and provides a list of tags used to denote each of those categories. The commonly used PoS tagsets include Penn Treebank PoS Tagset\footnote{\url{http://www.comp.leeds.ac.uk/amalgam/tagsets/upenn.html}} \parencite{Marcus:1993:BLA:972470.972475, santorini93penntagset}, the British National Corpus (BNC) Tagset and the BNC Enriched Tagset\footnote{\url{http://www.natcorp.ox.ac.uk/docs/gramtag.html}} \parencite{leech1994claws}. The difference between the text annotated with Penn Treebank PoS Tagset and BNC (Basic) Tagset is exemplified in the following sentence (i) and (ii), respectively.

\begin{enumerate}[label=(\roman*)]
\item \textit{I}/\texttt{PRP} \textit{saw}/\texttt{VBD} \textit{a}/\texttt{DT} \textit{boy}/\texttt{NN} \textit{with}/\texttt{IN} \textit{a}/\texttt{DT} \textit{dog}/\texttt{NN} \textit{.}/\texttt{.}
\item \textit{I}/\texttt{PNP} \textit{saw}/\texttt{VVD} \textit{a}/\texttt{AT0} \textit{boy}/\texttt{NN1} \textit{with}/\texttt{PRP} \textit{a}/\texttt{AT0} \textit{dog}/\texttt{NN1} \textit{.}/\texttt{PUN}
\end{enumerate}

There are a total of 48 tags in the Penn Treebank PoS Tagset, while the BNC Basic Tagset, also known as the C5 Tagset, distinguishes a total of 61 categories. Notably, the C5 Tagset includes separate categories for the various forms of the verbs \textit{be}, \textit{do} and \textit{have}.

The Penn Treebank PoS Tagset is used in the Stanford CoreNLP tool suite\footnote{\url{http://stanfordnlp.github.io/CoreNLP/}} \parencite{manning-EtAl:2014:P14-5}. Meanwhile, the TextPro tool suite\footnote{\url{http://hlt-services2.fbk.eu/textpro/}} \parencite{PIANTA08.645}, which is the one mainly used in our research, employs the BNC Basic Tagset.

\subsection{Syntactic Analysis}

\paragraph{Syntactic Parsing} 
\begin{displayquote}
\textit{Parsing means taking an input and producing some sort of linguistic structure for it.} --- \textcite{Jurafsky:2000:SLP:555733}
\end{displayquote}

\begin{figure}[t]
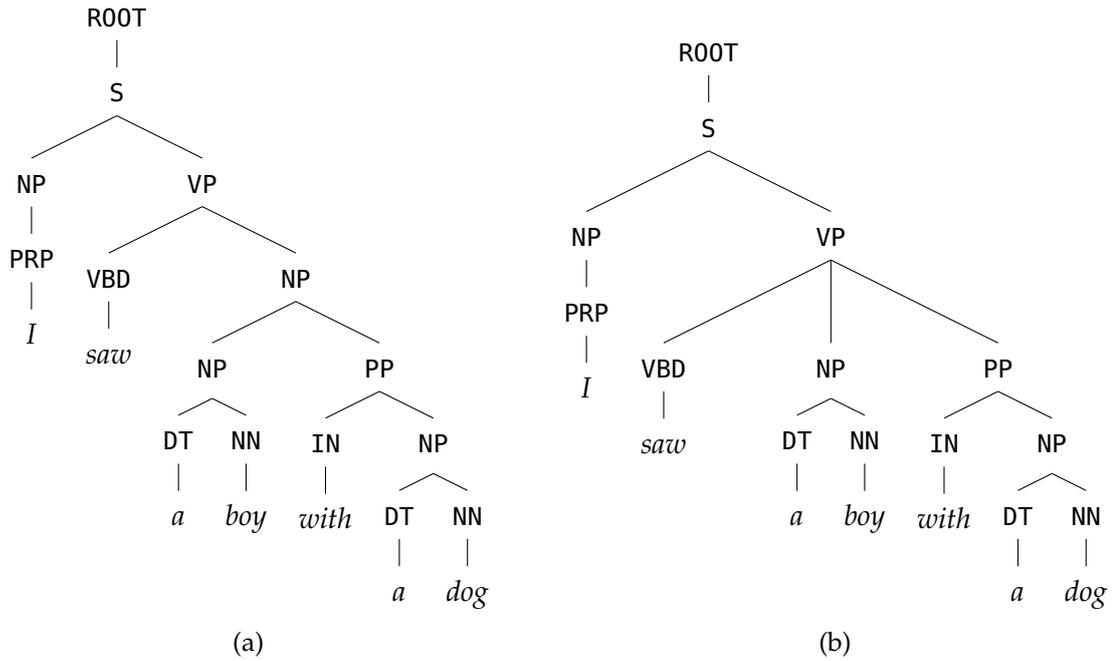

\begin{subfigure}[b]{0.5\textwidth}
\Tree [.\texttt{ROOT} [.\texttt{S} [.\texttt{NP} [.\texttt{PRP} \textit{I} ] ] [.\texttt{VP} [.\texttt{VBD} \textit{saw} ] [.\texttt{NP} [.\texttt{NP} [.\texttt{DT} \textit{a} ] [.\texttt{NN} \textit{boy} ] ] [.\texttt{PP} [.\texttt{IN} \textit{with} ] [.\texttt{NP} [.\texttt{DT} \textit{a} ] [.\texttt{NN} \textit{dog} ] ] ] ] ] ] ]
\caption{}
\label{fig:syntactic-tree-a}
\end{subfigure}
\begin{subfigure}[b]{0.5\textwidth}
\Tree [.\texttt{ROOT} [.\texttt{S} [.\texttt{NP} [.\texttt{PRP} \textit{I} ] ] [.\texttt{VP} [.\texttt{VBD} \textit{saw} ] [.\texttt{NP} [.\texttt{DT} \textit{a} ] [.\texttt{NN} \textit{boy} ] ] [.\texttt{PP} [.\texttt{IN} \textit{with} ] [.\texttt{NP} [.\texttt{DT} \textit{a} ] [.\texttt{NN} \textit{dog} ] ] ] ] ] ]
\caption{}
\label{fig:syntactic-tree-b}
\end{subfigure}
\caption{Two variants of syntactic parse trees for ``I saw a boy with a dog''. The interpretation represented by (a) is the most likely semantic representation and means that ``\textit{the boy}'' was ``\textit{with a dog}''.}
\label{fig:syntactic-tree}
\end{figure}

A syntactic parser takes a sentence as input and produces a syntactic structure that corresponds to a semantic interpretation of the sentence. For example, the sentence ``\textit{I saw a boy with a dog}'' can be parsed in two different ways (Figure~\ref{fig:syntactic-tree}). This divergence is caused by two possible interpretations of the sentence: (a) and (b). While both are grammatically correct, they reflect two different meanings: (a) the phrase ``\textit{a dog}'' is attached to ``\textit{a boy}'' which means \textit{accompanied}; (b) the phrase ``\textit{a dog}'' is attached to ``\textit{saw}'' which means a tool used to make the observation. The major challenge for a syntactic parser is to find the correct parse(s) from an exponential number of possible parses.

In terms of its overall structure, the parse tree is always rooted at a node \texttt{ROOT}, with the terminal elements that relate to actual words in the sentence. Each of its sub-parses, or \textit{internal nodes}, spans over several tokens, and is characterized by a set of syntactic types (e.g., NP and VP, which denote noun and verb phrases, resp.). The most important word in that span is called the \textit{head word}. In this work we will also refer to \textit{syntactically dominant} and \textit{governing} verbs. Syntactically dominant verbs are the verbs that are located closer to the root of the entire parse tree. For a number of words in a textual span, governing verbs are the verbs in verb phrases that are the roots of the corresponding sub-trees. For example, for the sentence in Figure~\ref{fig:syntactic-tree}, the verb ``\textit{saw}'' is the syntactically dominant verb of the sentence, and the governing verb for the textual span ``\textit{a boy with a dog}''.

\begin{figure}[th!]
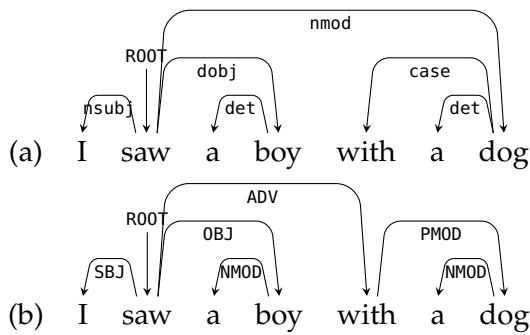

\centering
\begin{dependency}[text only label, label style={below}]
\begin{deptext}[column sep=.7em]
(a) \& I \& saw \& a \& boy \& with \& a \& dog \\
\end{deptext}
\deproot[edge unit distance=2ex]{3}{\texttt{ROOT}}
\depedge{3}{2}{\texttt{nsubj}}
\depedge{5}{4}{\texttt{det}}
\depedge{8}{7}{\texttt{det}}
\depedge{3}{5}{\texttt{dobj}}
\depedge{8}{6}{\texttt{case}}
\depedge[edge unit distance=1.8ex]{3}{8}{\texttt{nmod}}
\end{dependency}

\begin{dependency}[text only label, label style={below}]
\begin{deptext}[column sep=.7em]
(b) \& I \& saw \& a \& boy \& with \& a \& dog \\
\end{deptext}
\deproot[edge unit distance=2ex]{3}{\texttt{ROOT}}
\depedge{3}{2}{\texttt{SBJ}}
\depedge{5}{4}{\texttt{NMOD}}
\depedge{8}{7}{\texttt{NMOD}}
\depedge{3}{5}{\texttt{OBJ}}
\depedge{6}{8}{\texttt{PMOD}}
\depedge{3}{6}{\texttt{ADV}}
\end{dependency}
\caption{Dependency trees for a sentence ``I saw a boy with a dog'', using (a) Stanford CoreNLP tool suite and (b) Mate tools.}
\label{fig:dependency-tree}
\end{figure}

\paragraph{Dependency Parsing} In contrast to syntactic parsing, where the linguistic structure is formulated by the grammar that organizes the sentences' words into phrases, word dependency formalism orders them according to binary dependency relations between the words (as between a \textit{head} and a \textit{dependent}). Word dependency formalism is often referenced as an effective mean to represent the linguistic structures of languages with a relatively free word-order.

Examples of dependency parses for a sentence ``\textit{I saw a boy with a dog}'' are presented in Figure~\ref{fig:dependency-tree}. There are several dependency representations, such as (a) Stanford (Typed) Dependencies~\parencite{demarneffe:2008:stanford} used in Stanford CoreNLP, and (b) CoNLL-2008 Shared Task Syntactic Dependencies~\parencite{surdeanu-EtAl:2008:CONLL} used in Mate tools\footnote{\url{https://code.google.com/archive/p/mate-tools/}}~\parencite{bjorkelund-EtAl:2010:COLING-DEMOS}.

\subsection{Information Extraction}

Information extraction is a broad research field that uses computer algorithms to extract predefined structured information from natural language text, where elements of the structure relate to textual spans in the input. With the exception of temporal information processing, which will be explained further in Chapter~\ref{ch:auto-event-extraction}, the different tasks of information extraction are listed in the following sections.

\paragraph{Named-Entity Recognition} Named-entity recognition is a task of information extraction that categorizes single textual elements in text in terms of a set of common criterion (persons, organizations, locations, times, numbers, etc.).

\begin{displayquote}
\textit{The violent clashes between the security forces and protesters have lasted [\wordattr{two days}{\texttt{Date}}] in [\wordattr{Cairo}{\texttt{Location}}] and other cities.}
\end{displayquote}

In the example, the textual span ``\textit{two days}'' is identified and classified as an instance of \texttt{Date}, while the span of ``\textit{Cairo}'' is identified and classified as an instance of \texttt{Location}.

\paragraph{Word-Sense Disambiguation} The task of word-sense disambiguation is to assign a label to every noun phrase, (non-auxiliary) verb phrase, adverb and adjective in a text. This label indicates the meaning of its attached word, and is chosen from a dictionary of meanings for a large number of phrases.

\begin{displayquote}
\textit{The [\wordattr{violent}{\texttt{violent.01}}] [\wordattr{clashes}{\texttt{clash.04}}] between the [\wordattr{security}{\texttt{security.03}}] [\wordattr{forces}{\texttt{force.01}}] and [\wordattr{protesters}{\texttt{protester.02}}] have [\wordattr{lasted}{\texttt{last.01}}] [\wordattr{two}{\texttt{two.01}}] [\wordattr{days}{\texttt{day.04}}] in [\wordattr{Cairo}{\texttt{Cairo.02}}] and [\wordattr{other}{\texttt{other.01}}] [\wordattr{cities}{\texttt{city.01}}].}
\end{displayquote}

In the example, the meanings are assigned the labels of synsets in the WordNet lexical database~\parencite{Fellbaum-98}, e.g., the word ``\textit{clashes}'' receives the label \texttt{clash.04} which means ``\textit{fight}'' or ``\textit{fighting}'', whereas the most common sense \texttt{clash.01} stands for ``\textit{clang}'' or ``\textit{noise}''.


\paragraph{Semantic Role Labelling} Semantic Role Labelling (SRL) consists of the detection of the semantic arguments associated with the \textit{predicate} or \textit{verb} of a sentence, and their classification into their specific \textit{roles}. For example, given a sentence like \textit{``Mary sold the car to John''}, the task would be to recognize the verb \textit{``to sell''} as the predicate, \textit{``Mary''} as the seller (agent), \textit{``the car''} as the goods (theme) and \textit{``John''} as the recipient. The task is seen as an important step towards making sense of the meaning of a sentence, which is at a higher-level of abstraction than a syntactic tree. For instance, \textit{``The car has been sold by Mary to John''} has a different syntactic form, but the same semantic roles.


The FrameNet project \parencite{baker-fillmore-lowe:1998:ACLCOLING} produced the first major computational lexicon that systematically described many predicates and their corresponding roles. \textcite{Gildea:2002:ALS:643092.643093} developed the first automatic semantic role labeling system based on FrameNet. FrameNet additionally captures relationships between different frames, including among others: \textit{Precedes}, which captures a temporal order that holds between subframes of a complex scenario, and \textit{Causative\_of}, which expresses causality between frames.

Another project related to semantic role labelling is the PropBank project \parencite{J05-1004}, which added semantic role---or predicate-argument relations---annotations to the syntactic tree of the Penn Treebank corpus \parencite{PRASAD08.754}. The PropBank annotation is exemplified in the following sentence:

\begin{displayquote}
\textit{[\wordattr{The violent clashes}{\texttt{Arg1}}] between the security forces and protesters have [\wordattr{lasted}{\texttt{last.01}}] [\wordattr{two days}{\texttt{Arg2}}] in [\wordattr{Cairo and other cities}{\texttt{Arg-Loc}}].}
\end{displayquote}

Here the verb ``\textit{lasted}'' has a predicate label \texttt{last.01}, which means ``\textit{extend for some period of time}''. The related words have semantic roles:
\begin{itemize}
\item \texttt{Arg1} for ``\textit{The violent clashes}'', denoting \textit{thing that lasts}
\item \texttt{Arg2} for ``\textit{two days}'', denoting \textit{period of time}
\item \texttt{Arg-Loc} for ``\textit{Cairo and other cities}'', denoting \textit{location}
\end{itemize}



\paragraph{Coreference Resolution} Given a sentence or larger chunk of text, the task is to determine which words---\textit{mentions}---refer to the same objects---\textit{entities}. Anaphora resolution is a special case of this task, which is concerned with matching up pronouns with the nouns or names that they refer to. 

Another typical coreference problem is to find links between previously-extracted named entities. For example, ``\textit{International Business Machines}'' and ``\textit{IBM}'' might refer to the same real-world entity. If we take the two sentences ``\textit{M. Smith likes fishing. But he doesn't like biking}'', it would be beneficial to detect that ``\textit{he}'' is referring to the previously detected person ``\textit{M. Smith}''.

\paragraph{Relationship Extraction} This task basically deals with the identification of relations between entities, including:
\begin{itemize}
\item Compound noun relations: recognition of relations between two nouns.
\item (Geo)spatial analysis: recognition of trajectors, landmarks, frames of reference, paths, regions, directions and motions, and relations between them.
\item Discourse analysis: recognition of non-overlapping text spans and discourse relations between them.
\end{itemize}

\section{Techniques for Information Extraction}

\subsection{Rule-based Methods}

Rule-based methods are the earliest ones used in information extraction. A rule-based system makes use of a database of predefined and hand-crafted rules that specify knowledge typically in form of \textit{regular expressions}. Regular expressions are a linguistic formalism that is based on a regular grammar---one of the simplest classes of formal language grammars \parencite{journals/iandc/Chomsky59a}. 

Regular expressions are a declarative mechanism for specifying declarative languages based on regular grammars. Regular grammars are recognized by a computation device, called \textit{finite state automaton} (FSA). A finite state automaton \parencite{Hopcroft:1969:FLR:1096945} is a five-tuple $(\Theta, \theta_0, \Sigma, \delta, F)$, where $\Theta$ is a finite set of states, $\theta_0$ is the initial state, $\Sigma$ is a finite set of alphabet symbols, $\delta : \Theta \times \Sigma \times \times \Theta $ is  a relation from states and alphabet symbols to states, and $F \subseteq \Theta$ is a set of final states. The extension of $\delta$ to handle input strings is standard and denoted by $\delta^\ast$. $\delta^\ast(\theta, a)$ denotes the state reached from $\theta$ on reading the string $a$. A string $a$ is said to be accepted by an FSA if $\delta^\ast(\theta, a) \in F$. The language $A_L$ is the set of all strings accepted by $A_L$'s FSA. Strings that are not accepted by $A_L$'s FSA are outside of the language $A_L$.

Systems based on regular expressions are considered as rule-based systems in which knowledge about the domain is encoded in regular expressions. If the input string is accepted, i.e., it matches one of the regular expressions, it is labelled with a class label associated with that particular rule. In natural language processing, rule-based approaches were applied for, among others, \textit{tokenization}---identifying the spans of single tokens in a text, \textit{stemming}---finding the stem of a token, and \textit{Part-of-Speech (PoS) tagging}.

\begin{figure}
\small
\texttt{time = \textasciicircum((0?[0-9]|1[012])([:.][0-9]{2})?(\textbackslash s?[ap]m)|([01]?[0-9]|2[0-3])([:.][0-9]{2})?)\$ \\
date = \textasciicircum[1-9]|[1-2][0-9]|3[0-1])\$ \\
}
\caption{Regular expressions for extracting \textit{time} and \textit{date} in the POSIX Extended Regular Expression (ERE) syntax.}
\label{fig:regex}
\end{figure}

Figure~\ref{fig:regex} shows regular expression examples in the POSIX Extended Regular Expressions (ERE) syntax to extract \textit{time} and \textit{date} from a text.

Always traditionally popular, rule-based techniques have long been utilized for small-size applications and applications for new domains. However, with the development of large annotated corpora, machine learning techniques have grown increasingly popular, with users beginning to compare their performance to rule-based methods. These comparative studies have found that rule-based systems are very difficult to maintain, and that such systems are not well-scalable. Nevertheless, there are problems which can only be solved by the rule-based approach. Main reasons to still employ rule-based systems are:
\begin{itemize}[itemsep=1px]
\item New, small or restricted application domains.
\item Short development time for a set of generally applicable and observable rules.
\item Absence of annotated training data.
\item Poor quality of training data.
\end{itemize}

\subsection{Supervised Machine Learning}

Since rule-writing requires enormous human effort, an easier approach would be to utilize existing examples, i.e. \textit{annotations}, to extract the rules automatically; or to use statistics, which can predict the labels of words, phrases, sentences or even the entire document. In the following sections we describe a number of state-of-the-art supervised machine learning methods that are currently used in the field of natural language processing and information extraction. The focus of supervised approaches in NLP has hitherto been limited to feature extraction (how an object under consideration is represented in a numerical way as a vector of features), and selecting appropriate machine learning methods.

\paragraph{Formal Definitions} In terms of supervised machine learning, the labelling task can be defined as: given a set of $n$ observations ${x_1, x_2, ..., x_n}$ with their corresponding target class value $y_1, y_2, ..., y_n$, the goal is to predict the value of $y$ for an unseen instance $x$. More formally, it can be defined as $Func(x):x\rightarrow y$, where each instance $x$ is represented as a vector of feature values, i.e., $x=[f_1, f_2, ..., f_m]$, with $m$ being the total number of features used in the representation. Depending on the number of distinct target values of $y$, one distinguishes between \textit{binary} (with two target values) and \textit{multi-class} classifications. In the following sections we describe the commonly used machine learning methods to model the prediction function $Func(x)$.

\paragraph{Support Vector Machines} Support Vector Machines (SVMs) \parencite{Cortes:1995:SN:218919.218929} is a well-known discriminative machine learning classifier that models the data as points in a high-dimensional space, and spatially separates them as far as possible. Technically, an SVM constructs a hyperplane, which can be used for classification, regression or other tasks. The best separation of data points is achieved by the hyperplane that has the largest distance to the nearest data point of any class.

\begin{figure}
\centering
\includegraphics[scale=0.9]{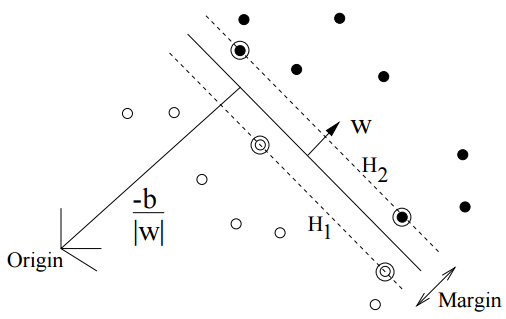}
\caption{Support Vector Machines with two characteristics hyperplanes $H_1$ and $H_2$ \parencite{Burges:1998:TSV:593419.593463}. The data points $\mathbf{x}$ that lie on the hyperplanes $H_1$ and $H_2$ are called \textit{support vectors} (circled), satisfying $\textbf{w} \cdot \mathbf{x} + b=0$, where $\textbf{w}$ is normal to the hyperplane, $\frac{|b|}{\|w\|}$ is the perpendicular distance from the hyperplane to the origin, and $\|w\|$ is the Euclidean norm of $\textbf{w}$.}
\label{fig:svm}
\end{figure}

Formally, an SVM is defined as: given a set of observations ${x_1, x_2, \dots, x_n}$ with a corresponding set of labels ${y_1, y_2, \dots, y_n}$, where $y_i \in {-1, +1}$, the separating hyperplane $H_0$ that divides the data points in space can be defined as:
\begin{equation}
\textbf{w} \cdot \mathbf{x} + b = 0
\end{equation}
where $\textbf{w}$ is the normal vector to the hyperplane, $\mathbf{x}$ is a set of points $x_i$ that lie on the hyperplane, and $\cdot$ denotes the dot product (see Figure~\ref{fig:svm}).

We can select two others hyperplanes $H_1$ and $H_2$ which also separate the data and defined as:
\begin{equation}
\textbf{w} \cdot \mathbf{x_i} + b \geq 1 \textrm{ for } y_i = +1
\label{eq:h1}
\end{equation}
and
\begin{equation}
\mathbf{w} \cdot \mathbf{x_i} + b \leq -1 \textrm{ for } y_i = -1
\label{eq:h2}
\end{equation}
so that $H_0$ is equidistant from $H_1$ and $H_2$, and taking into consideration the constraint that there is no data point between the two hyperplanes. Equation (~\ref{eq:h1}) and (\ref{eq:h2}) can be combined into a single constraint:
\begin{equation}
y_i \left( \mathbf{w} \cdot \mathbf{x_i} + b \right) \geq 1 \textrm{ for all } 1 \leq i \leq n
\label{eq:svm-constraint}
\end{equation}

The optimal hyperplane $\mathbf{w_0} \cdot \mathbf{x} + b_0 = 0$ is the unique one that separates the training data with a maximal margin, i.e., the distance $\frac{2}{\|w_0\|}$ between the two hyperplanes $H_1$ and $H_2$ is maximal. This means that the optimal hyperplane is the unique one that minimizes $\mathbf{w} \cdot \mathbf{w}$ under the constraint (\ref{eq:svm-constraint}). 

Consider the case where the training data cannot be separated without error. In this case
one may want to separate the training set with a minimal number of errors. To express this
formally some non-negative variables $\xi_i \geq 0, i=1 \dots l$, are introduced. The problem of finding the optimal \textit{soft-margin} hyperplane is then defined as:
\begin{equation}
\begin{split}
\min_{\mathbf{w},b,\xi}\hspace{5pt}&\frac{1}{2}\mathbf{w} \cdot \mathbf{w} + C \sum_{i=1}^l \xi_i\\
\textrm{subject to}\hspace{5pt}&y_i \left( \mathbf{w} \cdot \phi(\mathbf{x_i}) + b \right) \geq 1 -\xi_i,\\
&\xi_i \geq 0, i = 1, \dots, l\\
\end{split}
\end{equation}
where the training vectors $\mathbf{x_i}$ are mapped to a higher dimensional space by the function $\phi$. $C > 0$ is the penalty parameter of the error term. Furthermore, $K(\mathbf{x_i}, \mathbf{x_j}) \equiv \phi(\mathbf{x_i}) \cdot \phi(\mathbf{x_j})$ is called the kernel function. Though new
kernels are being proposed by researchers, the basic kernels include:
\begin{itemize}
\item linear: $K(\mathbf{x_i}, \mathbf{x_j}) = \mathbf{x_i} \cdot \mathbf{x_j}$
\item polynomial: $K(\mathbf{x_i}, \mathbf{x_j}) = (\gamma \mathbf{x_i} \cdot \mathbf{x_j} + r)^d, \gamma > 0$
\item radial basis function (RBF): $K(\mathbf{x_i}, \mathbf{x_j}) = \exp (-\gamma\|\mathbf{x_i} - \mathbf{x_j}\|^2), \gamma > 0$
\item sigmoid: $\tanh (\gamma \mathbf{x_i} \cdot \mathbf{x_j} + r)$
\end{itemize}

The earliest used implementation for SVM multi-class classification is probably the \textit{one-against-all} method. It construct $k$ SVM models where $k$ is the number of classes. The $m$th SVM is trained with all of the examples in the $m$th class with positive labels, and all other examples with negative labels. Thus, given $l$ training data $(x_1, y_1), \dots (x_l, y_l)$, where $x_i \in R^n$, $i= 1, \dots, l$ and $y_i \in {1, \dots, k}$ is the class of $x_i$, the $m$th SVM solves the following problem:
\begin{equation}
\begin{split}
\min_{\mathbf{w^m},b^m,\xi^m}\hspace{5pt}&\frac{1}{2}\mathbf{w^m} \cdot \mathbf{w^m} + C \sum_{i=1}^l \xi_i^m\\
\textrm{subject to}\hspace{5pt}&\left( \mathbf{w^m} \cdot \phi(\mathbf{x_i}) + b^m \right) \geq 1 -\xi_i^m,\hspace{5pt}&\textrm{if } y_i=m,\\
&\left( \mathbf{w^m} \cdot \phi(\mathbf{x_i}) + b^m \right) \leq -1 +\xi_i^m,\hspace{5pt}&\textrm{if } y_i \neq m,\\
&\xi_i^m \geq 0.\\
\end{split}
\label{eq:svm-one-vs-all}
\end{equation}

After solving (\ref{eq:svm-one-vs-all}), there are $k$ decision functions: $\mathbf{w^1} \cdot \phi(\mathbf{x})+b^1, \dots, \mathbf{w^k} \cdot \phi(\mathbf{x})+b^k$. We say $x$ is in the class which has the largest value of the decision function:
\begin{equation}
\textrm{class of } x \equiv \textrm{arg}\max\nolimits_{m=1,\dots,k}(\mathbf{w^m} \cdot \phi(\mathbf{x}) + b^m)
\end{equation}

Another major method is called the \textit{one-against-one} method, introduced by \textcite{Knerr1990}. This method constructs $k(k-1)/2$ classifiers where each one is trained on data from two classes. For the training data from the $i$th and the $j$th classes, we solve the following binary classification problem:
\begin{equation}
\begin{split}
\min_{\mathbf{w^{ij}},b^{ij},\xi^{ij}}\hspace{5pt}&\frac{1}{2}\mathbf{w^{ij}} \cdot \mathbf{w^{ij}} + C \sum_{t} \xi_t^{ij}\\
\textrm{subject to}\hspace{5pt}&\left( \mathbf{w^{ij}} \cdot \phi(\mathbf{x_t}) + b^{ij} \right) \geq 1 -\xi_t^{ij},\hspace{5pt}&\textrm{if } y_t=i,\\
&\left( \mathbf{w^{ij}} \cdot \phi(\mathbf{x_t}) + b^{ij} \right) \leq -1 +\xi_t^{ij},\hspace{5pt}&\textrm{if } y_t = j,\\
&\xi_t^{ij} \geq 0.\\
\end{split}
\label{eq:svm-one-vs-one}
\end{equation}
There are different methods for doing the future testing after all $k(k-1)/2$ classifiers are constructed. For instance, the following voting strategy suggested by \textcite{Friedman:96} may be used: if $\textrm{sign}(\mathbf{w^{ij}} \cdot \phi(\mathbf{x_t}) + b^{ij})$ says $x$ is in the $i$th class, then the vote for the $i$th class is added by one. Otherwise, the $j$th is increased by one. Then we predict $x$ is in the class with the largest vote. This voting approach is also called the ``Max Wins'' strategy. In case that two classes have identical votes, the one with the smaller index is usually selected, though it may not be a good strategy.


\paragraph{Logistic Regression} \textit{Logistic regression}\footnote{We took the explanations about Logistic Regression from \url{http://www.cs.cmu.edu/~tom/NewChapters.html} by Tom Mitchell.} is an approach to learning functions of the form $f : X \rightarrow Y$, or $P(Y|X)$ in the case where $Y$ is discrete-valued, and $X = \langle X_1 \dots X_n \rangle$ is any vector containing discrete or continuous variables.

\begin{figure}
\centering
\includegraphics[scale=0.8]{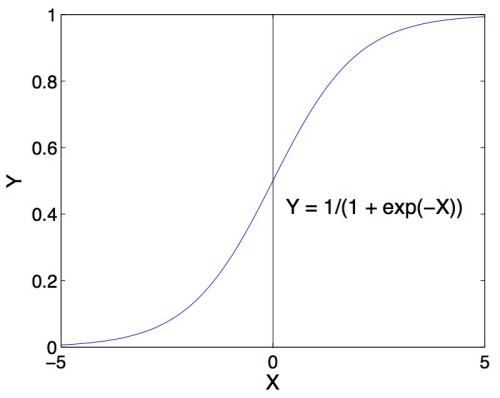}
\caption{Form of logistic function. In Logistic Regression, $P(Y|X)$ is assumed to follow this form.}
\label{fig:svm}
\end{figure}

Logistic Regression assumes a parametric form for the distribution $P(Y|X)$, then directly estimates its parameters from the training data. The parametric model assumed by Logistic Regression in the case where $Y$ is boolean is:
\begin{equation}
P(Y=1|X)=\frac{1}{1 + \exp(w_0 + \sum_{i=1}^n w_i X_i)}
\label{eq:logit1}
\end{equation}
and
\begin{equation}
P(Y=0|X)=\frac{\exp(w_0 + \sum_{i=1}^n w_i X_i)}{1 + \exp(w_0 + \sum_{i=1}^n w_i X_i)}
\label{eq:logit2}
\end{equation}
Note that equation (\ref{eq:logit2}) follows directly from equation (\ref{eq:logit1}), because the sum of these two probabilities must equal to 1.

One highly convenient property of this form for $P(Y|X)$ is that it leads to a simple linear expression for classification. To classify any given $X$ we generally want to assign the value $y_k$ maximizing $P(Y=y_k|X)$. Put another way, we assign the label $Y=0$ if the following condition holds:
\begin{equation}
1 < \frac{P(Y=0|X)}{P(Y=1|X)}
\end{equation}
substituting from equations (\ref{eq:logit1}) and (\ref{eq:logit2}), this becomes
\begin{equation}
1 < exp(w_0 + \sum_{i=1}^n w_i X_i)
\end{equation}
and taking the natural log of both sides we have a linear classification rule that assigns label $Y=0$ if $X$ satisfies
\begin{equation}
0 < w_0 + \sum_{i=1}^n w_i X_i
\end{equation}
and assigns $Y=1$ otherwise.

One reasonable approach to train a logistic regression model is to choose parameter values that maximize the conditional data likelihood. The conditional data likelihood is the probability of the observed $Y$ values in the training data, conditioned on their corresponding $X$ values. We choose parameters $W$ that satisfy
\begin{equation}
W \leftarrow \textrm{arg} \max_W \prod_l P(Y^l|X^l, W)
\end{equation}
where $W = \langle w_0, w_1 \dots w_n \rangle$ is the vector of parameters to be estimated, $Y^l$ denotes the observed value of $Y$ in the $l$th training example, and $X^l$ denotes the observed value of $X$ in the $l$th training example. The expression to the right of the $\textrm{arg} \max$ is the conditional data likelihood. Equivalently, we can work with the log of the conditional likelihood:
\begin{equation}
W \leftarrow \textrm{arg} \max_W \sum_l \ln P(Y^l|X^l, W)
\end{equation}

Above we considered using Logistic Regression to learn $P(Y|X)$ only for the case where $Y$ is a boolean variable, i.e. binary classification. If $Y$ can take on any of the discrete values ${y_1, \dots y_K}$, then the form of $P(Y=y_k|X)$ for $Y=y_1, Y=y_2, \dots Y=y_{K-1}$ is:
\begin{equation}
P(Y=y_k|X) = \frac{\exp (w_{k0} + \sum_{i=1}^n w_{ki} X_i)}{1 + \sum_{j=1}^{K-1} \exp (w_{j0} + \sum_{i=1}^n w_{ji}X_i)}
\end{equation}
When $Y=y_K$, it is
\begin{equation}
P(Y=y_K|X) = \frac{1}{1 + \sum_{j=1}^{K-1} \exp (w_{j0} + \sum_{i=1}^n w_{ji}X_i)}
\end{equation}
Here $w_{ji}$ denotes the weight associated with the $j$th class $Y=y_j$ and with input $X_i$. It is easy to see that our earlier expressions for the case where $Y$ is boolean (equation (\ref{eq:logit1}) and (\ref{eq:logit2})) are a special case of the above expressions. Note also that the form of the expression for $P(Y=y_k|X)$ assures that $\left[ \sum\nolimits_{k=1}^K P(Y=y_k|X) \right[ = 1$. 

The primary difference between these expressions and those for boolean $Y$ is that when $Y$ takes on $K$ possible values, we construct $K-1$ different linear expressions to capture the distributions for for the different values of $Y$. The distribution for the final, $K$th, value of $Y$ is simply one minus the probabilities of the first $K-1$ values.

\subsection{Hybrid Approaches}

Hybrid approaches are another kind of method employed in natural language processing, which combine rule-based with machine learning methods. Hybrid approaches are considered as a reasonable solution for a number of problems for which the training data exhibit irregularities and exceptions.

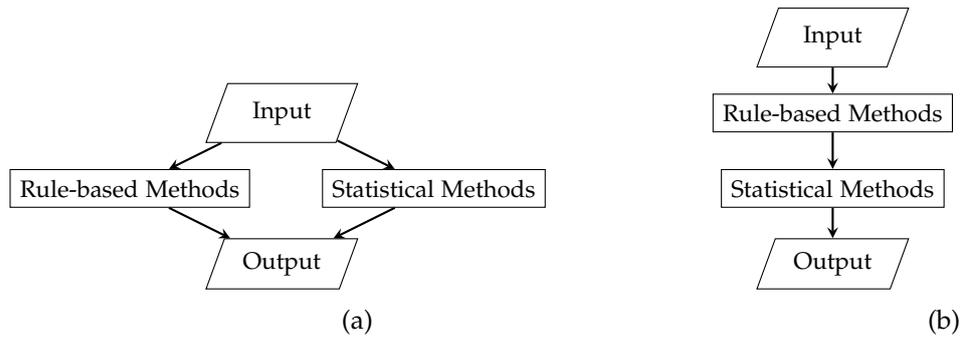
\begin{figure}[t]
\begin{subfigure}[b]{0.6\textwidth}
\begin{tikzpicture}
\tikzstyle{io} = [trapezium, trapezium left angle=70, trapezium right angle=110, minimum width=2cm, minimum height=0.5cm, text centered, draw=black]
\tikzstyle{process} = [rectangle, minimum width=2cm, minimum height=0.5cm, text centered, draw=black]
\tikzstyle{arrow} = [thick,->,>=stealth]

\node (in) [io] at (0,0) {\footnotesize{Input}};
\node (pro1) [process] at (-2,-1) {\footnotesize{Rule-based Methods}};
\node (pro2) [process] at (2,-1) {\footnotesize{Statistical Methods}};
\node (out) [io] at (0,-2) {\footnotesize{Output}};

\draw [arrow] (in) -- (pro1);
\draw [arrow] (in) -- (pro2);
\draw [arrow] (pro1) -- (out);
\draw [arrow] (pro2) -- (out);
\end{tikzpicture}
\caption{}
\end{subfigure}
\begin{subfigure}[b]{0.4\textwidth}
\begin{tikzpicture}
\tikzstyle{io} = [trapezium, trapezium left angle=70, trapezium right angle=110, minimum width=2cm, minimum height=0.5cm, text centered, draw=black]
\tikzstyle{process} = [rectangle, minimum width=2cm, minimum height=0.5cm, text centered, draw=black]
\tikzstyle{arrow} = [thick,->,>=stealth]

\node (in) [io] at (0,0) {\footnotesize{Input}};
\node (pro1) [process] at (0,-1) {\footnotesize{Rule-based Methods}};
\node (pro2) [process] at (0,-2) {\footnotesize{Statistical Methods}};
\node (out) [io] at (0,-3) {\footnotesize{Output}};

\draw [arrow] (in) -- (pro1);
\draw [arrow] (pro1) -- (pro2);
\draw [arrow] (pro2) -- (out);
\end{tikzpicture}
\caption{}
\end{subfigure}
\caption{Examples of hybrid architecture for information processing.}
\label{fig:hybrid}
\end{figure}

Figure \ref{fig:hybrid} exemplifies two hybrid architectures: (a) a concurrent information processing pipeline in which different tasks are performed by either rule-based or statistical methods, and (b) an information processing pipeline in which the output of the one family of methods is used as input for the other. Hybrid approaches are very popular in NLP applications such as machine translation, parsing, information extraction, etc. \textcite{schaefer2007} provides a good overview of integrating deep and shallow NLP components into hybrid architectures.

\subsection{Semi-supervised Machine Learning}

As the name suggests, \textit{semi-supervised learning}\footnote{We took the explanations about Semi-supervised Learning from \textcite{Zhu:2009:ISL:1717872}.} is somewhere between unsupervised and supervised learning. In fact, most semi-supervised learning strategies are based on extending either unsupervised or supervised learning to include additional information typical of the other learning paradigm. Specifically, semi-supervised learning encompasses several different settings, including:
\begin{itemize}
\item \textit{Semi-supervised classification}. Also known as classification with labelled and unlabelled data (or partially labelled data), this is an extension to the supervised classification problem. The training data consists of both $l$ labelled instances ${\lbrace(\mathbf{x_i}, y_i)\rbrace}_{i=1}^l$ and $u$ unlabelled instances ${\lbrace \mathbf{x_j}\rbrace}_{j=l+1}^{l+u}$. One typically assumes that there is much more unlabelled data than labelled data, i.e., $u\gg l$. The goal of semi-supervised classification is to train a classifier $f$ from both the labelled and unlabelled
data, such that it is better than the supervised classifier trained on the labelled data alone.
\item \textit{Constrained clustering}. This is an extension to unsupervised clustering. The training data consists of unlabelled instances ${\lbrace \mathbf{x_i} \rbrace}_{j=1}^n$, as well as some ``supervised information'' about the clusters. For example, such information can be so-called \textit{must-link} constraints, that two instances $\mathbf{x_i}, \mathbf{x_j}$ must be in the same cluster; and \textit{cannot-link} constraints, that $\mathbf{x_i}, \mathbf{x_j}$ cannot be in the same cluster. One can also constrain the size of the clusters. The goal of constrained clustering is to obtain better clustering than the clustering from unlabelled data alone.
\end{itemize}

Semi-supervised learning has tremendous practical value. In many tasks, there is a paucity of labelled data. The labels $y$ may be difficult to obtain because they require human annotators, special devices, or expensive and slow experiments. In this thesis, we will focus on a simple semi-supervised classification model: \textit{self-training}.

\paragraph{Self-training} Self-training is characterized by the fact that the learning process uses its own predictions to teach
itself. For this reason, it is also called self-teaching or bootstrapping (not to be confused with the statistical procedure with the same name). Self-training can be either inductive or transductive, depending on the nature of the predictor $f$. The algorithm for self-training is as follows:

\begin{algorithmic}[1]
\renewcommand{\algorithmicrequire}{\textbf{Input:}}
\REQUIRE labelled data ${\lbrace(\mathbf{x_i}, y_i)\rbrace}_{i=1}^l$, unlabelled instances ${\lbrace \mathbf{x_j}\rbrace}_{j=l+1}^{l+u}$.
\STATE Initially, let $L={\lbrace(\mathbf{x_i}, y_i)\rbrace}_{i=1}^l$ and $U={\lbrace \mathbf{x_j}\rbrace}_{j=l+1}^{l+u}$.
\REPEAT
\STATE Train $f$ from $L$ using supervised learning.
\STATE Apply $f$ to the unlabelled instances in $U$.
\STATE Remove a subset $S$ from $U$; add $\lbrace(\mathbf{x}, f(\mathbf{x}))|\mathbf{x} \in S \rbrace$ to $L$.
\UNTIL {U is empty.}
\end{algorithmic}

The main idea is to first train $f$ on labelled data. The function $f$ is then used to predict the labels for the unlabelled data. A subset $S$ of the unlabelled data, together with their predicted labels, are then selected to augment the labelled data. Typically, $S$ consists of the few unlabelled instances with the most confident $f$ predictions. The function $f$ is re-trained on the now larger set of labelled data, and the procedure repeats. It is also possible for $S$ to be the whole unlabelled data set. In this case, $L$ and $U$ remain the whole training sample, but the assigned labels on unlabelled instances might vary from iteration to iteration.

\begin{quote}
\textbf{Self-Training Assumption} \textit{The assumption of self-training is that its own predictions, at least the high confidence ones, tend to be correct.}
\end{quote}

The major advantages of self-training are its simplicity and the fact that it is a \textit{wrapper} method. This means that the choice of learner for $f$ in step 3 is left completely open. 
The self-training procedure ``wraps'' around the learner without changing its inner workings. This is important for many real world tasks related to natural language processing, where the learners can be complicated black boxes not amenable to changes.

On the other hand, it is conceivable that an early mistake made by $f$ (which is not perfect to start with, due to a small initial $L$) can reinforce itself by generating incorrectly labelled data. Re-training with this data will lead to an even worse $f$ in the next iteration. 

\subsection{Word Embeddings}
\label{sec:background-embeddings}

Image and audio processing systems typically work with rich, high-dimensional datasets encoded as vectors, e.g., the individual raw pixel-intensities for image data, or power spectral density coefficients for audio data. For tasks like object or speech recognition we know that all the information required to successfully perform the task is encoded in the data. However, natural language processing systems traditionally treat words as discrete atomic symbols, and therefore, provide no useful information to the system regarding the relationships that may exist between the individual symbols. This means that a model can leverage very little of what it has learned about \textit{cat} when it is processing data about \textit{dog}, for instance, that they are both animals, four-legged, pets, and so on. This kind of representations could lead to data sparsity, and usually means that we may need more data in order to successfully train statistical models. Vector representations of words can overcome these obstacles.

It has been shown that for words in the same language, the more often two words can be substituted into the same contexts the more similar in meaning they are judged to be \parencite{miller1991}. This phenomenon that words that occur in similar contexts tend to have similar meanings has been widely known as \textit{Distributional Hypothesis} \parencite{harris54}, which can be stated in the following way:
\begin{quote}
\textbf{Distributional Hypothesis} \textit{The degree of semantic similarity between two linguistic expressions $A$ and $B$ is a function of the similarity of the linguistic contexts in which $A$ and $B$ can appear.}
\end{quote}

This hypothesis is the core behind the application of vector-based models for semantic representation of words, which are variously known as \textit{word space} \parencite{sahlgren2006}, \textit{semantic spaces} \parencite{mitchell2010}, \textit{vector space models} (VSMs) \parencite{turney2010} or \textit{distributional semantic models} (DSMs) \parencite{baroni2010}. 

To have better illustration about distributional hypothesis, consider a foreign word such as \textit{wampimuk}, occurring in these two sentences: (1) \textit{He filled the wampimuk, passed it around and we all drunk some}, and (2) \textit{We found a little, hairy wampimuk sleeping behind the tree}. We could infer that the meaning of \textit{wampimuk} is either 'cup' or 'animal', heavily depends on its context which is either sentence (1) or (2) respectively.

The different approaches that leverage this principle can be divided into two categories \parencite{baroni-dinu-kruszewski:2014:P14-1}: (i) \textit{count-based models}, e.g. Latent Semantic Analysis (LSA) \parencite{pa:deerwester90indexing}, and (ii) \textit{predictive models}, e.g. neural probabilistic language models \parencite{journals/corr/abs-1301-3781}.

\paragraph{Count-based models} Count-based models compute the statistics of how often some word co-occurs with its neighbouring words in a large text corpus, and then map these count-statistics down to a small, dense vector for each word \parencite{sahlgren2006,pado2007,
bullinaria2007,agirre-EtAl:2009:NAACLHLT09,
baroni2010}.

One widely known algorithm falls under this category is \textit{GloVe}\footnote{We took the explanations about GloVe from \url{http://nlp.stanford.edu/projects/glove/}.} \parencite{pennington2014glove}. GloVe is essentially a log-bilinear model with a weighted least-squares objective. The main intuition underlying the model is the simple observation that ratios of word-word co-occurrence probabilities have the potential for encoding some form of meaning. For example, consider the co-occurrence probabilities for target words \textit{ice} and \textit{steam} with various probe words from the vocabulary. Here are some actual probabilities from a 6 billion word corpus:

\begin{table}[h!]
\small
\centering
\begin{tabular} {lcccc}
\hline
Probability and Ratio & $k=\textrm{\textit{solid}}$ & $k=\textrm{\textit{gas}}$ & $k=\textrm{\textit{water}}$ & $k=\textrm{\textit{fashion}}$ \\
\hline
$P(k|\textrm{\textit{ice}})$ & $1.9 \times 10^-4$ & $6.6 \times 10^-5$ & $3.0 \times 10^-3$ & $1.7 \times 10^-5$ \\
$P(k|\textrm{\textit{steam}})$ & $2.2 \times 10^-5$ & $7.8 \times 10^-4$ & $2.2 \times 10^-3$ & $1.8 \times 10^-5$ \\
$P(k|\textrm{\textit{ice}})/P(k|\textrm{\textit{steam}})$ & $8.9$ & $8.5 \times 10^-2$ & $1.36$ & $0.96$ \\
\hline
\end{tabular}
\end{table}

As one might expect, \textit{ice} co-occurs more frequently with \textit{solid} than it does with \textit{gas}, whereas \textit{steam} co-occurs more frequently with \textit{gas} than it does with \textit{solid}. Both words co-occur with their shared property \textit{water} frequently, and both co-occur with the unrelated word \textit{fashion} infrequently. Only in the ratio of probabilities does noise from non-discriminative words like \textit{water} and \textit{fashion} cancel out, so that large values (much greater than 1) correlate well with properties specific to \textit{ice}, and small values (much less than 1) correlate well with properties specific of \textit{steam}. In this way, the ratio of probabilities encodes some crude form of meaning associated with the abstract concept of thermodynamic phase. 

The training objective of GloVe is to learn word vectors such that their dot product equals the logarithm of the words' probability of co-occurrence. Owing to the fact that the logarithm of a ratio equals the difference of logarithms, this objective associates (the logarithm of) ratios of co-occurrence probabilities with vector differences in the word vector space. Because these ratios can encode some form of meaning, this information gets encoded as vector differences as well.

\paragraph{Predictive models} In predictive models, instead of first collecting context vectors and then re-weighting these vectors based on various criteria, the vector weights are directly set to optimally
predict the contexts in which the corresponding words tend to appear \parencite{Bengio:2003:NPL:944919.944966,Collobert:2008:UAN:1390156.1390177,Collobert:2011:NLP:1953048.2078186,huang-EtAl:2012:ACL20122,journals/corr/abs-1301-3781,turian-ratinov-bengio:2010:ACL}. 

\textit{Word2Vec}\footnote{We took the explanations about Word2Vec from \url{http://www.tensorflow.org/versions/r0.7/tutorials/word2vec/}.} \parencite{journals/corr/abs-1301-3781} is a particularly computationally-efficient predictive model for learning word embeddings from raw text. It comes in two flavours, the \textit{Continuous Bag-of-Words model (CBOW)} and the \textit{Skip-Gram model}. Algorithmically, these models are similar, except that CBOW predicts target words (`mat') from source context words (`the cat sits on the'), whereas the skip-gram does the inverse and predict source context words from the target words. This inversion might seem like an arbitrary choice, but statistically it has the effect that CBOW smoothes over a lot of the distributional information, by treating an entire context as one observation. For the most part, this turns out to be a useful feature for smaller datasets. On the other hand, skip-gram treats each context-target pair as a new observation, and this tends to do better when we have larger datasets.

Neural probabilistic language models are traditionally trained using the \textit{maximum likelihood (ML)} principle to maximize the probability of the target word $w_t$ given the previous words (history) $h$ in terms of a \textit{softmax} function:
\begin{equation}
\begin{split}
P(w_t|h)&=\textrm{softmax}(\textrm{score}(w_t, h))\\
&=\frac{\exp\lbrace\textrm{score}(w_t, h)\rbrace}{\sum_{\textrm{Word }w'\textrm{ in Vocab}} \exp\lbrace\textrm{score}(w', h)\rbrace}\\
\end{split}
\end{equation}
where $\textrm{score}(w_t,h)$ computes the compatibility of word $w_t$ with the context $h$, typically using a dot product. The model is trained by maximizing its log-likelihood on the training set, i.e. by optimizing:
\begin{equation}
\begin{split}
J_{ML}&=\log P(w_t|h)\\
&=\textrm{score}(w_t, h)-\log \left( \sum_{\textrm{Word }w'\textrm{ in Vocab}} \exp\lbrace\textrm{score}(w',h)\rbrace \right)\\
\end{split}
\end{equation}

This yields a properly normalized probabilistic model for language modelling. However, this is very expensive, because we need to compute and normalize each probability using the score for all other words $w'$ in the current context $h$, at every training step.

In Word2Vec, a full probabilistic model is not needed. Instead, the CBOW and skip-gram models are trained using a binary classification objective (logistic regression) to discriminate the real target words $w_t$ from $k$ imaginary (noise) words $\tilde{w}$, in the same context. Mathematically, the objective is to maximize:
\begin{equation}
J_{NEG}=\log Q_\theta(D=1|w_t, h) + k \underset{\tilde{w}\sim P_{\textrm{noise}}}{\mathbb{E}} \left[ \log Q_\theta(D=0|\tilde{w}, h) \right]
\end{equation}
where $Q_\theta (D=1|w, h)$ is the binary logistic regression probability under the model of seeing the word $w$ in the context $h$ in the dataset $D$, calculated in terms of the learned embedding vectors $\theta$. The expectation is approximated by drawing $k$ contrastive words from the noise distribution.

This objective is maximized when the model assigns high probabilities to the real words, and low probabilities to noise words. Technically, this is called \textit{Negative Sampling}, and there is a good mathematical motivation to use this loss function, i.e., the updates it proposes approximate the updates of the softmax function in the limit. But computationally it is especially appealing because computing the loss function now scales only with the number of noise words that are selected ($k$) instead of the size of the vocabulary.


\chapter{Temporal Information Processing}\label{ch:auto-event-extraction}
\minitoc
\begingroup\emergencystretch=.3em


Newspapers and narrative texts are often used to describe events that occur in a certain time and specify the temporal order of these events. To comprehend such texts, the capability to extract temporally relevant information is clearly required, which includes: (i) identifying events and (ii) temporally linking them to build event timelines. This capability is crucial to
a wide range of NLP applications, such as personalized news systems, question answering and document summarization. In this chapter we provide an introduction to temporal information processing that comprises annotation standards, annotated corpora and related evaluation campaigns. We also give a brief overview of the state-of-the-art methods for extracting temporal information from text.

\section{Modelling Temporal Information}
\label{sec:model-temp-info}

In NLP, the definition of an event can be varied depending on the target application. In topic detection and tracking \parencite{allan2002}, the term \textit{event} is used interchangeably with \textit{topic}, which describes something that happens and is usually used to identify a cluster of documents, e.g., \textit{Olympics}, \textit{wars}. On the other hand, information extraction provides finer granularity of event definitions, in which events are entities that happen/occur within the scope of a document.  

Events, especially within a narrative text, are naturally anchored to temporal attributes, which are often expressed with \textit{time expressions} such as `two days ago' or `Friday the $13^{th}$'. However, an event can also have non-temporal attributes such as \textit{event participants} and the \textit{location} where the event took place. Here is where event modelling plays its part in automatic event extraction, to define the structure of events one wants to extract from a text.

There are several annotation frameworks for events and time expressions that can be viewed as \textit{event models}, \textit{TimeML}~\parencite{pustejovsky2003} and \textit{ACE}~\parencite{ldc2005} being the prominent ones. There are other event models based on web ontology (RDFS+OWL) such as \textit{LODE}~\parencite{shaw2009}, \textit{SEM}~\parencite{vanhage2011} and \textit{DOLCE}~\parencite{gangemi2002}, which encode knowledge about events as triples. While event triples can be seen as ways to store the extracted knowledge to perform reasoning on, event annotations and the corresponding annotated corpora are geared towards automatic event extraction from texts in natural language.

\paragraph{TimeML} TimeML is a language specification for \textit{events} and \textit{time expressions}, which was developed in the context of the TERQAS workshop\footnote{\url{http://www.timeml.org/site/terqas/index.html}} supported by the AQUAINT program. The main purpose is to identify and extract events and their temporal anchoring from a text, such that it can be used to support a question answering system in answering temporally-based questions like ``In which year did Iraq finally pull out of Kuwait during the war in the 1990s?''. 

\paragraph{ACE} The ACE annotation framework was introduced by the ACE program, which provides annotated data, evaluation tools, and periodic evaluation exercises for a variety of information extraction tasks. It covers the annotation of five basic kinds of extraction targets including \textit{entities}, \textit{values}, \textit{time expressions}, \textit{relations} (between entities) and \textit{events} \parencite{ace2005}.

\paragraph{TimeML vs ACE} Both TimeML and ACE define an event as \textit{something that happens/occurs} or \textit{a state that holds true}, which can be expressed by a verb, a noun, an adjective, as well as a nominalization either from verbs or adjectives. However, both event models are designed for different purposes, hence, resulting in different annotation of events. In addition to basic features of events existing in both models (tense, aspect, polarity and modality), ACE events have more complex structures involving \textit{event arguments}, which can either be \textit{event participants} (entities participating in the corresponding events) or \textit{event attributes} (place and time of the corresponding events) ~\parencite{ldc2005}. 

While in TimeML all events are annotated, because every event takes part in the temporal network\footnote{Except for \textit{generics} as in ``\textbf{Use} of corporate jets for political \textbf{travel} is legal.'' \parencite{timeml2006}}, in ACE only 'interesting' events falling into a set of particular types and subtypes are annotated.  

In annotating temporal expressions, ACE and TimeML use similar temporal annotations. ACE uses TIMEX2~\parencite{ferro2001} model, which was developed under DARPA's Translingual Information Detection, Extraction and Summarization (TIDES) program, whereas TimeML introduces TIMEX3 annotation modelled on TIMEX~\parencite{setzer2001} as well as TIMEX2.

The most important attribute of TimeML that differs from ACE is the separation of the representation of events and time expressions from the \textit{anchoring} or \textit{ordering} dependencies. Instead of treating a time expression as an event argument, TimeML introduces \textit{temporal link} annotations to establish dependencies (temporal relations) between events and time expressions \parencite{pustejovsky2003}. This annotation is important in (i) anchoring an event to a time expression (event time-stamping) and (ii) determining the temporal order between events. This distinctive feature was the main reason why we chose TimeML as the event model for our research. 

\paragraph{Definitions} According to TimeML, we can formalize the definitions of temporal information as follows:
\begin{itemize}
	\item \textit{Events} are expressions in text denoting situations that \textit{happen} or \textit{occur}, or predicates describing \textit{states} or \textit{circumstances} in which something obtains or holds true. They can be punctual or last for a period of time.
	\item \textit{Time expressions}, \textit{temporal expressions}, or simply \textit{timexes} are expressions in text denoting time ``when'' something happens, how often it happens, or how long it lasts.
	\item \textit{Temporal relations} represent the temporal order holding between two arguments, i.e., event and event, event and timex, or timex and timex.
	\item \textit{Temporal signals} are specific types of word indicating or providing a cue of an explicit temporal relation between two arguments.
\end{itemize}

\section{TimeML Annotation Standard}
\label{sec:timeml-standard}

TimeML introduces 4 major data structures: \texttt{EVENT} for events, \texttt{TIMEX3} for time expressions, \texttt{SIGNAL} for temporal signals, and \texttt{LINK} for relations among \texttt{EVENT}s and \texttt{TIMEX3}s \parencite{pustejovsky2003,timeml2006}. There are three types of \texttt{LINK} tags: \texttt{TLINK}, \texttt{SLINK} and \texttt{ALINK}, which will be further explained in the following section. Note that TimeML \texttt{EVENT}s never participate in a link. Instead, their
corresponding event instance IDs, which are realized through the \texttt{MAKEINSTANCE} tag, are used.

\paragraph{}For the clarity purposes, henceforth, snippets of text annotated with \event{events}, \timex{timexes} and \signal{temporal signals} serving as examples will be in the respective forms. For example, ``John \event{drove} \signal{for} \timex{5 hours}.''

\subsection{TimeML Tags}
\label{sec:timeml-tags}

\paragraph{\uppercase{\texttt{\textbf{EVENT}}}} Events in a text can be expressed by tensed or untensed (phrasal) verbs (1), nominalizations (2), adjectives (3), predicative clauses (4), or prepositional phrases (5). 
\begin{enumerate}
	\item Foreign ministers of member-states \textit{has \event{agreed}} \textit{to \event{set} up} a seven-member panel to investigate who shot down Rwandan President Juvenal Habyarimana's plane.
	\item The financial \event{assistance} from the World Bank and the International Monetary Fund are not helping.
	\item Philippine volcano, \event{dormant} for six centuries, began exploding with searing gases, thick ash and deadly debris.
	\item Those observers looking for a battle between uncompromising representatives and very different ideologies \textit{will}, in all likelihood, \textit{be \event{disappointed}}.
	\item All 75 people \event{on board} the Aeroflot Airbus died.
\end{enumerate}

Note that some events may be sequentially discontinuous in some context as exhibited in (4). In order to simplify the annotation process, only the word considered as the \textit{syntactic head} is annotated, shown with bold letters in the examples, except for prepositional phrases. 

The attributes for the \texttt{EVENT} tag includes:
\begin{itemize}
	\item \texttt{eid} -- unique ID number.
	\item \texttt{class} -- class of the event: \texttt{REPORTING}, \texttt{PERCEPTION}, \texttt{ASPECTUAL}, \texttt{I\_ACTION}, \texttt{I\_STATE}, \texttt{STATE} or \texttt{OCCURRENCE}.
	\item \texttt{stem} -- stem of the event's head.
\end{itemize}

\paragraph{\uppercase{\texttt{\textbf{MAKEINSTANCE}}}} The \texttt{MAKEINSTANCE} tag is an auxiliary tag used to distinguish event tokens from event instances. The typical example of its usage is: to annotate the markable `taught' in ``He \event{taught} on \timex{Monday} and \timex{Tuesday}.'' as two event instances happened in different time. The attributes for this tag include:
\begin{itemize}
	\item \texttt{eiid} -- unique ID number.
	\item \texttt{eventID} -- unique ID to the referenced EVENT found in the text.
	\item \texttt{tense} -- tense of the event: \texttt{PAST}, \texttt{PRESENT}, \texttt{FUTURE}, \texttt{INFINITIVE}, \texttt{PRESPART}, \texttt{PASTPART} or \texttt{NONE}.
	\item \texttt{aspect} -- aspect of the event: \texttt{PROGRESSIVE}, \texttt{PERFECTIVE}, \texttt{PERFECTIVE\_PROGRESSIVE} or \texttt{NONE}.
	\item \texttt{pos} -- part-of-speech tag of the event: \texttt{ADJECTIVE}, \texttt{NOUN}, \texttt{VERB}, \texttt{PREPOSITION} or \texttt{OTHER}.
	\item \texttt{polarity} -- polarity of the event: \texttt{POS} or \texttt{NEG}.
	\item \texttt{modality} -- the modal word modifying the event (if exists).
\end{itemize}

\paragraph{\uppercase{\texttt{\textbf{TIMEX3}}}} The \texttt{TIMEX3} tag is used to mark up explicit temporal expressions, including dates, times, durations, and sets of dates and times. There are three major types of \texttt{TIMEX3} expressions: (i) fully specified
timexes, e.g., \textit{June 11 1989}, \textit{summer 2002}; (ii) underspecified timexes, e.g.
\textit{Monday}, \textit{next month}, \textit{two days ago}; (iii) durations, e.g., \textit{three months}. 

This tag allows specification of a temporal anchor, which facilitates the use of temporal functions to calculate the value of an underspecified timex. For example, within an article with a document creation time such as `January 3, 2006', the temporal expression `today' may occur. By anchoring the \texttt{TIMEX3} for `today' to the document creation time, we can determine the exact value of the \texttt{TIMEX3}.

The attributes of the \texttt{TIMEX3} tag, which are of particular interest in the scope of this work, include:\footnote{The full set of attributes with their descriptions can be found in \parencite{timeml2006}}
 
\begin{itemize}
	\item \texttt{tid} -- unique ID number.
	\item \texttt{type} -- type of timex: \texttt{DATE}, \texttt{TIME}, \texttt{DURATION} or \texttt{SET}.
	\item \texttt{value} -- normalized temporal value of the annotated timex represented in an extended ISO 8601 format.
	\item \texttt{functionInDocument} -- function of a \texttt{TIMEX3} in providing a temporal anchor for other temporal expressions in the document: \texttt{CREATION\_TIME}, \texttt{MODIFICATION\_TIME}, \texttt{PUBLICATION\_TIME}, \texttt{RELEASE\_TIME}, \texttt{RECEPTION\_TIME}, \texttt{EXPIRATION\_TIME} or \texttt{NONE}.
	\item \texttt{anchorTimeID} -- (optional) the timex ID of the timex to which the \texttt{TIMEX3} markable is temporally anchored.
\end{itemize}

\paragraph{\uppercase{\texttt{\textbf{SIGNAL}}}} The \texttt{SIGNAL} tag is used to mark up textual elements that make relations holding between two temporal elements explicit, which are generally:
\begin{itemize}
	\item Temporal prepositions: \textit{on}, \textit{in}, \textit{at}, \textit{from}, \textit{to}, \textit{during}, etc.
	\item Temporal conjunctions: \textit{before}, \textit{after}, \textit{while}, \textit{when}, etc.
	\item Prepositions signaling modality: \textit{to}.
	\item Special characters: `-' and `/', in temporal expressions denoting ranges, e.g., \textit{September 4-6} or \textit{Apr. 1999/Jul. 1999}.
\end{itemize}

The only attribute for the \texttt{SIGNAL} tag is \texttt{sid}, corresponding to the unique ID number.

\paragraph{\uppercase{\texttt{\textbf{TLINK}}}} The \texttt{TLINK}, Temporal Link, tag is used to (i) establish
a temporal order between two events (event-event pair), (ii) anchor an event to a time expression (event-timex pair), and (iii) establish a temporal order between two time expressions (timex-timex pair). Each temporal link has a temporal relation type assigned to it. The temporal relation types are modelled based on Allen's interval algebra between two intervals \parencite{allen1983}. Table~\ref{tab:allen} shows the TimeML temporal relation types corresponding to relation types existing in Allen's interval logic.

\begin{table}
\begin{tabular}{llll}
\hline
\textbf{Interval interpretation} & \textbf{Allen's Relation} & & \textbf{TimeML \texttt{TLINK} Type}\\[2pt] \hline
$\bullet \mbox{---} x \mbox{---} \bullet ~~~~ \circ \mbox{---} y \mbox{---} \circ$ & $x\ <\ y, y\ >\ x $ & $x$ \textit{before} $y$ & \texttt{BEFORE}, \texttt{AFTER}\\ \hline
$\bullet \mbox{---} x \mbox{---} \bullet$ & \multirow{2}{*}{$x\ m\ y, y\ m^{-1}\ x$} & \multirow{2}{*}{$x$ \textit{meets} $y$} & \multirow{2}{*}{\texttt{IBEFORE}, \texttt{IAFTER}}\\
$~~~~~~~~~~~~\circ \mbox{---} y \mbox{---} \circ$ \\ \hline
$\bullet \mbox{---} x \mbox{---} \bullet$  & \multirow{2}{*}{$x\ o\ y, y\ o^{-1}\ x$} & \multirow{2}{*}{$x$ \textit{overlaps with} $y$} &  \multirow{2}{*}{-}\\
$~~~~~~~~\circ \mbox{---} y \mbox{---} \circ$ \\ \hline
$\bullet \mbox{---} x \mbox{---} \bullet$  & \multirow{2}{*}{$x\ s\ y, y\ s^{-1}\ x$} & \multirow{2}{*}{$x$ \textit{starts} $y$} & \multirow{2}{*}{\texttt{BEGINS}, \texttt{BEGUN\_BY}} \\
$\circ \mbox{------} y \mbox{------} \circ$ \\ \hline
$~~~\bullet \mbox{---} x \mbox{---} \bullet$  & \multirow{2}{*}{$x\ d\ y, y\ d^{-1}\ x$} & \multirow{2}{*}{$x$ \textit{during} $y$} & \texttt{DURING}, \texttt{DURING\_INV} \\
$\circ \mbox{------} y \mbox{------} \circ$ &  & & (\texttt{IS\_INCLUDED}, \texttt{INCLUDES})\\ \hline
$~~~~~~~\bullet \mbox{---} x \mbox{---} \bullet$  & \multirow{2}{*}{$x\ f\ y, y\ f^{-1}\ x$} & \multirow{2}{*}{$x$ \textit{finishes} $y$} & \multirow{2}{*}{\texttt{ENDS}, \texttt{ENDED}\_BY}\\
$\circ \mbox{------} y \mbox{------} \circ$ \\[4pt] \hline
$\bullet \mbox{------} x \mbox{------} \bullet$  & \multirow{2}{*}{$x\ =\ y, y\ =\ x$} & \multirow{2}{*}{$x$ \textit{is equal to} $y$} & \multirow{2}{*}{\texttt{SIMULTANEOUS}} \\
$\circ \mbox{------} y \mbox{------} \circ$ \\[4pt] \hline
\end{tabular}
	\caption{Allen's atomic relations, their semantics when interpreted over the real line, and their corresponding TimeML \texttt{TLINK} type}
  	\label{tab:allen}
\end{table}

The Allen's \textit{overlap} relation is not represented in TimeML. However, TimeML introduces three more types of temporal relations (\texttt{IDENTITY}, \texttt{INCLUDES} and \texttt{IS\_INCLUDED}), resulting in a set of 14 relation types. \texttt{IDENTITY} relation is used to encode event co-reference as exhibited in the following sentence, ``John \event{drove} to Boston. During his \event{drive} he ate a donut.''

According to TimeML 1.2.1 annotation guidelines \parencite{timeml2006}, the difference between \texttt{DURING} and \texttt{IS\_INCLUDED} (also their inverses) is that \textit{\texttt{DURING}} relation is specified when an event persists throughout a temporal duration (1), while \texttt{IS\_INCLUDED} is specified when an event happens within a temporal expression (2). Moreover, \texttt{INCLUDES} and \texttt{IS\_INCLUDED} relations are used to specify a set/subset relationship between events (3).

\begin{enumerate}
	\item John \event{drove} for \timex{5 hours}.
	\item John \event{arrived} on \timex{Tuesday}.
	\item The police looked into the \event{slayings} of 14 women. In six of the \event{cases} suspects have already been arrested.
\end{enumerate}

The attributes of the \texttt{TLINK} tag include:
\begin{itemize}
	\item \texttt{lid} -- unique ID number.
	\item \texttt{eventInstanceID} or \texttt{timeID} -- unique ID of the annotated \texttt{MAKEINSTANCE} or \texttt{TIMEX3} involved in the temporal link.
	\item \texttt{relatedToEventInstance} or \texttt{relatedToTime} -- unique ID of the annotated \texttt{MAKEINSTANCE} or \texttt{TIMEX3} that is being related to.
	\item \texttt{relType} -- temporal relation holding between the elements: \texttt{BEFORE}, \texttt{AFTER}, \texttt{INCLUDES}, \texttt{IS\_INCLUDED}, \texttt{DURING}, \texttt{DURING\_INV}, \texttt{SIMULTANEOUS}, \texttt{IAFTER}, \texttt{IBEFORE}, \texttt{IDENTITY}, \texttt{BEGINS}, \texttt{ENDS}, \texttt{BEGUN\_BY} or \texttt{ENDED\_BY}.
	\item \texttt{signalID} -- (optional) the ID of \texttt{SIGNAL} explicitly signalling the temporal relation.
\end{itemize}

\paragraph{\uppercase{\texttt{\textbf{SLINK}}}} The \texttt{SLINK}, Subordination Link, tag is used to introduce a directional relation going from the main to the subordinated verb (indicated with \textsubscript{\textsc{s}}), which can be in one of the following contexts:
\begin{itemize}
	\item \textit{Modal}, e.g., ``Mary \event{wanted} John to \eventattr{buy}{s} some wine.''
	\item \textit{Factive}, e.g., ``John \event{managed} to \eventattr{go}{s} to the supermarket.''
	\item \textit{Counter-factive}, e.g., ``John \event{forgot} to \eventattr{buy}{s} some wine''.
	\item \textit{Evidential}, e.g., ``Mary \event{saw} John only \eventattr{carrying}{s} beer.''
	\item \textit{Negative-evidential}, e.g., ``John \event{denied} he \eventattr{bought}{s} only beer.''
	\item \textit{Conditional}, e.g., ``If John \eventattr{brings}{s} only beer, Mary will \event{buy} some wine.''
\end{itemize}

The attributes of the \texttt{SLINK} tag include:
\begin{itemize}
	\item \texttt{lid} -- unique ID number.
	\item \texttt{eventInstanceID} -- unique ID of the annotated \texttt{MAKEINSTANCE} involved in the subordination link.
	\item \texttt{subordinatedEventInstance} -- unique ID of the subordinated \texttt{MAKEINSTANCE} that is being related to.
	\item \texttt{relType} -- subordination relation holding between the event instances: \texttt{MODAL}, \texttt{EVIDENTIAL}, \texttt{NEG\_EVIDENTIAL}, \texttt{FACTIVE}, \texttt{COUNTER\_FACTIVE} or \texttt{CONDITIONAL}.
	\item \texttt{signalID} -- (optional) the ID of \texttt{SIGNAL} explicitly signalling the subordination relation.
\end{itemize}

\paragraph{\uppercase{\texttt{\textbf{ALINK}}}} The \texttt{ALINK}, Aspectual Link, tag represents the relationship between an aspectual event (indicated with \textsubscript{\textsc{a}}) and its argument event, belonging to one of the following:
\begin{itemize}
	\item \textit{Initiation}, e.g., ``John \eventattr{started}{a} to \event{read}.''
	\item \textit{Culmination}, e.g., ``John \eventattr{finished}{a} \event{assembling} the table.''
	\item \textit{Termination}, e.g., ``John \eventattr{stopped}{a} \event{talking}.''
	\item \textit{Continuation}, e.g., ``John \eventattr{kept}{a} \event{talking}.''
	\item \textit{Reinitiation}, e.g., ``John \eventattr{resumed}{a} \event{talking}.''
\end{itemize}

The attributes of the \texttt{ALINK} tag include:
\begin{itemize}
	\item \texttt{lid} -- unique ID number.
	\item \texttt{eventInstanceID} -- unique ID of the annotated (aspectual) \texttt{MAKEINSTANCE} involved in the aspectual link.
	\item \texttt{relatedToEventInstance} -- unique ID of the \texttt{MAKEINSTANCE} that is being related to.
	\item \texttt{relType} -- relation holding between the event instances: \texttt{INITIATES}, \texttt{CULMINATES}, \texttt{TERMINATES}, \texttt{CONTINUES} or \texttt{REINITIATES}.
	\item \texttt{signalID} -- (optional) the ID of \texttt{SIGNAL} explicitly signalling the relation.
\end{itemize}

\paragraph{Example} Figure~\ref{fig:example-timeml} shows an excerpt of news text annotated with temporal entities and temporal relations in TimeML annotation standard.

\begin{figure}
\noindent\fbox{%
\begin{minipage}{\textwidth}
\texttt{<TimeML>\\
<DOCID>wsj\_0679</DOCID>\\\\
<DCT><TIMEX3 tid="t0" type="DATE" value="1989-10-30" temporalFunction="false" functionInDocument="CREATION\_TIME">1989-10-30</TIMEX3></DCT>\\\\
<TEXT>\\
According to the filing, Hewlett-Packard <EVENT eid="e24" class="OCCURRENCE">acquired</EVENT> 730,070 common shares from Octel as a result of an <TIMEX3 tid="t25" type="DATE" value="1988-08-10" functionInDocument="NONE">Aug. 10, 1988</TIMEX3>, stock purchase <EVENT eid="e26" class="I\_ACTION">agreement</EVENT>.
 That <EVENT eid="e27" class="I\_ACTION">accord</EVENT> also <EVENT eid="e28" class="I\_ACTION">called</EVENT> for Hewlett-Packard to <EVENT eid="e29" class="OCCURRENCE">buy</EVENT> 730,070 Octel shares in the open market <SIGNAL sid=s30>within</SIGNAL> <TIMEX3 tid="t31" type="DURATION" value="P18M" functionInDocument="NONE">18 months</TIMEX3>.\\
</TEXT>\\\\
<MAKEINSTANCE eventID="e24" eiid="ei24" tense="PAST" aspect="NONE" polarity="POS" pos="VERB"/>\\
<MAKEINSTANCE eventID="e26" eiid="ei26" tense="NONE" aspect="NONE" polarity="POS" pos="NOUN"/>\\
<MAKEINSTANCE eventID="e27" eiid="ei27" tense="NONE" aspect="NONE" polarity="POS" pos="NOUN"/>\\
<MAKEINSTANCE eventID="e28" eiid="ei28" tense="PAST" aspect="NONE" polarity="POS" pos="VERB"/>\\
<MAKEINSTANCE eventID="e29" eiid="ei29" tense="INFINITIVE" aspect="NONE" polarity="POS" pos="VERB"/>\\\\
<TLINK lid="l21" relType="AFTER" timeID="t31" relatedToTime="t25"/>\\
<TLINK lid="l22" relType="DURING" eventInstanceID="ei29" relatedToTime="t31" signalID="s30"/>\\
<TLINK lid="l23" relType="AFTER" eventInstanceID="ei23" relatedToEventInstance="ei26"/>\\\\
</TimeML>}

\end{minipage}
}

\caption{Text excerpt annotated with temporal entities and temporal relations in TimeML standard.}
\label{fig:example-timeml}
\end{figure}

\subsection{ISO Related Standards}
\label{sec:iso-standard}

\paragraph{ISO 8601} ISO 8601 is an international standard providing an unambiguous method for representing dates and times, which is used by TimeML to represent the \texttt{value} attribute of the \texttt{TIMEX3} tag. The standard is based on the following principles:\footnote{Note that in the formulation, the emphasized letters denote place-holders for number or designated letter values.}
\begin{itemize}
	\item Date, time and duration unit values are organized from the most to the least significant: year, month (or week), day, hour, minute, second and fraction of second. 		
	\item \textit{Calendar dates} are represented in the form of \texttt{\textit{YYYY}-\textit{MM}-\textit{DD}}, where \texttt{\textit{YYYY}}, \texttt{\textit{MM}} and \texttt{\textit{DD}} are the place-holders for the year, month and day number values respectively, or \texttt{\textit{YYYY}-W\textit{ww}} with \texttt{\textit{ww}} for the week-of-year number value. 
	\item \textit{Times} are represented with respect to the 24-hour clock system and follow the format of \texttt{\textit{hh}:\textit{mm}:\textit{ss}.\textit{ff}}, with \texttt{\textit{hh}}, \texttt{\textit{mm}}, \texttt{\textit{ss}} and \texttt{\textit{ff}} for the hour, minute, second and fraction of second values respectively.
	\item The combination of a date and a time is represented in the format of: \\
		\texttt{\textit{YYYY}-\textit{MM}-\textit{DD}T\textit{hh}:\textit{mm}:\textit{ss}.\textit{ff}}.
	\item \textit{Durations} follow the format of \texttt{P\textit{n}\textit{X}}, where \texttt{\textit{n}} is the duration number value, and \texttt{\textit{X}} is the duration unit which can be one of the following units: \texttt{Y}, \texttt{M}, \texttt{W}, \texttt{D}, \texttt{H}, \texttt{M} and \texttt{S} for year, month, week, day, hour, minute and second respectively. 
	\item The combination of duration units follows the format of: \texttt{P\textit{n}Y\textit{n}M\textit{n}DT\textit{n}H\textit{n}M\textit{n}S} or \texttt{P\textit{n}W}.
\end{itemize}

An extended version of ISO 8601 was proposed in the TIDES annotation standard \parencite{ferro2001} to address the ambiguity and vagueness of natural language:
\begin{itemize}
	\item Parts of day, weekend, seasons, decades and centuries were introduced as new concepts. For example, \texttt{\textit{YYYY}-W\textit{ww}-WE} (where \texttt{WE} indicates weekend), \texttt{\textit{YYYY}-\textit{MM}-\textit{DD}T\textit{PoD}} (where \texttt{\textit{PoD}} takes one of the following values: \texttt{NI}, \texttt{MO}, \texttt{MI}, \texttt{AF} and \texttt{EV} for night, morning, midday, afternoon and evening respectively), etc.  
	\item Additional temporal values are used for temporal expressions such as `nowadays' to refer to either the past, present or future , i.e., \texttt{PAST\_REF}, \texttt{PRESENT\_REF} or \texttt{FUTURE\_REF} respectively. 
\end{itemize}

\paragraph{ISO-TimeML} Adopting the existing TimeML annotation standard, ISO-TimeML aims to define a mark-up language for annotating documents with information about time and events. Several changes to TimeML have been proposed to address capturing temporal semantics in text: (i) stand-off annotations rather than in-line annotations that do not modify the text being annotated, and (ii) the introduction of a new link for measuring out events (\texttt{MLINK}), which characterizes a temporal expression of \texttt{DURATION} type as a temporal measurement of an event. The resulting standard, ISO 24617-1:2012, SemAF-Time, specifies a formalized XML-based mark-up language facilitating the exchange of temporal information \parencite{PUSTEJOVSKY10.55}.

\section{TimeML Annotated Corpora}
\label{sec:timeml-corpora}

We list below several corpora annotated with either simplified or extended TimeML annotation standard, which are freely available for research purposes\footnote{To access the Clinical TempEval corpus, users must agree to handle the data appropriately, formalized in the requirement that users must sign a data use agreement with the Mayo Clinic.}. Most corpora are in English, but few other corpora are in other languages, such as Chinese, French, Italian, Korean and Spanish.

\paragraph{TimeBank 1.2} The TimeBank 1.2 corpus \parencite{timebank} is an annotated corpus of temporal semantics that was created following the TimeML 1.2.1 specification \parencite{timeml2006}. It contains 183 news articles, with just over 61,000 non-punctuation tokens,  coming from a variety of news report, specifically from the ACE program and PropBank. The ones taken from the ACE program are originally transcribed broadcast news from the following sources: ABC, CNN, PRI, VOA and news-wire from AP and NYT. Meanwhile, PropBank contains articles from the Wall Stree Journal. TimeBank 1.2 is freely distributed by the Linguistic Data Consortium.\footnote{\url{http://www.ldc.upenn.edu/Catalog/CatalogEntry.jsp?catalogId=LDC2006T08}}

\paragraph{AQUAINT TimeML Corpus} The AQUAINT corpus contains 73 news report document, and freely available for download.\footnote{\url{http://timeml.org/site/timebank/aquaint-timeml/aquaint_timeml_1.0.tar.gz}}. It is often referred to as the \textit{Opinion corpus}.

\paragraph{TempEval Related Corpora} The corpora released in the context of TempEval evaluation campaigns (see Section~\ref{sec:tempeval}), which serve as development and evaluation datasets, are mostly based on the TimeML annotation standard, with some exception in the early tasks for the purpose of simplification:
\begin{itemize}
	\item The corpus created for the first TempEval task~\parencite{verhagen-EtAl:2007:SemEval-2007} at SemEval-2007 employs a simplified version of TimeML. For example, there is no event instance annotation (realized with the \texttt{MAKEINSTANCE} tag), and the \texttt{TLINK} types include only three core relations (\texttt{BEFORE}, \texttt{AFTER} and \texttt{OVERLAP}), two less specific relations (\texttt{BEFORE-OR-OVERLAP} and \texttt{OVERLAP-OR-AFTER}) for ambiguous cases, and \texttt{VAGUE} for where no particular relation can be established.

	\item As the TempEval-2 task~\parencite{verhagen-EtAl:2010:SemEval} at SemEval-2010 attempted to address multilinguality, the corpus released within this task includes texts in Chinese, English, French, Italian, Korean and Spanish. The annotation contains the same set of \texttt{TLINK} types used in the previous TempEval.

	\item The TempEval-3 corpus created for the TempEval-3 task~\parencite{uzzaman-EtAl:2013:SemEval-2013} at SemEval-2013, however, is based on the latest TimeML annotation guideline version 1.2.1~\parencite{timeml2006}, with the complete set of 14 TLINK types. The corpus contains (i) the enhanced existing corpora, TimeBank 1.2 and AQUAINT, resulting in \textit{TBAQ-cleaned} as the development data for the task, (ii) the \textit{TempEval-3 silver} corpus\footnote{The TempEval-3 silver corpus is obtained by running automatic annotation systems, TIPSem and TIPSem-B~\parencite{llorens-saquete-navarro:2010:SemEval} and TRIOS~\parencite{uzzaman-allen:2010:SemEval}, on 600K word corpus collected from Gigaword.}, and (iii) the newly released \textit{TE3-Platinum} as the evaluation corpus.

	\item The creation of evaluation corpus for QA-TempEval~\parencite{llorens-EtAl:2015:SemEval} does not require manual annotation of all TimeML elements in the documents. The annotators created temporal-related questions from the documents, such as, ``Will Manchester United and Liverpool \event{play} each other after they \event{topped} their respective groups?'', provided the correct \textit{yes/no} answers, then annotated the corresponding entities and relations in the text following the TimeML annotation format. There are 294 questions in total, coming from 28 documents belonging to three different domains: \textit{news articles}, \textit{Wikipedia articles} (history, biographical) and \textit{informal blog posts} (narrative).
	
	\item The Clinical TempEval corpus~\parencite{bethard-EtAl:2015:SemEval} comprises 600 clinical notes and pathology reports from cancer patients at the Mayo clinic. The documents are annotated using an extended TimeML annotation framework, which includes new temporal expression types (e.g., \textsc{PrePostOp} for \timex{post-operative}), new \texttt{EVENT} attributes (e.g., \texttt{degree=LITTLE} for \event{slight nausea}) and new temporal relation type (\texttt{CONTAINS}).
	
	\item The evaluation dataset released for the ``TimeLine: Cross-Document Event Ordering'' task~\parencite{minard-EtAl:2015:SemEval} consists of 90 Wikinews articles within specific topics (e.g., \textit{Airbus}, \textit{General Motors}, \textit{Stock Market}) surrounding the target entities for which the event timelines are created. An event timeline is represented as ordered events, which are anchored to time with granularity ranging from \texttt{DAY} to \texttt{YEAR}. There are 37 event timelines for target entities of type \texttt{PERSON} (e.g., \textit{Steve Jobs}), \texttt{ORGANISATION} (e.g., Apple Inc.), \texttt{PRODUCT} (e.g., Airbus A380) and \texttt{FINANCIAL} (e.g., Nasdaq), with around 24 events and 18 event chains per timeline in average.
 
 \end{itemize}
 
\paragraph{Ita-TimeBank} The Ita-TimeBank corpus \parencite{caselli-EtAl:2011:LAW} is composed of two corpora (more than 150K tokens) that have been developed in parallel following the It-TimeML annotation scheme for Italian language. The two corpora are (i) the CELCT corpus containing news articles taken from the Italian Content Annotation Bank (I-CAB) \parencite{magnini:2006:lrec2006}, and (ii) the ILC corpus, which consists of 171 news articles collected from the Italian Syntactic-Semantic Treebank, the PAROLE corpus and the web.
 
 \paragraph{TimeBank-Dense} The TimeBank-Dense corpus~\parencite{chambers-etal:2014:TACL} is created to address the sparsity issue in the existing TimeML corpora. Using a specialized annotation tool, annotators are prompted to label all pairs of events and time expressions in the same sentence, all pairs of events and time expressions in the immediately following sentence, and all pairs of events and the document creation time. The \texttt{VAGUE} relation introduced at the first TempEval task \parencite{verhagen-EtAl:2007:SemEval-2007} is adopted to cope with ambiguous temporal relations, or to indicate pairs for which no clear temporal relation exists. The resulting corpus contains 12,715 temporal relations, under the labels \texttt{BEFORE}, \texttt{AFTER}, \texttt{INCLUDES}, \texttt{IS\_INCLUDED}, \texttt{SIMULTANEOUS} and \texttt{VAGUE}, over 36 documents taken from TimeBank.\footnote{This is significantly in contrast with the TimeBank corpus containing only 6,418 temporal relations over 183 documents.}

\section{TempEval Evaluation Campaigns}
\label{sec:tempeval}

TempEval is a series of evaluation campaigns, which are part of SemEval (Semantic Evaluation), an ongoing series of evaluations of computational semantic analysis systems. The ultimate goal of TempEval is the automatic identification of temporal expressions (timexes), events, and temporal relations within a text as specified in TimeML annotation \parencite{pustejovsky2003}. However, since addressing this aim in a first evaluation challenge was deemed too difficult, a staged approach was employed.

\paragraph{TempEval-1} The first TempEval \parencite{verhagen-EtAl:2007:SemEval-2007} focuses only on the categorization of temporal relations into simplified TimeML \texttt{TLINK} types, and only for English. There were three tasks proposed, each is a task of determining the \texttt{TLINK} type of:
\begin{itemize}
	\item pairs of event and timex within the same sentence, 
	\item pairs of event and DCT (Document Creation Time), and
	\item pairs of main events of adjacent sentences, where \textit{main event} is usually the syntactically dominant verb in a sentence.
\end{itemize}    

\paragraph{TempEval-2} TempEval-2 \parencite{verhagen-EtAl:2010:SemEval} extended the first TempEval, growing into a multilingual task, and adding three more tasks: 
\begin{itemize}
	\item determining the extent of time expressions (\texttt{TIMEX3} tagging) and the attribute values for \texttt{type} and \texttt{value}, 
	\item determining the extent of events (\texttt{EVENT} tagging) and the attribute values for \texttt{class}, \texttt{tense}, \texttt{aspect}, \texttt{polarity} and \texttt{modality}, 
	\item determining the \texttt{TLINK} type of pairs of events where one event syntactically dominate the other event.
\end{itemize}

\paragraph{TempEval-3} TempEval-3 \parencite{uzzaman-EtAl:2013:SemEval-2013} is different from its predecessor in several aspects:
\begin{itemize}
	\item \textit{Dataset} \hspace{7pt} In terms of size, the task provided 100K word gold standard data and 600K word silver standard data for training, compared to 50K word corpus used in TempEval-1 and TempEval-2. A new evaluation dataset was developed, \textit{TE3-Platinum}, based on manual annotations by experts over new text.
	\item \textit{End-to-end extraction task} \hspace{7pt} The temporal information extraction tasks are performed on raw text. Participants need to recognize \texttt{EVENT}s and \texttt{TIMEX3}s first, determine which ones to link, then label the links with \texttt{TLINK} types. In previous TempEvals, gold annotated \texttt{EVENT}s, \texttt{TIMEX3}s and \texttt{TLINK}s (without \texttt{type}) were given.
	\item \textit{\texttt{TLINK} types} \hspace{7pt} The full set of relation types according to TimeML is used, as opposed to the simplified one used in earlier TempEvals.
	\item \textit{Evaluation} \hspace{7pt} A single score, \textit{temporal awareness score}, was reported to rank the participating systems.
\end{itemize}
There were three main tasks proposed in TempEval-3 focusing on TimeML entities and relations:
\begin{itemize}
	\item \textit{Task A} \hspace{7pt} Determine the extent of timexes in a text as defined by the \texttt{TIMEX3} tag, and determine the value of their \texttt{type} and \texttt{value} attributes.
	\item \textit{Task B} \hspace{7pt} Determine the extent of events in a text as defined by the \texttt{EVENT} tag, and assign the value of the \texttt{class} attribute.
	\item \textit{Task ABC} \hspace{7pt} The end-to-end task that goes from raw text to TimeML annotation of \texttt{EVENT}s, \texttt{TIMEX3}s and \texttt{TLINK}s, which entails performing tasks A and B.
\end{itemize}

In addition to the main tasks, two extra temporal relation tasks were also included:
\begin{itemize}
	\item \textit{Task C} \hspace{7pt} Given gold annotated \texttt{EVENT}s and \texttt{TIMEX3}s, identify the pairs of entities having temporal link (\texttt{TLINK}) and classify the relation type.
	\item \textit{Task C relation type only} \hspace{7pt} Given gold annotated \texttt{EVENT}s, \texttt{TIMEX3}s and \texttt{TLINK}s (without \texttt{type}), classify the relation type.
\end{itemize}

\paragraph{TempEval Continuation} At the last SemEval-2015, there are several tasks related to temporal processing, taking it further into different directions: \textit{cross-document event ordering}~\parencite{minard-EtAl:2015:SemEval}, \textit{temporal-related question answering}~\parencite{llorens-EtAl:2015:SemEval} and \textit{clinical domain}~\parencite{bethard-EtAl:2015:SemEval}. 

We are particularly interested in the QA TempEval task~\parencite{llorens-EtAl:2015:SemEval}, which requires the participants to perform end-to-end TimeML annotation from the plain text in the same way as in TempEval-3 (Task ABC), but evaluates the systems in terms of correctly answered questions instead of using common information extraction performance measures. The task focuses on answering yes/no questions in the following format: \texttt{IS <entity1> <RELATION> <entity2> ?}, e.g., Is \event{event1} \texttt{BEFORE} \event{event2} ?. The systems are ranked based on the accuracy in answering the questions. 

\section{State-of-the-art Methods}
\label{sec:sota-methods}

The problem of temporal information processing can be decomposed into several sub-problems, as has been defined in TempEval-3. Hence, the best participating systems in TempEval-3 for each task, i.e., \textit{timex extraction} (Task A), \textit{event extraction} (Task B) and \textit{temporal relation extraction} (Task C), can be perceived as the state-of-the-art systems (see Table~\ref{tab:te3-state-of-the-art}).\footnote{Note that we only report the high-performing participating systems (and their best system runs) in TempEval-3.} Apart from TempEval-3, the efforts towards complete temporal information processing are still ongoing, so we also report systems claiming to be better than the best systems in TempEval-3. For some tasks, TIPSem~\parencite{llorens-saquete-navarro:2010:SemEval}, the best performing system in TempEval-2, also performed best in TempEval-3. However, it was used by the annotators to pre-label the evaluation corpus, so it was excluded from the ranking.

\begin{table}[h!]
\centering
\small
\begin{tabular}{lcccc|cc}
\hline
 & & & & \textbf{strict} & \textbf{value} & \textbf{class}\\
\textbf{System} & \textbf{F1} & \textbf{P} & \textbf{R} & \textbf{F1} & \textbf{F1} & \textbf{F1}\\
\hline
\multicolumn{5}{l|}{\textbf{Timex Extraction}} & & \\
HeidelTime-t~\parencite{strotgen2013} & 90.30 & 93.08 & 87.68 & 81.34 & \textbf{77.61} & \\
NavyTime-1,2~\parencite{chambers:2013:SemEval-2013} & \textbf{90.32} & 89.36 & \textbf{91.30} & 79.57 & 70.97 & \\
SUTime~\parencite{chang-manning:2012:SUTime} & \textbf{90.32} & 89.36 & \textbf{91.30} & 79.57 & 67.38 & \\
ClearTK-1,2~\parencite{bethard:2013:EMNLP} & 90.23 & \textbf{93.75} & 86.96 & \textbf{82.71} & 64.66 & \\
\hline
\multicolumn{5}{l|}{\textbf{Event Extraction}} & & \\
\textit{TIPSem~\parencite{llorens-saquete-navarro:2010:SemEval}} & \textit{82.89} & \textit{83.51} & \textit{82.28} & & & \textit{75.59}\\
ATT-1~\parencite{jung2013} & \textbf{81.05} & \textbf{81.44} & \textbf{80.67} & & & \textbf{71.88}\\
KUL & 79.32 & 80.69 & 77.99 & & & 70.17\\
\hline
\multicolumn{5}{l|}{\textbf{Temporal Relation Extraction}} & & \\
\textit{TIPSem~\parencite{llorens-saquete-navarro:2010:SemEval}} & \textit{44.25} & \textit{39.71} & \textit{49.94} & & & \\
ClearTK-2~\parencite{bethard:2013:EMNLP} & \textbf{36.26} & \textbf{37.32} & 35.25 & & & \\
UTTime-5~\parencite{laokulrat-EtAl:2013:SemEval-2013} & 34.90 & 35.94 & 33.92 & & & \\
NavyTime-1~\parencite{chambers:2013:SemEval-2013} & 31.06 & 35.48 & 27.62 & & & \\
UTTime-1~\parencite{laokulrat-EtAl:2013:SemEval-2013} & 24.65 & 15.18 & \textbf{65.64} & & & \\
\multicolumn{5}{l|}{\textbf{Temporal Relation Type Classification}} & & \\
UTTime-1,4~\parencite{laokulrat-EtAl:2013:SemEval-2013} & \textbf{56.45} & \textbf{55.58} & \textbf{57.35} & & & \\
\hline
\multicolumn{5}{l|}{\textbf{Temporal Information Processing}} & & \\
\textit{TIPSem~\parencite{llorens-saquete-navarro:2010:SemEval}} & \textit{42.39} & \textit{38.79} & \textit{46.74} & & & \\
ClearTK-2~\parencite{bethard:2013:EMNLP} & \textbf{30.98} & \textbf{34.08} & \textbf{28.40} & & & \\
\hline
\end{tabular}
\caption{\label{tab:te3-state-of-the-art}State-of-the-art temporal information processing systems according to TempEval-3}
\end{table}

\subsection{Timex Extraction}
\label{sec:timex-extraction}

In terms of recognizing the extent of timexes in a text, both rule-based and data-driven strategies are equally good. The rule-engineering systems HeidelTime, NavyTime and SUTime performed best at relaxed matching with 90.3\%, 90.32\% and 90.32\% F1-score respectively, while the statistical system ClearTK performed best at strict matching with 82.71\% F1-score. Strict match is when there is an exact match between the system entity and gold entity, e.g., \timex{sunday morning} vs \timex{sunday morning}, whereas relaxed match is when there is at least an overlap between the system entity and gold entity, e.g., \timex{sunday} vs \timex{sunday morning}.

The rule-engineering systems commonly rely on regular expression (regex) matching to find a temporal expressions in a text, whereas the data-driven approaches regard the problem as a BIO token-chunking task, building a classifier to decide whether a token is at the B(eginning) of, I(nside) of or O(utside) of a timex. 

In TempEval-3, the timex recognition task also includes \textit{determining the \texttt{type} of a timex} (\texttt{DATE}, \texttt{TIME}, \texttt{DURATION} of \texttt{SET}) and \textit{normalize its value}, e.g., \timex{the day before yesterday} would be normalized into \texttt{2015-12-30} (assuming that today is \texttt{2016-01-01}). The normalization task is currently (and perhaps inherently) done best by rule-engineered systems, HeidelTime being the best with 	77.61\% F1-score.\footnote{The F1-score for \textsc{value} captures the performance of extracting timex and identifying the attribute \texttt{value} together, i.e., $value_{f1} = timex_{f1} * value_{accuracy}$} ClearTK included another classifier to determine timex types, but used TIMEN~\parencite{LLORENS12.128.L12-1015}, which is rule-based, to normalize timex values.

Most rule-based approaches for timex normalization use string-to-string translation approach, i.e., each word in the expression is looked up in a normalization lexicon, then the resulting sequence is mapped directly to the normalized form. Both HeidelTime and TIMEN follow this approach. A drawback of this approach is that there are different rules for each expression, e.g., \timex{yesterday}, \timex{the day before yesterday}, regardless of the compositional nature that may hold, that \timex{the day before yesterday} is \signal{one day before} \timex{yesterday}. 

TimeNorm~\parencite{bethard:2013:EMNLP} exploits a synchronous context free grammar for timex normalization to address these shortcomings. Synchronous rules map the source language
to formally defined operators for manipulating
times. Time expressions are then parsed using
an extended CYK+ algorithm, and converted
to a normalized form by applying the operators
recursively. UWTime~\parencite{lee-EtAl:2014:P14-1} uses a Combinatory Categorial Grammar (CCG) to construct compositional meaning representations, while also considering contextual cues (e.g. the document creation time, the governing verb's tense) to compute the normalized value of a timex.

Evaluated on the TempEval-3 evaluation corpus, UWTime achieved 82.4\% F1-score on the \texttt{value} resolution task, while TimeNorm achieved 81.6\% accuracy given gold annotated timex extents, compared with 78.5\% and 74.1\% accuracies achieved by HeidelTime and TIMEN, respectively.

\subsection{Event Extraction}
\label{sec:event-extraction}

All high performing systems for event recognition in TempEval-3 used machine learning approaches. Typically a system consists of different classifiers each for recognizing events and determining their \texttt{class} attribute. For event extent recognition, since in TimeML the annotated events are usually single-word events, the problem is often regarded as a binary token-classification.\footnote{Some systems in TempEval3 also modelled the problem as a BIO token-chunking task, e.g., ClearTK, as in for timex recognition.} Meanwhile, since there are 7 event classes in TimeML, the task of determining the \texttt{class} attribute is modelled as a multi-class classification task.

The best performing system is ATT with 81.05\% and 71.88\%, followed by KUL with 79.32\% and 70.17\%.\footnote{The F1-score for \textsc{class} captures the performance of extracting event and identifying the attribute \texttt{class} together, i.e., $class_{f1} = event{f1} * class_{accuracy}$} These systems, also TIPSem, use semantic information obtained through semantic role labelling as features, which proves to play an important role in event recognition.

\subsection{Temporal Relation Extraction}
\label{sec:temp-rel-extraction}

In TempEval-3, identifying which pair of entities are connected by a temporal relation is a new task in the series of TempEval challenges; in TempEval and TempEval-2, the pair of entities are given and limited to specific syntactic constructs. TempEval-3 participants approached the problem with rule-based, data-driven and also hybrid methods. The rules are typically based on the possible TLINK candidates enumerated in the task description: (i) main events of consecutive sentences, (ii) pairs of events in the same sentence, (iii) event and timex in the same sentence and (iv) event and document creation time. 

For (ii) candidate pairs, TIPSem only considered pairs of events where one is subordinated by the other. ClearTK included three different multi-class classification models (for (ii), (iii) and (iv) candidate pairs) for temporal relation identification, as well as temporal relation type classification. Given a pair of entities, the classifiers have to predict the temporal relation type (\texttt{BEFORE}, \texttt{AFTER}, \texttt{SIMULTANEOUS}, etc.) or \texttt{NORELATION} if there is no relation exists. UTTime-1 only relied on rules to consider candidate pairs as having temporal relations. However, UTTime-5 used re-trained classifiers with an additional relation type \texttt{UNKNOWN} to filter the candidate pairs, in the same way as ClearTK. 

The hybrid method, employed by NavyTime, combines candidate-pair-rules with four binary classifiers (for (i), (ii), (iii) and (iv) candidate pairs) that decide whether a candidate pair is having temporal relation or not.

On the other hand, for classifying the temporal relation types, all participants resort to data-driven approaches. Both TIPSem and UTTime used sentence-level semantic information as features, obtained via semantic role labelling and deep syntactic parsing, respectively.

Regarding the classifiers used, ClearTK relied on Mallet\footnote{\url{http://mallet.cs.umass.edu/}} MaxEnt, OpenNLP\footnote{\url{http://opennlp.apache.org/}} MaxEnt, and LIBLINEAR~\parencite{REF08a}, and picked the final classifiers by running a grid search over models and parameters on the training data. UTTime  used two LIBLINEAR~\parencite{REF08a} classifiers (L2-regularized logistic regression); one for \textit{event-event} pairs, i.e., (i) and (ii) candidate pairs, and another one for \textit{event-timex} pairs, i.e., (iii) and (iv) candidate pairs. In addition to four binary classifiers for identifying candidate pairs having temporal links, NavyTime trained four MaxEnt classifiers for temporal relation classification. 

For both temporal relation identification and temporal relation type classification tasks, ClearTK is the best performing system with 36.26\% F1-score. The organizers also provided the gold annotated temporal links to measure the performance of systems in classifying the temporal relation types (Task C relation type only). UTTime with semantic features performed best with 56.45\% F1-score. Using only rules to determine the candidate pairs, UTTime-1 achieved the highest recall (65.64\%) at the expense of precision (15.18\%). UTTime-5 can obtain a better F1-score by reducing the recall significantly, but still in the second place after ClearTK with 34.90\% F1-score.

\subsection{Temporal Information Processing}
\label{sec:temp-info-proc}

For complete temporal annotation from raw text, which is the Task ABC in TempEval-3, the best performing system is ClearTK, with 30.98\% F1-score.

\section{Conclusions}

We presented an introduction to temporal information processing, particularly in using TimeML as the annotation framework. The separation of temporal entities and temporal anchoring/dependency representations in TimeML, also the fact that events are not limited to specific types, were the main reason why we chose TimeML over ACE for temporal information modelling in our research. The TimeML annotation standard has been described (Section~\ref{sec:timeml-standard}), along with several corpora annotated with TimeML (Section~\ref{sec:timeml-corpora}).

We have also given an overview of state-of-the-arts methods for extracting temporal information from text (Section~\ref{sec:sota-methods}), according to TempEval evaluation campaigns (Section~\ref{sec:tempeval}). 
TempEval-3 results reported by~\textcite{uzzaman-EtAl:2013:SemEval-2013} show that even though the performances of systems for extracting TimeML entities are quite good (>80\% F1-score), the overall performance of end-to-end temporal information extraction systems suffers due to the low performance on extracting temporal relations. The state-of-the-art performance on the temporal relation extraction task yields only around 36\% F1-score. This is the main reason underlying our choice to focus this work on the extraction of temporal relations.

Identifying temporal relations in a full discourse is a task that is difficult to define. In general it involves the classification of temporal relations between every possible pair of events and timexes. Hence, without a completely labelled graph of events and timexes, we cannot speak about true extraction, but rather about matching human labelling decisions that were constrained by time and effort. The TimeBank-Dense corpus mentioned in Section~\ref{sec:timeml-corpora} is created to cope with such problem. 

Several tasks in line with TempEval (Section~\ref{sec:tempeval}) approach the problem by changing the evaluation scheme used. In QA TempEval, the task is no longer about annotation accuracy, but rather the accuracy for targeted questions. The ``TimeLine: Cross-Document Event Ordering'' task limited the extraction of event timelines only to events related to specific target entities.

\endgroup

\chapter{Temporal Relation Extraction}\label{ch:temp-rel-type}
\begin{flushright}
\scriptsize
\textit{Our thoughts have an order, not of themselves, but because the mind generates the spatio-temporal relationships involved in every experience.} --- Robert Lanza
\end{flushright}
\minitoc

In the previous chapter we have shown that the low performance of state-of-the-art extraction systems for temporal information processing (30.98\% F1-score) is mainly due to the low performance of temporal relation extraction (36.26\% F1-score). Hence our decision to focus on this particular task. In this chapter, we describe our efforts in building an improved temporal relation extraction system. The system is evaluated with the TempEval-3 evaluation scheme, to be comparable with the reported state-of-the-art systems. 

Moreover, as has been discussed in the previous chapter, given the sparse annotation in the TempEval-3 corpus, it is not trivial to determine whether an unannotated temporal link is actually missed by the annotator or is simply having no relation. TimeBank-Dense (Section~\ref{sec:timeml-corpora}) was created to address the problem by building a complete labelled graph of temporal links, shifting the evaluation scheme from identifying (and classifying) \textit{some} relations to \textit{all} relations. On the other hand, QA TempEval (Section~\ref{sec:tempeval}) shifted the evaluation methodology towards a more extrinsic goal of question answering, evaluating systems based on how well the extracted temporal relations can be used to answer questions about the text. In this chapter, we also report our system's performances evaluated with both evaluation methodologies.

\section{Introduction}
\label{sec:temp-rel-type-intro}

Temporal relations, or temporal links, are annotations that bring together pieces of markable temporal information in a text, and make formal representation of temporally ordered events possible. In TimeML, temporal relation types have been modelled based on  Allen's interval algebra between two intervals \parencite{allen1983}. In Table~\ref{tab:allen} in Section~\ref{sec:timeml-tags} we show the relation types defined in Allen's interval logic, along with the corresponding \texttt{TLINK} types in TimeML.

TimeML \parencite{pustejovsky2003} is the annotation framework used in the TempEval series, evaluation exercises focused on temporal information processing, i.e. the extraction of temporal expressions (timexes), events and temporal relations in a text (see Section~\ref{sec:timeml-standard} and Section~\ref{sec:tempeval}). According to TempEval-3 results reported by \textcite{uzzaman-EtAl:2013:SemEval-2013}, while systems for timex extraction and event extraction tasks yield quite high performances with over 80\% F1-scores, the best performing system achieved very low performance on the temporal relation extraction task, bringing down the overall performance on the end-to-end temporal information processing task to only around 30\% F1-score. This is the main reason why we focus our attention on the automatic extraction of temporal relations.

Identifying temporal relations in a full discourse is a very difficult task. In general it involves the classification of temporal relations between every possible pair of events and timexes. With $n$ markable elements in a text, the total number of possible temporal links is $n^2-n$. Most of the research done so far focused on estimating the relation type, given an annotated pair of temporal events and timexes. In TempEval-1 and TempEval-2, participants were given gold annotated temporal links, which are missing the \texttt{type} annotation, between temporal entities following predefined syntactic constructs, e.g. pairs of main events of adjacent sentences. 

In TempEval-3, participants were required to identify pairs of temporal entities connected by a temporal link (\texttt{TLINK}), but possible \texttt{TLINK} candidates were only: (i) main events of consecutive sentences, (ii) pairs of events in the same sentence, (iii) event and timex in the same sentence and (iv) event and document creation time. Moreover, compared to earlier TempEval campaigns, TempEval-3 required the recognition of the full set of temporal relations in TimeML (14 \texttt{TLINK} types) instead of a simplified set, increasing the task complexity. 

In this chapter, we describe our methods in building an improved temporal relation extraction system (Section~\ref{sec:temp-rel-method}), then evaluate our system following the TempEval-3 evaluation scheme to be able to compare it with the state-of-the-art systems (Section~\ref{sec:temp-rel-tempeval3}). 

However, the sparse annotation of temporal relations in the TempEval corpora makes it difficult to build an automatic extraction system and evaluate the system regarding its performance, particularly on identifying temporal links. The best system in TempEval-3 for labelling the temporal links with 14 temporal relation types (Task C classifying only), UTTime~\parencite{laokulrat-EtAl:2013:SemEval-2013}, achieved around 56\% F1-score. When the system is evaluated on the temporal relation extraction task (Task C: identifying + classifying), its performance dropped to 24.65\% F1-score, even though it gained a very high recall of 65.64\%. The best performing system for Task C in TempEval-3, ClearTK~\parencite{bethard:2013:EMNLP}, optimized
only relation classification and intentionally
left many pairs unlabelled, balancing the precision and recall into 36.26\% F1-score.

The TimeBank-Dense corpus (Section~\ref{sec:timeml-corpora}) is created to cope with this sparsity issue. Using a specialized annotation tool, annotators are prompted to label all possible pairs of temporal entities, resulting in a complete graph of temporal relations. On the other hand, one of the continuation of the TempEval series, QA TempEval (Section~\ref{sec:tempeval}), approached the problem by changing the evaluation scheme used. The task is no longer about annotation accuracy, but rather the accuracy for answering targeted questions.

Therefore, we also evaluate our system following the TimeBank-Dense and QA TempEval evaluation methodologies (Section~\ref{sec:temp-rel-timebank-dense} and Section~\ref{sec:temp-rel-qa-tempeval}, respectively), to give a complete overview on how well our system can extract temporal relations between temporal entities in a text. 


\section{Related Work}
\label{sec:temp-rel-type-related-work}

Supervised learning for temporal relation extraction has already been explored in several earlier works. Most existing models formulate temporal ordering as a \textit{pairwise classification} task, where each pair of temporal entities is classified into temporal relation types~\parencite{Mani_2007.three, chambers-wang-jurafsky:2007:PosterDemo}.

Several works have tried to exploit an external \textit{temporal reasoning} module to improve the supervised learning models for temporal relation extraction, through training data expansion~\parencite{Mani_2007.three, tatu-srikanth:2008:PAPERS}, or testing data validation, i.e., replacing the inconsistencies in automatically identified relations (if any) with the next best relation types~\parencite{tatu-srikanth:2008:PAPERS}. Some other works tried to take advantage of global information to ensure that the pairwise classifications satisfy \textit{temporal logic transitivity constraints}, using frameworks like Integer Linear Programming and Markov Logic Networks~\parencite{chambers-jurafsky:2008:EMNLP, yoshikawa-EtAl:2009:ACLIJCNLP, uzzaman-allen:2010:SemEval}. The gains have been small, likely because of the disconnectedness that is common in sparsely
annotated corpora~\parencite{chambers-jurafsky:2008:EMNLP}.

In the context of TempEval evaluation campaigns (Section~\ref{sec:tempeval}), which is a series of evaluations of temporal information processing systems, our research on temporal relation extraction is based on the third instalment of the series, TempEval-3. For the tasks related to temporal relation extraction (Task C and Task C relation type only), there were five participants in total, including ClearTK~\parencite{bethard:2013:EMNLP}, UTTime~\parencite{laokulrat-EtAl:2013:SemEval-2013} and NavyTime~\parencite{chambers:2013:SemEval-2013}, which are reported in Section~\ref{sec:temp-rel-extraction}. These systems resorted to data-driven approaches for classifying the temporal relation types, using morphosyntactic information (e.g., PoS tags, syntactic parsing information) and lexical semantic information (e.g., WordNet synsets) as features. UTTime additionally used sentence-level semantic information (i.e., predicate-argument structure) as features.

Our proposed approach for temporal relation type classification is inspired by recent works on hybrid classification models~\parencite{dsouza-ng:2013:NAACL-HLT, chambers-etal:2014:TACL}. \textcite{dsouza-ng:2013:NAACL-HLT} introduce 437 hand-coded rules along with supervised classification models using lexical relation features (extracted from Merriam-Webster dictionary and WordNet), as well as semantic and discourse features. CAEVO, a CAscading EVent
Ordering architecture by \textcite{chambers-etal:2014:TACL}, combines rule-based and data-driven classifiers in a  sieve-based architecture for temporal ordering. The classifiers are ordered by their individual precision. After each classifier-sieve proposes its labels, the architecture infers transitive links from the new labels, adds them to the temporal label graph and informs the next classifier-sieve about this decision.

\section{Related Publications}
\label{sec:temp-rel-type-relation-pub}

In \textcite{mirza-tonelli:2014:EACL}, we argue that using a simple set of features, avoiding complex pre-processing steps (e.g., discourse parsing, deep syntactic parsing, semantic role labelling), combined with carefully selected contributing features, could result in a better performance compared with the work of \textcite{dsouza-ng:2013:NAACL-HLT} and the best system in TempEval-3 (Task C relation type only), UTTime~\parencite{laokulrat-EtAl:2013:SemEval-2013}.

For QA TempEval task, we submitted our temporal information processing system, HLT-FBK~\parencite{mirza-minard:2015:SemEval}, which ranked 1st in all three domains: News, Wikipedia and Blogs (informal narrative text).

Both works serve as the basis of our proposed temporal relation extraction system that will be described in the following sections. 

\section{Formal Task Definition}
\label{sec:temp-rel-type-formal-task}

For temporal relation extraction, we perform two tasks: \textit{identification} and \textit{classification}. We first identify pairs of temporal entities having temporal relations, then classify the temporal relation types of these pairs.

\paragraph{Temporal Relation Identification} Given a text annotated with a \textit{document creation time (DCT)} and temporal entities, which can be an \textit{event} or \textit{timex}, identify which entity pairs are considered as having temporal relations.

\paragraph{Temporal Relation Type Classification} Given an ordered pair of entities $(e_1, e_2)$, which can be a \textit{timex-timex} (T-T), \textit{event-DCT} (E-D), \textit{event-timex} (E-T) or \textit{event-event} (E-E) pair, assign a certain label to the pair, which can be one of the 14 \texttt{TLINK} types: \texttt{BEFORE}, \texttt{AFTER}, \texttt{INCLUDES}, \texttt{IS\_INCLUDED}, \texttt{DURING}, \texttt{DURING\_INV}, \texttt{SIMULTANEOUS}, \texttt{IAFTER}, \texttt{IBEFORE}, \texttt{IDENTITY}, \texttt{BEGINS}, \texttt{ENDS}, \texttt{BEGUN\_BY} or \texttt{ENDED\_BY}.

\paragraph{Example} Consider the following excerpt taken from the TimeBank corpus, annotated with events and temporal expressions:

\begin{quote}
DCT=\timexattr{1989-10-30}{t$_0$}

According to the filing, Hewlett - Packard \eventattr{acquired}{E$_{24}$} 730,070 common shares from Octel as a result of an \timexattr{Aug. 10, 1988}{T$_{25}$}, stock purchase \eventattr{agreement}{E$_{26}$}. That \eventattr{accord}{E$_{27}$} also \eventattr{called}{E$_{28}$} for Hewlett - Packard to \eventattr{buy}{E$_{29}$} 730,070 Octel shares in the open market within \timexattr{18 months}{T$_{30}$}.
\end{quote}

The temporal relation extraction system should be able to identify, among others:
\begin{itemize}[noitemsep]
\item timex-timex: [T$_0$ \texttt{AFTER} T$_{25}$], [T$_{30}$ \texttt{AFTER} T$_{25}$]
\item event-DCT: [E$_{24}$ \texttt{BEFORE} T$_{0}$, [E$_{28}$ \texttt{BEFORE} T$_{0}$]
\item event-timex: [E$_{26}$ \texttt{IS\_INCLUDED} T$_{25}$], [E$_{29}$ \texttt{DURING} T$_{30}$]
\item event-event: [E$_{24}$ \texttt{AFTER} T$_{26}$], [E$_{27}$ \texttt{INCLUDES} T$_{28}$]
\end{itemize}

\section{Method}
\label{sec:temp-rel-method}

\begin{figure}
\centering
\includegraphics[scale=0.9]{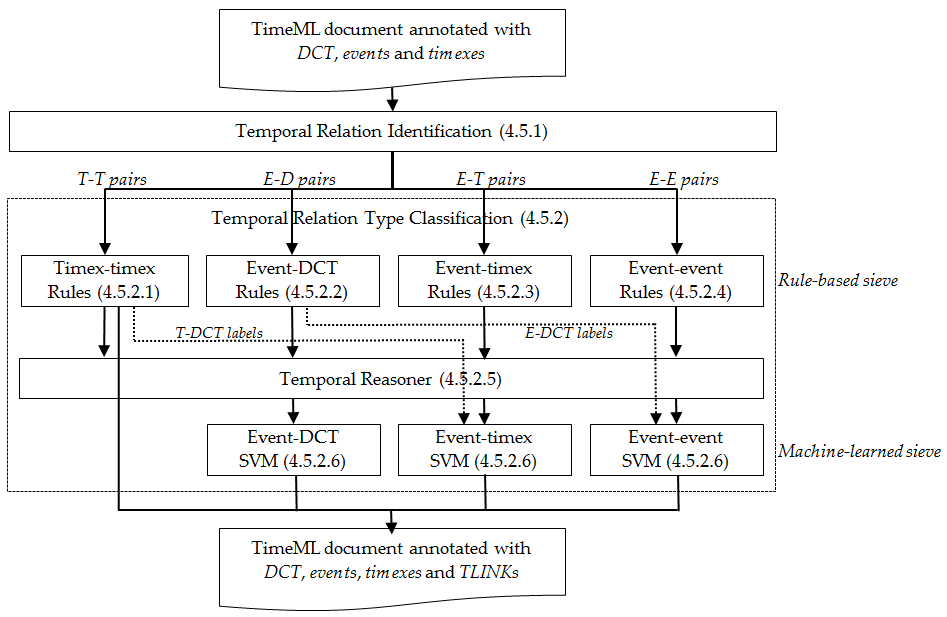}
\caption{Our proposed temporal relation extraction system, TempRelPro}
\label{fig:temporal-relation-extraction-system}
\end{figure}

We propose a hybrid approach for temporal relation extraction, as illustrated in Figure~\ref{fig:temporal-relation-extraction-system}. Our system, TempRelPro, is composed of two main modules: (i) \textit{temporal relation identification}, which is based on a simple set of rules, and (ii) \textit{temporal relation type classification}, which is a combination of rule-based and supervised classification modules, and a temporal reasoner component in between.

\subsection{Temporal Relation Identification}
\label{sec:temp-rel-method-identification}
All possible pairs having temporal relations according to the TempEval-3 task description are extracted using a set of simple rules; pairs of temporal entities satisfying one of the following rules are considered as having temporal links (\texttt{TLINK}s):
\begin{itemize}
\item pairs of main events of consecutive sentences
\item pairs of events in the same sentence
\item pairs of event and timex in the same sentence
\item pairs of event and document creation time
\item pairs of all possible timexes (including document creation time) linked with each other\footnote{Note that this is not included in the enumerated possible \texttt{TLINK}s in the TempEval-3 task description.}
\end{itemize}

These pairs are then grouped together into four different groups: \textit{timex-timex} (T-T), \textit{event-DCT} (E-D), \textit{event-timex} (E-T) and \textit{event-event} (E-E).

\subsection{Temporal Relation Type Classification}
\label{sec:temp-rel-method-classification}

Our approach for temporal relation type classification is inspired by CAEVO \parencite{chambers-etal:2014:TACL}, which combines rule-based and supervised classifiers in a sieve-based architecture. One of the benefits of this architecture is the seamless
enforcement of transitivity constraints, by inferring all transitive relations from each classifier-sieve’s output before the graph is passed on to the next one. The classifiers are ordered based on their precision. Hence, the most precise ones based on linguistic motivated rule-based approaches are executed first, followed by machine learned ones. 

We also follow the idea of a sieve-based architecture. However, our proposed system is different than CAEVO regarding the following:
\begin{itemize}
\item We consider all rule-based classifiers as one sieve component (rule-based sieve), and all Support Vector Machine (SVM) classifiers as another one (machine-learned sieve).
\item Instead of running transitive inference after each classifier, we run our \textit{temporal reasoner} module (Section~\ref{sec:temporal-reasoner}) on the output of the rule-based sieve, only once.
\item We use the output of the rule-based sieve as features for the machine-learned sieve, specifically:
\begin{itemize}
\item The timex-DCT link label proposed by the \textit{timex-timex rules} (Section~\ref{sec:timex-timex-rules}) are used as a feature in the \textit{event-timex SVM} (Section~\ref{sec:pair-classifier})
\item The event-DCT link label proposed by the \textit{event-DCT rules} (Section~\ref{sec:timex-timex-rules}) are used as a feature in the \textit{event-event SVM} (Section~\ref{sec:pair-classifier})
\end{itemize}
\item In Table~\ref{tab:caevo-vs-temprelpro} we report the comparison between CAEVO's sieves and ours. Several sieves are not implemented in our system. Some others are different in terms of generality. Note that the last sieve of CAEVO, which labels all unlabelled pairs with \texttt{VAGUE}, is not implemented in our system, but implicitly embedded in our machine-learned models.
\end{itemize}

\begin{table}[t]
\centering
\begin{adjustbox}{width=1\textwidth}
\begin{tabular}{lll}
\hline
\textbf{CAEVO} & \textbf{TempRelPro} & \textbf{\textit{Differences}}\\
\hline
Rules-Verb/Time Adjacent & Event-timex Rules & \\
Rules-TimeTime & Timex-timex Rules & \\
Rules-Reporting Governor & Event-event Rules & \\
Rules-Reichenbach & Event-event Rules & \\
Rules-General Governor & Event-event Rules & \\
Rules-WordNet & - & \\
\textit{Rules-Reporting DCT} & \textit{Event-DCT Rules} & \textit{TempRelPro considers all types of events, not only reporting events}\\
\textit{ML-E-T SameSent} & \textit{Event-timex SVM} & \textit{TempRelPro considers both intra- and inter- sentential pairs}\\
\textit{ML-E-E SameSent} & \textit{Event-event SVM} & \textit{TempRelPro considers both intra- and inter- sentential pairs}\\
ML-E-E Dominate & - & \\
ML-E-DCT & Event-DCT SVM & \\
Rules-AllVague & - & \\
\hline
\end{tabular}
\end{adjustbox}
\caption{\label{tab:caevo-vs-temprelpro}CAEVO's classifiers vs TempRelPro's classifiers}
\end{table}

\subsubsection{Timex-timex Rules}
\label{sec:timex-timex-rules}

Only temporal expressions of types \texttt{DATE} and \texttt{TIME} are considered in the hand-crafted set of rules, based on their \textit{normalized values}. For example, \timex{7 PM tonight} with \texttt{value} = \texttt{2015-12-12T19:00} \texttt{IS\_INCLUDED} in \timex{today} with \texttt{value} = \texttt{2015-12-12}. 

\subsubsection{Event-DCT Rules}
\label{sec:event-dct-rules}

The rules for E-D pairs are based on the tense and/or aspect of the event word:
\begin{itemize}
\item \textit{If} \texttt{tense}$_E$ = \texttt{PAST} \textit{and} \texttt{aspect}$_E$ = \texttt{PERFECTIVE} \textit{then} [E \texttt{BEFORE} D]
\item \textit{If} \texttt{tense}$_E$ = \texttt{PRESENT} \textit{and} \texttt{aspect}$_E$ = \texttt{PROGRESSIVE} \textit{then} [E \texttt{INCLUDES} D]
\item \textit{If} \texttt{tense}$_E$ = \texttt{PRESENT} \textit{and} \texttt{aspect}$_E$ = \texttt{PERFECTIVE\_PROGRESSIVE} \textit{then} [E \texttt{INCLUDES} D]
\item \textit{If} \texttt{tense}$_E$ = \texttt{FUTURE} \textit{then} [E \texttt{AFTER} D] 
\end{itemize}

\subsubsection{Event-timex Rules}
\label{sec:event-timex-rules}

Many prepositions in English have temporal senses, as has been discussed in The Preposition Project (TPP)~\parencite{Litkowski:Hargraves:06} and the Pattern Dictionary of English Prepositions (PDEP)~\parencite{litkowski:2014:P14-1}. We took the list of \textit{temporal prepositions}\footnote{\url{http://www.clres.com/db/classes/ClassTemporal.php}} and built a set of rules for E-T pairs based on their temporal senses (tsense). The rules are only applied whenever a temporal preposition establishes a temporal modifier relationship between an event (E) and a timex (T), based on the existing dependency path:
\begin{itemize}
\item \textit{If} \texttt{tsense} = \textsc{TimePoint} (e.g., \signal{in}, \signal{at}, \signal{on}) \textit{then} [E \texttt{IS\_INCLUDED} T]
\item \textit{If} \texttt{tsense} = \textsc{TimePreceding} (e.g., \signal{before}) \textit{then} [E \texttt{BEFORE} T]
\item \textit{If} \texttt{tsense} = \textsc{TimeFollowing} (e.g., \signal{after}) \textit{then} [E \texttt{AFTER} T]
\item \textit{If} \texttt{tsense} = \textsc{Duration} (e.g., \signal{during}, \signal{throughout}) \textit{then} [E \texttt{DURING} T]
\item \textit{If} \texttt{tsense} = \textsc{StartTime} (e.g., \signal{from}, \signal{since}) \textit{then} [E \texttt{BEGUN\_BY} T]
\item \textit{If} \texttt{tsense} = \textsc{EndTime} (e.g., \signal{until}) \textit{then} [E \texttt{ENDED\_BY} T]
\end{itemize}

In the absence of a temporal preposition, a timex might simply be a temporal modifier of an event, as exemplified in ``Police \eventattr{confirmed}{E} \timexattr{Friday}{T} that the body was found...''. In this case, we assume that [E \texttt{IS\_INCLUDED} T].

Moreover, sometimes events are modified by temporal expressions marking the starting time and ending time in a \textit{duration pattern}. For example, `between \texttt{tmx\_begin} and \texttt{tmx\_end}', `from \texttt{tmx\_begin} to/until \texttt{tmx\_end}' or `\texttt{tmx\_begin}-\texttt{tmx\_end}'. We define the rules as follow:
\begin{itemize}
\item \textit{If} T matches \texttt{tmx\_begin} \textit{then} [E \texttt{BEGUN\_BY} T]
\item \textit{If} T matches \texttt{tmx\_end} \textit{then} [E \texttt{ENDED\_BY} T]
\end{itemize}

\subsubsection{Event-event Rules}
\label{sec:event-event-rules}

The first set of rules applied to E-E pairs is based on the existing dependency path (\texttt{dep})\footnote{The dependency path syntax is according to The CoNLL-2008 Shared Task on Joint Parsing of Syntactic and Semantic Dependencies~\parencite{surdeanu-EtAl:2008:CONLL}.} between the first event (E$_1$) and the second event (E$_2$):
\begin{itemize}
\item \textit{If} E$_2$ is the \textit{logical subject} of E$_1$ (a passive verb), i.e., \texttt{dep} = \texttt{LGS-PMOD} \textit{then} [E$_1$ \texttt{AFTER} E$_2$], e.g., ``The disastrous chain reaction \eventattr{touched}{E$_1$} off by the \eventattr{collapse}{E$_2$} of Lehman Brothers...''
\item \textit{If} E$_2$ is the \textit{locative adverb} of E$_1$, i.e., \texttt{dep} = \texttt{LOC-PMOD} \textit{then} [E$_1$ \texttt{IS\_INCLUDED} E$_2$], e.g., ``China's current economic policies cause an enormous \eventattr{surge}{E$_1$} in coal \eventattr{consumption}{E$_2$}.''
\item \textit{If} E$_2$ is the \textit{predicative complement} of E$_1$ (a raising/control verb), i.e., \texttt{dep} = \texttt{OPRD-IM} or \texttt{dep} = \texttt{OPRD}:
\begin{itemize}
\item \textit{If} E$_1$ is an aspectual verb for \textit{initiation} (e.g., begin, start) \textit{then} [E$_1$ \texttt{BEGINS} E$_2$], e.g., ``The situation \eventattr{began}{E$_1$} to \eventattr{relax}{E$_2$} in the early 1990s.''
\item \textit{If} E$_1$ is an aspectual verb for \textit{culmination/termination} (e.g., finish, stop) \textit{then} [E$_1$ \texttt{ENDS} E$_2$], e.g., ``There 's some price at which we 'd \eventattr{stop}{E$_1$} \eventattr{bidding}{E$_2$}.''
\item \textit{If} E$_1$ is an aspectual verb for \textit{continuation} (e.g., continue, keep) \textit{then} [E$_1$ \texttt{INCLUDES} E$_2$], e.g., ``The maturing industry 's growth \eventattr{continues}{E$_1$} to \eventattr{slow}{E$_2$}.''
\item \textit{If} E$_1$ is a general verb \textit{and} \texttt{aspect}$_{E_1}$ = \texttt{PERFECTIVE\_PROGRESSIVE} \textit{then} [E$_1$ \texttt{SIMULTANEOUS} E$_2$], e.g., ``Hewlett-Packard have been \eventattr{working}{E$_1$} to \eventattr{develop}{E$_2$} quantum computers.''
\item \textit{If} E$_1$ is a general verb \textit{then} [E$_1$ \texttt{BEFORE} E$_2$], e.g., ``The AAR consortium \eventattr{attempted}{E$_1$} to \eventattr{block}{E$_2$} a drilling joint venture.''
\end{itemize}
\end{itemize}

The other sets of rules are taken from CAEVO~\parencite{chambers-etal:2014:TACL}, including:
\begin{itemize}
\item Rules for links between a reporting event and another event that is syntactically dominated by the reporting event, based on the \texttt{tense} and \texttt{aspect} of both events.
\item Reichenbach rules based on the analysis of the role played by various tenses of English verbs in conveying temporal discourse~\parencite{reichenbach47}.
\end{itemize}

\subsubsection{Temporal Reasoner}
\label{sec:temporal-reasoner}

Consider as an example the following news excerpt taken from the TimeBank corpus, annotated with events and temporal expressions:
\begin{quote}
She (Magdalene Albright, Ed.) then lavished praise, and the State Department's award for heroism, on embassy staffers before \eventattr{meeting}{e$_{116}$} with bombing victims at the Muhimbili Medical Center and with government officials. [\dots] (During the meeting, Ed.) Albright \eventattr{announced}{e$_{45}$} a gift of 500 pounds (225 kilograms) of medical supplies to Tanzania and Kenya from the Walter Reed Army Medical Center. She also \eventattr{pledged}{e$_{46}$} to \eventattr{ask}{e$_{47}$} Congress to approve [\dots] 
\end{quote}
The annotated temporal relations of the documents are the following: [e$_{45}$ \texttt{BEFORE} e$_{46}$], [e$_{46}$ \texttt{BEFORE} e$_{47}$], [e$_{116}$ \texttt{IS\_INCLUDED} e$_{45}$]\footnote{This is an annotation error that later causes an inconsistency in the temporal graph during the consistency checking.} and [e$_{116}$ \texttt{INCLUDES} e$_{46}$].

An annotated TimeML document can be mapped into a constraint problem according to how \texttt{TLINK}s are mapped into Allen relations (Table~\ref{tab:allen}). A possible mapping is as follows: 
\begin{itemize}
\item $<$ and $>$ for \texttt{BEFORE} and \texttt{AFTER}
\item $o$ and $o^{-1}$ for \texttt{DURING} and \texttt{DURING\_INV}
\item $d$ and $d^{-1}$ for \texttt{IS\_INCLUDED} and \texttt{INCLUDES}
\item $s$ and $s^{-1}$ for \texttt{BEGINS} and \texttt{BEGUN\_BY}
\item $f$ and $f^{-1}$ for \texttt{ENDS} and \texttt{ENDED\_BY}
\end{itemize}

For instance, the \texttt{TLINK}s in the previous excerpt can be mapped as follows: $<$ for \texttt{BEFORE} between \textit{announced} and \textit{pledged}; $d$ for \texttt{IS\_INCLUDED} between \textit{meeting} and \textit{announced}; and $d^{-1}$ for \texttt{INCLUDES} between \textit{meeting} and \textit{pledge}. Other mappings are possible, e.g., by relaxing the mapping of \texttt{BEFORE} and its inverse \texttt{AFTER} into $<\!\!\!m$ and  $<\!\!\!m^{-1}$, respectively, considering vagueness in interpreting temporal annotations. For example, $<$ for \texttt{BEFORE} between \textit{announced} and \textit{pledged} could be replaced by $<m$ in case of uncertainty whether one event is before or \textit{immediately} before the other.

These and other mappings are handled by the \textit{Service-oriented Qualitative Temporal Reasoner} (SQTR), which was developed for reasoning on TimeML documents within the TERENCE FP7 project (GA n.\ 257410) in a \textit{Service-Oriented Architecture} context~\parencite{erl2004service}. SQTR is used to check consistency and perform deduction, and relies on the \textit{Generic Qualitative Reasoner} (GQR), a fast solver for generic qualitative constraint problems, such as Allen constraint problems. The rationale of preferring GQR to other solutions, such as fast SAT solvers, is due to its scalability, simplicity of use and efficiency performances~\parencite{DBLP:conf/ijcai/WestphalW09}.

SQTR behaves as follows:
\begin{itemize}
  \item In case of \textit{consistency checking}, SQTR maps the TimeML document into a GQR constraint problem, invokes GQR, and returns a true/false value. In case of consistency, it also returns the mapping for which consistency is found for informing the deduction operation. If we consider the previous example, the system will detect an inconsistency, caused by the annotation of \texttt{IS\_INCLUDED} between \textit{meeting} and \textit{announced}, which should be \texttt{INCLUDES} instead for the set of \texttt{TLINK}s to be consistent.
  \item In case of \textit{deduction}, SQTR maps the TimeML document into a GQR constraint problem, invokes GQR, maps the GQR output to a TimeML document, marks the deduced \texttt{TLINK}s with an attribute \texttt{deduced} set to true, and returns such a document as the result. The system will deduce, for example, a new relation \texttt{BEFORE} between \textit{announced} and \textit{ask}, because the same relation holds between \textit{announced} and \textit{pledge} and between \textit{pledge} and \textit{ask}.
\end{itemize}

Note that the temporal reasoner only deduce new \texttt{TLINK}s if the TimeML document is found to be consistent.

\subsubsection{SVM Classifiers}
\label{sec:pair-classifier}

We built three supervised classification models each for event-DCT (E-D), event-timex (E-T) and event-event (E-E) pairs, using LIBLINEAR \parencite{REF08a} L2-regularized L2-loss linear SVM (dual), with default parameters, and one-vs-rest strategy for multi-class classification.

\paragraph{Tools and Resources} Several external tools and resources are used to extract features from each temporal entity pair, including:
\begin{itemize}
\item TextPro tool
suite\footnote{\url{http://textpro.fbk.eu/}}~\parencite{PIANTA08.645} to get the morphological analysis (PoS tags, shallow phrase chunk) of each token in the text.
\item Mate tools\footnote{\url{http://code.google.com/archive/p/mate-tools/}}~\parencite{bjorkelund-EtAl:2010:COLING-DEMOS} to extract the dependency path between tokens in the document.
\item WordNet similarity module\footnote{\url{http://ws4jdemo.appspot.com/}} to compute (Lin) semantic similarity/relatedness~\parencite{Lin:1998:IDS:645527.657297} between words.
\item Temporal signal lists as described in \textcite{mirza-tonelli:2014:EACL}. However, we further expand the lists using the Paraphrase Database \parencite{ganitkevitch2013ppdb}, and manually cluster some signals together, e.g. \{\textit{before}, \textit{prior to}, \textit{in advance of}\}. Finally, we have 50 timex-related and 138 event-related temporal signals in total, which are clustered into 27 and 35 clusters, respectively (Appendix~\ref{app:temporal-signals}).
\end{itemize}

\begin{savenotes}
\begin{table}[h!]
\centering
\small
\begin{adjustbox}{width=1\textwidth}
\begin{tabular}{lccccl}
\hline
\textbf{Feature} & \textbf{E-D} & \textbf{E-T} & \textbf{E-E} & \textbf{Rep.} & \textbf{Description}\\
\hline
\multicolumn{4}{l}{\textbf{Morphosyntactic information}} & \\
\hspace{5pt}PoS & x & x & x & one-hot & Part-of-speech tags of $e_1$ and $e_2$.\\
\hspace{5pt}phraseChunk & x & x & x & one-hot & Shallow phrase chunk of $e_1$ and $e_2$.\\
\hdashline
\hspace{5pt}samePoS & & x & x & binary & Whether $e_1$ and $e_2$ have the same PoS.\\
\hline
\multicolumn{4}{l}{\textbf{Textual context}} & \\
\hspace{5pt}entityOrder & & x & & binary & Appearance order of $e_1$ and $e_2$ in the text.\footnote{The order of $e_1$ and $e_2$ in event-event pairs is always according to the appearance order in the text, while in event-timex pairs, $e_2$ is always a timex regardless of the appearance order.}\\
\hspace{5pt}sentenceDistance & & x & x & binary & \texttt{0} if $e_1$ and $e_2$ are in the same sentence, \texttt{1} otherwise.\\
\hspace{5pt}entityDistance & & x & x & binary & \texttt{0} if $e_1$ and $e_2$ are adjacent, \texttt{1} otherwise.\\
\hline
\multicolumn{4}{l}{\textbf{\texttt{EVENT} attributes}} & \\
\hspace{5pt}class & x & x & x & one-hot & \multirow{4}{*}{\texttt{EVENT} attributes as specified in TimeML.}\\
\hspace{5pt}tense & x & x & x & one-hot & \\
\hspace{5pt}aspect & x & x & x & one-hot & \\
\hspace{5pt}polarity & x & x & x & one-hot & \\
\hdashline
\hspace{5pt}sameClass & & & x & binary & \multirow{3}{*}{Whether $e_1$ and $e_2$ have the same \texttt{EVENT} attributes.}\\
\hspace{5pt}sameTenseAspect & & & x & binary & \\
\hspace{5pt}samePolarity & & & x & binary & \\
\hline
\multicolumn{4}{l}{\textbf{\texttt{TIMEX3} attributes}} & \\
\hspace{5pt}type & x & x & & one-hot & \texttt{TIMEX3} attributes as specified in TimeML.\\
\hline
\multicolumn{4}{l}{\textbf{Dependency information}} & \\
\hspace{5pt}dependencyPath & & & x & one-hot & Dependency path between $e_1$ and $e_2$.\\
\hspace{5pt}isMainVerb & x & x & x & binary & Whether $e_1$/$e_2$ is the main verb of the sentence.\\
\hline
\multicolumn{4}{l}{\textbf{Temporal signals}} & \\
\hspace{5pt}signalTokens & & x & x & one-hot & Tokens (cluster) of temporal signal around $e_1$ and $e_2$.\\
\hspace{5pt}signalPosition & & x & x & one-hot & Temporal signal position w.r.t $e_1$/$e_2$, \\
 & & & & & e.g., \texttt{BETWEEN}, \texttt{BEFORE}, \texttt{BEGIN}, etc.\\
\hspace{5pt}signalDependency & & x & x & one-hot & Temporal signal dependency path between signal \\
 & & & & & tokens and $e_1$/$e_2$.\\
\hline
\multicolumn{4}{l}{\textbf{Lexical semantic information}} & \\
\hspace{5pt}wnSim & & & x & one-hot & WordNet similarity computed between the lemmas \\
 & & & & & of $e_1$ and $e_2$.\\
\hline
\multicolumn{4}{l}{\textbf{\texttt{TLINK} labels from the rule-based sieve}} & \\
\hspace{5pt}timex-DCT label & & x & & one-hot & The \texttt{TLINK} type of the $e_2$ (timex) and DCT pair (if any). \\
\hspace{5pt}event-DCT label & & & x & one-hot & The \texttt{TLINK} types of the $e_1$/$e_2$ and DCT pairs (if any). \\
\hline
\end{tabular}
\end{adjustbox}
\caption{\label{tab:feature-set}Feature set for event-DCT (E-D), event-timex (E-T) and event-event (E-E) classification models, along with each feature representation (Rep.) in the feature vector and feature descriptions.}
\end{table}
\end{savenotes}

\paragraph{Feature Set} The implemented features are listed in Table~\ref{tab:feature-set}. Some features are computed independently based on either $e_1$ or $e_2$ of the temporal entity pairs, while some others are pairwise features, which are computed based on
both entities. In order to have a feature vector of reasonable size, we simplified the possible values of some features during the one-hot encoding:
\begin{itemize}
\item \textit{dependencyPath} We only consider several dependency path between the event pairs denoting e.g. coordination, subordination, subject and object relations. 
\item \textit{signalTokens} The \textit{clusterID} of signal cluster, e.g., \{\textit{before}, \textit{prior to}, \textit{in advance of}\}, is considered as a feature instead of the signal tokens.
\item \textit{signalDependency} For each atomic label in a vector of syntactic dependency labels according to~\textcite{surdeanu-EtAl:2008:CONLL},\footnote{We manually selected a subset of such labels that are relevant for temporal modifier.} if the signal dependency path contains the atomic label, the value in the feature vector is flipped to 1. Hence, \texttt{TMP-SUB} and \texttt{SUB-TMP} will have the same one-hot representations.
\item \textit{wnSim} The value of WordNet similarity measure is discretized as follows: $sim\leq0.0$, $0.0<sim\leq0.5$, $0.5<sim\leq1.0$ and $sim>1.0$.
\end{itemize}
Note that several features from \textcite{mirza-tonelli:2014:EACL} such as \textit{string features} and \textit{temporal discourse connectives}\footnote{The information about discourse connectives was acquired using the addDiscourse tool~\parencite{pitler-nenkova:2009:Short}, which identifies connectives based on syntactic constructions, and assigns them to one of four semantic classes: Temporal, Expansion, Contingency and Comparison.} are not used. String features, i.e., \textit{token} and \textit{lemma} of temporal entities, are removed in order to increase the classifiers' robustness in dealing with completely new texts with different vocabularies. So instead, we include \textit{WordNet similarity} in the feature set. Temporal discourse connectives are no more included as features because it did not prove to be beneficial.

\paragraph{Label Simplification} During the feature extraction process for training the classification models, we collapse some labels, i.e., \texttt{IBEFORE} into \texttt{BEFORE}, \texttt{IAFTER} into \texttt{AFTER}, \texttt{DURING} and \texttt{DURING\_INV} into \texttt{SIMULTANEOUS}, in order to simplify the learning process, also considering the sparse annotation of such labels in the datasets.

\section{Evaluation}
\label{sec:temp-rel-evaluation}

\subsection{TempEval-3 Evaluation}
\label{sec:temp-rel-tempeval3}

\paragraph{Dataset} We use the same training and test data released in the context of Tempeval-3. Two types of training data were made available in the challenge: \textit{TBAQ-cleaned} and \textit{TE3-Silver-data}. The former includes the cleaned and improved version of the TimeBank 1.2 corpus and the AQUAINT TimeML corpus (see Section~\ref{sec:timeml-corpora}). TE3-Silver-data, instead, is a 600K word corpus annotated by the best performing systems at Tempeval-2, which we do not use because it was proven not so useful for temporal relation extraction task \parencite{uzzaman-EtAl:2013:SemEval-2013}. For evaluation, the newly created \textit{TempEval-3-platinum}
evaluation corpus is used. The distribution of the relation types in training and test datasets
is shown in Table~\ref{tab:tbaq-dataset}.

\begin{table}[t]
\centering
\small
\begin{tabular}{rrrr|c|rrrr}
\hline
\multicolumn{4}{c|}{\textbf{TBAQ-cleaned}} & \multirow{3}{*}{\textbf{Relation}} & \multicolumn{4}{c}{\textbf{TempEval-3-platinum}} \\
\multicolumn{4}{c|}{\textit{training}} & & \multicolumn{4}{c}{\textit{test}} \\
\cline{1-4}\cline{6-9}
T-T & E-D & E-T & E-E & & T-T & E-D & E-T & E-E\\ \hline
\textbf{168} & \textbf{1366} & 230 & \textbf{1797} & \texttt{BEFORE} & 7 & \textbf{90} & 5 & \textbf{210} \\
29 & 205 & 124 & 1141 & \texttt{AFTER} & 2 & 26 & 5 & 184\\
2 & 1 & 2 & 70 & \texttt{IBEFORE} & 2 &  & 5 & 2\\
1 & 1 & 4 & 33 & \texttt{IAFTER} & 1 &  & 7 & 1\\
111 &  & 5 & 742 & \texttt{IDENTITY} &  &  &  & 15\\
2 & 2 & 56 & 519 & \texttt{SIMULTANEOUS} & 4 & 0 & 6 & 81\\
107 & 554 & 141 & 462 & \texttt{INCLUDES} & \textbf{8} & 37 & 2 & 40\\
29 & 471 & \textbf{1782} & 262 & \texttt{IS\_INCLUDED} & 4 & 14 & \textbf{114} & 47\\
 & 61 & 119 & 38 & \texttt{DURING} &  &  &  & \\
 & 1 & 19 & 42 & \texttt{DURING\_INV} &  &  &  & 1\\
1 &  & 23 & 48 & \texttt{BEGINS} & 1 &  & 1 & 1\\
3 & 11 & 56 & 38 & \texttt{BEGUN\_BY} & 1 &  & 1 & \\
1 &  & 65 & 33 & \texttt{ENDS} &  &  & 2 & 1\\
14 & 8 & 47 & 46 & \texttt{ENDED\_BY} &  &  & 2 & \\
\hline
\textbf{468} & \textbf{2681} & \textbf{2673} & \textbf{5271} & \textbf{Total} & \textbf{30} & \textbf{167} & \textbf{150} & \textbf{583}\\
\hline
\end{tabular}
\caption{\label{tab:tbaq-dataset}The distribution of each relation type in the datasets for each type of temporal entity pairs: timex-timex (T-T), event-DCT (E-D), event-timex (E-T) and event-event (E-E). }
\end{table}




\paragraph{Evaluation Metrics} TempEval-3 introduced an evaluation metric \parencite{uzzaman-allen:2011:ACL-HLT2011} capturing temporal awareness in terms of precision, recall and F1-score. To compute precision and recall, the correctness of annotated temporal links is verified using temporal closure, by checking the existence of the identified relations in the closure graph. However, there is a minor variation of the formula, that the reduced graph of relations is considered instead of all relations of the system and reference.\footnote{Details can be found in Chapter 6 of \parencite{uzzaman-2012}.}
\[Precision=\frac{|Sys_{relation}^{-}\cap Ref_{relation}^{+}|}{|Sys_{relation}^{-}|}\]
\[Recall=\frac{|Ref{}_{relation}^{-}\cap Sys{}_{relation}^{+}|}{|Ref{}_{relation}^{-}|}\]
\textit{Precision} is the proportion of the number of reduced system relations ($Sys_{relation}^{-}$) that can be verified from the reference annotation temporal closure graph ($Ref_{relation}^{+}$), out of the number of temporal relations in the reduced system relations ($Sys_{relation}^{-}$). Similarly, \textit{Recall} is the proportion of number of reduced reference annotation relations ($Ref{}_{relation}^{-}$) that can be verified from the system's temporal closure graph ($Sys{}_{relation}^{+}$), out of the number of temporal relations in reduced reference annotation ($Ref{}_{relation}^{-}$). In order to replicate this type of evaluation, we use the scorer made available to the task participants.

\paragraph{Evaluation Results}


We compare in Table~\ref{tab:compare-tempeval-3} the performance of TempRelPro to the other systems participating in temporal relation tasks of TempEval-3, Task C and Task C relation type only, according to the figures reported in \parencite{uzzaman-EtAl:2013:SemEval-2013}. We also compare TempRelPro performance with our preliminary results reported in~\textcite{mirza-tonelli:2014:EACL} for Task C relation type only. 

For the temporal relation type classification task (Task C relation type only), TempRelPro achieves the best performance with 61.86\% F1-score. For the temporal relation extraction task (Task C), our approach is most similar to UTTime-1 with the highest recall in TempEval-3. In comparison with UTTime-1, we can double the precision without reducing too much the recall. TempRelPro achieves the best F1-score of 40.15\%, almost 4\% increase compared with the best system in TempEval-3, ClearTK-2.

\begin{table}[t]
\centering
\small
\begin{tabular} {lccc|ccc}
\hline
\multirow{2}{*}{\textbf{System}} & \multicolumn{3}{c|}{\textbf{Task C}} & \multicolumn{3}{c}{\textbf{Task C relation type only}} \\
& \textbf{P} & \textbf{R} & \textbf{F1} & \textbf{P} & \textbf{R} & \textbf{F1}\\ \hline 
\textbf{TempRelPro} & \textbf{30.30} & \textbf{59.49} & \textbf{40.15} & \textbf{62.13} & \textbf{61.59} & \textbf{61.86} \\
\hdashline
\textcite{mirza-tonelli:2014:EACL} & - & - & - & 58.80 & 58.17 & 58.48 \\
\hline
\textbf{TempEval-3} & & & & & & \\
ClearTK-2 & \textbf{37.32} & 35.25 & \textbf{36.26} & - & - & - \\ 
UTTime-5 & 35.94 & 33.92 & 34.90 & 53.85 & 55.58 & 54.70 \\ 
NavyTime-1 & 35.48 & 27.62 & 31.06 & 46.59 & 47.07 & 46.83\\ 
JU-CSE & 21.04 & 35.47 & 26.41 & 35.07 & 34.48 & 34.77\\ 
UTTime-1 & 15.18 & \textbf{65.64} & 24.65 & \textbf{55.58} & \textbf{57.43} & \textbf{56.45}\\
\hline
\end{tabular}
\caption{\label{tab:compare-tempeval-3}Tempeval-3 evaluation on temporal relation extraction tasks}
\end{table}

\begin{table}[t]
\centering
\small
\begin{tabular} {lcc|cc|cc|cc|ccc}
\hline
\multirow{2}{*}{\textbf{Sieve}} & \multicolumn{2}{c|}{\textbf{T-T}} & \multicolumn{2}{c|}{\textbf{E-D}} & \multicolumn{2}{c|}{\textbf{E-T}}  & \multicolumn{2}{c|}{\textbf{E-E}} & \multicolumn{3}{c}{\textbf{Overall}} \\
& \textbf{P} & \textbf{R} & \textbf{P} & \textbf{R} & \textbf{P} & \textbf{R} & \textbf{P} & \textbf{R} & \textbf{P} & \textbf{R} & \textbf{F1}\\ \hline 
\multicolumn{9}{l}{\textbf{Temporal Relation Identification}} \\
 & \textit{0.03} & 0.67 & 0.37 & 0.99 & 0.33 & 0.99 & 0.42 & 0.97 & 0.40 & 0.96 & 0.56 \\
\textit{Without T-T} & & & & & & & & & \textit{\textbf{0.53}} & \textit{\textbf{0.95}} & \textit{\textbf{0.68}} \\
\hline
\multicolumn{9}{l}{\textbf{Temporal Relation Type Classification}} \\
RB & 0.85 & 0.57 & 1 & 0.08 & 0.91 & 0.39 & 0.91 & 0.05 & 0.91 & 0.13 & 0.22 \\
ML &  &  & 0.77 & 0.76 & 0.73 & 0.72 & 0.53 & 0.51 & 0.61 & 0.58 & 0.59 \\
\hdashline
RB + TR  & 0.85 & 0.57 & 1 & 0.17 & 0.92 & 0.48 & 0.89 & 0.06 & 0.92 & 0.16 & 0.28 \\
RB + ML & 0.85 & 0.57 & 0.78 & 0.77 & 0.73 & 0.72 & 0.53 & 0.51 & 0.62 & 0.59 & 0.61 \\
\hdashline
\textbf{RB + TR + ML} & 0.85 & 0.57 & 0.79 & 0.78 & 0.75 & 0.81 & 0.53 & 0.51 & \textbf{0.63} & \textbf{0.61} & \textbf{0.62} \\
\hdashline
\textit{Majority labels} & \textit{0.35} & \textit{0.23} & \textit{0.55} & \textit{0.54} & \textit{0.77} & \textit{0.76} & \textit{0.37} & \textit{0.36} & \textit{0.47} & \textit{0.45} & \textit{0.46} \\
\hline
\end{tabular}
\caption{\label{tab:temprelpro-modules}TempRelPro performances per module on temporal relation identification and type classification, evaluated on the TempEval-3 evaluation corpus. RB = rule-based sieve, ML = machine-learned sieve and TR = temporal reasoner.}
\end{table}

We also report in Table~\ref{tab:temprelpro-modules} the performances of each module included in TempRelPro, evaluated on TempEval-3-platinum. The temporal relation identification module (Section~\ref{sec:temp-rel-method-identification}) obtains a very low precision for T-T pairs because the dataset contains very few annotated timex-timex links. If we remove the T-T pairs, we can increase the F1-score for the temporal relation identification task by 12\%. Therefore, in our final annotated TimeML documents for the TempEval-3 evaluation, T-T pairs are not included, even though they play a big role in the temporal relation type classification task. 

Regarding the temporal relation type classification modules (Section~\ref{sec:temp-rel-method-classification}), there is no significant improvement by combining rule-based and machine-learned sieves (RB + ML), compared with only using machine-learned classifiers (ML), particularly for E-T and E-E pairs. However, introducing the temporal reasoner in between (RB + TR + ML) results in significant improvement especially for E-T pairs, since recall increases from 72\% to 81\%. We also compare TempRelPro performances for this classification task to a majority class baseline for each temporal entity type according to the distribution of temporal relation types in the training data (Table~\ref{tab:tbaq-dataset}), i.e., \texttt{BEFORE} for T-T, \texttt{BEFORE} for E-D, \texttt{IS\_INCLUDED} for E-T and \texttt{BEFORE} for E-E pairs.

\subsection{TimeBank-Dense Evaluation}
\label{sec:temp-rel-timebank-dense}

\paragraph{Dataset} We follow the experimental setup in \textcite{chambers-etal:2014:TACL}, in which the TimeBank-Dense corpus (mentioned in Section~\ref{sec:timeml-corpora}) is split into a 22 document training set, a 5 document development set and a 9 document test set\footnote{Available at \url{http://www.usna.edu/Users/cs/nchamber/caevo/}.}. All the classification models for the machine-learned sieve are trained using the training set. We evaluate our system performances on the test set.

\paragraph{Adjustments} The set of \texttt{TLINK} types used in TimeBank-Dense corpus is different from the one  used in TempEval-3. Some relation types are not used, and the \texttt{VAGUE} relation introduced at the first TempEval task \parencite{verhagen-EtAl:2007:SemEval-2007} is adopted to cope with ambiguous temporal relations, or to indicate pairs for which no clear temporal relation exists. The final set of \texttt{TLINK} types in TimeBank-Dense includes: \texttt{BEFORE}, \texttt{AFTER}, \texttt{INCLUDES}, \texttt{IS\_INCLUDED}, \texttt{SIMULTANEOUS} and \texttt{VAGUE}. Therefore, we map the relation types of \texttt{TLINK}s labelled by TempRelPro as follows:\footnote{We tried different mappings in our experiments and found this mapping to be the one giving the best outcome.}
\begin{itemize}
\item \texttt{IBEFORE}, \texttt{BEGINS} and \texttt{ENDED\_BY} into \texttt{BEFORE}
\item \texttt{IAFTER}, \texttt{BEGUN\_BY} and \texttt{ENDS} into \texttt{AFTER}
\item \texttt{IDENTITY}, \texttt{DURING} and \texttt{DURING\_INV} into \texttt{SIMULTANEOUS}
\end{itemize}
Moreover, we introduce some rules for E-D and E-T pairs to recover the \texttt{VAGUE} relations, such as:
\begin{itemize}
\item If the PoS tag of the event in an E-D pair is an adjective then [E \texttt{VAGUE} D]
\item If the timex \texttt{value} in an E-T pair is \texttt{PAST\_REF}, \texttt{PRESENT\_REF} or \texttt{FUTURE\_REF} then [E \texttt{VAGUE} T]
\end{itemize}

\begin{table}[t]
\centering
\small
\begin{tabular} {lcccc|ccc}
\hline
\multirow{2}{*}{\textbf{System}} & {\textbf{T-T}} & {\textbf{E-D}} & {\textbf{E-T}}  & {\textbf{E-E}} & \multicolumn{3}{c}{\textbf{Overall}} \\
& \textbf{P/R/F1} & \textbf{P/R/F1} & \textbf{P/R/F1} & \textbf{P/R/F1} & \textbf{P} & \textbf{R} & \textbf{F1}\\ \hline 
\textbf{TempRelPro} & \textbf{0.780} & 0.518 & \textbf{0.556} & 0.487 & \textbf{0.512} & \textbf{0.510} & \textbf{0.511} \\
CAEVO & 0.712 & \textbf{0.553} & 0.494 & \textbf{0.494} & 0.508 & 0.506 & 0.507 \\
\hline
\end{tabular}
\caption{\label{tab:temprelpro-vs-caevo}TempRelPro performances evaluated on the TimeBank-Dense test set and compared with CAEVO.}
\end{table}

\paragraph{Evaluation Results} In Table~\ref{tab:temprelpro-vs-caevo} we report the performances of TempRelPro compared with CAEVO. We achieve a small improvement in the overall F1-score, i.e., 51.1\% vs 50.7\%. For each temporal entity pair type, since we label all possible links, precision and recall are the same. TempRelPro is significantly better than CAEVO in labelling T-T and E-T pairs.

\begin{table}[t]
\centering
\begin{adjustbox}{width=1\textwidth}
\begin{tabular} {lc|cc|cc|cc|ccc|ccc}
\hline
\multirow{2}{*}{\textbf{Sieve}} & {\textbf{T-T}} & \multicolumn{2}{c|}{\textbf{E-D}} & \multicolumn{2}{c|}{\textbf{E-T}}  & \multicolumn{2}{c|}{\textbf{E-E}} & \multicolumn{3}{c}{\textbf{Overall}} & \multicolumn{3}{c}{\textbf{CAEVO}} \\
& \textbf{P/R/F1} & \textbf{P} & \textbf{R} & \textbf{P} & \textbf{R} & \textbf{P} & \textbf{R} & \textbf{P} & \textbf{R} & \textbf{F1} & \textbf{P} & \textbf{R} & \textbf{F1}\\ \hline 
\multicolumn{9}{l}{\textbf{Temporal Relation Type Classification}} \\
RB & 0.780 & 0.667 & 0.070 & 0.705 & 0.073 & 0.609 & 0.010 & 0.727 & 0.049 & 0.092 \\
ML &  & \multicolumn{2}{c|}{0.473} & \multicolumn{2}{c|}{0.480} & \multicolumn{2}{c|}{0.488} & 0.484 & 0.471 & 0.478 & 0.458 & 0.202 & 0.280 \\
\hdashline
RB + TR  & 0.780 & 0.722 & 0.125 & 0.700 & 0.166 & 0.546 & 0.013 & 0.713 & 0.076 & 0.138 \\
RB + ML & 0.780 & \multicolumn{2}{c|}{0.495} &  \multicolumn{2}{c|}{0.480} & \multicolumn{2}{c|}{0.488} & 0.495 & 0.493 & 0.494 & 0.486 & 0.240 & 0.321 \\
\hdashline
\textbf{RB + TR + ML} & 0.780 & \multicolumn{2}{c|}{0.518} &  \multicolumn{2}{c|}{0.556} & \multicolumn{2}{c|}{0.487} & \textbf{0.512} & \textbf{0.510} & \textbf{0.511} & 0.505 & 0.328 & 0.398 \\
\hdashline
\textit{RB + TR + ML + AllVague} &  &  &  &  &  &  &  &  &  &  & \textit{0.507} & \textit{0.507} & \textit{0.507} \\
\hline
\end{tabular}
\end{adjustbox}
\caption{\label{tab:temprelpro-timebank-dense-modules}TempRelPro performances per module on temporal relation type classification, evaluated on the TimeBank-Dense test set, and compared with CAEVO. RB = rule-based sieve, ML = machine-learned sieve and TR = temporal reasoner.}
\end{table}

We also report in Table~\ref{tab:temprelpro-timebank-dense-modules} the performances of each module composing TempRelPro, evaluated on the TimeBank-Dense test set. Note that one of differences between TempRelPro and CAEVO is that in TempRelPro machine-learned sieve (ML) is the last sieve, while in CAEVO \textit{AllVague} is the last sieve. This explains the big difference of F1-score for the RB + TR + ML composition in TempRelPro and CAEVO, i.e., 51.1\% vs 39.8\%. 

In general, combining RB and ML modules results in a slight improvement (47.8\% to 49.4\% F1-score), especially for T-T (since there is no ML classifier for T-T pairs) and E-D pairs, but not for E-T and E-E pairs. Introducing the TR module in between (RB + TR + ML) is even more beneficial, resulting in overall 51.1\% F1-score, especially for E-T pairs with an increase from 48\% to 55.6\% F1-score. This is in line with the results of the TempEval-3 evaluation (Section~\ref{sec:temp-rel-tempeval3}). 

With only two sieves, TempRelPro is arguably more  efficient than CAEVO, because (i) the temporal closure inference over extracted \texttt{TLINK}s is run only once and (ii) we use less classifiers in general (see Table~\ref{tab:caevo-vs-temprelpro}). Our decision to consider all rule-based classifiers as one sieve is motivated by the hypothesis that entity pairs generated by each rule-based classifier, i.e. E-D, E-T and E-E pairs, are independent of each other. Using the consistency checking module of the temporal reasoner, we found out that all the documents in the test set, annotated by the rule-based classifiers, are consistent, which supports our hypothesis.

\subsection{QA TempEval Evaluation}
\label{sec:temp-rel-qa-tempeval}

\paragraph{Dataset} The training data set is the TimeML annotated data released by the task organizers, which includes TBAQ-cleaned and TE3-Platinum corpora reused
from the TempEval-3 task~\parencite{uzzaman-EtAl:2013:SemEval-2013}.
The test data are 30 plain texts extracted from news, wikipedia and blogs domains (10 documents each). For evaluating
the system, 294 temporal-based questions and the test data annotated with entities relevant for the
questions are used.

\paragraph{Temporal Entity Extraction System} We use the same systems reported in \textcite{mirza-minard:2015:SemEval} for timex and event extraction.

\paragraph{Evaluation System} Given the documents labelled by the participating systems, the evaluation process consists of three main steps~\parencite{llorens-EtAl:2015:SemEval}:
\begin{itemize}
\item \textit{ID normalization}: this step is performed because systems may provide different IDs to the same temporal entities annotated in the gold standard test data.
\item \textit{Timegraph generation}: Timegraph~\parencite{gerevini1995} is used to compute temporal closure as proposed by \textcite{journals/ci/MillerS90}. Timegraph is first initialized by adding the system's explicit \texttt{TLINK}s. Then the Timegraph’s reasoning mechanism infers implicit
relations through rules such as transitivity.
\item \textit{Question processing}: queries are converted to point-based queries in order to check the necessary point relations in Timegraph to verify an interval relation. For example, to answer the question ``is $e_1$ \texttt{AFTER} $e_2$'', the evaluation system verifies whether $start(e_1)>end(e_2)$; if it is verified then the answer is true (\texttt{YES}), if it conflicts with the Timegraph then it is false (\texttt{NO}), otherwise it is \texttt{UNKNOWN}.
\end{itemize} 

\paragraph{Evaluation Metrics} For each question the obtained answer from the Timegraph (created with system annotations) is compared with the expected answer (human annotated). 
\[\textrm{Precision (}P\textrm{)} =\frac{\textrm{num\_correct}}{\textrm{num\_answered}}\]
\[\textrm{Recall (}R\textrm{)} =\frac{\textrm{num\_correct}}{\textrm{num\_questions}}\]
\[\textrm{F1-score (}F1\textrm{)} =\frac{2*P*R}{P+R}\]
Recall (QA accuracy) is used as the main metrics to rank the systems, and F1-score is used in case of the same recall. Coverage is used to measure how many questions can be answered by a system, regardless of the correctness. 

\paragraph{Evaluation Results} We compare TempRelPro with our previous system submitted for QA TempEval, HLT-FBK~\parencite{mirza-minard:2015:SemEval}, in Table~\ref{tab:qa-tempeval-results}. HLT-FBK shows a significant improvement by including an event co-reference rule\footnote{Whenever two events co-refer, the E-E pair is excluded from the classifier and automatically labelled \texttt{SIMULTANEOUS}.} (HLT-FBK + coref). The event co-reference information was obtained from the NewsReader pipeline.\footnote{More information about the NewsReader pipeline, as well as a demo, are available on the project website \url{http://www.newsreader-project.eu/results/}.} For TempRelPro, we include the event co-reference rule in the rule-based sieve for E-E pairs (Section~\ref{sec:event-event-rules}). Using event co-reference, the overall performance of TempRelPro (TempRelPro + coref) is slightly improved, especially for Blogs domain. In general, HLT-FBK + coref is very good in covering the number of questions answered (Cov), but not in answering accurately.

\begin{table}[t]
\centering
\begin{adjustbox}{width=1\textwidth}
\begin{tabular}{l|cccc|cccc|cccc||c}
\hline
\textbf{System} & \multicolumn{4}{c|}{\textbf{News}} & \multicolumn{4}{c|}{\textbf{Wikipedia}} & \multicolumn{4}{c||}{\textbf{Blogs}}& \multicolumn{1}{c}{\textbf{All}}\\ 
 & \textbf{Cov} & \textbf{P} & \textbf{R} & \textbf{F1} & \textbf{Cov} & \textbf{P} & \textbf{R} & \textbf{F1} & \textbf{Cov} & \textbf{P} & \textbf{R} & \textbf{F1}  & \textbf{R}\\
 \hline
TempRelPro & 0.62 & 0.62 & 0.38 & 0.48 & 0.55 & 0.74 & 0.41 & 0.52 & 0.34 & 0.45 & 0.15 & 0.23 & 0.34\\
\textbf{TempRelPro + coref} & 0.61 & \textbf{0.63} & \textbf{0.38} & \textbf{0.48} & 0.55 & \textbf{0.74} & \textbf{0.41} & \textbf{0.52} & 0.37 & \textbf{0.50} & 0.18 & \textbf{0.27} & \textbf{0.35}\\
\hdashline
HLT-FBK & 0.36 & 0.56 & 0.20 & 0.30 & 0.29 & 0.58 & 0.17 & 0.26 & 0.29 & 0.47 & 0.14 & 0.21 & 0.17 \\
HLT-FBK + coref & \textbf{0.69} & 0.43 & 0.29 & 0.35 & \textbf{0.58} & 0.62 & 0.36 & 0.46 & \textbf{0.58} & 0.34 & \textbf{0.20} & 0.25 & 0.30\\
\hline
\end{tabular}
\end{adjustbox}
\caption{TempRelPro performances in terms of coverage (Cov), precision (P), recall (R) and F1-score (F1), compared with HLT-FBK\label{tab:qa-tempeval-results}.}
\end{table}

\begin{table}[t]
\centering
\small
\begin{tabular}{l|cccc}
\hline
\textbf{System} & \textbf{Cov} & \textbf{P} & \textbf{R} & \textbf{F1}\\ 
 \hline
TempRelPro & 0.53 & 0.65 & 0.34 & 0.45\\
\textbf{TempRelPro + coref} & 0.53 & 0.66 & \textbf{0.35} & \textbf{0.46}\\
\hdashline
HLT-FBK + trefl & 0.48 & 0.61 & 0.29 & 0.39\\
HLT-FBK + coref + trefl & \textbf{0.67} & 0.51 & 0.34 & 0.40\\
HITSZ-ICRC + trefl & 0.15 & 0.58 & 0.09 & 0.15\\
\hdashline
CAEVO + trefl & 0.36 & 0.60 & 0.21 & 0.32\\
TIPSemB + trefl & 0.37 & 0.64 & 0.24 & 0.35\\
TIPSem + trefl & 0.40 & \textbf{0.68} & 0.27 & 0.38\\
\hline
\end{tabular}
\caption{TempRelPro performances in terms of coverage (Cov), precision (P), recall (R) and F1-score (F1) for all domains, compared with systems in QA TempEval augmented with TREFL\label{tab:qa-tempeval-results-trefl}.}
\end{table}

The QA TempEval organizers also provide an extra evaluation, augmenting the participating systems with a time expression reasoner (TREFL) as a post-processing step~\parencite{llorens-EtAl:2015:SemEval}. The TREFL component adds \texttt{TLINK}s between timexes based on their resolved values. Note that TempRelPro already includes the T-T links in the final TimeML documents produced, based on the output of the rule-based sieve for T-T pairs (Section~\ref{sec:timex-timex-rules}). In Table~\ref{tab:qa-tempeval-results-trefl} we report the performance of TempRelPro compared with participating systems in QA TempEval, augmented with TREFL, as reported in \textcite{llorens-EtAl:2015:SemEval}. A comparison with off-the-shelf systems not optimized for the task, i.e., CAEVO~\parencite{chambers-etal:2014:TACL}, which is the same system reported in Section~\ref{sec:temp-rel-timebank-dense}, and TIPSemB and TIPSem~\parencite{llorens-saquete-navarro:2010:SemEval}, was also provided. TempRelPro + coref achieves the best performance with 35\% recall and 46\% F1-score.

\section{Conclusions}

Our decision to focus on temporal relation extraction is driven by the low performance of state-of-the-art systems in the TempEval-3 evaluation campaign for this particular task (36.26\% F1-score), compared with system performances for the temporal entity extraction tasks (>80\% F1-score). We have described our approach in building an improved temporal relation extraction system, TempRelPro, which is inspired by a sieve-based architecture for temporal ordering introduced by \textcite{chambers-etal:2014:TACL} with their system, CAEVO. However, our approach is different from CAEVO by adopting simpler architecture, considering all rule-based classifiers as one sieve and all machine-learned classifiers as another one. Hence, we run our temporal reasoner module only once, in between the two sieves we have. Moreover, we also introduced a novel method to include the rule-based sieve output, particularly the labels of timex-DCT and event-DCT links, as features for the supervised event-timex and event-event classifiers.

We have evaluated TempRelPro using the TempEval-3 evaluation scheme, which results in a significant improvement of 40.15\% F1-score, compared to the best performing system in TempEval-3, ClearTK-2, with 36.26\% F1-score. Unfortunately, building and evaluating an automatic temporal relation extraction system is not trivial, given the sparse annotated temporal relations as in TempEval-3. Without a completely labelled graph of temporal entities, we cannot speak of true extraction, but rather of matching human annotation decisions that were constrained by time and effort. This is shown by the low precision achieved by TempRelPro, since it extracts many \texttt{TLINK}s of which the real labels are unknown. Therefore, we also evaluated TempRelPro following the TimeBank-Dense evaluation methodology~\parencite{chambers-etal:2014:TACL} and QA TempEval~\parencite{llorens-saquete-navarro:2010:SemEval}. In general, TempRelPro performs best in both evaluation methodologies.

According to the TempEval-3 and TimeBank-Dense evaluation schemes, component-wise, combining rule-based and machine-learned sieves (RB + ML) results in a slight improvement. However, introducing the temporal reasoner module in between the sieves (RB + TR + ML) is quite beneficial, especially for E-T pairs. 

If we look into each type of temporal entity pairs, TempRelPro still performs poorly with E-E pairs. On TempEval-3 evaluation, TempRelPro performances when labelling T-T, E-D and E-T pairs are already above 70\% F1-scores, but only around 50\% F1-score for E-E pairs. On TimeBank-Dense evaluation, its performance for E-E pairs is still $<$50\% F1-score. Our efforts in improving the performance of the supervised classifier for E-E pairs, i.e., combining it with a rule-based classifier, introducing a temporal reasoner module, or including event-DCT labels as features, do not result in a better outcome.

There are several directions that we look into regarding this issue. The first one is by building a causal relation extraction system, because there is a temporal constraint in causal relations, so that the \textit{causing event} always happens \texttt{BEFORE} the \textit{resulting event}. In the following chapters we will discuss the interaction between these two types of relations, and whether extracting causal relations can help in improving the output of a temporal relation extraction system, especially for pairs of events (Chapter \ref{ch:integrated-system}).

The other direction would be to exploit the lexical semantic information about the event words in building a supervised classifier for E-E pairs, using word embeddings and deep learning techniques (Chapter \ref{ch:deep-learning}). Currently, the only lexical semantic information used by the event-event SVM classifier in TempRelPro is WordNet similarity measure between pairs of words.



\chapter{Annotating Causality between Events}\label{ch:annotating-causality}
\minitoc

In the previous chapter we have described our efforts in building an improved temporal relation extraction system, TempRelPro (Section~\ref{sec:temp-rel-method}). However, according to the TempRelPro evaluations (Section~\ref{sec:temp-rel-evaluation}), TempRelPro still performs poorly in labelling the temporal relation types of \textit{event-event} pairs. One direction to address this issue is to build a causal relation extraction system, since there is a temporal constraint in causality, that the \textit{cause} happens \texttt{BEFORE} the \textit{effect}. The system would then be used to support temporal relation type classification between events. 

Unfortunately, unlike for temporal relations, there is no available corpus yet to be used for building (and evaluating) a causal relation extraction system for event-event pairs. Moreover, while there is a wide consensus in the NLP community over the modelling of temporal relations between events, mainly based on Allen's temporal logic, the question on how to annotate other types of event relations, in particular causal relation, is still open. 

In this chapter, we present some annotation guidelines to capture explicit causality between event-event pairs, partly inspired by the TimeML annotation standard (Section~\ref{sec:timeml-standard}). Based on the guidelines, we manually annotated causality in the TimeBank corpus (Section~\ref{sec:timeml-corpora}) taken from the TempEval-3 evaluation campaign (Section~\ref{sec:tempeval}). We chose this corpus because gold annotated events were already present, between which we could add causal links. Finally, we report some statistics from the resulting causality-annotated corpus, Causal-TimeBank, on the behaviour of causal cues in a text.

\section{Introduction}
\label{sec:annotating-causality-intro}


While there is a wide consensus in the NLP community over the modeling of temporal relations between events, mainly based on Allen's interval algebra \parencite{allen1983}, the question on how to model other types of event relations is still open. In particular, linguistic annotation of causal relations, which have been widely investigated from a philosophical and logical point of view, are still under debate. This leads, in turn, to the lack of a standard benchmark to evaluate causal relation extraction systems, making it difficult to compare systems performances, and to identify the state-of-the-art approach for this particular task.

Although several resources exist in which causality has been annotated, they cover only few aspects of causality and do not model it in a global way, comparable to what has been proposed for temporal relations in TimeML. See for instance the annotation of causal arguments in PropBank \parencite{propbank} and of causal discourse relations in the Penn Discourse Treebank \parencite{PRASAD08.754}.

In Section~\ref{sec:annotating-causality-guidelines}, we propose annotation guidelines for explicit construction of causality inspired by TimeML, trying to take advantage of the clear definition of events, signals and relations proposed by \textcite{pustejovsky2003}. 
This is the first step towards the annotation of a TimeML corpus with causality. 

We annotated TimeBank, a freely available corpus, with the aim of making it available to the research community for further evaluations. Our annotation effort results in Causal-TimeBank, a TimeML corpus annotated with both temporal and causal information (Section~\ref{sec:annotating-causality-causal-timebank}). We chose TimeBank because it already contains gold annotated temporal information, including temporal entities (events and temporal expressions) and temporal relations. The other reason is because we want to investigate the strict connection between temporal and causal relations. In fact, there is a temporal constraint in causality, i.e. the \textit{cause} must occur \texttt{BEFORE} the \textit{effect}. We believe that investigating this precondition on a corpus basis can contribute to improving the performance of temporal and causal relation extraction systems.

\section{Related Work}
\label{sec:annotating-causality-related-work}

Unlike the \textit{temporal order} that has a clear definition, there is no consensus in the NLP community on how to define \textit{causality}. Causality is not a linguistic notion, meaning that although language can be used to express causality, causality exists as a psychological tool for understanding the world independently of language \parencite{neeleman2012}. In the psychology field, several models have been proposed to model causality, including the \textit{counterfactual model} \parencite{lewis1973}, \textit{probabilistic contrast model} \parencite{cheng1991,cheng1992} and the \textit{dynamics model} \parencite{wolff:2003,wolff:2005,wolff:2007}, which is based on Talmy's \textit{force dynamic} account of causality \parencite{talmy1985,talmy:1988}.

Several attempts have been made to annotate causal relations in texts.
A common approach is to look for specific cue phrases like \textit{because} or \textit{since} or to look for verbs that contain a cause as part of their meaning, such as \textit{break} (\textit{cause to be broken}) or \textit{kill} (\textit{cause to die}) \parencite{Khoo:2000:ECK:1075218.1075261,Sakaji2008.PAKM,girju-EtAl:2007:SemEval-2007}. In PropBank \parencite{propbank}, causal relations are annotated in the form of predicate-argument relations, 
where \textsc{argm-cau} is used to annotate `the reason for an action', for example:  
``They \role{\textsc{predicate}}{moved} to London \role{\textsc{argm-cau}}{because of the baby}.''

Another scheme annotates causal relations between discourse arguments, 
in the framework of the Penn Discourse Treebank (PDTB) \parencite{PRASAD08.754}. As opposed to PropBank, this kind of relations holds only between clauses, and does not involve predicates and their arguments. In PDTB, the \emph{Cause} relation type is classified as a subtype of \textit{Contingency}.
 
Causal relations have also been annotated as relations between events in a restricted set of linguistic constructions \parencite{BETHARD08.229},  between clauses in text from novels \parencite{GRIVAZ10.145}, or in noun-noun compounds \parencite{girju-EtAl:2007:SemEval-2007}.

Several types of annotation guidelines for causal relations have been presented, with varying degrees of reliability.
One of the simpler approaches asks annotators to check whether the sentence containing event pairs conjoined by `and' can be paraphrased using a connective phrase such as `and as a result' or `and as a consequence' \parencite{BETHARD08.229}. For example, 
``Fuel tanks had \event{leaked} and \event{contaminated} the soil.''
could be rephrased as
``Fuel tanks had \event{leaked} and as a result \event{contaminated} the soil.''
This approach is relatively simple for annotators, but agreement is only moderate (kappa of 0.556), in part because there are both causal and non-causal readings of such connective phrases. Another approach to annotate causal relations tries to combine linguistic tests with semantic reasoning tests.
In the work of \textcite{GRIVAZ10.145}, the linguistic paraphrasing suggested by \textcite{BETHARD08.229} is augmented with rules that take into account other semantic constraints, for instance if the potential cause occurs before or after the potential effect.

\textcite{do-chan-roth:2011:EMNLP} developed an evaluation corpus by collecting 20 news articles from CNN, allowing the detection of causality between \textit{verb-verb}, \textit{verb-noun}, and \textit{noun-noun} triggered event pairs. The most recent work of \textcite{riaz-girju:2013:SIGDIAL} focuses on the identification of causal relations between verbal events. They rely on the unambiguous discourse markers \textit{because} and \textit{but} to automatically collect training instances of cause and non-cause event pairs, respectively. The result is a knowledge base of causal associations of verbs, which contains three classes of verb pairs: \textit{strongly causal}, \textit{ambiguous} and \textit{strongly non-causal}.


\section{Related Publications}
\label{sec:temp-rel-type-relation-pub}

In \textcite{mirza-EtAl:2014:CAtoCL}, we have presented the annotation guidelines to capture causality between event pairs, inspired by TimeML. Based on the automatic annotation performed, we also reported some statistics on the behavior of causal cues in text and perform
a preliminary investigation on the interaction
between causal and temporal relations.

\section{TimeML-based Causality Annotation Guidelines}
\label{sec:annotating-causality-guidelines}

As part of a wider annotation effort aimed to annotate texts at the semantic level \parencite{NewsReaderTechRep}, within the NewsReader project\footnote{\url{http://www.newsreader-project.eu/}}, we propose guidelines for the annotation of causal information. In particular, we define causal relations between events based on the TimeML definition of events \parencite{PUSTEJOVSKY10.55}, as including all types of actions (punctual and durative) and states.
Syntactically, events can be realized by a wide range of linguistic expressions such as verbs, nouns (which can realize eventualities in different ways, for example through a nominalization process of a verb or by possessing an eventive meaning), adjectives and prepositional constructions. 


Following TimeML, our annotation of causal relations is realized with a \texttt{LINK} tag, i.e., \texttt{CLINK}, parallel with the \texttt{TLINK} tag in TimeML for temporal relations. The annotation of \texttt{CLINK} also includes the \texttt{csignalID} attribute, which refers to the ID of the causal signal, realized with a \texttt{CSIGNAL} tag marking a cue for an explicit causal relation.

\paragraph{}For the sake of clarity, henceforth, snippets of text annotated with \event{events} and \signal{causal markers} (either causative verbs or causal signals) serving as examples will be in the respective forms. For example, ``\event{Tsunami} \signal{caused} an financial \event{crisis}'' or ``The financial crisis \event{deepened} \signal{due to} the \event{tsunami}.''


\paragraph{\texttt{CSIGNAL}} Parallel to the \texttt{SIGNAL} tag in TimeML, which marks a cue for an explicit temporal relations, we introduce the notion of causal signals through the \texttt{CSIGNAL} tag to mark-up textual elements indicating the presence of a causal relations. Such elements include all causal uses of: 
\begin{itemize}
\item prepositions, e.g. \textit{because of, on account of, as a result
of, in response to, due to, from, by};
\item conjunctions, e.g. \textit{because, since, so that, as};
\item adverbial connectors, e.g. \textit{as a result, so, therefore, thus, hence, thereby, consequently}; and
\item clause-integrated expressions, e.g. \textit{the result is, the reason why, that's why}.
\end{itemize}

The extent of \texttt{CSIGNAL}s corresponds to the whole expression, so multi-token extensions are allowed. The only attribute for the \texttt{CSIGNAL} tag is \texttt{id}, corresponding to a unique ID number.

\paragraph{\texttt{CLINK}} The \texttt{CLINK} tag is a directional one-to-one relation where the causing event is the source (the first argument, indicated with \textsubscript{\textsc{s}} in the examples) and the caused event is the \emph{target} (the second argument, indicated with \textsubscript{\textsc{t}}). 

A seminal research in cognitive psychology based on the force dynamics theory \parencite{talmy:1988} has shown that causation covers three main kinds of causal concepts \parencite{wolff:2007}, which are \texttt{CAUSE}, \texttt{ENABLE}, and \texttt{PREVENT}, and that these causal concepts can be lexicalized as verbs \parencite{wolff:2003}: 
\begin{itemize}
	\item \texttt{CAUSE}-type verbs -- \textit{bribe, cause, compel, convince, drive, have, impel, incite, induce, influence, inspire, lead, move, persuade, prompt, push, force, get, make, rouse, send, set, spur, start, stimulate};
	\item \texttt{ENABLE}-type verbs -- \textit{aid, allow, enable, help, leave, let,
permit};
	\item \texttt{PREVENT}-type verbs -- \textit{bar, block, constrain, deter, discourage, dissuade, hamper, hinder, hold, impede, keep, prevent, protect, restrain, restrict, save, stop}. 
\end{itemize}
\texttt{CAUSE}, \texttt{ENABLE}, and \texttt{PREVENT} categories of causation and the corresponding verbs are taken into account in our guidelines. 

As causal relations are often not overtly expressed in text \parencite{wolff:2005}, we restrict the annotation of \texttt{CLINK}s to the presence of an explicit causal construction linking two events in the same sentence\footnote{A typical example of implicit causal construction is represented by lexical causatives; for example, \textit{kill} has the embedded meaning of causing someone to die \parencite{huang:2012}. In the present guidelines, these cases are not included.}, as detailed below: 
\begin{itemize}
\item \textit{Basic constructions} for \texttt{CAUSE}, \texttt{ENABLE} and \texttt{PREVENT} categories of causation as shown in the following examples:\\
The \eventattr{purchase}{s} \signal{caused} the \eventattr{creation}{t} of the current building.\\ 
The \eventattr{purchase}{s} \signal{enabled} the \eventattr{diversification}{t} of their business.\\ 
The \eventattr{purchase}{s} \signal{prevented} a future \eventattr{transfer}{t}.

\item Expressions containing \textit{affect verbs}, such as \textit{affect}, \textit{influence}, \textit{determine}, and \textit{change}. They can be usually rephrased using \textit{cause}, \textit{enable}, or \textit{prevent}:\\ 
Ogun ACN \eventattr{crisis}{s} \signal{affects} the \eventattr{launch}{t} of the All Progressives Congress. $\rightarrow$ Ogun ACN \eventattr{crisis}{s} \signal{causes/enables/prevents} the \eventattr{launch}{t} of the All Progressives Congress.

\item Expressions containing \textit{link verbs}, such as \textit{link (to/with)}, \textit{lead (to)}, and \textit{depend on}. They can usually be replaced only with \textit{cause} and \textit{enable}:\\
An \eventattr{earthquake}{t} in North America was \signal{linked} to a \eventattr{tsunami}{s} in Japan. $\rightarrow$ An \eventattr{earthquake}{t} in North America was \signal{caused/enabled} by a \eventattr{tsunami}{s} in Japan.\\ 
*An \eventattr{earthquake}{t} in North America was \signal{prevented} by a \eventattr{tsunami}{s} in Japan.

\item \textit{Periphrastic causatives} are generally composed of a verb that takes an embedded clause or predicate as a complement; for example, in the sentence ``The \eventattr{blast}{s} \signal{caused} the boat to \eventattr{heel}{t} violently,'' the verb (i.e. \textit{caused}) expresses the notion of \texttt{CAUSE} while the embedded verb (i.e. \textit{heel)} expresses a particular result. Note that the notion of \texttt{CAUSE} can be expressed by verbs belonging to the three categories previously mentioned (which are \texttt{CAUSE}-type verbs, \texttt{ENABLE}-type verbs and \texttt{PREVENT}-type verbs), and the same sets of verbs are taken into consideration.

\item Expressions containing \textit{causative conjunctions and prepositions} as previously listed in the \texttt{CSIGNAL} section. Causative conjunctions and prepositions are annotated as \texttt{CSIGNAL}s and their ID is to be reported in the \texttt{csignalID} attribute of the \texttt{CLINK}.\footnote{The absence of a value for the \texttt{csignalID} attribute means that the causal relation is encoded by a causative verb.}
\end{itemize}

In some contexts, the coordinating conjunction \textit{and}, or the temporal conjunctions \textit{since} and \textit{as}, can also imply causation. We decided to annotate these ambiguous conjunctions, given the causation context, as \texttt{CAUSAL}s. 
However, even though the temporal conjunctions \textit{after} and \textit{when} can also implicitly assert a causal relation, they should not be annotated as \texttt{CSIGNAL}s and no \texttt{CLINK}s are to be created (temporal relations have to be created instead).

The recognition of \texttt{ENABLE}-type causal relations is not always straightforward. The suggestion is to try rephrasing the sentence using the \textit{cause} verb:
\begin{enumerate}[label=(\roman*)]
	\item The board \event{authorized} the \event{purchase} of the stocks.
	\item The \event{authorization} of the board \signal{caused} the stocks to be \event{purchased}.
\end{enumerate}
The verb authorize proves to be an \texttt{ENABLE}-type verb. In sentence (i), a \texttt{CLINK} is established between \textit{authorize} and \textit{purchase}, while in sentence (ii), a \texttt{CLINK} is annotated between \textit{authorization} and \textit{purchased}.

The attributes of the \texttt{CLINK} tag include:
\begin{itemize}
	\item \texttt{id} -- unique ID number.
	\item \texttt{source} -- unique ID of the annotated \texttt{EVENT} involved in the causal link as the causing event.
	\item \texttt{target} -- unique ID of the annotated \texttt{EVENT} involved in the causal link as the resulting event.
	\item \texttt{csignalID} -- (optional) the ID of \texttt{CSIGNAL} explicitly marking the causal link.
\end{itemize}

\paragraph{Example} Consider our previous example of text excerpt annotated with temporal entities and temporal relations in TimeML annotation standard (Figure~\ref{fig:example-timeml}), in Figure~\ref{fig:example-causal-timeml} we present the same text also annotated with causal signals and causal relations following the previously explained annotation guidelines.

\begin{figure}
\noindent\fbox{%
\begin{minipage}{\textwidth}
\texttt{<TimeML>\\
<DOCID>wsj\_0679</DOCID>\\\\
<DCT><TIMEX3 tid="t0" type="DATE" value="1989-10-30" temporalFunction="false" functionInDocument="CREATION\_TIME">1989-10-30</TIMEX3></DCT>\\\\
<TEXT>\\
According to the filing, Hewlett-Packard <EVENT eid="e24" class="OCCURRENCE">acquired</EVENT> 730,070 common shares from Octel \textbf{<CSIGNAL id="cs32">as a result of</CSIGNAL>} an <TIMEX3 tid="t25" type="DATE" value="1988-08-10" functionInDocument="NONE">Aug. 10, 1988</TIMEX3>, stock purchase <EVENT eid="e26" class="I\_ACTION">agreement</EVENT>.
 That <EVENT eid="e27" class="I\_ACTION">accord</EVENT> also <EVENT eid="e28" class="I\_ACTION">called</EVENT> for Hewlett-Packard to <EVENT eid="e29" class="OCCURRENCE">buy</EVENT> 730,070 Octel shares in the open market <SIGNAL sid=s30>within</SIGNAL> <TIMEX3 tid="t31" type="DURATION" value="P18M" functionInDocument="NONE">18 months</TIMEX3>.\\
</TEXT>\\\\
<MAKEINSTANCE eventID="e24" eiid="ei24" tense="PAST" aspect="NONE" polarity="POS" pos="VERB"/>\\
<MAKEINSTANCE eventID="e26" eiid="ei26" tense="NONE" aspect="NONE" polarity="POS" pos="NOUN"/>\\
<MAKEINSTANCE eventID="e27" eiid="ei27" tense="NONE" aspect="NONE" polarity="POS" pos="NOUN"/>\\
<MAKEINSTANCE eventID="e28" eiid="ei28" tense="PAST" aspect="NONE" polarity="POS" pos="VERB"/>\\
<MAKEINSTANCE eventID="e29" eiid="ei29" tense="INFINITIVE" aspect="NONE" polarity="POS" pos="VERB"/>\\\\
<TLINK lid="l21" relType="AFTER" timeID="t31" relatedToTime="t25"/>\\
<TLINK lid="l22" relType="DURING" eventInstanceID="ei29" relatedToTime="t31" signalID="s30"/>\\
<TLINK lid="l23" relType="AFTER" eventInstanceID="ei23" relatedToEventInstance="ei26"/>\\
\textbf{<CLINK id="l24" source="ei26" target="ei24" csignalID="cs32">}\\\\
</TimeML>}

\end{minipage}
}

\caption{Text excerpt annotated with temporal entities and temporal relations in TimeML standard, as well as causal signals and causal relations following our annotation guidelines.}
\label{fig:example-causal-timeml}
\end{figure}

\section{Causal-TimeBank}
\label{sec:annotating-causality-causal-timebank}

Based on the guidelines detailed in Section~\ref{sec:annotating-causality-guidelines}, we manually annotated causality in the TimeBank corpus (Section~\ref{sec:timeml-corpora}) taken from TempEval-3 (Section~\ref{sec:tempeval}), containing 183 documents with 6,811 annotated events in total. We chose this corpus because gold events were already present, between which we could add causal links. Besides, one of our research goals is the analysis of the interaction between temporal and causal information, and TimeBank already presents full manual annotation of temporal information according to the TimeML standard. The resulting corpus, Causal-TimeBank, is made available to the research community for further evaluations.\footnote{\url{http://hlt.fbk.eu/technologies/causal-timebank}} 


However, during the annotation process, we noticed that some events involved in causal relations were not annotated, probably because the corpus was originally built focusing on events involved in temporal relations. Therefore, we annotated also 137 new events, which led to around 56\% increase in the number of annotated \texttt{CLINK}s. 

Annotation was performed using the CAT tool \parencite{cat:2012}, a web-based application with a plugin to import annotated data in TimeML and add information on top of it. The agreement reached by two annotators on a subset of 5 documents is 0.844 Dice's coefficient on \texttt{CSIGNAL}s (micro-average over markables) and 0.73 on \texttt{CLINK}s. 

\subsection{Corpus Statistics}
\label{causal-timebank-statistics}

\begin{table}[h!]
\centering
\begin{adjustbox}{width=1\textwidth}
\begin{tabular}{llccc||llccc}
\hline
 \textbf{Type} & \textbf{Signals} & \texttt{\textbf{CSIGNAL}} & \texttt{\textbf{CLINK}} & \textbf{All} & \textbf{Type} & \textbf{Verbs} & \textbf{Periphr.} & \texttt{\textbf{CLINK}} & \textbf{All}\\ 
\hline
prepositions & as a result of & 7 & 9 & 8 & \texttt{CAUSE} & cause & 2 & 6 & 9\\
 & as factors in &  & 1 & 1 &  & convince & 1 & 1 & 2\\
 & as punishment for & 1 & 1 & 1 &  & enforce & 0 & 1 & 2\\
 & because of & 24 & 23 & 26 &  & \textit{force} & \textit{6} & \textit{2} & \textit{10}\\
 & due mostly to & 1 & 1 & 1 &  & fuel & 0 & 1 & 2\\
 & due to & 1 & 1 & 4 &  & \textit{make} & \textit{22} & \textit{8} & \textit{72}\\
 & in connection with & 1 & 1 & 9 &  & persuade & 2 & 1 & 3\\
 & in exchange for & 2 & 2 & 2 &  & prompt & 1 & 1 & 4\\
 & in punishment for & 1 & 1 & 1 &  & provoke & 0 & 1 & 1\\
 & in response to & 5 & 5 & 5 &  & reignite & 0 & 1 & 1\\
 & in the wake of & 4 & 6 & 5 &  & \textit{send} & \textit{1} & \textit{1} & \textit{11}\\
 & pursuant to & 1 & 1 & 1 &  & spark & 0 & 1 & 2\\
 & responsible for &  & 2 & 4 &  & touch off & 0 & 1 & 1\\
 & thanks in part to & 1 & 1 & 1 &  & trigger & 0 & 1 & 5\\
\cline{6-10}
 & the combined effect of & 1 & 1 & 1 & \texttt{ENABLE} & \textit{allow} & \textit{16} & \textit{4} & \textit{19}\\
 & the main reason for & 1 & 1 & 1 &  & authorize & 1 & 1 & 5\\
 & the result of & 2 & 4 & 2 &  & ensure & 0 & 4 & 5\\
 & \textit{by} & \textit{20} & \textit{21} & \textit{296} &  & guarantee & 0 & 1 & 1\\
 & \textit{from} & \textit{10} & \textit{11} & \textit{360} &  & \textit{help} & \textit{18} & \textit{11} & \textit{25}\\
 & \textit{for} & \textit{1} & \textit{1} & \textit{591} &  & let & 2 & 2 & 2\\
 & \textit{with} & \textit{4} & \textit{3} & \textit{284} &  & permit & 2 & 2 & 3\\
\hline
conjunctions & \textit{and} & \textit{5} & \textit{5} & \textit{1045} & \texttt{PREVENT} & avoid & 1 & 1 & 3\\
 & \textit{as} & \textit{9} & \textit{9} & \textit{205} &  & \textit{block} & \textit{3} & \textit{4} & \textit{10}\\
 & because & 40 & 40 & 43 &  & \textit{keep} & \textit{2} & \textit{3} & \textit{17}\\
 & \textit{since} & \textit{3} & \textit{3} & \textit{31} &  & prevent & 4 & 1 & 4\\
 & so that & 1 & 0 & 3 &  & \textit{protect} & \textit{5} & \textit{3} & \textit{9}\\
\hline
adverbial & as a result & 4 & 5 & 4 & \texttt{AFFECT} & undermine &  & 1 & 4\\
connectors & for a reason & 1 & 1 & 1 & & affect &  & 1 & 8\\ 
\cline{6-10}
 & in exchange & 1 & 1 & 1 & \texttt{LINK} & follow &  & 7 & 24\\
 & consequently & 1 & 1 & 1 &  & lead (to) &  & 5 & 24\\
 & \textit{so} & \textit{9} & \textit{9} &  \textit{39} &  & link (with) &  & 1 & 5\\
 & therefore & 2 & 1 & 4 &  & reflect &  & 17 & 22\\
 & thus & 2 & 2 & 5 &  & related (to) &  & 1 & 13\\
\cline{1-5}
clause- & that 's why & 1 & 1 & 1 &  & rely (on) &  & 2 & 3\\
integrated & that , he said , is why & 1 & 1 & 1 &  & result (from) &  & 3 & \multirow{2}{*}{10}\\
expressions & is a principal reason & 1 & 1 & 1 &  & result (in) &  & 8 & \\
 & the reason (why) & 2 & 3 & 3 &  & stem (from) &  & 5 & 4\\
\hline
 &  & \textbf{171} & \textbf{180} &  &  &  &  & \textbf{115} & \\
\hline
\end{tabular}
\end{adjustbox}

\caption{\label{tab:causal-timebank}Statistics of causal markers found in Causal-TimeBank, including causal signals and causative verbs, and the corresponding numbers of \texttt{CLINK}s.}
\end{table}

In the Causal-TimeBank corpus, the total number of annotated \texttt{CSIGNAL}s is 171 and there are 318 \texttt{CLINK}s, much less than the number of \texttt{TLINK}s---particularly of event-event pairs---found in the corpus, which is 2,519. Besides, not all documents contain causality relations between events. From the total number of documents in TimeBank, only 109 (around 60\%) of them contain explicit causal links and only 87 (around 47\%) of them contain \texttt{CSIGNAL}s.

In Table~\ref{tab:causal-timebank} we report the statistics of causal signals and causative verbs found in the corpus, along with the corresponding numbers of \texttt{CLINK}s associated with them. 

\paragraph{Causal Signals} There are 180 \texttt{CLINK}s explicitly cued by causal signals. Note that only 169 \texttt{CLINK}s are actually annotated with the \texttt{csignalID} attribute referring to the annotated \texttt{CSIGNAL}s, the rest are missed by the annotators. Several \texttt{CLINK}s are annotated based on the presence of causal markers, but the markers are missed to be annotated as \texttt{CSIGNAL}s, i.e., \textit{as factors in}, \textit{responsible for}. Some \texttt{CSIGNAL}s may correspond to several \texttt{CLINK}s as always the case when the events are in coordinating clauses, e.g. ``The company said its shipments \eventattr{declined}{t} \signal{as a result of} a \eventattr{reduction}{s} in inventories by service centers and \eventattr{increasing}{s} competitive pressures in the construction market.''

Several causal markers can be perceived as variations from basic ones, e.g., \textit{\signal{due} mostly \signal{to}}, \textit{\signal{thanks} in part \signal{to}}, \textit{\signal{the} combined \signal{effect of}}, \textit{\signal{the} main \signal{reason for}}, \textit{\signal{that}, he said, \signal{is why}}, or \textit{\signal{is a} principal \signal{reason}}, mostly by an insertion of adjectives or adverbs. These variations need to be taken into account in building an automatic extraction system for causal signals, or causal relations.

Some prepositions (e.g., \textit{by}, \textit{from}, \textit{for}, \textit{with}), conjunctions (e.g., \textit{and}, \textit{as}, \textit{since}) and adverbs (e.g., \textit{so}) are highly ambiguous (indicated with \textit{italic} in Table~\ref{tab:causal-timebank}). They are very abundant in the corpus, but only a few of them that can be regarded as causal signals, depending on the context.     

\paragraph{Causative Verbs} There are 115 \texttt{CLINK}s resulting from the causative verb constructions. Some verbs of \texttt{CAUSE}, \texttt{ENABLE} and \texttt{PREVENT} types do not have to be involved in a periphrastic construction to cue a causal relation, since they already own a strong causation sense, e.g., \textit{cause}, \textit{provoke}, \textit{trigger}, \textit{ensure}, \textit{guarantee}, \textit{avoid}, \textit{prevent}. Some others occur more often in a text, but they must be in a periphrastic construction to carry a causation meaning, e.g., \textit{make}, \textit{send}, \textit{allow}, \textit{help}, \textit{keep}, \textit{protect} (indicated with \textit{italic} in Table~\ref{tab:causal-timebank}). Furthermore, even though these verbs are involved in a periphrastic construction, if their subjects are not events, they cannot be considered as causal markers. 

Most of the verbs classified as \textit{link verbs} need to be followed by a specific prepositions in order to carry a causal meaning, e.g., \textit{lead (to)}, \textit{link (to/with)}, \textit{result (from/in)}, \textit{stem (from)}. The verbs \textit{follow}, as in ``A steep \eventattr{rise}{t} in world oil prices \signal{followed} the Kuwait \eventattr{invasion}{s}'', do not always assert a causal relation, since it could be that only temporal order is realized by this verb. The (transitive) verb \textit{reflect} is highly ambiguous, because it could mean\footnote{Merriam-Webster.com \url{http://www.merriam-webster.com} (4 March 2016)}:
\begin{enumerate}[topsep=0pt,itemsep=-1ex,partopsep=1ex,parsep=1ex,label=(\roman*)]
\item to make manifest or apparent, e.g., ``The third-quarter net income \eventattr{fell}{t} 22 \%, \signal{reflecting} the \eventattr{damages}{s} from Hurricane Hugo.''
\item to bring or cast as a result, e.g., ``The \eventattr{success}{s} of the project \signal{reflected} great \eventattr{credit}{t} on all the staff.''
\item to give back or exhibit as an image, likeness, or outline, e.g., ``The water reflects the color of the sky.''	
\end{enumerate}
which shows that it carries different senses of causal direction in (i) and (ii), depending on its semantic, and carries no causation meaning at all in (iii). 




The rest of around 23 \texttt{CLINK}s, which are not involved in a construction with causal signals or causative verbs, are related to some adjectives or verbs that can be perceived to carry a causal meaning depending on the context. For example, ``The \eventattr{downturn}{t} all across Asia \signal{means} that people are not \eventattr{spending}{s} here'' or ``The Bush administration considers the \eventattr{sanctions}{s} \signal{essential} to \eventattr{keeping}{t} Saddam Hussein under control.''


\section{Conclusions}
\label{sec:annotating-causality-conclusion}

We have presented our guidelines for annotating causality between events (Section~\ref{sec:annotating-causality-guidelines}), strongly inspired by TimeML. In fact, we inherit the concept of events, event relations and signals. 

We manually annotated TimeBank, a freely available corpus, with causality information according to the presented annotation guidelines, with the aim of making it available to the research community for further evaluations, and supporting supervised learning for automatically extracting causal relations between events.

During the annotation process, we realized that some events that are actually involved in causal relations were not annotated, probably because TimeBank was created focusing only on temporal relations. By annotating additional events, we get around 56\% more \texttt{CLINK}s compared with only using original TimeBank's events.

The resulting causality corpus, Causal-TimeBank (Section~\ref{sec:annotating-causality-causal-timebank}), contains 171 \texttt{CSIGNAL} and 318 \texttt{CLINK} annotations, so much less compared with 2,519 \texttt{TLINK}s, particularly of event-event (E-E) pairs, found in the corpus. This shows that causal relations, particularly the explicit ones, appear very rarely in a text. 

From the statistics presented in Section~\ref{causal-timebank-statistics}, we can observe some ambiguous causal markers, either causal signals or causative verbs, which occur abundantly in a text but do not always carry a causation sense. Some causative verbs can be easily disambiguated based on the construction they appear at, i.e., periphrastic construction, or when both subject and object of such verbs are events. 

On the other hand, it is not trivial to disambiguate causal signals such as \textit{from}, \textit{and} or \textit{since}. \textcite{BETHARD08.229} attempted to disambiguate the conjunction \textit{and} by asking the annotators to try paraphrasing it with \textit{and as a result} or \textit{and as a consequence}. This approach is relatively simple for annotators, but the agreement is only moderate (kappa of 0.556), showing that it is a difficult task.

Having an annotated data for causality between events, the next step would be to exploit the corpus to build (and evaluate) an automatic extraction system for causal relations. The statistics obtained regarding the behaviour of causal markers in a text would definitely help us in building our system.

\chapter{Causal Relation Extraction}\label{ch:caus-rel-recognition}
\begin{flushright}
\scriptsize
\textit{I would rather discover a single causal connection than win the throne of Persia.} --- Democritus
\end{flushright}
\minitoc

As explained in Chapter~\ref{ch:temp-rel-type}, one direction to address the low performance of the event-event pair classifier for temporal relations is to build a causal relation extraction system, since there is a temporal constraint in causality, that the \textit{cause} happens \texttt{BEFORE} the \textit{effect}. 

Apart from the efforts to improve the temporal relation extraction system, the recognition of causal relations holding between events in a text is crucial to reconstruct the causal chain of events in a story. This could be exploited in question answering, decision support systems and even predicting future events given a chain of past events.

In this chapter we describe an approach for building an automatic system for identifying causal links between events, making use of the annotated causality corpus Causal-TimeBank, resulted from our annotation effort for causality between events (Chapter~\ref{ch:annotating-causality}). We also use the same corpus to evaluate the performance of the developed causal relation extraction system.

\section{Introduction}

An important part of text understanding arises from
understanding if and how two events are related semantically.
For instance, when given a sentence ``The police
arrested him because he killed someone,'' humans
understand that there are two events, triggered by
the words \textit{arrested} and \textit{killed}, and that there is a causality relationship between these two events.

Besides being an important component of discourse
understanding, automatically identifying causal relations between events is important for various NLP applications such as question answering, decision support systems or predicting future events given a chain of past events. 

In this chapter, we take advantage of the developed causality corpus, Causal-TimeBank (Section~\ref{sec:annotating-causality-causal-timebank}), to build an automatic extraction system for identifying causal links between events in a text (Section~\ref{causal-rel-recog-method}), and to perform an evaluation on the system's performance (Section~\ref{sec:causal-rel-evaluation}). Causal-TimeBank is the TimeBank corpus, containing the annotation of temporal entities and temporal relations, annotated with the formalisation of causality information inspired by TimeML (Section~\ref{sec:annotating-causality-guidelines}). This causality corpus provides annotations of a range of expressions for explicit causality which was never considered before, and hence, gives us a broader view over causal relations between events found in a text.

Since the annotated corpus contains only causal links that are explicitly cued by either causal signals or causal verbs, we focus our effort on extracting causal links that are overtly expressed in the text. The approaches for building the causal relation extraction system is heavily influenced by the hybrid approach in a sieve-based architecture that proved to be beneficial for the temporal relation extraction task (Section~\ref{sec:temp-rel-method}). We combine the rule-base methods presented in \textcite{mirza-EtAl:2014:CAtoCL} with the statistical-based methods presented in \textcite{mirza-tonelli:2014:Coling} in a similar fashion as for temporal relation extraction. Moreover, we tried to address one of the issues explained in  \textcite{mirza-tonelli:2014:Coling} related to the dependency parser errors, by using another parser which has a better coverage for long-range dependencies.


\section{Related Work}

The problem of detecting causality between events is as challenging as recognizing their temporal order, but less analyzed from an NLP perspective. Besides, it has mostly focused on specific types of event pairs and causal expressions in text, and has failed to provide a global account of causal phenomena that can be captured with NLP techniques. \textit{SemEval-2007 Task 4 Classification of Semantic Relations between Nominals} \parencite{girju-EtAl:2007:SemEval-2007} gives access to a corpus containing nominal causal relations among others, as causality is one of the considered semantic relations in the task. 

\textcite{BETHARD08.229} collected 1,000 conjoined event pairs connected by \textit{and} from the Wall Street Journal corpus. The event pairs were annotated manually with both temporal ({\sc before}, {\sc after}, {\sc no-rel}) and causal relations ({\sc cause}, {\sc no-rel}). They use 697 event pairs to train a classification model for causal relations, and use the rest for evaluating the system, which results in 37.4\% F-score. \textcite{rink:2010} perform textual graph classification using the same corpus, and make use of manually annotated temporal relation types as a feature to build a classification model for causal relations between events. This results in 57.9\% F-score, 15\% improvement in performance compared with the system without the additional feature of temporal relations. 

\textcite{do-chan-roth:2011:EMNLP} developed an evaluation corpus by collecting 20 news articles from CNN, allowing the detection of causality between \textit{verb-verb}, \textit{verb-noun}, and \textit{noun-noun} triggered event pairs. Causality between event pairs is measured by taking into account Point-wise Mutual Information (PMI) between the cause and the effect. They also incorporate discourse information, specifically the connective types extracted from the Penn Discourse TreeBank (PDTB), and achieve a performance of 46.9\% F-score. 
\textcite{DBLP:conf/nldb/IttooB11} presented a minimally-supervised algorithm that extracts explicit and implicit causal relations based on syntactic-structure-based causal patterns.

The most recent work of \textcite{riaz-girju:2013:SIGDIAL} focuses on the identification of causal relations between
verbal events. They rely on the unambiguous discourse markers \textit{because} and \textit{but} to automatically collect training instances of cause and non-cause event pairs, respectively. The result is a knowledge base of causal associations of verbs, which contains three classes of verb pairs: \textit{strongly causal}, \textit{ambiguous} and \textit{strongly non-causal}.

\section{Related Publications}

In \textcite{mirza-EtAl:2014:CAtoCL} we presented a simple rule-based system to extract (explicit) event causality from a text. The rule-based system relies on an algorithm that, given a term \textit{t} belonging to \textit{affect}, \textit{link}, \textit{causative verbs} or \textit{causal signals} (as listed in the annotation guidelines presented in Section~\ref{sec:annotating-causality-guidelines}), looks for specific dependency constructions where \textit{t} is connected to the two observed events. If such dependencies are found, a \texttt{CLINK} is automatically set between the two events.

In \textcite{mirza-tonelli:2014:Coling} we presented a  data-driven approach to extract causal relations between events, making use of the Causal-TimeBank corpus (Section~\ref{sec:annotating-causality-causal-timebank}). The system is a pipeline of two classifiers: (i) \texttt{CSIGNAL} labeller and (ii) \texttt{CLINK} classifier.

\section{Formal Task Definition}
\label{sec:causal-rel-recog-formal-task}

We can formulate the task of recognizing event causality in a text as a classification task. Given an ordered candidate event pair $(e_1, e_2)$ the classifier has to decide whether there is a causal relation between them or not. However, since we also consider the direction of the causal link, i.e. identifying \textit{source} and \textit{target}, an event pair $(e_1, e_2)$ is classified into 3 classes: (i) \texttt{CLINK}, where $e_1$ is the source and $e_2$ is the target, meaning $e_1$ \texttt{cause} $e_2$; (ii) \texttt{CLINK-R}, with the reverse order of source and target ($e_2$ and $e_1$, resp.), meaning $e_1$ \texttt{is\_caused\_by} $e_2$; and (iii) \texttt{NO-REL}, for when there is no causal relation.



\paragraph{Example} Consider the following excerpt taken from the TimeBank corpus, annotated with events:

\begin{quote}
DCT=\timexattr{1989-10-30}{t$_0$}

Other market-maker gripes: Program \eventattr{trading}{E$_{11}$} also \eventattr{causes}{E$_{12}$} the Nasdaq Composite Index to \eventattr{lose}{E$_{13}$} ground against other segments of the stock market. Peter DaPuzzo, head of retail equity trading at Shearson Lehman Hutton, \eventattr{acknowledges}{E$_{14}$} that he wasn't \eventattr{troubled}{E$_{15}$} by program \eventattr{trading}{E$_{16}$} when it \eventattr{began}{E$_{17}$} in the pre-crash bull market because it \eventattr{added}{E$_{18}$} liquidity and people were \eventattr{pleased}{E$_{19}$} to \eventattr{see}{E$_{20}$} stock prices \eventattr{rising}{E$_{21}$}.
\end{quote}

The causal relation extraction system should be able to identify: [E$_{11}$ \texttt{CLINK} E$_{13}$], [E$_{15}$ \texttt{CLINK-R} E$_{18}$] and [E$_{15}$ \texttt{CLINK-R} E$_{19}$].

\section{Method}
\label{causal-rel-recog-method}

Following the success of a hybrid approach in a sieve-based architecture for the temporal relation extraction task, we decided to combine in a similar way the rule-based methods in \textcite{mirza-EtAl:2014:CAtoCL} with the statistical methods in \textcite{mirza-tonelli:2014:Coling}, for identifying explicit causal links (\texttt{CLINK}s) in a text. We can define two main problems included in this task:
\begin{enumerate}[label=(\roman*)]
\item Recognizing \texttt{CLINK}s established by \textit{affect}, \textit{link} and \textit{causative verbs} (\texttt{CAUSE}-, \texttt{ENABLE}- and \texttt{PREVENT}-type verbs), hereinafter simply addressed as \textit{causal verbs}; and
\item Recognizing \texttt{CLINK}s marked by \textit{causal signals}.
\end{enumerate}

The causal constructions containing causal verbs are quite straightforward: assuming verb $v$ belongs to such verbs, the first event must be the \textit{subject} of $v$ and the second event must be either the \textit{object} or the \textit{predicative complement} of $v$. Considering that such relations between the events and causal verbs are usually embedded in their dependency paths, we can easily approach problem (i) with a rule-based method.

Meanwhile, some causal signals can be ambiguous, and the dependency paths between causal signals and events can be varied. Moreover, the position of causal signals with respect to the two events is crucial to determine the causal direction, e.g., ``The building \eventattr{collapsed}{t} \signal{because of} the \eventattr{earthquake}{s}'' vs ``\signal{Because of} the \eventattr{earthquake}{s} the building \eventattr{collapsed}{t}''. Using the available Causal-TimeBank corpus, we believe that for problem (ii) a classification model can be learned to discover the regularities in which event pairs, connected by causal signals, are identified as having causal links.

We propose a hybrid approach for causal relation extraction, as illustrated in Figure~\ref{fig:temporal-relation-extraction-system}. Our system, CauseRelPro, is a combination of rule-based and supervised classification modules, in a sieve-based architecture.

\begin{figure}
\centering
\includegraphics[scale=0.9]{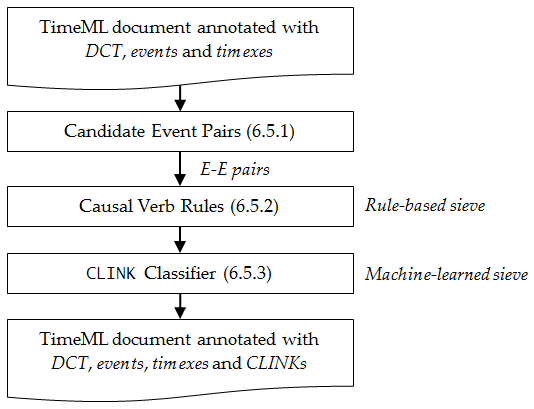}
\caption{Our proposed causal relation extraction system, CauseRelPro}
\label{fig:causal-relation-extraction-system}
\end{figure}

\subsection{Candidate Event Pairs}
\label{sec:causerelpro-candidate-event-pairs}

Given a document already annotated with events, we take into account every possible combination of events in a sentence in a forward manner as \textit{candidate event pairs}. For example, if we have a sentence ``$e_1$, triggered by $e_2$, cause them to $e_3$,'' the candidate event pairs are ($e_1$,$e_2$), ($e_1$,$e_3$) and ($e_2$,$e_3$). We also include as candidate event pairs the combination of each event in a sentence with events in the following one. This is necessary to account for inter-sentential causality, under the simplifying assumption that causality may occur only between events in two consecutive sentences.

\subsection{Causal Verb Rules}
\label{sec:causal-verb-rules}

\paragraph{Causal Verb List} We take lists of affect, link and causative verbs presented in the annotation guidelines (Section~\ref{sec:annotating-causality-guidelines}) as the causal verb list. We further expand the list, which contains 56 verbs in total, using the Paraphrase Database \parencite{ganitkevitch2013ppdb} and original verbs as seeds, resulting in a total of 97 verbs.

Looking at the statistics of causal verbs in Causal-TimeBank (Section~\ref{causal-timebank-statistics}), we can identify ambiguous \texttt{CAUSAL}-, \texttt{ENABLE}- and \texttt{PREVENT}-type verbs, e.g., \textit{make}, \textit{send}, \textit{allow}, \textit{help}, \textit{keep}, \textit{protect}, which must be in a periphrastic construction in order to carry a causation meaning. We make a separate list for such verbs, distinguishing them from \texttt{CAUSAL}-, \texttt{ENABLE}- and \texttt{PREVENT}-type verbs with a strong causation sense, e.g., \textit{cause}, \textit{provoke}, \textit{trigger}, \textit{ensure}, \textit{guarantee}, \textit{avoid}, \textit{prevent}.

Most of the link verbs need to be followed by a specific prepositions in order to carry a causal meaning. Moreover, \textit{result in} and \textit{result from} carry different sense of causal directions, i.e., the causing event is the subject of \textit{result in}, but instead the object of \textit{result from}. Therefore, such verb-preposition combinations are distinct items in the list, and each corresponds to the carried sense of causal direction. The verb \textit{follow} and \textit{reflect} are excluded from the list because they are highly ambiguous.

In the end, our causal verb list (Appendix~\ref{app:causal-verbs-signals}) contains 96 verbs belonging to 8 verb-types, including: \texttt{AFFECT}, \texttt{LINK}, \texttt{CAUSE}, \texttt{CAUSE-AMBIGUOUS}, \texttt{ENABLE}, \texttt{ENABLE-AMBIGUOUS}, \texttt{PREVENT} and \texttt{PREVENT-AMBIGUOUS}. Different from all of the \texttt{AFFECT}, \texttt{CAUSE}, \texttt{ENABLE} and \texttt{PREVENT} verbs that have the normal causal direction ($e_1$ \texttt{CLINK} $e_2$), most of the \texttt{LINK} verbs have the inverse causal direction ($e_1$ \texttt{CLINK-R} $e_2$), such as \textit{link(ed)-with}, \textit{stem-from}, \textit{result-from}, etc.

\paragraph{Candidate Event Pair Filtering} We only consider candidate event pairs in which causal verbs $v$ occur between the two events $(e_1, e_2)$ in the text, e.g., ``The \eventattr{blast}{$e_1$} \signalattr{caused}{$v$} the boat to \eventattr{heel}{$e_2$} violently.''

\begin{table}
\centering
\begin{adjustbox}{width=1\textwidth}
\begin{tabular}{lll}
\hline
\textbf{Relation} & \textbf{Path} & \textbf{Example} \\
\hline
\textbf{between $v$ and $e_1$} & \textbf{\texttt{dep1}} & \\
$e_1$ is subject of $v$ & \texttt{SBJ} & \textit{The Pope's \eventattr{visit}{$e_1$} \signalattr{persuades}{$v$} Cubans...} \\ 
$v$ is predicative complement of $e_1$ & \texttt{PRD-IM} & \textit{The \eventattr{roundup}{$e_1$} was to \signalattr{prevent}{$v$} them...} \\
$v$ is modifier of $e_1$ (nominal) & \texttt{NMOD} & \textit{An \eventattr{agreement}{$e_1$} that \signalattr{permits}{$v$} the Russian...} \\
$v$ is apposition of $e_1$ & \texttt{APPO} & \textit{..., with the \eventattr{crisis}{$e_1$} \signalattr{triggered}{$v$} by...} \\
$v$ is general adverbial of $e_1$ & \texttt{ADV} & \textit{The number \eventattr{increased}{$e_1$}, \signalattr{prompting}{$v$}...} \\
$v$ is adverbial of purpose/reason of $e_1$ & \texttt{PRP-IM} & \textit{The major \eventattr{allocated}{$e_1$} funds to \signalattr{help}{$v$}...} \\
\hline
\textbf{between $v$ and $e_2$} & \textbf{\texttt{dep2}} & \\
$e_2$ is object of $v$ & \texttt{OBJ} & \textit{...have \signalattr{provoked}{$v$} widespread \eventattr{violence}{$e_2$}.} \\
$e_2$ is logical subject of $v$ (passive verb) & \texttt{LGS-PMOD} & \textit{...\signalattr{triggered}{$v$} by the \eventattr{end}{$e_2$} of the...} \\
\multirow{2}{*}{$e_2$ is predicative complement of $v$ (raising/control verb)} & \texttt{OPRD} & \textit{...funds to \signalattr{help}{$v$} \eventattr{build}{$e_2$} a museum.} \\
 & \texttt{OPRD-IM} & \textit{...\signalattr{persuades}{$v$} Cubans to \eventattr{break}{$e_2$} loose.} \\
$e_2$ is general adverbial of $v$ & \texttt{ADV-PMOD} & \textit{...\signalattr{protect}{$v$} them from unspecified \eventattr{threats}{$e_2$}.} \\
$e_2$ is adverbial of direction of $v$ & \texttt{DIR-PMOD} & \textit{...\signalattr{lead to}{$v$} a \eventattr{surge}{$e_2$} of inexpensive imports.} \\
$e_2$ is modifier of $v$ (adjective or adverbial) & \texttt{AMOD-PMOD} & \textit{...\signalattr{related to}{$v$} \eventattr{problems}{$e_2$} under a contract.} \\
\hline
\end{tabular}
\end{adjustbox}
\caption{\label{tab:dependency-relations} Possible dependency relations between $v$ and $e_1$/$e_2$, their corresponding paths according to \textcite{surdeanu-EtAl:2008:CONLL} and examples in texts.}
\end{table}

\paragraph{} The set of rules applied to the filtered candidate event pairs is based on (i) the category of the causal verb $v$, (ii) the possible existing dependency relations between $v$ and $e_1$ (\texttt{dep1}), and between $v$ and $e_2$ (\texttt{dep2}), as listed in Table~\ref{tab:dependency-relations}, and (iii) the causal direction sense ($dir$) embedded in $v$:
\begin{itemize}
\item \textit{If} $v$ is an \texttt{AFFECT} verb:
\begin{itemize}
\item \textit{If} \texttt{dep1} is one of all possible relations \textit{and} \texttt{dep2 = OBJ} \textit{then} [$e_1$ \texttt{CLINK} $e_2$]
\end{itemize}
\item \textit{If} $v$ is a \texttt{LINK} verb:
\begin{itemize}
\item \textit{If} \texttt{dep1} is one of all possible relations \textit{and} \texttt{dep2 = OBJ/ADV-PMOD/DIR-PMOD/AMOD-PMOD} \textit{and} \texttt{dir = CLINK} \textit{then} [$e_1$ \texttt{CLINK} $e_2$]
\item \textit{If} \texttt{dep1} is one of all possible relations \textit{and} \texttt{dep2 = OBJ/ADV-PMOD/DIR-PMOD/AMOD-PMOD} \textit{and} \texttt{dir = CLINK-R} \textit{then} [$e_1$ \texttt{CLINK-R} $e_2$]
\end{itemize}
\item \textit{If} $v$ is a \texttt{CAUSE}, \texttt{ENABLE} \textit{or} \texttt{PREVENT} verb
\begin{itemize}
\item \textit{If} \texttt{dep1} is one of all possible relations \textit{and} \texttt{dep2 = OBJ/OPRD/OPRD-IM/ADV-PMOD} \textit{then} [$e_1$ \texttt{CLINK} $e_2$]
\item \textit{If} \texttt{dep1} is one of all possible relations \textit{and} \texttt{dep2 = LGS-PMOD} \textit{then} [$e_1$ \texttt{CLINK-R} $e_2$]
\end{itemize}
\item \textit{If} $v$ is a \texttt{CAUSE-AMBIGUOUS}, \texttt{ENABLE-AMBIGUOUS} \textit{or} \texttt{PREVENT-AMBIGUOUS} verb
\begin{itemize}
\item \textit{If} \texttt{dep1} is one of all possible relations \textit{and} \texttt{dep2 = OPRD/OPRD-IM/ADV-PMOD} \textit{then} [$e_1$ \texttt{CLINK} $e_2$]
\end{itemize}
\end{itemize}

\subsection{\texttt{CLINK} Classifier}
\label{sec:clink-classifier}

For recognizing (and determining the causal direction of) \texttt{CLINK}s in a text, we built a classification model using LIBLINEAR \parencite{REF08a} L2-regularized L2-loss linear SVM (dual), with default parameters, and one-vs-rest strategy for multi-class classification.

\paragraph{Tools and Resources} The same external tools and resources for building the classifiers for temporal relation extraction (Section~\ref{sec:pair-classifier}) are used to extract features from each event pair, such as PoS tags, shallow phrase chunk, dependency path and WordNet (Lin) semantic similarity/relatedness.

Additionally, we take the list of causal signals from the annotation guidelines (Section~\ref{sec:annotating-causality-guidelines}) as the \textit{causal signal list}. Again we expand the list using the Paraphrase Database \parencite{ganitkevitch2013ppdb}, resulting in a total of 200 signals. We also manually cluster some signals together, e.g. \{\textit{therefore}, \textit{thereby}, \textit{hence}, \textit{consequently}\}, as we did for temporal signals. Note that we exclude \textit{and}, \textit{for} and \textit{with} from the list because they are highly ambiguous. 

For some causal signals, instead of the signal text we put regular expression patterns in the list to cover possible variations of causal signals. For example, 
\begin{quotation}
\texttt{due ([a-z]+\textbackslash\textbackslash s)?to}
\end{quotation}
for \{\textit{`due to'}, \textit{`due mainly to'}, \textit{`due mostly to'}, ...\}; or 
\begin{quotation}
\texttt{th[ai][st] (, ([a-z]+\textbackslash\textbackslash s)+, )*[i']s ([a-z]+\textbackslash\textbackslash s)*why}
\end{quotation} 
for \{\textit{`this is why'}, \textit{`that 's exactly why'}, \textit{`that , he said , is why'}, ...\}.

In the end, our causal signal list (Appendix~\ref{app:causal-verbs-signals}) contains 66 causal signals belonging to 19 clusters. Among the signals in the list, 45 are actually causal signal patterns covering more variety of causal signals.

\paragraph{Candidate Event Pair Filtering} We only take into account a candidate event pair $(e_1, e_2)$ in which:
\begin{itemize}
\item a causal signal $s$ occur in the sentence where $e_1$ and/or $e_2$ take place, and that $s$ is connected, i.e. there exists a dependency path, to either $e_1$ or $e_2$, or both;
\item there exists no dependency relation between $e_1$ and $e_2$ that of type: subject (\texttt{SBJ}), object (\texttt{OBJ}), coordination (\texttt{COORD-CONJ}), verb chain (\texttt{VC}), locative adverbial (\texttt{LOC-PMOD}) or predicative complement of raising/control verb (\texttt{OPRD-(IM)}); and
\item the entity distance between $e_1$ and $e_2$ is less than 5.
\end{itemize} 

\begin{savenotes}
\begin{table}[h!]
\centering
\small
\begin{adjustbox}{width=1\textwidth}
\begin{tabular}{lcl}
\hline
\textbf{Feature} & \textbf{Rep.} & \textbf{Description}\\
\hline
\multicolumn{3}{l}{\textbf{Morphosyntactic information}} \\
\hspace{5pt}PoS & one-hot & Part-of-speech tags of $e_1$ and $e_2$.\\
\hspace{5pt}phraseChunk & one-hot & Shallow phrase chunk of $e_1$ and $e_2$.\\
\hdashline
\hspace{5pt}samePoS & binary & Whether $e_1$ and $e_2$ have the same PoS.\\
\hline
\multicolumn{3}{l}{\textbf{Textual context}} \\
\hspace{5pt}sentenceDistance & binary & \texttt{0} if $e_1$ and $e_2$ are in the same sentence, \texttt{1} otherwise.\\
\hspace{5pt}entityDistance & binary & \texttt{0} if $e_1$ and $e_2$ are adjacent, \texttt{1} otherwise.\\
\hline
\multicolumn{3}{l}{\textbf{\texttt{EVENT} attributes}} \\
\hspace{5pt}class & one-hot & \multirow{4}{*}{\texttt{EVENT} attributes as specified in TimeML.}\\
\hspace{5pt}tense & one-hot & \\
\hspace{5pt}aspect & one-hot & \\
\hspace{5pt}polarity & one-hot & \\
\hdashline
\hspace{5pt}sameClass & binary & \multirow{3}{*}{Whether $e_1$ and $e_2$ have the same \texttt{EVENT} attributes.}\\
\hspace{5pt}sameTenseAspect & binary & \\
\hspace{5pt}samePolarity & binary & \\
\hline
\multicolumn{3}{l}{\textbf{Dependency information}} \\
\hspace{5pt}dependencyPath & one-hot & Dependency path between $e_1$ and $e_2$.\\
\hspace{5pt}isMainVerb & binary & Whether $e_1$/$e_2$ is the main verb of the sentence.\\
\hline
\multicolumn{3}{l}{\textbf{Temporal signals}} \\
\hspace{5pt}tempSignalTokens & one-hot & Tokens (cluster) of temporal signal around $e_1$ and $e_2$.\\
\hspace{5pt}tempSignalPosition & one-hot & Temporal signal position w.r.t $e_1$/$e_2$, e.g., \texttt{BETWEEN}, \texttt{BEFORE}, \texttt{BEGIN}, etc.\\
\hspace{5pt}tempSignalDependency & one-hot & Temporal signal dependency path between signal tokens and $e_1$/$e_2$.\\
\hline
\multicolumn{3}{l}{\textbf{Causal signals}} \\
\hspace{5pt}causSignalTokens & one-hot & Tokens (cluster) of causal signal around $e_1$ and $e_2$.\\
\hspace{5pt}causSignalPosition & one-hot & Causal signal position w.r.t $e_1$/$e_2$, e.g., \texttt{BETWEEN}, \texttt{BEFORE}, \texttt{BEGIN}, etc.\\
\hspace{5pt}causSignalDependency & one-hot & Causal signal dependency path between signal tokens and $e_1$/$e_2$.\\
\hline
\multicolumn{3}{l}{\textbf{Lexical semantic information}} \\
\hspace{5pt}wnSim & one-hot & WordNet similarity computed between the lemmas of $e_1$ and $e_2$.\\
\hline
\end{tabular}
\end{adjustbox}
\caption{\label{tab:feature-set-causal}Feature set for the \texttt{CLINK} classification model, along with each feature representation (Rep.) in the feature vector and feature descriptions.}
\end{table}
\end{savenotes}

\paragraph{Feature Set} The implemented features are listed in Table~\ref{tab:feature-set-causal}. All of the features for the event-event (E-E) classification model for temporal relation extraction (Table~\ref{tab:feature-set}) are re-used, along with additional features related to causal signals. As for temporal relation extraction (Section~\ref{sec:pair-classifier}), we also simplified the possible values of causal signal features during the one-hot encoding:
\begin{itemize}
\item \textit{causSignalTokens} The \textit{clusterID} of signal cluster, e.g., \{\textit{therefore}, \textit{thereby}, \textit{hence}, \textit{consequently}\}, is considered as a feature instead of the signal tokens.
\item \textit{causSignalDependency} For each atomic label in a vector of syntactic dependency labels according to~\textcite{surdeanu-EtAl:2008:CONLL}, if the signal dependency path contains the atomic label, the value in the feature vector is flipped to 1. Hence, \texttt{PRP-PMOD} and \texttt{PMOD-PRP} will have the same one-hot representations.
\end{itemize} 

\section{Evaluation}
\label{sec:causal-rel-evaluation}

We evaluate CauseRelPro using the causality corpus, Causal-TimeBank (Section~\ref{sec:annotating-causality-causal-timebank}), in \textit{stratified 10-fold cross-validation}. The stratified cross-validation scheme is chosen to account for the highly imbalanced dataset as illustrated in Table~\ref{tab:causerelpro-modules}, i.e., from the total of 28,058 candidate event pairs, only 318 are under the \texttt{CLINK/CLINK-R} class while the rest are under the \texttt{NO-REL} class. With this scheme, the proportion of \texttt{CLINK}, \texttt{CLINK-R} and \texttt{NO-REL} instances are approximately the same in all 10 folds.

\begin{table}[t]
\centering
\small
\begin{tabular} {lr|rrr|rr|ccc}
\hline
 & \textbf{\#Cand.} & \textbf{\#\texttt{CLINK}} & \textbf{\#\texttt{CLINK-R}} & \textbf{Total} & \textbf{TP} & \textbf{FP} & \textbf{P} & \textbf{R} & \textbf{F1}\\
\hline
\multicolumn{2}{l|}{Causal Verb Rules} &  &  &  &  &  &  &  & \\
 & 1143 & 73 & 21 & 94 & 50 & 11 & 0.8197 & 0.1572 & 0.2639\\
 \hline
\multicolumn{2}{l|}{\texttt{CLINK} Classifier} &  &  &  &  &  &  &  & \\
\hspace{10pt}without filter & 27997 & 109 & 158 & 267 & 25 & 17 & 0.5952 & 0.0786 & 0.1389\\
\hspace{10pt}with filter & 3262 & 34 & 128 & 162 & 43 & 41 & 0.5119 & 0.1352 & 0.2139\\
\hline
\multicolumn{7}{l|}{\textbf{CauseRelPro: Causal Verb Rules + \texttt{CLINK} Classifier}} &  &  & \\
\hspace{10pt}without filter & 28058 & 151 & 167 & 318 & 75 & 28 & 0.7282 & 0.2358 & 0.3563\\
\hspace{10pt}\textbf{with filter} & 3323 & 76 & 136 & 212 & 93 & 52 & \textbf{0.6414} & \textbf{0.2925} & \textbf{0.4017}\\
\hdashline
\hspace{10pt}\textit{corrected FP} &  &  &  &  & \textit{100} & \textit{45} & \textit{0.6897} & \textit{0.3145} & \textit{\textbf{0.4320}}\\
\hline
\multicolumn{7}{l|}{\textcite{mirza-EtAl:2014:CAtoCL} rule-based system} & 0.3679 & 0.1226 & 0.1840\\
\multicolumn{7}{l|}{\textcite{mirza-tonelli:2014:Coling} data-driven system} & 0.6729 & 0.2264 & 0.3388\\
\hline
\end{tabular}
\caption{\label{tab:causerelpro-modules}CauseRelPro micro-averaged performances per module on causal relation extraction, evaluated with stratified 10-fold cross-validation on the Causal-TimeBank corpus. \#Cand. = number of candidate event pairs, TP = number of true positives and FP = number of false positives.}
\end{table}  

We report in Table~\ref{tab:causerelpro-modules} the micro-averaged performances of each module included in CauseRelPro. Out of the 94 pairs under the \texttt{CLINK}/\texttt{CLINK-R} class in the candidate pairs, the causal verb rules can correctly recognize 50 causal links, summing up into 26.39\% F1-score.

For the CLINK classifier, we compare the performance of the module with and without the \textit{candidate event pair filtering rules} (Section~\ref{sec:clink-classifier}). Without the filtering rules, the proportion of positive (\texttt{CLINK}/\texttt{CLINK-R}) and negative (\texttt{NO-REL}) instances is around 1:103, hence the classifier's very low recall of 7.86\% and F1-score of 13.89\%. If we enforce the filtering rules, the proportion of positive and negative instances is around 1:19, and therefore, the classifier performs much better with 21.39\% F1-score.

We combine the causal verb rules and the \texttt{CLINK} classifier in a sieve-based architecture. All candidate event pairs that are not identified as having causal links (\texttt{CLINK}/\texttt{CLINK-R}) by the causal verb rules, are passed to the \texttt{CLINK} classifier. If we enforce the filtering rules of the \texttt{CLINK} classifier, we basically ignore 106 annotated causal links in the Causal-TimeBank, and only try to recognize the rest of 212 causal links. Out of the 212 considered causal links, our system can correctly identifies 93 event pairs as having causal links.

Finally, CauseRelPro, which is the combination of the causal verb rules and the \texttt{CLINK} classifier (with candidate event pair filtering rules), achieves 40.17\% F1-score in recognizing causality between event pairs in the Causal-TimeBank corpus. We manually evaluated the false positives extracted by the causal verb rules, and we found that 7 out of 11 event pairs are actually having causal links but not annotated. If we consider the corrected numbers of false positives and true positives, CauseRelPro achieves 43.2\% F1-score instead. Some false positive examples are reported below:

\begin{enumerate}[label=(\roman*)]
\item The white house yesterday \event{disclosed} that Kuwait's ousted government has formally \event{asked} the U.S. to \eventattr{enforce}{s} the total trade \event{embargo} the United Nations has \event{imposed} on Iraq, \signal{allowing} the U.S. and other nations to immediately \eventattr{begin}{t} stopping ships carrying Iraqi goods.
\item The Oklahoma City energy and defense concern said it will record a \$7.5 million reserve for its defense group, including a \$4.7 million \eventattr{charge}{s} \signal{related to} \event{problems} under a fixed-price development contract and \$2.8 million overhead \eventattr{costs}{t} that won't be reimbursed.
\end{enumerate}

Sentences (i) and (ii) contains the causal links that are extracted by the causal verb rules, but not annotated (false positives). In sentence (i), despite of the long sentence and many possibilities for the source event, the dependency parser managed to pick the correct source event, and therefore, establish the correct causal link. Unfortunately, this causal link is missed by the annotators. Meanwhile, in sentence (ii), the dependency parser mistakenly identify \event{cost} as the coordinating noun of \event{problems} instead of \event{charge}, resulting in a wrong causal link.

\begin{enumerate}[label=(\roman*)]
\setcounter{enumi}{2}
\item StatesWest Airlines, Phoenix, Ariz., said it \eventattr{withdrew}{t} its offer to \event{acquire} Mesa Airlines \signal{because} the Farmington, N.M., carrier didn't \eventattr{respond}{s} to its offer...
\end{enumerate}

\textcite{mirza-tonelli:2014:Coling} stated that one of the contributing factors for the low performance is the dependency errors from the parser used, i.e. Stanford CoreNLP dependency parser. One of the reported mistakes is exemplified in sentence (iii), where the causal link is established between \event{acquire} and \event{respond}, instead of \event{withdrew} and \event{respond}. Using a different dependency parser, in our case namely the parser of the Mate tools, resolves this problem since the dependency parser correctly connects \event{withdrew} and \event{respond}, given the causal marker \signal{because}.

We also compare in Table~\ref{tab:causerelpro-modules} the performance of CauseRelPro to the performances of \textcite{mirza-EtAl:2014:CAtoCL} rule-based system (18.40\% F1-score) and \textcite{mirza-tonelli:2014:Coling} data-driven system (33.88\% F1-score), even though they are not directly comparable since the two latter systems used 5-fold cross-validation as the evaluation scheme.

\section{Conclusions}

We have presented our approach to building an improved causal relation extraction system, CauseRelPro, which is inspired by the sieve-based architecture for the temporal relation extraction task (Section~\ref{sec:temp-rel-method}), which is proven to bring advantages. CauseRelPro is a combination of the rule-based methods presented in \textcite{mirza-EtAl:2014:CAtoCL} and the statistical-based methods presented in \textcite{mirza-tonelli:2014:Coling}, with some modifications regarding the tools and resources used, including the dependency parser (Stanford CoreNLP vs Mate tools), the lists of causal signals and verbs (augmented with paraphrases from PPDB, and clustered) and the algorithm for the classifier (YamCha SVM vs LIBLINEAR SVM).

We have also evaluated CauseRelPro using the Causal-TimeBank corpus in stratified 10-fold cross-validation, resulting in 40.95\% F1-score, much better than the previous two systems reported in \textcite{mirza-EtAl:2014:CAtoCL} and \textcite{mirza-tonelli:2014:Coling}, even though they are not directly comparable because 5-fold cross-validation was used instead. Furthermore, we manually evaluated the output of the causal verb rules, and found that among the 11 false positives, 7 causal links are actually correct. The wrongly extracted ones are due to the dependency parser errors. However, we found that compared with the previous dependency parser used, i.e. Stanford CoreNLP dependency parser, the dependency parser of Mate tools performs better in connecting the events involved in causal relations.

As has been explained in Chapter~\ref{ch:annotating-causality}, we intentionally added causality annotation on the TimeBank corpus, which is layered with the annotation of temporal entities and temporal relations, because we want to investigate the strict connection between temporal and causal relations. In fact, there is a temporal constraint in causality, i.e. the \textit{cause} must occur \texttt{BEFORE} the \textit{effect}. 

In the following chapter (Chapter \ref{ch:integrated-system}) we will discuss the interaction between these two types of relations. \textcite{bethard-martin:2008:ACLShort, rink:2010} showed that including temporal relation information in detecting causal links results in improved classification performance. We will investigate whether the same will hold for our causal relation extraction system, and whether extracting causal relations can help in improving the output of a temporal relation extraction system, especially for pairs of events. 

\chapter{Integrated System for Temporal and Causal Relations}\label{ch:integrated-system}
\begin{flushright}
\scriptsize
\textit{Post hoc, ergo propter hoc --- After this, therefore, because of this.}
\end{flushright}
\minitoc

Given a resource annotated with temporal entities, i.e. events and temporal expressions, temporal and causal relations (Chapter~\ref{ch:annotating-causality}), a temporal relation extraction system (Chapter~\ref{ch:temp-rel-type}) and a causal relation extraction system (Chapter~\ref{ch:caus-rel-recognition}), our next step is to build an integrated system for extracting both temporal and causal relations between events in texts. We start from the premises that there is a temporal constraint in causality, i.e., the causing event must happen \texttt{BEFORE} the resulting event, and that a system for extracting both temporal and causal relations may benefit from integrating this presumption.

In this chapter we first investigate the interaction between temporal and causal relations in the text, by looking at the constructed Causal-TimeBank corpus (Section~\ref{sec:annotating-causality-causal-timebank}). We also investigate the effects of using each type of relations as features for the supervised classifiers used to extract temporal and causal relations. Next, we propose a way to combine the temporal relation and causal relation extraction systems into one integrated system.  

\section{Introduction}
\label{sec:temp-cause-rel-intro}


We have seen in the previous chapters (Chapter~\ref{ch:auto-event-extraction}, Chapter~\ref{ch:temp-rel-type} and Chapter~\ref{ch:caus-rel-recognition}) the performances of automatic extraction systems for events and event relations, specifically temporal and causal relations. 
Now, given the temporal constraint of causality, we want to investigate the interaction between temporal and causal relations, which is made possible by the corpus annotated with both relations, Causal-TimeBank (Section~\ref{sec:annotating-causality-causal-timebank}). Causal-TimeBank is the result of our causality annotation effort (Chapter~\ref{ch:annotating-causality}) on TimeBank, a corpus widely used by the research community working on temporal information processing. Different from several previous works, we aim at providing a more comprehensive account of how causal relations can be explicitly expressed in a text, and we do not limit our analysis to specific connectives. Our investigation into the interaction between temporal and causal relations between events in a text is reported in Section~\ref{sec:tlink-vs-clink}.

Several works related to temporal and causal relations have shown that temporal information is crucial in identifying causal relations \parencite{bethard-martin:2008:ACLShort, rink:2010, mirza-tonelli:2014:Coling}. In this chapter we will investigate the effects of using temporal relation types as features for the causal relation extraction task, and vice versa, causal links as features for the temporal relation extraction task (Section~\ref{sec:tlink-vs-clink-as-features}). However, we expect that given the sparsity of explicit causal links, the reverse impact may not hold.


Even though the explicit causal information may not be so beneficial for data-driven temporal relation extraction, we want to explore other options for exploiting causal information, namely post-editing rules to correct misclassified temporal relations using the output of the causal relation extraction system. In Section~\ref{sec:temp-cause-rel-method}, we describe our proposed approach for the integrated system for temporal and causal relations. We perform an evaluation of the proposed approach in Section~\ref{sec:temp-cause-rel-eval}.

\section{Related Work}
\label{sec:temp-cause-rel-related-work}

\textcite{GRIVAZ10.145} presented an experiment that elicits the intuitive features or tests of causation that are consciously used in causal reasoning. Temporal order is shown to be one of the features that helped to rule out non-causal occurrences. 

\textcite{BETHARD08.229} collected 1,000 conjoined event pairs connected by \textit{and} from the Wall Street Journal corpus. The event pairs were annotated manually with both temporal (\texttt{BEFORE}, \texttt{AFTER}, \texttt{NO-REL}) and causal relations (\texttt{CAUSAL}, \texttt{NO-REL}). A corpus analysis is reported to show the ties between the temporal and causal annotations, which includes the fact that 32\% of \texttt{CAUSAL} relations in the corpus did not have an underlying \texttt{BEFORE} relations. \textcite{bethard-martin:2008:ACLShort} trained machine learning models using this corpus of parallel temporal and causal relations, achieving 49\% F1-score for temporal relations and 52.4\% F1-score for causal relations. The performance of the causal relation classifier is boosted by exploiting gold-standard temporal labels as features.

\textcite{rink:2010} performed textual graph classification using the same corpus, and make use of manually annotated temporal relation types as a feature to build a classification model for causal relations between events. This results in 57.9\% F-score, 15\% improvement in performance compared with the system without the additional feature of temporal relations.

\section{Related Publications}
\label{sec:temp-cause-rel-related-pub}

In \textcite{mirza-tonelli:2014:Coling} we proposed a data-driven approach for extracting causality between events, using the manually annotated causality corpus, Causal-TimeBank. The evaluation and analysis of the system's performance provides an insight into explicit causality in texts and the connection between temporal and causal relations.

We have submitted a research paper containing the description of our integrated temporal and causal relation extraction system \parencite{mirza-tonelli:2016:ACL}. 

\section{Temporal and Causal Links in Causal-TimeBank}
\label{sec:tlink-vs-clink}

We provide in Table~\ref{tab:clink-tlink} some statistics on the overlaps between causal links and temporal relation types in the Causal-TimeBank corpus (Section~\ref{sec:annotating-causality-causal-timebank}). The \textit{Others} class in the table includes \texttt{SIMULTANEOUS}, \texttt{IS\_INCLUDED}, \texttt{BEGUN\_BY} and \texttt{DURING\_INV} relations. In total, only around 32\% of 318 annotated causal links have an underlying temporal relation. Examples of explicit event causality found in the text are few in comparison with the number of annotated temporal relations, particularly of event-event (E-E) pairs, i.e. 318 \texttt{CLINK}s vs 2,519 (E-E) \texttt{TLINK}s. This means that only around 4\% of the total (E-E) \texttt{TLINK}s overlap with \texttt{CLINK}s. Note that the annotators could not see the temporal links already present in the data, therefore they were not biased by \texttt{TLINK}s when assessing causal links.

\begin{table}[h]
\centering
\begin{adjustbox}{width=1\textwidth}
\begin{tabular}{llrrrrrr|rr}
\hline
 & & \textbf{\texttt{BEFORE}} & \textbf{\texttt{AFTER}} & \textbf{\texttt{IBEFORE}} & \textbf{\texttt{IAFTER}} & \textbf{Others} & \textbf{\#Overlap} & \textbf{\#\texttt{CLINK}s} & \textbf{\#(E-E) \texttt{TLINK}s}\\
\hline
Causal-TimeBank & \texttt{CLINK} & 15 & 5 & 0 & 0 & 4 & 24 & \multirow{2}{*}{318} & \multirow{2}{*}{2,519} \\
 & \texttt{CLINK-R} & 1 & 67 & 0 & 3 & 8 & 79\\
\hdashline
Causal-TimeBank + TR & \texttt{CLINK} & 26 & 6 & 0 & 0 & 5 & 37 & \multirow{2}{*}{318} & \multirow{2}{*}{10,226} \\
 & \texttt{CLINK-R} & 3 & 74 & 0 & 3 & 9 & 89\\
\hline
\end{tabular}
\end{adjustbox}
\caption{\label{tab:clink-tlink} Statistics of \texttt{CLINK}s overlapping with \texttt{TLINK}s. TR = temporal reasoner.}
\end{table}

We also run the temporal reasoner module (Section~\ref{sec:temporal-reasoner}) on the Causal-TimeBank corpus to enrich the temporal relation annotation. The number of (E-E) \texttt{TLINK}s greatly increases, around 4 times of the original number. However, the number of causal links overlapping with temporal relation annotation does not improve significantly, i.e. from 32.39\% to 39.62\%.

The data confirm our intuition that temporal information is a strong constraint in causal relations, with the \texttt{BEFORE} class having the most overlaps with \texttt{CLINK} and \texttt{AFTER} with \texttt{CLINK-R}. We found that the few cases where \texttt{CLINK}s overlap with \texttt{AFTER} relation are not due to annotation mistakes, as in the example ``But some analysts \eventattr{questioned}{t} how much of an impact the retirement package will have, \signal{because} few jobs will \eventattr{end}{s} up being eliminated.'' This shows that the concept of causality is more abstract than the concept of temporal order of events in a text. Here, the causing event is not the future event `jobs will \event{end} up being eliminated', but rather the \textit{knowledge} that `jobs will \event{end} up being eliminated'. Annotating causality between events in a text is indeed a much more challenging task.




\section{Temporal and Causal Links as Features}
\label{sec:tlink-vs-clink-as-features}

In the following sections we will investigate the effects of using temporal relation types as features for the causal relation extraction task, and vice versa, causal links as features for the temporal relation extraction task.

\subsection{\texttt{TLINK}s for Causal Relation Extraction}
\label{sec:temp-cause-rel-tlink}

We take the causal relation extraction system described in Section~\ref{causal-rel-recog-method}, CauseRelPro, for the experiment. The feature set for the \texttt{CLINK} classifier (Section~\ref{sec:clink-classifier}) is augmented with temporal order information of the event pairs, which is taken from the gold annotated temporal relations. Specifically, for an event pair $(e_1, e_2)$, if there exists a temporal link connecting $e_1$ and $e_2$, the \texttt{TLINK} type (e.g. \texttt{BEFORE}, \texttt{AFTER}, \texttt{SIMULTANEOUS}, etc.) is added to the feature set, \texttt{NONE} is added otherwise. Note that we use the enriched temporal relation annotation using the temporal reasoner module, i.e., the \textit{Causal-TimeBank + TR} corpus in Table~\ref{tab:clink-tlink}, since it provides more information about the underlying temporal relations.

\begin{table}[t]
\centering
\small
\begin{tabular} {lrrr|ccc}
\hline
 & \textbf{Total} & \textbf{TP} & \textbf{FP} & \textbf{P} & \textbf{R} & \textbf{F1}\\
\hline
CauseRelPro & 318 & 100 & 45 & 0.6897 & 0.2925 & 0.4017\\
\textbf{CauseRelPro + \texttt{TLINK}} & \textbf{318} & \textbf{109} & \textbf{48} & \textbf{0.6943} & \textbf{0.3428} & \textbf{0.4589}\\
\hline
\end{tabular}
\caption{\label{tab:causerelpro-tlink}CauseRelPro micro-averaged performances on causal relation extraction, evaluated with stratified 10-fold cross-validation on the Causal-TimeBank corpus. + \texttt{TLINK} = with \texttt{TLINK} types as features, TP = number of true positives and FP = number of false positives.}
\end{table}  

The experiment is done following the evaluation methodology explained in Section~\ref{sec:causal-rel-evaluation}, which is a stratified 10-fold cross-validation on the Causal-TimeBank corpus. The results are reported in Table~\ref{tab:causerelpro-tlink}, with a quite significant improvement---particularly in recall---if \texttt{TLINK} types are included in the feature set, resulting in 45.89\% F1-score. The reported results are based on the corrected number of false positives, since we found that 7 out of 11 event pairs are actually having causal links but not annotated (Section~\ref{sec:causal-rel-evaluation}).

The significant improvement by adding \texttt{TLINK} types as features supports the previous finding by the related works, that temporal information can boost the performance in recognizing causality between events in a text.

\subsection{\texttt{CLINK}s for Temporal Relation Extraction}
\label{sec:temp-cause-rel-clink}

For the experiment, we take the supervised classification model for event-event (E-E) pairs included in TempRelPro described in Section~\ref{sec:temp-rel-method}. The feature set for the E-E classifier (Section~\ref{sec:pair-classifier}) is augmented with causality information of the event pairs, which is taken from the gold annotated \texttt{CLINK}s in the Causal-TimeBank corpus. Specifically, for an event pair $(e_1, e_2)$, if there exists a causal link connecting $e_1$ and $e_2$, the causal direction (i.e. \texttt{CLINK} or \texttt{CLINK-R}) is added to the feature set, \texttt{NONE} is added otherwise.

\begin{table}[t]
\centering
\small
\begin{tabular} {lrr|c}
\hline
 & \textbf{Total} & \textbf{TP} & \textbf{P/R/F1}\\
\hline
E-E Classifier & 2,519 & 1160 & 0.4605\\
\textbf{E-E Classifier + \texttt{CLINK}} & \textbf{2,519} & \textbf{1159} & \textbf{0.4601}\\
\hline
\end{tabular}
\caption{\label{tab:temprelpro-clink}TempRelPro E-E Classifier micro-averaged performances on temporal relation type classification, evaluated with stratified 10-fold cross-validation on the Causal-TimeBank corpus. + \texttt{CLINK} = with \texttt{CLINK} directions as features and TP = number of true positives.}
\end{table} 

We evaluate the E-E classifier on the task of classifying the temporal relation types, with stratified 10-fold cross-validation on the Causal-TimeBank corpus, under the assumption that the temporal links (\texttt{TLINK}s) are already established between events. From the total of 2,519 event-event (E-E) pairs in the corpus, the E-E classifier augmented with \texttt{CLINK} information as features yields one less true positives than the classifier without \texttt{CLINK}. Note that since we classify all \texttt{TLINK}s---of event-event pairs---in the corpus, the precision and recall are the same. This result confirms our intuition, that while temporal order information can benefit the performance of causal relation extraction, the converse does not hold because the explicit causal links found in texts are very sparse.
 
\section{Integrated System - CATENA}
\label{sec:temp-cause-rel-method}

\begin{figure}
\centering
\includegraphics[scale=0.7]{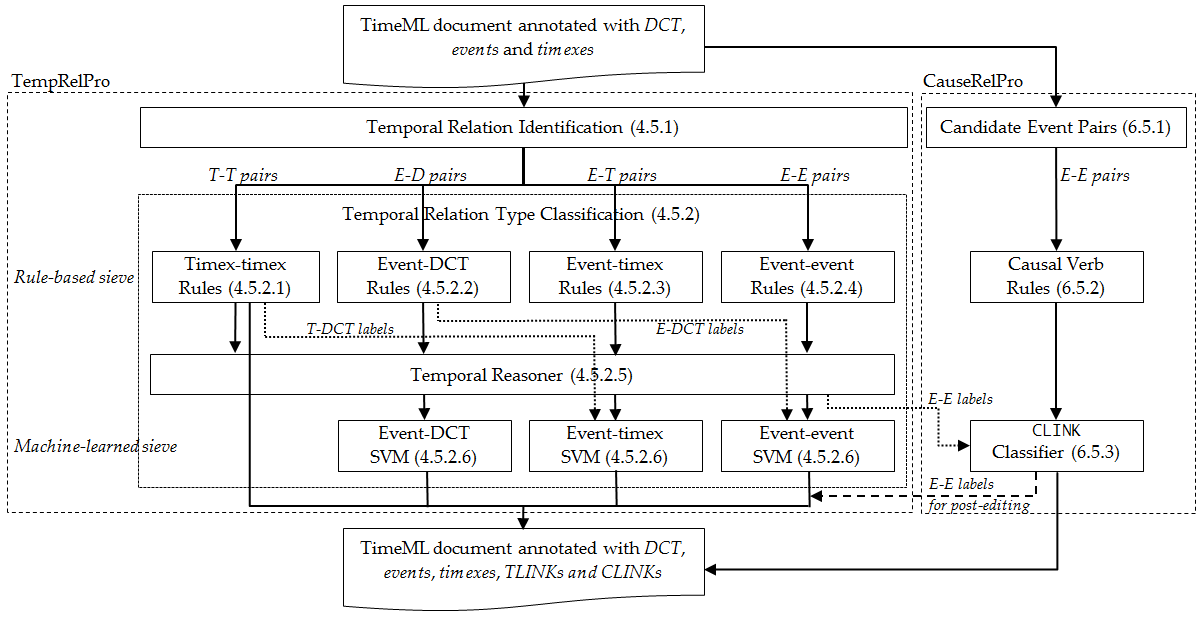}
\caption{CATENA, CAusal and Temporal relation Extraction from NAtural language texts}
\label{fig:catena}
\end{figure}

CATENA---CAusal and Temporal relation Extraction from NAtural language texts---is an integrated system for extracting temporal and causal relations from texts. It includes two main extraction modules, TempRelPro (Section~\ref{sec:temp-rel-method}) and CauseRelPro (Section~\ref{causal-rel-recog-method}), for temporal and causal relation extraction, respectively. As shown in Figure~\ref{fig:catena}, both modules take as input a document annotated with document creation time (DCT), events and temporal expressions (so-called temporal entities) according to TimeML. The output is the same document with temporal and causal links set between pairs of temporal entities.

The modules for temporal and causal relation extraction rely both on a sieve-based architecture, in which the output of a rule-based component is fed into a supervised classifier. Although some steps can be run in parallel, the two modules interact, based on the assumption that the notion of causality is tightly connected with the temporal dimension and that information from one module can be used to improve or check the consistency of the other. In particular, (i) \texttt{TLINK} labels (e.g. \texttt{BEFORE}, \texttt{AFTER}, \texttt{SIMULTANEOUS}, etc.) for event-event (E-E) pairs, as a result of rule-based sieve + temporal reasoner modules in TempRelPro, are used as features for the \texttt{CLINK} classifier in CauseRelPro; and (ii) \texttt{CLINK} labels (i.e. \texttt{CLINK} and \texttt{CLINK-R}) are used as a post-editing method for correcting the wrong labelled event pairs by the E-E classifier in TempRelPro. The post-editing method relies on a set of rules based on the temporal constraint of causality, that the \textit{cause} must precede the \textit{effect}:
\begin{itemize}
\item \textit{If} an event pair $(e_1, e_2)$ is found to have a causal link with normal direction (\texttt{CLINK}), \textit{then} [$e_1$ \texttt{BEFORE} $e_2$] 
\item \textit{If} an event pair $(e_1, e_2)$ is found to have a causal link with reverse direction (\texttt{CLINK-R}), \textit{then} [$e_1$ \texttt{AFTER} $e_2$] 
\end{itemize} 

\section{Evaluation}
\label{sec:temp-cause-rel-eval}

The purpose of the evaluation is twofold: (i) to evaluate the quality of individual modules for temporal and causal relations; and (ii) to investigate whether the post-editing method improves the performance of the temporal relation extraction system. 
We perform the same TempEval-3 Evaluation previously presented in Section~\ref{sec:temp-rel-tempeval3} for evaluating temporal relation extraction.


\paragraph{Dataset} For TempRelPro, we use the same training data released by the organizers, i.e. the TBAQ-Cleaned corpus. Meanwhile, Causal-TimeBank is used as the training data for CauseRelPro. The evaluation data is the TempEval-3-platinum corpus used as the evaluation corpus in TempEval-3. We manually annotated the 20 evaluation documents with causal links following the annotation guidelines of Causal-TimeBank, described in Chapter~\ref{ch:annotating-causality}. Causal relations are much sparser than temporal ones, and we found only 26 \textsc{clink}s. Note that we only annotated causal links where events involved in causality are also overtly annotated in the corpus, even though we found several cases where there is causality but involved events are not annotated, as what was found during the annotation of Causal-TimeBank.  

\paragraph{Evaluation Results} In order to measure the contribution of each component to the overall performance of the system, we evaluate the performance of each sieve both in the temporal and in the causal module. Results are reported in Table \ref{tab:temporal-per-sieve}. For temporal relation extraction, we have presented the evaluation in Section~\ref{sec:temp-rel-tempeval3}. 

For causal relation extraction, the combination of rule-based and machine-learned sieves (RB + ML) achieves 0.62 F1-score in TempEval-3 evaluation, with the ML component contributing to increase the recall of the highly precise RB component. 

\begin{table}[t]
\centering
\small
\begin{tabular} {lccc}
\hline
Sieve & P & R & F1\\
 \hline
 \multicolumn{4}{l}{\textbf{Temporal Relation Identification}} \\
 & 0.53 & 0.95 & 0.68 \\
\hline
\multicolumn{4}{l}{\textbf{Temporal Relation Type Classification}} \\
RB & 0.91 & 0.13 & 0.22 \\
ML & 0.61 & 0.58 & 0.59 \\
\hdashline
RB + TR  & 0.92 & 0.16 & 0.28 \\
RB + ML & 0.62 & 0.59 & 0.61 \\
\hdashline
\textbf{RB + TR + ML} & \textbf{0.62} & \textbf{0.62} & \textbf{0.62} \\
\hline
\multicolumn{4}{l}{\textbf{Causal Relation Extraction}} \\
RB & 0.92 & 0.42 & 0.58 \\
ML & 0.43 & 0.16 & 0.18 \\
\hdashline
\textbf{RB + ML}  & \textbf{0.74} & \textbf{0.54} & \textbf{0.62} \\
\hspace{10pt}-\texttt{TLINK} feature & 0.75 & 0.46 & 0.57\\
\hline
\end{tabular}
\caption{\label{tab:temporal-per-sieve}CATENA performances per sieve and sieve combination. RB: rule-based sieve, ML: machine-learned sieve and TR: temporal reasoner.}
\end{table}

\begin{enumerate}[label=(\roman*)]
\item ``An Israeli \eventattr{raid}{s} on the ship \signal{left} nine passengers \eventattr{dead}{t}, ...''
\item ``The FAA on Friday announced it will \eventattr{close}{t} 149 regional airport control towers \signal{because of} forced spending \eventattr{cuts}{s} ...''
\item ``The \eventattr{incident}{s} \signal{provoked} an international \eventattr{outcry}{t} and \signal{led to} a major \eventattr{deterioration}{t} in relations between Turkey and Israel.''
\item ``But the \eventattr{tie-up}{s} with Rosneft will \eventattr{keep}{s} BP in Russia, \signal{allowing} it to \eventattr{continue}{t} to explore and exploit ...''
\end{enumerate}

The system can identify causal links marked by both causal verbs (i) and causal signals (ii). The dependency relations allow the system to recognize causality in coordinating clauses (iii). Moreover, the dependency paths also enable the extraction of a causal chain between \event{tie-up}, \event{keep} and \event{continue} (iv). Unfortunately, the dependencies are also the cause of mistakenly identified causal links, as in ``The FAA on Friday announced it will \eventattr{close}{t} 149 regional airport control towers \signal{because of} forced spending cuts -- sparing 40 others that the FAA had been \eventattr{expected}{s} to shutter.'' Other mistakes are related to the ambiguous causal signals that the classifier failed to disambiguate, such as \textit{from} in ``But with 20 female senators now in \eventattr{office}{t} women have morphed \signal{from} the \eventattr{curiosity}{s} they were for much of the 20th century.''




As shown in Figure~\ref{fig:catena}, E-E labels returned by the temporal reasoner are used by the \texttt{CLINK} classifier as features, whose causal relations are then used to post-edit \texttt{TLINK} labels. 

We evaluate the impact of the first step through an ablation test, by removing \texttt{TLINK} types from the features used by the \texttt{TLINK} classifier. 
Without \texttt{TLINK} types, the F1-score drops from 0.62 to 0.57, with a significant recall drop from 0.54 to 0.46. This shows that temporal information is beneficial to the classification of causal relations between events, especially in terms of recall. 

As for the evaluation of \texttt{TLINK} post-editing using \texttt{TLINK}s, the system identifies in the test set 19 causal links, 4 of which are passed to the temporal module (the others are already consistent with \texttt{BEFORE/AFTER} labels). 
We manually evaluated the 4 links: 3 of them would add new correct \texttt{TLINK}s that are currently not annotated in the evaluation corpus. The fourth would add a \texttt{BEFORE} label between \event{cloaked} and \event{coughing} in ``A haze akin to volcanic fumes \eventattr{cloaked}{s} the capital, \signal{causing} convulsive \eventattr{coughing}{t} ...''. This relation is labelled as \texttt{INCLUDES} in the gold standard, but we believe that \texttt{BEFORE} would be correct as well.

\section{Conclusions}

Following the analysis of the connection between temporal and (explicit) causal relations in a text, we have presented our proposed approach for integrating our temporal and causal relation extraction systems. The integrated system, CATENA, is a combination of previously explained TempRelPro (Section~\ref{sec:temp-rel-method}) and CauseRelPro (Section~\ref{causal-rel-recog-method}), which takes an input document annotated with temporal entities, i.e. event and timexes, and produces the same document augmented with temporal and causal links.

The integrated system exploits the presumption about event precedence when two events are connected by causality, that the causing event must happen \texttt{BEFORE} the resulting event. With this presumption, the interaction between TempRelPro and CauseRelPro is manifested by (i) using the \texttt{TLINK} labels---resulted from the rule-based sieve + temporal reasoner modules of TempRelPro---as features for the \texttt{CLINK} classifier and (ii) using the output of the \texttt{CLINK} classifier as a post-editing method for correcting wrong temporal labels of event-event pairs.

From the evaluation, we found that the causal relation extraction module achieves 62\% F1-score, correctly identified 14 out of the 26 gold annnotated \texttt{CLINK}s in the evaluation corpus. The mistakes are mostly due to dependency parsing mistakes and issues in disambiguating signals such as \textit{from}. \textcite{BETHARD08.229} attempted to disambiguate the conjunction \textit{and} by asking the annotators to try paraphrasing it with \textit{and as a result} or \textit{and as a consequence}. This approach could be adopted for other ambiguous causal signals such as \textit{from}, \textit{by}, \textit{as} and \textit{since}.

We also found that the post-editing rules would improve the output of temporal relation labelling, but this phenomenon is not captured by the TempEval-3 evaluation, due to the sparse \texttt{TLINK} annotation in the evaluation corpus.  

Nevertheless, explicit causality found in a text is very sparse, and hence, cannot contribute much in improving the performance of the temporal relation extraction system. Extracting implicit causality from texts and investigating the concept of causal-transitivity are two directions that can be explored, so that the amount of causality information will be significant enough to contribute more in boosting the \texttt{TLINK} labeller's performance. 

\chapter{Word Embeddings for Temporal Relations}\label{ch:deep-learning}
\begin{flushright}
\scriptsize
\textit{You shall know a word by the company it keeps.} --- \textcite{firth57synopsis}
\end{flushright}
\minitoc

It has been discussed in Chapter~\ref{ch:temp-rel-type} that our improved temporal relation extraction system, TempRelPro (Section~\ref{sec:temp-rel-method}), performs poorly in labelling the temporal relation types of event-event pairs. While morpho-syntactic and context features are sufficient for classifying timex-timex and event-timex pairs, we believe that exploiting the lexical semantic information about the event words can benefit the supervised classifier for event-event pairs. Currently, the only lexical semantic information used by the event-event SVM classifier in TempRelPro is WordNet similarity measure between the pair of words.

In this chapter we explore the possibilities of using word embeddings as lexical semantic features of event words for temporal relation type classification between event pairs.\footnote{Joint work with Ilija Ilievski, Prof. Min-Yen Kan and Prof. Hwee Tou Ng, National University of Singapore (NUS).}

\section{Introduction}

The identification of discourse relations between events, namely temporal and causal relations, is relatively straightforward when there is an explicit marker (e.g. \textit{before}, \textit{because}) connecting the two events.	The tense, aspect and modality of event words, as well as specific syntactic constructions, could also play a big role in determining the temporal order of events. It becomes more challenging when such an overt indicator is lacking, which is often the case when two events take place in different sentences, as exemplified in (i), where the label for the event pair $(e_1,e_2)$ is \texttt{BEFORE}. This type of event pairs can not be disregarded since for example, in the TempEval-3 evaluation corpus, 32.76\% of the event pairs do not occur in the same sentences.

\begin{enumerate}[label=(\roman*)]
\item \textit{When Wong Kwan \eventattr{spent}{$e_1$} seventy million dollars for this house, he thought it was a great deal. He \eventattr{sold}{$e_2$} the property to five buyers and said he'd double his money.}
\end{enumerate}

Most research on implicit relations, tracing back to \textcite{marcu-echihabi:2002:ACL}, incorporate word-based information in the form of word pair features. Such word pairs are often encoded in a one-hot representation, in which each possible word pair corresponds to a single component of a very high-dimensional vector. From a machine learning point of view, this type of sparse representation makes parameter estimation extremely difficult and prone to over-fitting. It is also very challenging to achieve any interesting semantic generalization with this representation. Consider for example, (\textit{attack}, \textit{injured}) that would be at equal distance from a synonymic pair (\textit{raid}, \textit{hurt}) and an antonymic pair (\textit{died}, \textit{shooting}). 

Recently there has been an increasing interest in using word embeddings as an alternative to traditional hand-crafted features. Word embeddings represent (embed) the semantic of a word in a continuous vector space, where semantically similar words are mapped to nearby points. The underlying principle is the \textit{Distributional Hypothesis} \parencite{harris54}, which states that words which are similar in meaning occur in similar contexts. \textcite{baroni-dinu-kruszewski:2014:P14-1} draws a distinction of approaches leveraging this principle into two categories: (i) \textit{count-based} models and (ii) \textit{predictive} models. \textit{GloVe} \parencite{pennington2014glove} and \textit{Word2Vec} \parencite{journals/corr/abs-1301-3781} are the two popular word embedding algorithms recently, each represents (i) and (ii), respectively. We have briefly explained the two algorithms in Section~\ref{sec:background-embeddings}.

\section{Related Work}

Most works on implicit discourse relations focused on the Penn Discourse Treebank (PDTB) \parencite{PRASAD08.754}, in which relations are annotated at the discourse level. There are five distinct groups of relations: \textit{implicit}, \textit{explicit}, \textit{alternative lexicalizations}, \textit{entity relations} and \textit{no relation}; each could carry multiple relation types that are organized into a three-level hierarchy. The top level relations, for example, includes \textit{Temporal}, \textit{Contingency}, \textit{Comparison} and \textit{Expansion}. \textcite{braud-denis:2015:EMNLP} presented a detailed comparative studies for assessing the benefit of unsupervised word representations, i.e. one-hot word pair representations against low-dimensional ones based on Brown cluster and word embeddings, for identifying implicit discourse relations in PDTB. 


\textcite{baroni-dinu-kruszewski:2014:P14-1} provides a systematic comparison between count-based and predictive word embeddings, on a wide range of lexical semantic tasks, including semantic relatedness, synonym detection, concept categorization, selectional preferences and analogy. The main takeaway is that the predictive models are shown to perform better than count-based ones.

\section{Experiments}

Our objective is to assess the usefulness of different vector representations of word pairs for temporal relation type classification of event-event pairs. Specifically, we want to establish (i) whether predictive models are better than count-based ones for this particular task, (ii) which vector combination schemes are more suitable for the task, and finally, (iii) whether traditional features are still relevant in the presence of dense representations of word pairs. 

The problem of temporal relation type classification can be formally defined as:
\begin{quote}
Given an ordered pair of events $(e_1, e_2)$, assign a certain label to the pair, which can be one of the 14 TimeML temporal relation (\texttt{TLINK}) types: \texttt{BEFORE}, \texttt{AFTER}, \texttt{INCLUDES}, \texttt{IS\_INCLUDED}, \texttt{DURING}, \texttt{DURING\_INV}, \texttt{SIMULTANEOUS}, \texttt{IAFTER}, \texttt{IBEFORE}, \texttt{IDENTITY}, \texttt{BEGINS}, \texttt{ENDS}, \texttt{BEGUN\_BY} or \texttt{ENDED\_BY}.
\end{quote}
However, we collapse some labels as explained in Section~\ref{sec:pair-classifier}, i.e., \texttt{IBEFORE} into \texttt{BEFORE}, \texttt{IAFTER} into \texttt{AFTER}, \texttt{DURING} and \texttt{DURING\_INV} into \texttt{SIMULTANEOUS}, considering the sparse annotation of such labels in the datasets, leaving 10 \texttt{TLINK} types as possible labels.

\paragraph{Pre-trained Word Vectors} We take pre-trained word vectors from GloVe\footnote{\url{http://nlp.stanford.edu/data/glove.6B.zip}} and Word2Vec\footnote{\url{http://drive.google.com/file/d/0B7XkCwpI5KDYNlNUTTlSS21pQmM/}} representing count-based and predictive word embeddings, respectively. The GloVe embeddings are 300-dimensional vectors trained upon 6 billion tokens of Wikipedia articles (2014) and English Gigaword (5$^{\textrm{th}}$ edition) with 400,000 uncased vocabularies. The Word2Vec embeddings are 300-dimensional vectors for 3 million words and phrases trained on part of Google News dataset (about 100 billion words). 

Given an event pair $(e_1, e_2)$, we retrieve the pair of word vectors $(\vec{w}_1, \vec{w}_2)$ based on vector look-up for the head words of $e_1$ and $e_2$ in the pre-trained word vectors.

\paragraph{Vector combinations} Given a pair of word vectors $(\vec{w}_1, \vec{w}_2)$, we consider (i) \textit{concatenation} $(\vec{w}_1 \oplus \vec{w}_2)$, (ii) \textit{addition} $(\vec{w}_1 + \vec{w}_2)$ and (iii) \textit{subtraction} $(\vec{w}_1 - \vec{w}_2)$, as vector combination schemes. Note that in (i) the word ordering information is retained, which is not the case in (ii) and (iii).

\paragraph{Traditional features} Considered as the \textit{traditional features} are the same features described in Section~\ref{sec:pair-classifier} for event-event (E-E) pairs. 

\paragraph{Classifier} We use LIBLINEAR \parencite{REF08a} L2-regularized logistic regression (dual), with default parameters, as the classifier. We have run the same set of experiments using LIBLINEAR L2-regularized L2-loss linear SVM (dual), and found that logistic regression yields better results than SVM.

\paragraph{Experimental setup} Experiments were run using the \textit{TBAQ-cleaned} corpus (Section~\ref{sec:annotating-causality-causal-timebank}), in \textit{stratified 10-fold cross-validation}. TBAQ-cleaned is the training corpus released in the context of TempEval-3 challenge, which includes (i) the AQUAINT TimeML corpus, containing 73 news report documents, and (ii) the TimeBank corpus, with 183 news articles. The stratified cross-validation scheme is chosen to account for the imbalanced distribution of relation types as illustrated in Table~\ref{tab:tbaq-dataset}, e.g., the event pairs under the majority type \texttt{BEFORE} makes 34\% of the data. With this scheme, the proportion of instances under each \texttt{TLINK} type are approximately the same in all 10 folds.

Two experimental settings are considered:
\begin{enumerate}[label=S\arabic*]
\item Given a pair of word vectors $(\vec{w}_1, \vec{w}_2)$, which is retrieved from either GloVe or Word2Vec pre-trained vectors, $(\vec{w}_1 \oplus \vec{w}_2)$, $(\vec{w}_1 + \vec{w}_2)$ or $(\vec{w}_1 - \vec{w}_2)$ is considered as the feature set for the classifier.
\item Given a pair of word vectors $(\vec{w}_1, \vec{w}_2)$, which is the best performing embeddings in S1, and the traditional feature set $\vec{f}$,  $((\vec{w}_1 \oplus \vec{w}_2) \oplus \vec{f})$, $((\vec{w}_1 + \vec{w}_2) \oplus \vec{f})$ or $((\vec{w}_1 - \vec{w}_2) \oplus \vec{f})$ is considered as the feature set for the classifier.
\end{enumerate}

\begin{table}[t]
\centering
\small
\begin{tabular} {lrcccc|c}
\hline
 & \multicolumn{2}{c}{\textbf{Feature vector}} & \textbf{Total} & \textbf{TP} & \textbf{FP} & \textbf{F1}\\
\hline
& Traditional features & $\vec{f}$ & 5271 & 2717 & 2554 & \textbf{0.5155} \\
\hline
\textbf{S1} & GloVe & $(\vec{w}_1 \oplus \vec{w}_2)$ & 5271 & 2388 & 2883 & 0.4530 \\
 & & $(\vec{w}_1 + \vec{w}_2)$ & 5271 & 2131 & 3140 & 0.4043 \\
 & & $(\vec{w}_1 - \vec{w}_2)$ & 5271 & 2070 & 3201 & 0.3927 \\
\hdashline
 & Word2Vec & $(\vec{w}_1 \oplus \vec{w}_2)$ & 5271 & 2609 & 2662 & \textbf{0.4950} \\
 & & $(\vec{w}_1 + \vec{w}_2)$ & 5271 & 2266 & 3005 & 0.4299 \\
 & & $(\vec{w}_1 - \vec{w}_2)$ & 5271 & 2258 & 3013 & 0.4284 \\
\hline
\textbf{S2} & Word2Vec & $((\vec{w}_1 \oplus \vec{w}_2) \oplus \vec{f})$ & 5271 & 3036 & 2235 & \textbf{0.5760} \\
 & & $((\vec{w}_1 + \vec{w}_2) \oplus \vec{f})$ & 5271 & 2901 & 2370 & 0.5504 \\
 & & $((\vec{w}_1 - \vec{w}_2) \oplus \vec{f})$ & 5271 & 2887 & 2384 & 0.5477 \\
\hline
\end{tabular}
\caption{\label{tab:embedding}Classifier performances (F1-scores) in different experimental settings S1 and S2, compared with using only traditional features. TP: true positives and FP: false positives.}
\end{table}

\begin{table}[t]
\centering
\begin{adjustbox}{width=1\textwidth}
\begin{tabular} {lccc|c|ccc}
\hline
\textbf{\texttt{TLINK} type} & \textbf{$(\vec{w}_1 \oplus \vec{w}_2)$} & \textbf{$(\vec{w}_1 + \vec{w}_2)$} & \textbf{$(\vec{w}_1 - \vec{w}_2)$} & \textbf{$\vec{f}$} & \textbf{$((\vec{w}_1 \oplus \vec{w}_2) \oplus \vec{f})$} & \textbf{$((\vec{w}_1 + \vec{w}_2) \oplus \vec{f})$} & \textbf{$((\vec{w}_1 - \vec{w}_2) \oplus \vec{f})$}\\
\hline
\texttt{BEFORE} & \textbf{0.6120} & 0.5755 & 0.5406 & 0.6156 & \textbf{0.6718} & 0.6440 & 0.6491\\
\texttt{AFTER} & \textbf{0.4674} & 0.3258 & 0.4450 & 0.5294 & \textbf{0.5800} & 0.5486 & 0.5680\\
\texttt{IDENTITY} & 0.5142 & 0.4528 & \textbf{0.5201} & 0.6262 & \textbf{0.6650} & 0.6456 & 0.6479\\
\texttt{SIMULTANEOUS} & \textbf{0.2571} & 0.2375 & 0.1809 & 0.1589 & 0.3056 & \textbf{0.3114} & 0.1932\\
\texttt{INCLUDES} & \textbf{0.3526} & 0.2348 & 0.3278 & 0.3022 & \textbf{0.4131} & 0.3627 & 0.3598\\
\texttt{IS\_INCLUDED} & \textbf{0.2436} & 0.0769 & 0.2268 & 0.2273 & \textbf{0.3455} & 0.3077 & 0.2527\\
\texttt{BEGINS} & 0 & 0 & \textbf{0.0494} & 0.0741 & \textbf{0.1071} & 0.1000 & 0.1053\\
\texttt{BEGUN\_BY} & 0.0513 & 0 & \textbf{0.1481} & 0 & \textbf{0.1395} & 0.0930 & 0.0976\\
\texttt{ENDS} & 0 & 0 & \textbf{0.0303} & 0 & \textbf{0.2000} & 0.1500 & 0.1500\\
\texttt{ENDED\_BY} & \textbf{0.3051} & 0.2540 & 0.1982 & 0.0727 & \textbf{0.2807} & 0.2712 & 0.0784\\
\hline
\textbf{Overall} & \textbf{0.4950} & \textbf{0.4299} & \textbf{0.4284} & \textbf{0.5155} & \textbf{0.5760} & \textbf{0.5504} & \textbf{0.5477}\\
\hline
\end{tabular}
\end{adjustbox}
\caption{\label{tab:embedding-per-tlink}F1-scores per \texttt{TLINK} type with different feature vectors. Pairs of word vectors ($\vec{w}_1$, $\vec{w}_2$) are retrieved from Word2Vec pre-trained vectors.}
\end{table}

\paragraph{Experiment Results} In Table~\ref{tab:embedding} we report the performances of the classifier in different experimental settings S1 and S2, compared with the classifier performance using only traditional features. Since we classify all possible event pairs in the dataset, precision and recall are the same.

We found that pre-trained word vectors from Word2Vec perform better than the ones from GloVe. Based on this we may conclude that word embeddings from predictive models are superior than those from count-based models for this particular task. However, we should take into account that the word vectors from GloVe are only partially trained on news texts, while Word2Vec embeddings are fully trained on news dataset. Given that the TBAQ-cleaned corpus used in these experiments contains news articles, the different source of texts used for training the embeddings may contribute to the vary performance. Moreover, we acknowledge the different size of data used to train the embeddings, i.e. 6 billion tokens for GloVe vs 100 billion words for Word2Vec, which may also be the cause of GloVe embeddings' loss.

From the different vector combinations, concatenation $(\vec{w}_1 \oplus \vec{w}_2)$ is shown to be the best combination. Using the concatenated Word2Vec embeddings $(\vec{w}_1 \oplus \vec{w}_2)$ as features results in 0.4950 F1-score, almost as good as using only traditional features (0.5155 F1-score). The fact that this representation retain the word order information may be the reason why it beats the other vector combinations. With the exception of \texttt{IDENTITY} and \texttt{SIMULTANEOUS}, all of the other \texttt{TLINK} types are asymmetric, e.g. \texttt{BEFORE}/\texttt{AFTER}, \texttt{INCLUDES}/\texttt{IS\_INCLUDED}.

Combining the word embeddings with traditional features (S2) yields significant improvement. With $((\vec{w}_1 \oplus \vec{w}_2) \oplus \vec{f})$ as features, the classifier achieves 0.5760 F1-score, around 6\% improvement compared to using only $\vec{f}$ as features and around 8\% if compared to using only $(\vec{w}_1 \oplus \vec{w}_2)$ as features. The rationale behind this improvement would be that in this setting, the classifier could utilize both morpho-syntactic and lexical-semantic features to learn the temporal ordering between events in texts.

We detail in Table~\ref{tab:embedding-per-tlink}, the performances of the classifier (in terms of F1-scores) with different feature vectors. Pairs of word vectors $(\vec{w}_1, \vec{w}_2)$ are retrieved from Word2Vec pre-trained vectors. $(\vec{w}_1 - \vec{w}_2)$ is shown to be the best in identifying \texttt{IDENTITY}, \texttt{BEGINS}, \texttt{BEGUN\_BY} and \texttt{ENDS} relation types, while the rest are best identified by $(\vec{w}_1 \oplus \vec{w}_2)$. Combining $(\vec{w}_1 \oplus \vec{w}_2)$ and $\vec{f}$ improves the identification of all \texttt{TLINK} types in general, particularly \texttt{BEGINS}/\texttt{BEGUN\_BY} and \texttt{ENDS}/\texttt{ENDED\_BY} types, which were barely identified when $(\vec{w}_1 \oplus \vec{w}_2)$ or $\vec{f}$ is used individually as features.

\section{Evaluation}

\paragraph{Dataset} We use the same training and evaluation data released in the context of Tempeval-3, i.e., TBAQ-cleaned and TempEval-3-platinum.

\begin{table}[t]
\centering
\small
\begin{adjustbox}{width=1\textwidth}
\begin{tabular} {lccc|ccc|ccc}
\hline
\multirow{2}{*}{\textbf{\texttt{TLINK} type}} & \multicolumn{3}{c|}{\textbf{$(\vec{w}_1 \oplus \vec{w}_2)$}} & \multicolumn{3}{c|}{\textbf{$\vec{f}$}} & \multicolumn{3}{c}{\textbf{$((\vec{w}_1 \oplus \vec{w}_2) \oplus \vec{f})$}}\\
 & \textbf{P} & \textbf{R} & \textbf{F1} & \textbf{P} & \textbf{R} & \textbf{F1} & \textbf{P} & \textbf{R} & \textbf{F1}\\
\hline
\texttt{\textbf{BEFORE}} & 0.4548 & 0.7123 & 0.5551 & 0.5420 & 0.7381 & \textbf{0.6250} & 0.5278 & 0.7170 & 0.6080\\
\texttt{\textbf{AFTER}} & 0.5548 & 0.4649 & 0.5059 & 0.5907 & 0.6196 & \textbf{0.6048} & 0.6099 & 0.6000 & \textbf{0.6049}\\
\texttt{\textbf{IDENTITY}} & 0.0175 & 0.0667 & 0.0278 & 0.2245 & 0.7333 & 0.3438 & 0.2444 & 0.7333 & \textbf{0.3667}\\
\texttt{\textbf{SIMULTANEOUS}} & 0.3529 & 0.0732 & 0.1212 & 0.1667 & 0.0370 & 0.0606 & 0.2308 & 0.0732 & \textbf{0.1111}\\
\texttt{\textbf{INCLUDES}} & 0.1765 & 0.0750 & 0.1053 & 0.3077 & 0.2000 & \textbf{0.2424} & 0.1852 & 0.1250 & 0.1493\\
\texttt{\textbf{IS\_INCLUDED}} & 0.4000 & 0.0426 & 0.0769 & 0.3333 & 0.0638 & 0.1071 & 0.3846 & 0.1064 & \textbf{0.1667}\\
\texttt{\textbf{BEGINS}} & 0 & 0 & 0 & 0 & 0 & 0 & 0 & 0 & 0\\
\texttt{\textbf{BEGUN\_BY}} & 0 & 0 & 0 & 0 & 0 & 0 & 0 & 0 & 0\\
\texttt{\textbf{ENDS}} & 0 & 0 & 0 & 0 & 0 & 0 & 0 & 0 & 0\\
\texttt{\textbf{ENDED\_BY}} & 0 & 0 & 0 & 0 & 0 & 0 & 0 & 0 & 0\\
\hline
\textbf{Overall} & \multicolumn{3}{c|}{\textbf{0.4271}} & \multicolumn{3}{c|}{\textbf{0.5043}} & \multicolumn{3}{c}{\textbf{0.4974}} \\
\hline
\end{tabular}
\end{adjustbox}
\caption{\label{tab:embedding-per-tlink-eval}Classifier performance per \texttt{TLINK} type with different feature vectors, evaluated on TempEval-3-platinum. Pairs of word vectors ($\vec{w}_1$, $\vec{w}_2$) are retrieved from Word2Vec pre-trained vectors.}
\end{table}

\paragraph{} Table~\ref{tab:embedding-per-tlink-eval} shows the classifier performances with different feature vectors, evaluated on the TempEval-3-platinum corpus. In general, using only $(\vec{w}_1 \oplus \vec{w}_2)$ as features does not give any benefit since the performance is significantly worse compared to using only traditional features $\vec{f}$, i.e. 0.4271 vs 0.5043 F1-scores. Combining the word embedding and traditional features $((\vec{w}_1 \oplus \vec{w}_2) \oplus \vec{f})$ also does not improve the classifier performance in general. However, if we look into each \texttt{TLINK} type, the classifier performance in identifying \texttt{IDENTITY}, \texttt{SIMULTANEOUS} and \texttt{IS\_INCLUDED} is improved, and quite significantly for \texttt{SIMULTANEOUS}. 

\section{Conclusions}

We have presented a preliminary investigation into the potentiality of exploiting word embeddings for temporal relation type classification between event pairs. We compared two existing pre-trained word vectors from the two popular word embedding algorithms, GloVe and Word2Vec, representing count-based and predictive models, resp. Word2Vec embeddings are found to be the better word representation for this particular task. However, we cannot make a strong judgement towards predictive models being better than count-based models. This is because of the different training criteria, such as source of texts and size of the training set, used to build the two embeddings.

We also found that concatenation is the best way to combine the word vectors, although subtraction may also bring advantages for some \texttt{TLINK} types such as \texttt{IDENTITY} and \texttt{BEGINS}/\texttt{BEGUN\_BY}.

In the 10-fold cross-validation setting, combining word embedding and traditional features $((\vec{w}_1 \oplus \vec{w}_2) \oplus \vec{f})$ results in significant improvement. However, using the same feature vector evaluated on the TempEval-3 evaluation corpus, the classifier does not improve in general, although for some \texttt{TLINK} types (\texttt{IDENTITY}, \texttt{SIMULTANEOUS} and \texttt{IS\_INCLUDED}) we observe a performance gain. 

These results shed some light on how word embeddings can potentially improve a classifier performance for temporal relation extraction. In the future we would like to explore the impact of ensemble learning, \textit{stacking} method in particular, in which a super-classifier is trained to combine the predictions of several other classifiers, e.g., classification models with $(\vec{w}_1 \oplus \vec{w}_2)$, $(\vec{w}_1 + \vec{w}_2)$, $(\vec{w}_1 - \vec{w}_2)$ and $\vec{f}$ as features.

Furthermore, instead of using general-purpose word embeddings, several works presented methods for building task-specific word embeddings \parencite{hashimoto-EtAl:2015:CoNLL,boros-EtAl:2014:EMNLP2014,nguyen-grishman:2014:P14-2,tang-EtAl:2014:SemEval}, which may also be beneficial for temporal ordering task.

\chapter{Training Data Expansion}\label{ch:training-data-expansion}
\begin{flushright}
\scriptsize
\textit{We don’t have better algorithms. We just have more data.} --- Google’s Research Director Peter Norvig
\end{flushright}
\minitoc

\section{Introduction}

The scarcity of annotated data is often an issue in building an extraction system with supervised learning approach. \textcite{Halevy:2009:UED:1525642.1525689} argue that linear classifiers trained on millions of specific features outperform more elaborate models that try to learn general rules but are trained on less data. One of the widely known approaches to gain more training examples is semi-supervised learning, since many machine-learning researchers have found that unlabelled data, when used in conjunction with a small amount of labelled data, can produce considerable improvement in learning accuracy.

In this chapter we explore two approaches to expand the training data for temporal and causal relation extraction, namely (i) \textit{temporal reasoning on demand} for temporal relation type classification\footnote{Joint work with Rosella Gennari, Free University of Bozen-Bolzano and Pierpaolo Vittorini, Università degli Studi dell'Aquila.} and (ii) \textit{self-training}, a wrapper method for semi-supervised learning, for causal relation extraction. Each approach will be discussed in details in Section~\ref{sec:expansion-tlink} and Section~\ref{sec:expansion-clink}, respectively.

\section{Related Work}

Several systems have been proposed to classify temporal relations between temporal entities according to TimeML specifications, as shown in the TempEval evaluation campaigns \parencite{verhagen-EtAl:2007:SemEval-2007,verhagen-EtAl:2010:SemEval,uzzaman-EtAl:2013:SemEval-2013}. However, only a few of them dealt with the problem of combining classification and temporal reasoning \parencite{mani-EtAl:2006:COLACL}. Such a combination can have three main advantages: (i) to detect and possibly discard  documents in the training data that are inconsistent, (ii) to automatically analyse and partially complete documents with missing relevant data before running a classifier, (iii) to improve the output of the classifier by adding data or spotting inconsistencies in the classifier output \parencite{tatu-srikanth:2008:PAPERS}.

Research work has recently focused on (ii) and (iii), using mainly Integer Linear Programming \parencite{chambers-jurafsky:2008:EMNLP,do2012} and Markov Logic Networks \parencite{yoshikawa-EtAl:2009:ACLIJCNLP} to optimize the classification output. To our knowledge, however, no studies have so far tackled the problem (i) to improve on the quality of temporal relation classification. The only work partially related to this topic is by \textcite{DERCZYNSKI10.546}, whose CAVaT tool is able to perform error checking and validation of TimeML documents. However, the impact of error checking and validation on automated classification performances is not experimentally assessed. 

As for (ii), the annotation of event-event relations following TimeML guidelines has focused on events in the same or close sentences, therefore, most machine-learning based classifiers learn to annotate following this principle. However, deducing additional relevant relations can be beneficial to a classifier performance. \textcite{laokulrat-miwa-tsuruoka:2014:TextGraphs-9} partially complete documents by deducing relations before running a classifier: they extract data by using timegraphs, and apply a stacked learning method to temporal relation classification.  However, for performance reasons, they do not deduce all possible relations, e.g., they arbitrarily limit the number of deduced  relations for each document to 10,000.
More generally concerning the costs of (ii), no past work has taken them into account from a statistical experimental viewpoint.

Semi-supervised setting for temporal relation extraction were made possible in the TempEval-3 shared task. The task organizers released the \textit{TempEval-3 silver} corpus, a 600K corpus collected from Gigaword, which is automatically annotated by best performing systems in the preceding TempEval task. In TempEval-3, none of the participants submitted the systems trained on this silver data, most probably because the amount of gold data released within the task was enough to achieve good results in a fully supervised setting.
 
\textcite{fisher-simmons:2015:ACL-IJCNLP} presents a semi-supervised spectral model for a sequential relation labelling task for discourse parsing, in which \textit{Temporal} and \textit{Contingency/Causal} relations are included in the possible relation types. The empirical evaluation on the Penn Discourse TreeBank (PDTB) \parencite{pdtb2007} dataset yields 0.485 F1-score, around 7-9 percentage point improvement over
approaches that do not utilize unlabelled training
data.

\section{Temporal Reasoning on Demand}
\label{sec:expansion-tlink}

In an attempt to improve the performance of temporal relation type classification between event pairs, we introduce a method to enrich the training data via temporal reasoner. However, instead of running the reasoner over the whole training set, we argue that this step can be enhanced by invoking the reasoner only when it is estimated to be effective for improving the classifier's performance. The estimation is based on document parameters that are easy to calculate. 

Moreover, the gold standard corpora are found to contain inconsistent documents with respect to the temporal graph built from annotated temporal relations using a temporal closure algorithm \parencite{DBLP:journals/lre/Verhagen05,DBLP:journals/jdiq/GennariTV15}. As \textcite{DBLP:journals/lre/Verhagen05} pointed out, ``it is hard to annotate a one-page document without introducing inconsistencies'' and ``even trained annotators are liable to introduce relations that clash with previous choices''. Therefore, we also investigate the impact of excluding inconsistent documents from the training set in order to improve not only the quantity but also the quality of the training data.

\paragraph{Method} Given a training set with TimeML documents, the following steps are performed: 

\begin{enumerate}[label=Step \arabic*,leftmargin=4\parindent]
\item The temporal reasoner checks the consistency of each document. If the document is found to be \textit{consistent}, it reports so; else it tries with a more relaxed \texttt{TLINK} mapping into Allen relation (see Section~\ref{sec:temporal-reasoner}) or, if no other mapping is available, it terminates reporting \textit{inconsistency}.
\item Documents that are inconsistent with respect to all the considered mappings are discarded from the training set.
\item For each document, the number of temporal relations (\texttt{TLINK}s) that can be deduced is empirically predicted.
\item The decision whether deduction should be run or not on a document depends on whether the predicted number of deduced \texttt{TLINK}s falls below a certain \textit{threshold}. Accordingly, the temporal reasoner runs the deduction (or not).
\item The initial training set, now enriched with deduced \texttt{TLINK}s, is used to train the classifiers for determining the \texttt{TLINK} types of event-event (E-E) and event-timex (E-T) pairs.
\end{enumerate}

For the temporal reasoning process, including inconsistency checking (Step 1) and new \texttt{TLINK} deduction (Step 4), the temporal reasoner (SQTR) explained in Section~\ref{sec:temporal-reasoner} is used. The method for estimating the number of deduced \texttt{TLINK}s (Step 3) will be further detailed in the following section (Section~\ref{sec:expansion-tlink-predict-deduced}). The threshold for Step 4 is estimated in the experiments reported in Section~\ref{sec:expansion-tlink-experiments}.

The \texttt{TLINK} type classification system reported in this chapter is a hybrid classification system, which determines the temporal relation type of a given pair of ordered temporal entities $(e_1, e_2)$. Included are two supervised classifiers for event-event (E-E) and event-timex (E-T) pairs, and a set of rules for timex-timex (T-T) pairs.

The supervised classifiers for E-E and E-T pairs (Step 5) are basically the prior version of the E-E classifier and E-T classifier included in TempRelPro (Section~\ref{sec:pair-classifier}). Meanwhile, the set of rules for T-T pairs is the same set of rules explained in Section~\ref{sec:timex-timex-rules}. Henceforth, we will address the \texttt{TLINK} type classification system used in this chapter as \textit{TempRelPro-beta}. Several features that make TempRelPro-beta different from the current version include:
\begin{itemize}
\item \textit{dependencyPath} The considered dependency path were not limited to any specific constructions. 
\item \textit{signalTokens} For TempRelPro-beta, temporal signals were not clustered, hence, the tokens of \textit{temporal signals} or \textit{temporal discourse markers} appearing around the event pairs are used as features instead of the \textit{clusterID} of signal cluster. The discourse markers were obtained using addDiscourse \parencite{pitler-nenkova:2009:Short}, retaining only those labelled as \textit{Temporal}.
\item \textit{signalDependency} This feature was not included in TempRelPro-beta.
\end{itemize}
Furthermore, in TempRelPro-beta, the \textit{label simplification} method mentioned in Section~\ref{sec:pair-classifier} is not employed, so the classifiers label the entity pairs with the full set of 14 \texttt{TLINK} types.

Since it has been shown that the current system for temporal relation extraction performs poorly in classifying the temporal relation types of event-event pairs, in this chapter the automated temporal reasoning is considered only for event-event pairs, albeit the method design is in principle more general. Nevertheless, the classification experiments presented in the following sections cover event-event, event-timex and timex-timex pairs, using TempRelPro-beta, in line with the tasks proposed in the TempEval campaigns.

\subsection{Predicting Number of Deduced \texttt{TLINK}s}
\label{sec:expansion-tlink-predict-deduced}

As suggested by \textcite{DBLP:journals/lre/Verhagen05}, there should be a relation between the \textit{size of the maximal connected components} (SMCC) of the graph of the document and the number of deducible \texttt{TLINK}s. This could also be related to number of \texttt{EVENT}s and \texttt{TLINK}s of the input document. All such variables were computed for each consistent document in the training corpus of TempEval-3~\cite{uzzaman-EtAl:2013:SemEval-2013} and a regression analysis was conducted with: (i) the number of deduced \texttt{TLINK}s as dependent variable, and (ii) the number of \texttt{EVENT}s, the number of \texttt{TLINK}s, and SMCC as independent variables. Table \ref{tab:regress1} shows the results of the analysis.

\begin{table}[t]
\centering
\small
\begin{tabular}{lrr} \hline
\multirow{2}{*}{\bf Variables} & \multicolumn{2}{c}{\textbf{\#Deducible \texttt{TLINK}s}} \\ 
                               & \bf Estimates & \bf p-value \\ \hline
\#\texttt{TLINK}s                       & $12.8$        & $0.000$ \\
\#\texttt{EVENT}s                       & $-17.6$       & $0.006$ \\
SMCC                           & $17.1$        & $0.007$ \\ \hdashline
Intercept                      & $-10.0$       & $0.079$ \\ \hline
\end{tabular}
\caption{Regression analysis ($p=0.000$, R$^2=0.633$)}
\label{tab:regress1}
\end{table}

Since the model proves to fit with good reliability (R$^2$=0.633), one can assume that a linear relation exists between the number of \texttt{TLINK}s, \texttt{EVENT}s, SMCC, and the number of deducible \texttt{TLINK}s. According to the estimates in Table~\ref{tab:regress1}, SMCC together with the number of \texttt{TLINK}s are likely to have a large impact on the number of deducible \texttt{TLINK}s. 
On the contrary, increasing the number of \texttt{EVENT}s has an adverse effect. 

Such results confirm that SMCC, the number of \texttt{TLINK}s and \texttt{EVENT}s of the input document can predict the number of deducible \texttt{TLINK}s (e.g., if \#\texttt{TLINK}s = $15$, \#\texttt{EVENT}s = $10$, SMCC = $10$, one may expect to deduce $\approx$ $43, CI_{95\%}=[32,54]$ \texttt{TLINK}s). Moreover they enable Step 4, which relies on the following research hypothesis: if the estimated number of deducible \texttt{TLINK}s falls below a threshold, deducing \texttt{TLINK}s is likely to be effective for classification performances; else not. The threshold is estimated in the experiments reported in the following section. 

\subsection{Experiments}
\label{sec:expansion-tlink-experiments}

Two sets of experiments were run. The first experiment focuses on the effect of consistency checking and/or deduction on classification performances. The second experiment instead evaluates the threshold foreseen in Step 4 returning the best classification performances.

\paragraph{Dataset} Experiments were run using two widely used corpora as the training set, which were made available during the Tempeval-3 challenge, \textit{TBAQ-cleaned}, which includes (i) the AQUAINT TimeML corpus, containing 73 news report documents, and (ii) the TimeBank corpus, with 183 news articles. The test data is the newly created \textit{TempEval-3-platinum} evaluation corpus that was annotated and reviewed by the Tempeval-3 task organizers.


\paragraph{Consistency Checking and Deduction} Classification performances of TempRelPro-beta were computed, comparing four different experimental settings. In all settings, Step 4 of the workflow is disabled, since there were no available experimental data concerning the threshold. Several experimental settings are considered:

\begin{enumerate}[label=S\arabic*]
\item The temporal relation classifiers are trained using only the TimeBank and AQUAINT corpora (TBAQ) without running SQTR.
\item The training set is obtained by discarding the 9 inconsistent documents detected by SQTR (TBAQ$^{c}$), i.e., removing around 11.61\% event-event pairs and 5.1\% event-timex pairs. Note that although consistency checking concerns only event-event pairs, when a document is inconsistent it is discarded as a whole, hence, e.g., also event-timex pairs are removed.
\item The training set is obtained after computing the deductive closure on event-event pairs with SQTR (TBAQ$^{d-all}$).
\item The training set is obtained with deduction after discarding inconsistent documents, i.e., TBAQ$^{c}$ after deduction  (TBAQ$^{cd-all}$). After deduction, the number of event-event pairs is almost three times as many as in annotated input documents, with new 9,750 pairs addition.
\end{enumerate}

Classification results are reported in Table~\ref{tab:classifier}.  
Precision, recall and F1-score are computed using the graph-based evaluation metrics of the official TempEval-3 scorer, while accuracy is given by the \texttt{predict} function of LIBLINEAR. 

\begin{table}[t]
	\centering
	\small
	\begin{tabular}{l|c c|c c c}
	\hline
    \multirow{2}{*}{\textbf{Training Set}} & \multicolumn{2}{|c|}{\textbf{Accuracy}} & \multirow{2}{*}{\textbf{P}} & \multirow{2}{*}{\textbf{R}} & \multirow{2}{*}{\textbf{F1}} \\ 
    & E-E & E-T & & & \\ \hline
    \textbf{TBAQ} & .4734 & .7697 & .5935 & \textbf{.6026} & \textbf{.5980} \\
    TBAQ$^{c}$ & .4683 & .7666 & .5891 & .5971 & .5931 \\ \hline
    TBAQ$^{d-all}$ & .\textbf{4820} & .7697 & \textbf{.5944} & .5993 & .5969 \\
    TBAQ$^{cd-all}$ & .4786 & .7666 & .5925 & .5960 & .5943 \\
    \hline
    \end{tabular}
	\caption{Classifier performances for event-event (ee) and event-timex (et) pairs with different training data}
    \label{tab:classifier}
\end{table}

Results are in line with the literature and show that consistency checking and deduction are not always beneficial for classification performances. In particular, removing inconsistent documents slightly decreases performances (possibly because less data is given to the classifier), even if this improves the quality of training data for classification.

To compare the effect of deduction, the performances are compared based on the same training data, i.e., either all documents (TBAQ vs. TBAQ$^{d-all}$)  or only consistent documents (TBAQ$^{c}$ vs. TBAQ$^{cd-all}$). Accordingly, deduction improves precision, lowers recall, improves F1-score when the training data is made up only of consistent documents, and lowers F1-score when all documents are included in the training set.


As mentioned earlier, consistency checking discards  document as a whole, even if only two relations are inconsistent, and thus also discarding consistent annotations in the document that the classifier may use. Such an issue has space for improvements, e.g., by implementing  in SQTR a further method that extracts consistent annotations from an inconsistent document. Concerning deduction, the main reason for the lowered recall is that the already skewed data become more unbalanced after computing deduction (e.g., there are $~$6000 instances of \texttt{BEFORE} type vs $<$100 instances of \texttt{IBEFORE} type). 

\subsubsection{Threshold.}
%
For each document, the estimated number of deducible \texttt{TLINK}s between events is computed (Step 3). Next, a routine was implemented for finding the lowest threshold which yields the highest F-score, where the threshold corresponds to the number of maximum deducible \texttt{TLINK}s for each document.

\begin{figure}
\centering
  \includegraphics[width=.8\textwidth]{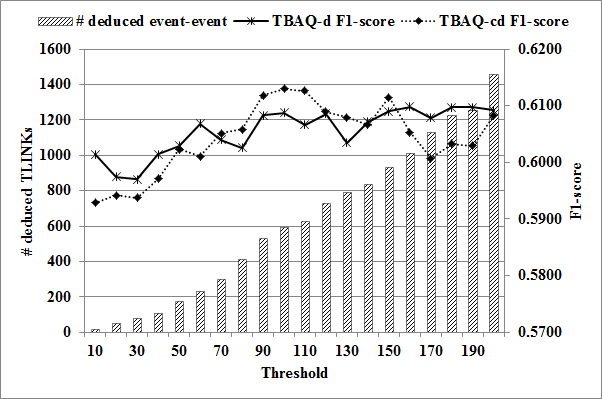}
  \caption{Classifier performances using TBAQ and TBAQ$^{c}$ as training set expanded with deduced \texttt{TLINK}s using different deduction thresholds}
  \label{fig:threshold}
\end{figure}

The routine starts from the threshold value $t=10$, going up until $t=200$ with interval $i=10$. Figure \ref{fig:threshold} shows the best threshold value, before the F1-scores decreases or becomes stagnant. Again, the impact of Step 4 is assessed by applying deduction on event-event pairs from TBAQ and TBAQ$^{c}$. This analysis shows that classification performance using TBAQ$^{d}$ or TBAQ$^{cd}$ is sensitive to the threshold. 





The classification experiment using TBAQ$^{d}$ and TBAQ$^{cd}$ was then re-run with their best thresholds ($160$ and $100$ respectively). Results are reported in Table~\ref{tab:classifier2}.

\begin{table}
	\centering
	\small
	\begin{tabular}{l|c c|c c c}
	\hline
    \multirow{2}{*}{\textbf{Training Set}} & \multicolumn{2}{|c|}{\textbf{Accuracy}} & \multirow{2}{*}{\textbf{P}} & \multirow{2}{*}{\textbf{R}} & \multirow{2}{*}{\textbf{F1}} \\ 
    & ee & et & & & \\ \hline
    TBAQ$^{d-160}$ & .4837 & .7697 & .6070 & .6126 & .6098 \\
    \textbf{TBAQ$^{cd-100}$} & .4854 & .7666 & .6112 & .6148 & \textbf{.6130} \\ 
    \hline
    \end{tabular}
	\caption{Classifier performances with TBAQ$^{d}$ and TBAQ$^{cd}$ training data and their best deduction threshold, i.e., 160 and 100 respectively}
    \label{tab:classifier2}
\end{table}

The outcome of the second set of experiments sheds light on when deduction should be invoked: in a flexible setting, with deduction being called only for specific documents, classification performances achieve state-of-the art results. In other words, deduction proves useful to increase TLINKs between events only for specific documents. Also discarding inconsistent documents helps in this setting. These findings corroborate our initial hypothesis that a flexible system for temporal relation classification may be useful to address consistency and data sparseness problems of the training data, improving both quality of data and classification performances. 

\section{Semi-supervised Learning}
\label{sec:expansion-clink}

Given the sparse annotated \texttt{CLINK}s in the Causal-TimeBank corpus, we introduce a bootstrap method to get more labelled examples. Specifically, we employ \textit{self-training}, a wrapper method for \textit{semi-supervised learning}. We first train a \texttt{CLINK} classifier based on labelled data in the Causal-TimeBank. The classifier is then applied to unlabelled data to generate more examples to be used as input for the \texttt{CLINK} classifier.

Furthermore, given the nature of news texts, that often describe the same set of events just by rewording the same underlying story, we make the following assumptions:
\begin{enumerate}[label=(\roman*)]
\item for the same two event pairs in different news articles, the causal relations (if any) may be expressed with different causal markers, or be implicit; and
\item if we set a causal relation between two events in a news, the same holds every time the two events are mentioned in similar news. 
\end{enumerate}
Therefore, we can bootstrap new training data by propagating the causal relation through event co-reference. In these new training instances, causality may be expressed differently from the original news.

\begin{figure}
\centering
\includegraphics[width=0.9\textwidth]{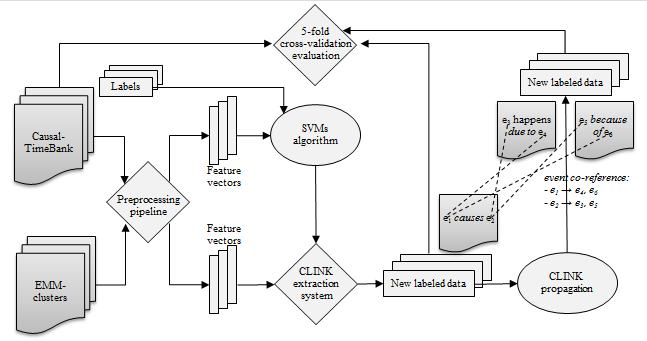}
\caption{The work-flow of the \texttt{CLINK} extraction system and bootstrapping method.}
\label{fig:flowchart}
\end{figure}

In addition to the labelled data obtained via self-training, we bootstrap further training examples through the so-called \textit{\texttt{CLINK} propagation} method, following the assumptions above. The workflow of our \texttt{CLINK} extraction system and bootstrapping method is reported in Figure~\ref{fig:flowchart}.

Note that the \texttt{CLINK} extraction system reported in this chapter is slightly different than our current CauseRelPro (Section~\ref{causal-rel-recog-method}), which will be explained further in Section~\ref{sec:expansion-clink-extraction}, and henceforth named as \textit{CauseRelPro-beta}. We will detail our training data expansion experiments through self-training and CLINK propagation in Section~\ref{sec:expansion-clink-experiments}.

\subsection{\texttt{CLINK} Extraction System}
\label{sec:expansion-clink-extraction}

\paragraph{Preprocessing} Several modules and tools are used in the preprocessing steps to leverage syntactic and semantic information needed for the automatic extraction part. In particular, we employ the following systems:
\begin{itemize}
\item \textit{NewsReader NLP pipeline}, which is an NLP pipeline developed as part of the NewsReader project.\footnote{\url{http://www.newsreader-project.eu/results/software/}}, with the goal to integrate several analysis layers in a single NLP suite.  We are particularly interested in the output of the event annotation module, since we want to detect causality between events. However, the other modules of the pipeline such as part-of-speech tagging, constituency parser, dependency parser, event co-reference and temporal relation extraction also play a great role in producing feature vectors for to-be-classified event pairs.
\item \textit{SuperSense tagger} \parencite{Ciaramita:2006}, which annotates verbs and nouns in a text with 41 semantic categories (WordNet supersense).
\item \textit{addDiscourse} \parencite{pitler-nenkova:2009:Short}, which identifies discourse connectives and assigns them to one of four semantic classes: \textit{Temporal}, \textit{Expansion}, \textit{Contingency} and \textit{Comparison}. Causal connectives are included in the \textit{Contingency} class.
\end{itemize}

CauseRelPro-beta includes a classification model based on Support Vector Machines (SVMs) with a \textit{one-vs-one} strategy for multi-class classification and polynomial kernel. Considered as candidate event pairs are the same candidate pairs described in Section~\ref{sec:causerelpro-candidate-event-pairs}. The classification model is mainly based on the \texttt{CLINK} classifier introduced in \textcite{mirza-tonelli:2014:Coling}, with several novelties:
\begin{enumerate}[label=(\roman*)]
\item We consider WordNet supersenses as token-level features instead of tokens or lemmas to better generalize our model and avoid over-fitting.
\item We avoid a two-step procedure, in which causal signals are first automatically identified as a separate task and then included as features in the classification model for event relations. Instead, \textit{causal markers features}, which will be detailed below, are included in the set together with event features and temporal information.
\item Co-reference between event pairs is considered, since no causal relation can hold between two co-referring events. This information is provided by the NewsReader pipeline, which includes an implementation of the event coreference tool by \textcite{CYBULSKA:2013}.
\end{enumerate}


\paragraph{Causal marker features} We consider three types of causal markers that can cue a causal relation between events:
\begin{enumerate}
\item \textit{Causal signals}. We rely on the list of causal signals used to annotate \texttt{CSIGNAL}s in the Causal-TimeBank corpus. For some causal signals that are discontinuous, e.g. \textit{due (mostly) to}, we include their regular expression patterns in the list, e.g. \textit{/due .*to/}.
\item \textit{Causal connectives}, i.e. the discourse connectives under the \textit{Contingency} class according to the output of the addDiscourse tool.
\item \textit{Causal verbs}. The three types of verbs lexicalizing causal concepts as listed in \textcite{wolff:2003}: \textit{i)} \texttt{CAUSE}-type verbs, e.g. \textit{cause, prompt, force}; \textit{ii)} \texttt{ENABLE}-type verbs, e.g. \textit{allow, enable, help}; and \textit{iii)} \texttt{PREVENT}-type verbs, e.g. \textit{block, prevent, restrain}.
\end{enumerate}

Based on the existence of causal markers around $e_1$ and $e_2$, exactly in that priority order\footnote{We first look for causal signals. If we do not find any, then we continue looking for causal connectives. And so on.}, we include as features:
\begin{itemize}[leftmargin=*,topsep=2pt]
\setlength{\itemsep}{2pt}
\setlength{\parskip}{0pt}
\setlength{\parsep}{0pt}
\item causal marker string;
\item causal marker position, i.e. \textit{between} $e_1$ and $e_2$, \textit{before} $e_1$, or at the \textit{beginning} of the sentence where $e_1$/$e_2$ is in; and
\item dependency path between the causal marker and $e_1$/$e_2$. 
\end{itemize}

Since the previously mentioned causal marker features are based on non-exhaustive lists of causal signals and causal verbs, not to mention the imperfect output of the addDiscourse tool, we also consider possible causal signals and causal verbs derived from dependency relations.

\begin{figure}
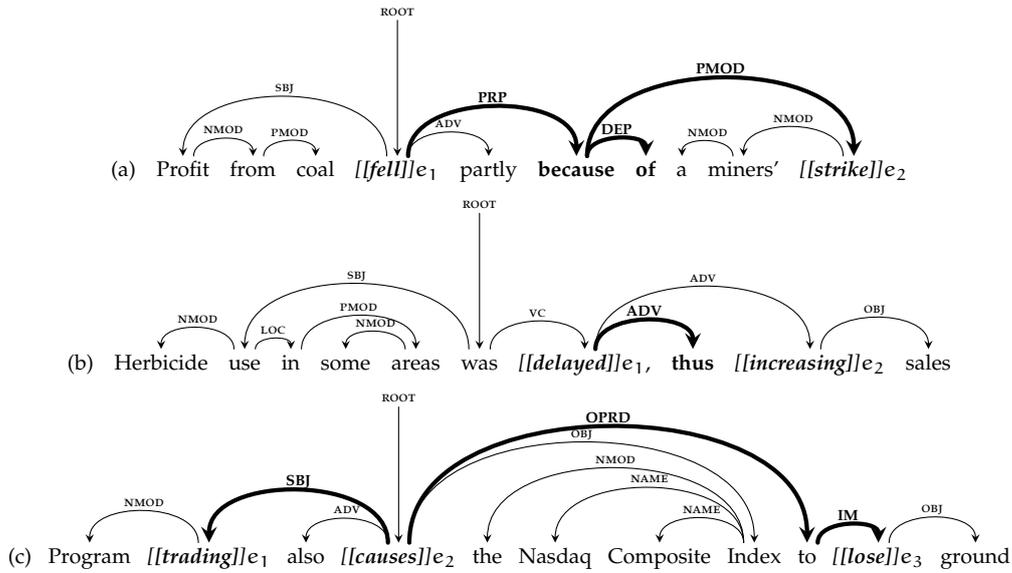

\centering
\scriptsize
\begin{dependency}[arc edge, arc angle=90, text only label, label style={above}]
\begin{deptext}[column sep=.5em]
(a) \& Profit \& from \& coal \& \entity{[fell]}{$e_1$} \& partly \& \textbf{because} \& \textbf{of} \& a \& miners' \& \entity{[strike]}{$e_2$} \\
\end{deptext}
\deproot{5}{\sc root}
\depedge{2}{3}{\sc nmod}
\depedge{3}{4}{\sc pmod}
\depedge{5}{2}{\sc sbj}
\depedge{5}{6}{\sc adv}
\depedge[edge style={ultra thick}]{7}{8}{\textbf{DEP}}
\depedge{10}{9}{\sc nmod}
\depedge{11}{10}{\sc nmod}
\depedge[edge style={ultra thick}]{5}{7}{\textbf{PRP}}
\depedge[edge style={ultra thick}]{7}{11}{\textbf{PMOD}}
\end{dependency}

\begin{dependency}[arc edge, arc angle=90, text only label, label style={above}]
\begin{deptext}[column sep=.5em]
(b) \& Herbicide \& use \& in \& some \& areas \& was \& \entity{[delayed]}{$e_1$}, \& \textbf{thus} \& \entity{[increasing]}{$e_2$} \& sales \\
\end{deptext}
\deproot{7}{\sc root}
\depedge{3}{2}{\sc nmod}
\depedge{3}{4}{\sc loc}
\depedge{6}{5}{\sc nmod}
\depedge{7}{8}{\sc vc}
\depedge[edge style={ultra thick}]{8}{9}{\textbf{ADV}}
\depedge{10}{11}{\sc obj}
\depedge{4}{6}{\sc pmod}
\depedge{7}{3}{\sc sbj}
\depedge{8}{10}{\sc adv}
\end{dependency}

\begin{dependency}[arc edge, arc angle=90, text only label, label style={above}]
\begin{deptext}[column sep=.3em]
(c) \& Program \& \entity{[trading]}{$e_1$} \& also \& \entity{[\textbf{causes}]}{$e_2$} \& the \& Nasdaq \& Composite \& Index \& to \& \entity{[lose]}{$e_3$} \& ground \\
\end{deptext}
\deproot{5}{\sc root}
\depedge{3}{2}{\sc nmod}
\depedge{5}{4}{\sc adv}
\depedge[edge style={ultra thick}]{5}{3}{\textbf{SBJ}}
\depedge{9}{8}{\sc name}
\depedge{9}{7}{\sc name}
\depedge{9}{6}{\sc nmod}
\depedge{5}{9}{\sc obj}
\depedge[edge style={ultra thick}]{5}{10}{\textbf{OPRD}}
\depedge[edge style={ultra thick}]{10}{11}{\textbf{IM}}
\depedge{11}{12}{\sc obj}
\end{dependency}

\caption{Examples of common dependency paths between events and causal signals or causal verbs}
\label{fig:dep-path}
\end{figure}

We observed common dependency paths connecting events with causal signals or causal verbs, as illustrated in Figure~\ref{fig:dep-path}. The possible causal signals are identified using dependency path pattern matching, e.g. \textit{e}$\rightarrow${\sc prp}$\rightarrow${\sc dep} (because of), {\sc pmod}$\rightarrow$\textit{e} (because), and \textit{e}$\rightarrow${\sc adv} (thus). Pattern priority rules are employed, e.g. \textit{e}$\rightarrow${\sc prp}$\rightarrow${\sc dep} $>$ \textit{e}$\rightarrow${\sc adv}, to avoid extracting \textit{partly} (via \textit{e}$\rightarrow${\sc adv}) in sentence (a). We also consider PoS tags to avoid extracting \textit{increasing} (via \textit{e}$\rightarrow${\sc adv}) in sentence (b). The same method is applied to identify possible causal verbs, e.g. {\sc sbj}$\rightarrow$\textit{e} (causes) and {\sc oprd}$\rightarrow${\sc im}$\rightarrow$\textit{e} (causes).

In the end, we include as features:
\begin{itemize}[leftmargin=*,topsep=2pt]
\setlength{\itemsep}{2pt}
\setlength{\parskip}{0pt}
\setlength{\parsep}{0pt}
\item possible causal signals via dependency path pattern matching for $e_1$/$e_2$;
\item their positions w.r.t $e_1$/$e_2$ (\textit{before} or \textit{after}); 
\item possible causal verbs via dependency path pattern matching for $e_1$/$e_2$; and
\item their positions w.r.t $e_1$/$e_2$ (\textit{before} or \textit{after});
\end{itemize}

\subsection{Experiments}
\label{sec:expansion-clink-experiments}

\paragraph{Dataset} \textit{Causal-TimeBank} (Section~\ref{sec:annotating-causality-causal-timebank}) is used as the initial training set. The unlabelled dataset is collected from the EMM NewsBrief platform\footnote{\url{http://emm.newsbrief.eu}}, which performs multilingual news aggregation and analysis from news portals world-wide, updated every 10 minutes. The news are automatically clustered according to subjects. 

We collect 16 RSS feeds of subjects containing more than 15 news articles from the 'top stories' list in English from January 20 to January 21, 2015. For each subject, we select only the news published within the same hour, to ensure the high similarity between the news in the same cluster. The news articles are automatically fetched from the links listed in the RSS feeds, cleaned, and then converted into the annotation format required by the NewsReader NLP pipeline. In the end, we have 16 clusters of 15 documents each, for a total of 240 documents in the newly created corpus, i.e. \textit{EMM-clusters}.

\paragraph{CauseRelPro-beta} We first run a classification experiment aimed at comparing the performance of CauseRelPro-beta in a supervised setting with the existing system by \textcite{mirza-tonelli:2014:Coling} as the baseline. We adopt the same five-fold cross-validation setting and use only the Causal-TimeBank corpus as training and test set. The goal of this first experiment is to assess the impact of the new features presented in Section \ref{sec:expansion-clink-extraction}.


\begin{table}
\centering
\small
\begin{tabular}{lccc}
\hline
\textbf{System} & \textbf{P} & \textbf{R} & \textbf{F1}\\
\hline
\textcite{mirza-tonelli:2014:Coling} & 0.6729 & 0.2264 & 0.3388\\
\textbf{CauseRelPro-beta} & \textbf{0.5985} & \textbf{0.2484} & \textbf{0.3511}\\
\hline
\end{tabular}
\caption{\label{tab:clink} CauseRelPro-beta's micro-averaged scores.}
\end{table}

Table~\ref{tab:clink} shows the micro-averaged performance of the system, compared with the baseline. 
The main difference between CauseRelPro-beta and that reported in \textcite{mirza-tonelli:2014:Coling} is the elimination of the middle step in which causal signals are identified. This, together with the use of supersenses, contributes to increasing recall. However, using token-based features and having a specific step to label \texttt{CSIGNAL}s yield better precision. 

We also conduct some experiments by using different degrees of polynomial kernel (Table~\ref{tab:degrees}). Note that even though the best F1-score is achieved by using the 3$^{rd}$ degree of polynomial kernel, the best precision of 0.6337 is achieved with degree 4.

\begin{table}
\centering
\small
\begin{tabular}{lccc}
\hline
 & \textbf{P} & \textbf{R} & \textbf{F1}\\
\hline
2$^{nd}$ degree & 0.5031 & 0.2516 & 0.3354\\
3$^{rd}$ degree & 0.5985 & 0.2484 & 0.3511\\
4$^{th}$ degree & 0.6337 & 0.2013 & 0.3055\\
\hline
\end{tabular}
\caption{\label{tab:degrees} CauseRelPro-beta's micro-averaged scores using different degrees of polynomial kernel.}
\end{table}

\paragraph{}The best performance achieves only 0.3511 F1-score. We argue that this low performance is strongly dependent on the lack of training data. Moreover, the proportion of positive and negative examples in the training data is highly imbalanced. In our previous experiment with a 5-fold cross-validation setting, on average, only 1.14\% of event pairs in the training data is labelled with causal relations. In order to tackle this issue, we first implement a self-training approach using the EMM clusters as additional corpus, and then apply \texttt{CLINK} propagation. Each step is evaluated separately in the following paragraphs.

\paragraph{Self-training} In order to boost the training data with more \texttt{CLINK}s to learn from, we adopt a basic configuration of the self-training approach. CauserelPro-beta is trained using the whole Causal-TimeBank corpus, which has gold annotated events and \texttt{TLINK}s. Since the model will be used to label new data to be included in the training set, we give preference to a higher precision over recall, therefore we use the 4$^{th}$ degree of polynomial kernel in building the classifier, in line with the findings reported in Table \ref{tab:degrees}. 

\begin{table}[t]
\centering
\begin{adjustbox}{width=1\textwidth}
\begin{tabular}{lcccccc}
\hline
 & \textbf{TP} & \textbf{FP} & \textbf{FN} & \textbf{P} & \textbf{R} & \textbf{F1}\\
\hline
Causal-TimeBank & 79 & 53 & 239 & 0.5985 & 0.2484 & 0.3511\\
Causal-TimeBank + EMM-clusters & 95 & 54 & 223 & 0.6376 & 0.2987 & 0.4069\\
\textbf{Causal-TimeBank + EMM-clusters + prop. \texttt{CLINK}s} & 108 & 74 & 210 & \textbf{0.5934} & \textbf{0.3396} & \textbf{0.4320}\\
\hline
\end{tabular}
\end{adjustbox}
\caption{\label{tab:bootstrap} Impact of increased training data, using EMM-clusters and propagated (prop.) \texttt{CLINK}s, on system performances (micro-averaged scores). TP: true positives, FP: false positives and FN: false negatives.}
\end{table}

Then, we run the model on the EMM-clusters dataset, which contains events and \texttt{TLINK}s annotated by the NewsReader pipeline. New labelled event pairs are obtained this way. Note that we only consider the positive examples as additional training data, as we want to reduce the imbalance in the dataset. In the end, we have 324 additional event pairs labelled with \texttt{CLINK}/\texttt{CLINK-R}.

The impact of having more training data is evaluated in the same five-fold cross-validation setting as before. The Causal-TimeBank corpus is used again for evaluation. For each fold, we append the training data with the new labeled event pairs. A new classifier is trained with the new training data using the 3$^{rd}$ degree of polynomial kernel, this time to obtain the best possible F1. 

As shown in Table~\ref{tab:bootstrap}, the appended training corpus (Causal-TimeBank + EMM-clusters) improves micro-averaged F1-scores of the system by 5.57\% significantly (p $<$ 0.01).

\paragraph{\texttt{CLINK} Propagation} Given the EMM-clusters dataset annotated with \texttt{CLINK}s, our \texttt{CLINK} propagation algorithm is as follows:

\begin{algorithmic}[1]
 \FOR {every cluster $C$ in EMM-clusters corpus}
  \FOR {every news article $N$ in $C$}
   \STATE $clinks \leftarrow$ all event pairs labelled with \texttt{CLINK}/\texttt{CLINK-R} above confidence threshold
   \FOR {every event pair $(e_1, e_2)$ in $clinks$}
    \STATE Look for event pair ($ec_1$, $ec_2$) in other news article in $C$, where $ec_1$ co-refers with $e_1$ and $ec_2$ co-refers with $e_2$
    \IF {($ec_1$, $ec_2$) is not labelled with \texttt{CLINK}/\texttt{CLINK-R}}
     \STATE Label ($ec_1$, $ec_2$) with the label of $(e_1, e_2)$ (establish the propagated \texttt{CLINK})
    \ENDIF
   \ENDFOR
  \ENDFOR
 \ENDFOR
\end{algorithmic}

\begin{figure}
\centering
\includegraphics[width=0.55\textwidth]{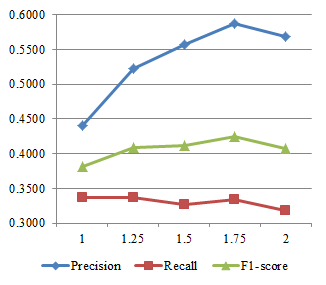}
\caption{Comparison of CauseRelPro-beta's micro-averaged scores with different values of confidence threshold in \texttt{CLINK} propagation.}
\label{fig:confidence}
\end{figure}

In order to decide which new \texttt{CLINK}s should be added to the training data, we did some experiments to select the confidence threshold returned by the classifier that maximizes the system's performance. Figure~\ref{fig:confidence} shows how the performance measures change at different cut-off values, ranging from 1 to 2, with 1.75 giving the best outcome. Using this threshold, we add 32 event pairs labelled with \texttt{CLINK}/\texttt{CLINK-R} to the training data.

We evaluate the performance of the system trained with the enriched training data from \texttt{CLINK} propagation in the same five-fold cross-validation setting as the previous experiment. As shown in Table~\ref{tab:bootstrap}, the micro-averaged F1-score increases by 8.1\% with respect to the initial configuration if we append propagated \texttt{CLINK}s as new labelled examples (Causal-TimeBank + EMM-clusters + prop. \texttt{CLINK}s). The improvement is statistically significant with p-value $<$ 0.001. As expected, the overall improvement is caused by the increased recall, even though the system precision drops.

As an example, we report below three text passages from different news in the same cluster of the EMM corpus. The \texttt{CLINK} in the first excerpt was correctly propagated to the other two, so that 2 additional event-pair instances were added to the training data. 

\begin{enumerate}[label=(\roman*)] 
\item U.S. Secretary of State John Kerry arrived in Nigeria on Sunday to urge its rival political camps to respect the outcome of a Feb. 14 presidential election, amid concerns that post-poll \eventattr{violence}{$e_1$} could \signal{undermine} the \eventattr{fight}{$e_2$} against Boko Haram militants.
\item Washington is concerned that post-poll \eventattr{violence}{$e_3$} could undermine the stability of Africa's top oil producer and \signal{hamper} efforts to \eventattr{tackle}{$e_4$} the Islamist militants of Boko Haram.
\item U.S. Secretary of State John Kerry arrived in the commercial capital Lagos on Sunday to urge the candidates and their supporters to respect the election outcome, underscoring U.S. concerns that post-poll \eventattr{violence}{$e_5$} could destabilise the country and \signal{undermine} the \eventattr{fight}{$e_6$} against Boko Haram.
\end{enumerate}

The event pair $(e_1, e_2)$ in (i) is labelled as having a causal relation with a high confidence score by the classifier trained with the Causal-TimeBank corpus. This is because there is a causal verb \textit{undermine}, which falls under the \texttt{PREVENT} causal type, connecting the two events.

Event pairs ($e_3$, $e_4$) and ($e_5$, $e_6$) in (ii) and (iii), respectively, are labelled with having causal relations through the \texttt{CLINK} propagation method, since $e_1$ co-refers with $e_3$ and $e_5$, and $e_2$ co-refers with $e_4$ and $e_6$.

If we apply the classification model learnt only from the Causal-TimeBank, instead, the event pair ($e_3$, $e_4$) is not labelled as having a causal relation, probably because there is no training instance including the causal verb \textit{to hamper}. Also the event pair ($e_5$, $e_6$) is not recognized as having a causal relation, probably because there is no direct syntactic connection between the causal verb \textit{undermine} and $e_5$ (\textit{violence}). 

\section{Conclusions}

\paragraph{Temporal Reasoning on Demand} We have presented an approach to improve temporal relation classification, by activating temporal reasoning on training data to improve not only the quantity but also the quality of the training data. However, the reasoning process, in particular deducing new \texttt{TLINK}s based on existing set of \texttt{TLINK}s, is only run when it is estimated to be effective, i.e. \textit{temporal reasoning on demand}. 

The result of a regression analysis showed that the number of \texttt{TLINK}s deducible from an annotated document can be estimated considering easy-to-measure parameters from the document. According to the experiments, deduction may be beneficial when the expected number of deducible \texttt{TLINK}s is inferior to a certain threshold, experimentally assessed. With this setting, removing inconsistent documents prior to deduction can also have a positive impact on classification performances. 

A first possible improvement may concern how to be more precise in estimating the number of deducible \texttt{TLINK}s, which in turns should lead to an improved prediction on when to run deduction. To this aim, one should increase the $R^2$ of the regression analysis, by finding further or different parameters that predict with a greater reliability the number of deducible \texttt{TLINK}s, and thus improving on the impact of deduction on classification performances.


Furthermore, as already mentioned, the current version of the reasoner identifies inconsistent documents as a whole, without providing any judgement at the level of temporal entities. Removing (only) inconsistent pairs is a very challenging task but with potential benefits for classification performances, which should be explored in the future (e.g., see \cite{mitra2006line} for a possible approach).

\paragraph{Semi-supervised Learning} Since the performance of the \texttt{CLINK} extraction system seems to be negatively affected by the scarcity of training examples, we proposed a self-training method to deal with this issue. Moreover, we proposed the causal link propagation method to further enrich the labelled data using event co-reference information in the news clusters dataset.

Our experiments show that self-training and \texttt{CLINK} propagation methods can significantly improve the performance of a \texttt{CLINK} extraction system, even if the semi-supervised learning is simplified (only one iteration is performed) and the unlabelled data contain only 240 documents. Despite the limited number of newly acquired training examples (324 through self-training and 32 through CLINK propagation), they still have a significant impact on the classifier performance, since the original training corpus contains only 318 causal links.

In the future, we would like to investigate the impact of our bootstrap method in a standard semi-supervised setting with many more unlabelled data and several iterations. 

\chapter{Multilinguality in Temporal Processing}\label{ch:multilinguality}
\begin{flushright}
\scriptsize
\textit{To have another language is to possess a second soul.} --- Charlemagne
\end{flushright}
\minitoc

\section{Introduction}
\label{multilinguality-intro}

Research on temporal information processing has been gaining a lot of attention from the NLP community in the recent years. However, most research efforts in temporal information processing have focused only on English.

TempEval-2, one of TempEval evaluation campaigns, attempted to address multilinguality in temporal information processing by releasing annotated TimeML corpora in 6 languages including English. The distribution over the six languages was highly uneven; out of 18 participating systems, only 3 were for Spanish, the rest were for English, and none for the other languages.

Apart from TempEval, HeidelTime\footnote{\url{http://heideltime.ifi.uni-heidelberg.de/heideltime/}}, a multilingual, domain-sensitive temporal tagger currently contains hand-crafted resources for 13 languages. In addition, the most recent version contains automatically created resources for more than 200 languages. 

In this chapter, we focus on temporal information processing for two languages other than English: Italian and Indonesian. For Italian, there was the EVENTI challenge for temporal information processing of Italian texts, which provides us a framework to evaluate our temporal information processing system for Italian. For Indonesian, our extension effort is only for the temporal expression extraction. This is because this task, especially timex normalization, typically requires a rule-engineering approach unlike event extraction or temporal relation extraction, which are commonly approached with data-driven methods. This will be the first step towards a complete temporal information processing for the Indonesian language.

\section{Related Work}
\label{multilinguality-related-work}

The second instalment of TempEval evaluation campaigns (Section~\ref{sec:tempeval}), TempEval-2~\parencite{verhagen-EtAl:2010:SemEval}, extended the first TempEval with a multilingual task. In addition to English, the organizers released TimeML annotated corpora in 5 other languages: Chinese, Italian, French, Korean and Spanish. All corpora include timex and event annotation in TimeML standard. However, not all corpora  contain data for all subtasks related to temporal relation extraction.

From the eighteen system runs submitted to TempEval-2, sixteen were for English, one for Spanish, i.e. UC3M for timex extraction, and two for both English and Spanish, i.e. TIPSem and TIPSem-B~\parencite{llorens-saquete-navarro:2010:SemEval}. 

For temporal expression extraction, HeidelTime~\parencite{stroetgen2014} is perhaps the temporal expression tagging system covering the most languages. HeidelTime currently understands documents in 11 languages, including English, German, Dutch, Vietnamese, Arabic, Spanish, Italian, French, Chinese, Russian, and Croatian. Even though the most recent work by \textcite{strotgen-gertz:2015:EMNLP} presented an automatic extension approach to cover around 200+ languages in the world, we believe that for low-resource (and less-explored) languages such as Indonesian, a manual extension effort is still required.

The recent work on temporal expression tagging for Indonesian documents~\parencite{simamora2013} only covers temporal expressions of \texttt{DATE} type. Moreover, the annotated documents are not in the TimeML annotation format, which is the widely used annotation format for temporal expression tagging. As far as we know, we are the first to implement a system for annotating temporal expressions in Indonesian documents with the TimeML format.

\section{Related Publications}

In \textcite{mirza-minard:2014:Evalita}, we summarized our attempts and approaches in building a complete extraction system for temporal expressions, events, and temporal relations in Italian documents, which participated in the EVENTI challenge. 

In \textcite{mirza:2015:Pacling}, we presented an automatic system for recognizing and normalizing the value of temporal expressions in Indonesian texts.

\section{Italian}


EVENTI\footnote{\url{https://sites.google.com/site/eventievalita2014/}}, one of the new tasks of Evalita 2014\footnote{\url{http://www.evalita.it/2014}}, was established to promote research in temporal information processing for Italian texts. Currently, even though there exist some independent modules for temporal expression extraction (e.g. HeidelTime \parencite{stroetgen2014}) and event extraction (e.g. \textcite{caselli2011}), there is no complete system for temporal information processing for Italian.

The main EVENTI task is composed of 4 subtasks for temporal expression (timex), event and temporal relation extraction from newspaper articles. The evaluation scheme follows the existing TempEval evaluation campaign for English (Section~\ref{sec:tempeval}), particularly TempEval-3. Additionally, a pilot task on temporal information processing of historical texts was also proposed.

Our temporal information extraction system, FBK-HLT-time, participated in both main task (for news articles) and pilot task (for historical texts), and was the only participant for the event extraction and temporal relation extraction tasks. For timex extraction in historical texts, FBK-HLT-time was the best performing system, showing that our approach is more robust to domain changes.

\subsection{Temporal Information Extraction System}
\label{sec:it-temp-info-extraction}
We developed an end-to-end temporal information extraction system to participate in the EVENTI evaluation campaign. It combines three subsystems: (i) time expression (timex) recognizer and normalizer (Section~\ref{sec:it-timex-extraction}), (ii) event extraction (Section~\ref{sec:it-event-extraction}) and (iii) temporal relation identification and classification (Section~\ref{sec:it-temporal-relation-extraction}). These subsystems have been first developed for English as part of the NewsReader project\footnote{\url{http://www.newsreader-project.eu/}} and then adapted to Italian. 

\paragraph{Data} We used the Ita-TimeBank released by the task organizers of EVENTI-Evalita 2014, containing 274 documents and around 112,385 tokens in total, for developing purposes. For the final end-to-end system submitted to EVALITA, we use this corpus as the training data for our classification models.

\paragraph{Tools and Resources} Several tools were used in developing our end-to-end system, especially for timex and temporal relation extraction:
\begin{itemize}
\item TextPro\footnote{\url{http://textpro.fbk.eu/}} \parencite{PIANTA08.645}, a suite of NLP tools for processing English and Italian texts. Among the modules we specifically use: lemmatizer, morphological analyzer, part-of-speech tagger, chunker, named entity tagger and dependency parser. 
\item YamCha\footnote{\url{http://chasen.org/~taku/software/yamcha/}}, a text chunker which uses SVMs algorithm. YamCha supports the dynamic features that are decided dynamically during the classification. It also supports multi-class classification using either \textit{one-vs-rest} or \textit{one-vs-one} strategies.
\item TimeNorm\footnote{\url{http://github.com/bethard/timenorm}}  \parencite{bethard:2013:EMNLP}, a library for converting natural language expressions of dates and times into their normalized form, based on \textit{synchronous context free grammars}.
\end{itemize}
Moreover, we also exploit several external resources, such as the list of temporal signals extracted from the  annotated Ita-TimeBank corpus. \textcite{mirza-tonelli:2014:EACL} show that the performance of their temporal relation classification system benefits from distinguishing event-related signals (e.g. \textit{mentre} [while], \textit{intanto} [in the meantime]) from timex-related signals (e.g. \textit{tra} [between], \textit{entro} [within]). Therefore we split the list of signals into two separate lists. Signals that are used in both cases (e.g. \textit{già} [before], \textit{dopo} [after], \textit{quando} [when]) are added to both lists. 

In the following sections we will explain our efforts in building a complete temporal information extraction system, excluding the event extraction module because we were not directly involved in its development.

\subsubsection{Timex Extraction System}
\label{sec:it-timex-extraction}

An automatic extraction system for temporal expressions typically consists of two modules, i.e., (i) timex extent recognition and type classification, and (ii) timex normalization. As has been discussed in Section~\ref{sec:timex-extraction}, for recognizing the timex extent in English texts, both data-driven and rule-based strategies are equally good; the statistical system ClearTK performed best at strict matching with 82.71\% F1-score. Meanwhile, the timex normalization task is currently done best by a rule-engineered system, TimeNorm, which achieves 81.6\% F1-score. 

\paragraph{Timex Extent and Type Identification} We decided to adopt the data-driven approach for recognizing the extent of a timex. However, unlike ClearTK, we combined both timex extent recognition and \texttt{type} classification\footnote{Temporal expressions are classified into 4 timex types in TimeML, i.e., \texttt{DATE}, \texttt{TIME}, \texttt{DURATION} and \texttt{SET}.} tasks as one text chunking task, using only one classification model. Since the extent of a timex can be expressed by a multi-word expression, we employ the BIO tagging to annotate the data. In the end, the classifier has to classify a token into 9 classes, including \texttt{B-DATE}, \texttt{I-DATE}, \texttt{B-TIME}, \texttt{I-TIME}, \texttt{B-DURATION}, \texttt{I-DURATION}, \texttt{B-SET}, \texttt{I-SET} and \texttt{O} (for other). 

The classification model is built using the Support Vector Machine (SVM) implementation provided by YamCha. The following features were defined to characterize a token:
\begin{itemize}
\item Token's text, lemma, part-of-speech (PoS) tags, flat constituent (noun phrase or verbal phrase), and the entity's type if the token is part of a named entity;
\item Whether a token matches regular expression patterns for unit (e.g. \textit{secondo} [second]), part of a day (e.g. \textit{mattina} [morning]), name of days, name of months, name of seasons, ordinal and cardinal numbers (e.g. \textit{31}, \textit{quindici} [fifteen], \textit{primo} [first]), year (e.g. \textit{'80}, \textit{2014}), time (e.g. \textit{08:30}), duration (e.g. \textit{1h3'}, \textit{50"}), adverbs (e.g. \textit{passato} [past], \textit{ieri} [yesterday]), names (e.g. \textit{Natale} [Christmas], \textit{Pasqua} [Easter]), or set (e.g. \textit{ogni} [every], \textit{mensile} [monthly]);
\item Whether a token matches regular expression patterns for SIGNAL (e.g. \textit{per} [for], \textit{dalle} [from]);
\item All of the above features for the preceding 2 and following 2 tokens, except the token's text;
\item The preceding 2 labels tagged by the classifier.
\end{itemize}


\begin{table}[t]
\centering
\begin{tabular} {lccc|ccc}
\hline
 & \multicolumn{3}{c|}{one-vs-one} & \multicolumn{3}{c}{one-vs-rest} \\
 & \textbf{P} & \textbf{R} & \textbf{F1} & \textbf{P} & \textbf{R} & \textbf{F1}\\
\hline
\texttt{DATE} & 0.781 & 0.739 & 0.759 & 0.827 & 0.806 & 0.816\\
\texttt{TIME} & 0.819 & 0.514 & 0.632 & 0.833 & 0.759 & 0.794\\
\texttt{DURATION} & 0.739 & 0.628 & 0.679 & 0.762 & 0.707 & 0.733\\
\texttt{SET} & 0.765 & 0.176 & 0.286 & 0.786 & 0.446 & 0.569\\
\hline
Overall & 0.776 & 0.667 & 0.717 & \textbf{0.813} & \textbf{0.766} & \textbf{0.789}\\
\hline
\end{tabular}
\caption{\label{tab:timex-detection}Classifier performance for timex extent and type identification with 5-fold cross-validation evaluation on the training data.}
\end{table}

In Table~\ref{tab:timex-detection} we report the classifier performance with 5-fold cross-validation scheme and a strict-match evaluation, comparing the one-vs-one method with one-vs-rest for multi-class classification. The one-vs-rest method gives better performance, especially for recognizing timex under the \texttt{SET} class.

\begin{figure*}[th!]
\small
\texttt{$[$Int:1Digit$]$ $|||$ due $|||$ 2 $|||$ 1.0\\
$[$Int:Hundred2Digit$]$ $|||$ cento $|||$ 0 0 $|||$ 1.0\\
$[$Int:3Digit$]$ $|||$ $[$Int:1Digit$]$ $[$Int:Hundred2Digit$]$ $|||$ $[$Int:1Digit$]$ $[$Int:Hundred2Digit$]$ $|||$ 1.0\\
$[$Period:Amount$]$ $|||$ $[$Int$]$ $[$Unit$]$ $|||$ Simple $[$Int$]$ $[$Unit$]$ $|||$ 1.0\\
$[$Period$]$ $|||$ $[$Period:Amount,1$]$ e $[$Period:Amount,2$]$ $|||$ Sum $[$Period:Amount,1$]$ $[$Period:Amount,2$]$ $|||$ 1.0\\
$[$TimeSpan:Regular$]$ $|||$ oggi $|||$ FindEnclosing PRESENT DAYS $|||$ 1.0\\
$[$TimeSpan:Regular$]$ $|||$ ieri $|||$ EndAtStartOf ( TimeSpan FindEnclosing PRESENT DAYS ) ( Period Simple 1 DAYS ) $|||$ 1.0\\}
\caption{Examples of the Italian grammar for TimeNorm}
\label{fig:timenorm-grammar}
\end{figure*}

\paragraph{Timex Value Normalization} For timex normalization, we decided to extend TimeNorm \parencite{bethard:2013:EMNLP} to cover Italian time expressions. For English, it is shown to have a better accuracy compared with other systems such as HeidelTime \parencite{strotgen2013} and TIMEN \parencite{LLORENS12.128.L12-1015}.

We translated and modified some of the existing rules of the English grammar into Italian. Figure \ref{fig:timenorm-grammar} presents some examples of the grammar for Italian. We also modified the TimeNorm code in order to support Italian language specificity:
\begin{itemize}
\item Normalizing accented letters, e.g. \textit{più} [more], \textit{ventitré} [twenty three].
\item Handling the token splitting for Italian numbers, because unlike in English, in Italian there is no space between digits, e.g. \textit{duemilaquattordici} [two thousand fourteen].
\item Detecting all forms (i.e. feminine, masculine, singular, plural) of articles (e.g. \textit{un'}, \textit{la}, \textit{gli}) or the combination of prepositions and articles (e.g. \textit{del}, \textit{alla}, \textit{dalle}, \textit{nei}), and convert them into a unified form. For example, \textit{del}, \textit{dello}, \textit{della}, \textit{dei}, \textit{degli} and \textit{delle} are all replaced with \textit{del}. This step is necessary for building a concise and simpler grammar.
\end{itemize} 

To process the annotated timex in TimeML format, some pre-processing and post-processing steps are needed before and after the normalizing process by TimeNorm. 

The pre-processing rules treat time expressions composed by only one number of one or two digits, and append a \textit{unit} or \textit{name of month}, which is inferred from closed timex (e.g. [ore 17$_{timex}$] - [23$_{timex}$] $\rightarrow$ [ore 23$_{timex}$]) or the document creation time (e.g. \textit{Siamo partiti il [7$_{timex}$]} ({\small \texttt{DCT=2014-09-23 tid="t0"}}) $\rightarrow$ [7 settembre$_{timex}$]).

TimeNorm returns a list of all possible values for given timex and anchor time. We defined a set of post-processing rules in order to select one of the returned values, that is most consistent with the timex type.  For example, if the timex is of type \texttt{DURATION}, the system selects the value starting with \texttt{P} (for Period of time). 

During the development of the grammar for Italian we noticed that the TimeNorm grammar does not support the normalization of the \textit{semester} or \textit{half-year} unit (e.g. \textit{il primo semestre} [the first semester]). 
We developed another set of  post-processing rules in order to cope with this issue. Despite all that, some expressions still cannot be normalized because they are too complex, e.g. \textit{ultimo trimestre dell'anno precedente} [last quarter of the previous year] or \textit{primi mesi dell'anno prossimo} [first months of next year].

If we always assume that the anchor time used for the timex normalization task is the document creation time (DCT), the system yields 0.753 accuracy on the training data. Note that the result is heavily biased by the fact that the system is tailored to improve its performance using the same corpus.

\paragraph{Empty Timex Identification}

The It-TimeML annotation guidelines adopted for the EVENTI task allow the creation of empty \texttt{TIMEX3} tags, whenever a temporal expression can be inferred from a text-consuming one.
For example, for the expression \textit{un mese fa} [one month ago], two \texttt{TIMEX3} tags are annotated: (i) one of type \texttt{DURATION} that strictly corresponds to the duration of one month (\texttt{P1M}) and (ii) one of type \texttt{DATE} that is not text consuming, referring to the date of one month ago (with the DCT as the anchor time).

As these timexes are not overtly expressed they cannot be discovered by the text chunking approach. We performed the recognition of the empty timexes using a set of simple post-processing rules and the output of the timex normalization module. 

\subsubsection{Event Extraction System}
\label{sec:it-event-extraction}

The subsystem used for the event extraction task is reported in \textcite{mirza-minard:2014:Evalita}, and was mainly developed by Anne-Lyse Minard from the Human Language Technology Group at the Fondazione Bruno Kessler, Trento, Italy.

\subsubsection{Temporal Relation Extraction System}
\label{sec:it-temporal-relation-extraction}

An automatic extraction system for temporal relations is typically composed of two modules for (i) temporal relation identification and (ii) temporal relation type classification. As has been discussed in Section~\ref{sec:temp-rel-extraction}, for English language, the TempEval-3 participants approached the task (i) with rule-based (e.g., UTTime), data-driven (e.g., ClearTK) and also hybrid methods (e.g., NavyTime). Meanwhile, for (ii), all participating systems resort to data-driven approaches.

\paragraph{Temporal Relation Identification} In the EVENTI task, the task of temporal link identification is restricted to \textit{event-event} (E-E) and \textit{event-timex} (E-T) pairs within the same sentence. We decided to adopt the hybrid approach for this task. First, we considered all combinations of E-E and E-T pairs within the same sentence (in a forward manner) as candidate temporal links. For example, if we have a sentence with entity order such as ``...$ev_1$...$ev_2$...$tmx_1$...$ev_3$...'', the candidate pairs are ($ev_1$, $ev_2$), ($ev_1$, $tmx_1$), ($ev_1$, $ev_3$), ($ev_2$, $tmx_1$), ($ev_2$, $ev_3$) and ($ev_3$, $tmx_1$).

Next, in order to filter the candidate temporal links, we trained a classifier to decide whether a given E-E or E-T pair is considered as having a temporal link (\texttt{REL}) or not (\texttt{O}). The classification models are built in the same way as in classifying the temporal relation types, using the same set of features, which will be explained in the following section.

\paragraph{Temporal Relation Type Classification}

A classification model is trained for each type of entity pair (E-E and E-T), as suggested in several previous works \parencite{mani-EtAl:2006:COLACL, chambers:2013:SemEval-2013}. Again, YamCha is used to build the classifiers. However, this time, a feature vector is built for each pair of entities $(e_1, e_2)$ and not for each token as in the previously mentioned temporal entity extraction tasks. 

Given an ordered pair of entities $(e_1, e_2)$ 
that could be either event/event or event/timex pair, 
the classifier has to assign a certain label, i.e., one of the 13 TimeML temporal relation types: \texttt{BEFORE}, \texttt{AFTER}, \texttt{IBEFORE}, \texttt{IAFTER}, \texttt{INCLUDES}, \texttt{IS\_INCLUDED}, \texttt{MEASURE}, \texttt{SIMULTANEOUS}, \texttt{BEGINS}, \texttt{BEGUN\_BY}, \texttt{ENDS}, \texttt{ENDED\_BY} and \texttt{IDENTITY}. The overall approach is largely inspired by an existing framework for the classification of temporal relations in English documents \parencite{mirza-tonelli:2014:EACL}. The implemented features are as follows:
\begin{itemize}
\item \textit{String and grammatical features}. Tokens, lemmas, PoS tags and NP-chunks of $e_1$ and $e_2$, along with a binary feature indicating whether $e_1$ and $e_2$ have the same PoS tags (only for event/event pairs).
\item \textit{Textual context}. Pair order (only for event/timex pairs, i.e. event/timex or timex/event), textual order (i.e. the appearance order of $e_1$ and $e_2$ in the text) and entity distance (i.e. the number of entities occurring between $e_1$ and $e_2$).
\item \textit{Entity attributes}. Event attributes (\textit{class}, \textit{tense}, \textit{aspect} and \textit{polarity})~\footnote{The event attributes \textit{tense}, \textit{aspect} and \textit{polarity} have been annotated using rules based on the EVENTI guidelines and using the morphological analyses of each token.}, and timex \textit{type} attribute~\footnote{The \textit{value} attribute tends to decrease the classifier performance as shown in \textcite{mirza-tonelli:2014:EACL}, and therefore, it is excluded from the feature set.} of $e_1$ and $e_2$ as specified in TimeML annotation. Four binary features are used to represent whether $e_1$ and $e_2$ have the same event attributes or not (only for event/event pairs).
\item \textit{Dependency information}. Dependency relation type existing between $e_1$ and $e_2$, dependency order (i.e. \textit{governor-dependent} or \textit{dependent-governor}), and binary features indicating whether $e_1$/$e_2$ is the \textit{root} of the sentence.
\item \textit{Temporal signals}. We take into account the list of temporal signals mentioned in Section~\ref{sec:it-temp-info-extraction}. Tokens of temporal signals occurring around $e_1$ and $e_2$ and their positions with respect to $e_1$ and $e_2$ (i.e. \textit{between} $e_1$ and $e_2$, \textit{before} $e_1$, or at the beginning of the sentence) are used as features.
\end{itemize}

In order to provide the classifier with more data to learn from, we bootstrap the training data with inverse relations (e.g., \texttt{BEFORE}/\texttt{AFTER}). By switching the order of the entities in a given pair and labelling the pair with the inverse relation type, we roughly double the size of the training corpus.

\begin{table}[t]
\centering
\begin{tabular} {lccc}
\hline
\textbf{System runs} & \textbf{P} & \textbf{R} & \textbf{F1}\\ \hline 
Run1 & 0.405 & 0.394 & 0.399\\ 
Run2 & 0.411 & 0.394 & 0.402\\ 
Run3 & 0.343 & 0.329 & 0.335\\
\hline
\end{tabular}
\caption{\label{tab:temprel-classification}Experiment results on the temporal relation type classification task, with 5-fold cross-validation evaluation on the training data.}
\end{table}

We evaluate the system's performance in classifying the temporal relation types with 5-fold cross-validation scheme on the training data. The evaluation scores are computed using the scorer\footnote{\url{http://sites.google.com/site/eventievalita2014/data-tools}} provided by EVENTI organizers. Table \ref{tab:temprel-classification} shows the evaluation results of each following system run:
\begin{itemize}
\item Run1: Two classifiers are used to determine the temporal relation types of E-E and E-T pairs.
\item Run2: The same as Run1, but we only consider the frequent relation types as classes for E-E pairs, meaning that we discarded E-E pairs under \texttt{IBEFORE}/\texttt{IAFTER}, \texttt{BEGINS}/ \texttt{BEGUN\_BY}, and \texttt{ENDS}/\texttt{ENDED\_BY} classes in building the classifier for E-E pairs.
\item Run3: The same as Run2, but we tried to incorporate the \texttt{TLINK} rules for E-T pairs which conforms to specific signal patterns as explained in the task guidelines\footnote{\url{http://sites.google.com/site/eventievalita2014/file-cabinet/specificheEvalita_v2.pdf}}. For example, \texttt{EVENT} + \textit{da/dalle/dal/dai/dall’} + \texttt{type=DATE} $\rightarrow$ \texttt{relType=BEGUN\_BY}. The E-T pairs matching the patterns are automatically assigned with relation types according to the rules, and do not need to be classified.
\end{itemize}

The number of training instances of E-E pairs under the \texttt{IBEFORE}/\texttt{IAFTER}, \texttt{BEGINS}/ \texttt{BEGUN\_BY}, and \texttt{ENDS}/\texttt{ENDED\_BY} classes is so few that removing them from the training corpus resulted in a slightly improved performance. Even when these classes are included in the training corpus, the classifier will still fail to classify the E-E pairs into these classes due to the heavily skewed dataset.

Similar phenomenon happens with E-T pairs. The classifier tends to classify an E-T pair into only three classes: \texttt{IS\_INCLUDED}, \texttt{INCLUDES} (when the pair order is timex/event) or \texttt{MEASURE}. We try to address this issue by incorporating the \texttt{TLINK} rules based on \texttt{EVENT}-signal-\texttt{TIMEX3} patterns listed in the task guidelines. Unfortunately, this solution does not help improving the system, perhaps because the rules were not strictly followed in the annotation process.

These phenomena of imbalanced dataset can also be observed in English TimeML corpora. For English, in the case of E-E pairs, we collapsed \texttt{IBEFORE}/\texttt{IAFTER} and \texttt{DURING}/ \texttt{DURING\_INV} relations into \texttt{BEFORE}/\texttt{AFTER} and \texttt{SIMULTANEOUS}, resp; and built a rule-based module to extract possible \texttt{BEGINS}/\texttt{BEGUN\_BY} and \texttt{ENDS}/\texttt{ENDED\_BY} relations. Note that compared with English, there is a slight difference in It-TimeML annotation guidelines, namely the introduction of the \texttt{TLINK} type \texttt{MEASURE} for event-timex pairs when the timex is of \texttt{DURATION} type.

\subsection{EVENTI (Evalita 2014) Evaluation}

Following the TempEval-3 evaluation scheme, the EVENTI task included 4 subtasks:
\begin{itemize}
\item Task A: Determine the extent and the normalization of temporal expressions according to the TimeML \texttt{TIMEX3} tag. Empty \texttt{TIMEX3} tags, as specified in It-TimeML annotation guidelines, will be taken into account as well.
\item Task B: Determine the extent and the \texttt{class} attribute value of events according to the TimeML \texttt{EVENT} tag.
\item Task C: Determine temporal relations from raw text. This involves performing Task A and Task B, and subsequently identifying pairs of temporal entities connected by a temporal relation (\texttt{TLINK}) and classifying their relation types.
\item Task D: Determine the temporal relation types given gold annotated pairs of temporal entities.
\end{itemize}

\paragraph{Dataset} The training data is the Ita-TimeBank released by the task organizers, containing 274 documents and around 112,385 tokens in total. For the evaluation stage, the organizers released two test corpora:
\begin{itemize}
\item Main task corpus, containing 92 documents from Ita-TimeBank.
\item Pilot task corpus, containing 10 documents of historical texts published in ``Il Trentino'' newspaper by the Italian statesman A. De Gasperi in 1914.
\end{itemize}

\paragraph{Evaluation Results} Table \ref{tab:eventi} shows the results of our system, FBK-HLT-time, on the two tasks of the EVENTI challenge, i.e. the main task (MT) and the pilot task (PT), and on the 4 subtasks. For the pilot task we report only the results obtained with the best system runs. For Task A, there were 3 participants and 6 unique runs in total; we also compare our system with HT 1.8, the extended version of HeidelTime~\parencite{stroetgen2014}, which achieved the highest score for the timex normalization (for the main task). For Task B, C and D, FBK-HLT-time was the only participant.

\begin{table}[t]
\begin{adjustbox}{width=1\textwidth}
\begin{tabular}{lllcccc|ccc}
\hline
\textbf{Subtask} & \textbf{Task} & \textbf{Run} & \textbf{F1} & \textbf{R} & \textbf{P} & \textbf{strict F1} & \textbf{\texttt{type} F1} & \textbf{\texttt{value} F1} & \textbf{\texttt{class} F1} \\           
\hline
Task A &\textbf{MT} & R1 & 0.886 & 0.841 & 0.936 & \textbf{0.827} & \textbf{0.800} & 0.665 \\
 & & HT 1.8 & \textbf{0.893} & 0.854 & 0.935 & 0.821 & 0.643 & \textbf{0.709} \\
\hdashline
 & \textbf{PT} & R1& \textbf{0.870} & 0.794 & 0.963 & \textbf{0.746} & \textbf{0.678} & \textbf{0.475} \\
 & & HT 1.8 & 0.788 & 0.691 & 0.918 & 0.671 & 0.624 & 0.459 \\
\hline
Task B & \textbf{MT}  & R1 & \textbf{0.884} & 0.868      & 0.902 & \textbf{0.867} &  &  & \textbf{0.671} \\
 & & R2 & 0.749 & 0.632 & 0.917 & 0.732 &              &  & 0.632 \\
 & & R3 & 0.875 & 0.838 & 0.915 & 0.858 &  &  & 0.670 \\
\hdashline                 
& \textbf{PT} & R1 & \textbf{0.843} & 0.793 & 0.900        & \textbf{0.834} &  &  & 0.604 \\
\hline
Task D & \textbf{MT} & R1 & 0.736 & 0.731 & 0.740 & 0.731  \\
& & R2 & 0.419 & 0.541 & 0.342 & 0.309  \\
& & \textit{R2*} & {\bf \textit{0.738}} & \textit{0.733} & \textit{0.742} & {\bf \textit{0.733}}  \\
\hdashline  
& \textbf{PT} & \textit{R1 \& R2} & {\bf \textit{0.588}} & \textit{0.588} & \textit{0.588} & {\bf \textit{0.570}} \\
\hline
Task C & \textbf{MT} & Ev R1 / Tr R1 & \textbf{0.264} & 0.238 & 0.296 & \textbf{0.341} \\
& & Ev R1 / Tr R2 & 0.253 & 0.241 & 0.265 & 0.325 \\
& & Ev R2 / Tr R1 & 0.209 & 0.167 & 0.282 & 0.267 \\
& & Ev R2 / Tr R2 & 0.203 & 0.168 & 0.255 & 0.258 \\
& & Ev R3 / Tr R1 & 0.247 & 0.211 & 0.297 & 0.327 \\
& & Ev R3 / Tr R2 & 0.247 & 0.211 & 0.297 & 0.327 \\
\hdashline  
&\textbf{PT} & Ev R1 / Tr R1 & \textbf{0.185} & 0.139 & 0.277 & \textbf{0.232} \\
\hline
\end{tabular}
\end{adjustbox}
\caption{FBK-HLT-time results on EVENTI (Evalita 2014) \small{(MT: Main Task; PT: Pilot Task; Ev Rn: run n of Task B; Tr Rn: run n of Task D)}. HT 1.8, extended HeidelTime, is one of competing participants.\label{tab:eventi}}
\end{table}

\paragraph{Task A: Timex Extraction} For recognizing the extent of timex in news articles, the system achieves 0.827 F1-score using strict-match scheme. The performances for determining the timex \texttt{type} and determining the timex \texttt{value} (timex normalization) are 0.8 F1-score and 0.665 F1-score, respectively. The performance for timex normalization is still considerably lower than the state-of-the-art system for English (TimeNorm), with 81.6\% F1-score. This suggests that the TimeNorm adaptation for Italian can still be improved, for example by including \textit{semester} or \textit{half-year} as a unit.

For the pilot task, in recognizing the extent of timex, the system achieves comparable scores with the main task. However, in determining the timex \texttt{type} and \texttt{value}, the accuracies drop considerably. With the assumption that the articles written with a gap of one century differ more at the lexical level than at the syntactic level, our take on this phenomena is that in recognizing the extent of timex, the system depends more on the syntactic features. Meanwhile, in determining the timex \texttt{type} and \texttt{value}, the system relies more on the lexical/semantic features and so the performances of the system decrease when it is applied to historical texts.

Compared with other participants, for the main task, our system performed best in strict matching and in classifying the timex types. A rule-engineering system, HT 1.8, which extended HeidelTime, performed better in relaxed matching\footnote{This is in line with the reported results for the timex extraction task for English texts in TempEval-3 (Section~\ref{sec:timex-extraction}).} with 0.893 F1-score, and in timex normalization with 0.709 F1-score. However, our system performed best for the pilot task, showing that our approach is more capable of domain adaptation.

\paragraph{Task B: Event Extraction} We observed that event classification performed better with the one-vs-one multi-class strategy (Run1), with a strict F1-score of 0.867 for event detection and an F1-score of 0.671 for event classification, than with the one-vs-rest one (Run2). Looking at the number of predicted events with both classifiers, the second classifier did not classify all the events found (1036 events were not classified). For this reason the precision is slightly better but the recall is much lower. 

On the pilot task data the results are a bit lower, with a strict F1-score of 0.834 for event detection and an F1-score of 0.604 for event classification. Note that for Run 3 we re-trained the model only on 80\% of the data due to a problem while training the model on all the training data.

\paragraph{Task D: Temporal Relation Extraction} The two runs submitted to EVALITA for this subtask, Run1 and Run2, corresponds to Run2 and Run3 explained in Section~\ref{sec:it-temporal-relation-extraction}, respectively. For the main task, there was a slight error in the format conversion for Run 2. Hence, we recomputed the scores of \textit{Run 2*} independently, which results in a slightly better performance compared with Run 1. The system (\textit{Run 2*}) yields 0.738 F1-score using TempEval-3 evaluation scheme.

For the pilot task (post-submission evaluation), both Run 1 and Run 2 have exactly the same F1-scores, i.e. 0.588. This suggests that in the pilot data there is no E-T pair matching the \texttt{EVENT}-signal-\texttt{TIMEX3} pattern rules listed in the task guidelines. Similar to the classification of timex types, the classifiers tend to rely more on lexical/semantic features, hence, the system performances decrease when they are applied on historical texts.

As the dataset is heavily skewed, we have decided to reduce the set of temporal relation types. It would be interesting to see if using patterns or trigger lists as a post-processing step can improve the system in the detection of the under-represented relations. For example, the relation type \texttt{IAFTER} (as a special case of the relation \texttt{AFTER}) can be recognized through the adjective \textit{immediato} [immediate].  

\paragraph{Task C: Temporal Awareness} This task involves performing Task A and Task B, and subsequently identifying pairs of temporal entities having a \texttt{TLINK} and classifying their temporal relation types (Task D). For this task, we combine the timex extraction system, the 3 system runs for event extraction (Ev), the system for identifying temporal links, and the 2 system runs for classifying temporal relation types (Tr). 

We found that for both main task and pilot task, the best performing system is the combination of the best run of task B (Ev Run 1) and the best run of task D (Tr Run 1), with 0.341 F1-score and 0.232 F1-score respectively (strict-match evaluation).

\section{Indonesian}





We propose a rule-based system for recognizing and normalizing temporal expressions for Indonesian documents. For normalizing temporal expressions, we extend an existing normalizer for English, TimeNorm~\parencite{bethard:2013:EMNLP}. We report some modifications of the tool required for Indonesian language, with respect to the different characteristics of Indonesian temporal expressions compared with English.

For recognizing (and determining the types of) temporal expressions, we build a finite state transducer heavily influenced by the crafted TimeNorm's time grammar for Indonesian. Even though it is shown that the machine learning approach can be as good as rule-engineering for recognizing temporal expressions, since there is no available Indonesian TimeML corpus yet, we resort to the rule-based approach. We believe that annotating sufficient data for the machine learning approach is more time-consuming than hand-crafting a transducer.

The evaluation is done on 25 news articles, containing 9,549 tokens (comparable with the TempEval-3 evaluation corpus with 9,833 tokens). The system yields 92.87\% F1-score in recognizing temporal expressions and 85.26\% F1-score in normalizing them.

\subsection{Challenges in Indonesian}

Bahasa Indonesia (or simply Indonesian) is the official language of Indonesia, which is the fourth most populous nation in the world. Of its large population, the majority speaks Indonesian, making it one of the most widely spoken languages in the world. Nevertheless, it is still highly under-represented in terms of NLP research. The lack of available annotated corpora makes it difficult to build (and evaluate) automatic NLP systems using data-driven approaches.

One of many characteristics that makes Indonesian different from other languages such as English or Italian is that the form of the verb does not change to indicate tense or aspect. A sentence ``\textit{Saya pergi ke kantor} [I go to office]'' carries no indication of whether the verb refers to a regular occurrence or to a single occurrence and, if the latter, when it happens in relation to the present. This is inferred from the context within which the utterance is made, by looking at either \textit{aspect markers} (e.g., \textit{sudah} [already], \textit{sedang} [in the process of], \textit{akan} [will]), or temporal expressions.

Regarding temporal expressions, there are several differences compared to English including, among others, the order of numbers in dates (e.g. \textit{3/21/2015} vs \textit{21/3/2015}), the punctuations used (e.g. \textit{5:30} vs \textit{05.30}, \textit{2.5} vs \textit{2,5}) and the fact that there are only two seasons (rainy or dry) known in Indonesia. Furthermore, since Indonesian is an agglutinative language, some of the temporal expressions of \texttt{DURATION} type contain affixes.

\subsection{Timex Extraction System}

The actual steps in the temporal expression (timex) extraction task are (i) recognizing the extent of a timex, (ii) determining its type, then (iii) normalizing the timex (resolving its value). However, during the development phase, we first develop the system to normalize temporal expressions based on an existing system for English. Then, based on the created time grammar, we develop a finite state transducer to do both recognizing temporal expressions' extents and determining their types in one step. In the following sections, we will organize the explanation of each module composing our timex extraction system, i.e., timex normalization, timex extent recognition and timex type classification, in such order.

The complete system, called \textit{IndoTimex}\footnote{\url{http://paramitamirza.ml/indotimex/}}, is implemented in Python and made available for download\footnote{\url{http://github.com/paramitamirza/IndoTimex}}. The system takes as input a TimeML document (or a collection of TimeML documents) and gives as output a TimeML document (or a collection of TimeML documents) annotated with temporal expressions (TIMEX3 tags).

\subsubsection{Timex Normalization}

\begin{figure}[t]
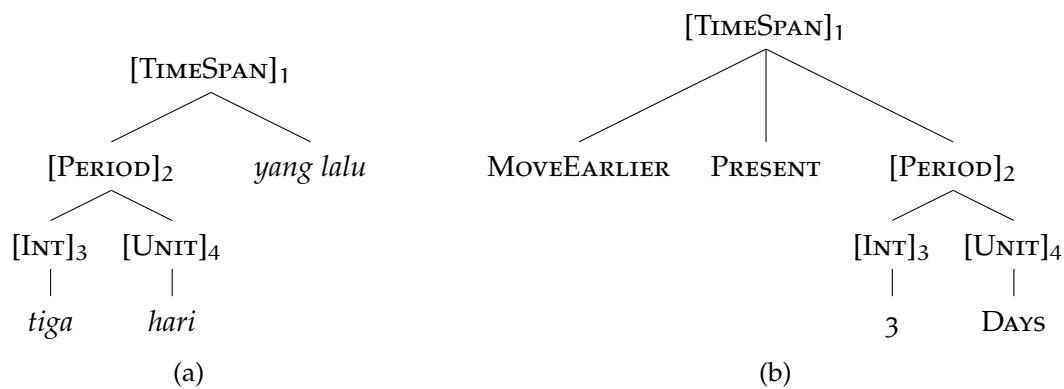

\begin{subfigure}[b]{0.4\textwidth}
\Tree [.\textsc{[TimeSpan]$_1$} [.\textsc{[Period]$_2$} [.\textsc{[Int]$_3$} \textit{tiga} ] [.\textsc{[Unit]$_4$} \textit{hari} ] ] \textit{yang lalu} ]
\caption{}
\label{time-parse-source}
\end{subfigure}
\begin{subfigure}[b]{0.6\textwidth}
\Tree [.\textsc{[TimeSpan]$_1$} \textsc{MoveEarlier} \textsc{Present} [.\textsc{[Period]$_2$} [.\textsc{[Int]$_3$} 3 ] [.\textsc{[Unit]$_4$} \textit{\textsc{Days}} ] ] ]
\caption{}
\label{time-parse-target}
\end{subfigure}
\caption{The synchronous parse from (a) the source language \textit{tiga hari yang lalu} [three days ago] to (b) the target formal time representation
\textsc{MoveEarlier}(PRESENT, \textsc{Simple}(3, \textsc{Days})). Subscripts on non-terminals indicate the alignment between the source and target parses.}
\label{time-parse}
\end{figure}

Temporal expressions in Indonesian language are quite similar with the ones in English. Therefore, for normalization, we decided to extend TimeNorm~\parencite{bethard:2013:EMNLP} to cover Indonesian temporal expressions. TimeNorm is a tool for normalizing temporal expressions based on a synchronous context free grammar, developed in Scala. Given an anchor time, TimeNorm parses time expressions and returns all possible normalized values of the expressions. A temporal expressions is parsed with an extended CYK+ algorithm, then converted to its normalized form by applying the operators recursively. The normalization value is determined as specified in TIDES(02).

The time grammar in TimeNorm, based on a synchronous context free grammar formalism, allows two trees (one in the source language and one in the target language) to be constructed simultaneously. Figure~\ref{time-parse} shows a synchronous parse for \textit{tiga hari yang lalu} [three days ago], where Figure~\ref{time-parse-source} is the source side (an Indonesian expression), Figure~\ref{time-parse-target} is the target side (a temporal operator expression), and the alignment is shown via subscripts. 

\begin{table}[t]
\centering
\begin{tabular}{rl}
\hline
\textbf{English (US)} & \textbf{Indonesian}\\
\hline
March (the) 21(st), 2015 & (tanggal) 21 Maret 2015\\
3/21/2015 & 21/3/2015\\
3-21-2015 & or 21-3-2015\\
\hdashline
'80s & tahun 80-an\\
1980s & tahun 1980-an\\
eighties & tahun delapan puluhan\\
\hdashline
21st century & abad ke-21\\
 & abad XXI\\
\hline
5:30 (am) & (pukul) 05.30\\
5:30 (pm) & (pukul) 17.30\\
\hdashline
1h 34' 56'' & 1.34.56 jam\\
2.5 hours & 2,5 jam\\
\hdashline
a \underline{\smash{year}} & se\underline{\smash{tahun}}\\
5 \underline{\smash{year}}s & 5 \underline{\smash{tahun}}\\
few \underline{\smash{year}}s & beberapa \underline{\smash{tahun}}\\
\underline{\smash{year}}s & ber\underline{\smash{tahun-tahun}}\\
\hline
\end{tabular}
\caption{Differences on expressing time in English and Indonesian.}
\label{diff}
\end{table}

Extending TimeNorm for a new language is very straightforward, we just need to translate the existing time grammar for English into Indonesian. However, there are some differences on expressing time in American English and Indonesian, as shown in Table~\ref{diff}. Therefore, several adjustments are required to cope with those differences, as well as to comply with the TIDES(02) standard:
 \begin{itemize}
 \item Dates are always in the Day-Month-Year order.
 \item Roman numerals are added since they are used in describing century (e.g. \textit{abad XVII} [17th century]).
 \item The expression for time is written with dot (.) instead of colon (:), and the same applies for time duration.
 \item The `am/pm' expression is not used since hours range from 0 to 24.
 \item Comma (,) is used as the decimal separator instead of dot (.).
 \item There is no distinction between plural and singular time units following quantifiers (e.g. \textit{tahun} [year] denotes both \textit{year} and \textit{years}).
 \item There are three time zones in Indonesia, namely \textit{WIB} (UTC+07:00), \textit{WITA} (UTC+08:00) and \textit{WIT} (UTC+09:00). In normalizing the temporal expression, we decided to ignore the time zones even though they are included in the extents.
 \item Indonesia has only two seasons, \textit{musim hujan} [rainy season] and \textit{musim kemarau} [dry season], which are not available in the standard. Hence, we normalize \textit{musim hujan} and \textit{musim kemarau} as \textsc{Winter} and \textsc{Summer} respectively.
 \item \textit{Sore} and \textit{petang} could mean both  `afternoon' and `evening'. We decided to normalize \textit{sore} as \textsc{Afternoon}, while \textit{petang} as \textsc{Evening}.
 \item The \textit{`\textsc{DayOfWeek} malam'} [\textsc{DayOfWeek} night] expression, can also be expressed with \textit{`malam \textsc{DayOfWeek}-after'}, e.g. \textit{malam Minggu} [night (of) Sunday] means \textit{Sabtu malam} [Saturday night]. A special rule is needed to handle this case, which is quite similar with the rule for `Christmas Eve' or `New Year's Eve'.
 \end{itemize}  

Apart from the grammar, there are several modifications of the TimeNorm code in order to support Indonesian temporal expressions:
\begin{itemize}
\item In Indonesian language, being an agglutinative language, some temporal expressions contain affixes. In the numerals, the prefix \textit{se-} when attached to a \textsc{Unit} (e.g. \textit{tahun} [year]) or a \textsc{PartOfDay} (e.g. \textit{pagi} [morning]) means one. Hence, \textit{se\underline{\smash{tahun}}} denotes a year and \textit{se\underline{\smash{pagi}}an} (with suffix \textit{-an}) a whole morning. Moreover, to make a \textsc{Unit} become plural, the prefix \textit{ber-} is added to the reduplicated \textsc{Unit}, e.g. \textit{ber\underline{\smash{jam-jam}}} [hours]. In order to have a concise grammar, we need to isolate the affixes from the root expressions before giving the temporal expressions to the parser.
\item The term \textit{minggu} is ambiguous, which could mean `week' (a \textsc{Unit}) or `Sunday' (a \textsc{DayOfWeek}). However, as in English, a \textsc{DayOfWeek} is always capitalised. Therefore, we disambiguate the term according to this rule before giving it to the parser.
\end{itemize}

\subsubsection{Recognizing Temporal Expressions and Determining The Types}

Based on the time grammar for TimeNorm, we construct regular expression rules to label tokens with \textsc{[Int]}, \textsc{[Unit]} or \textsc{[Field]}, e.g. \textit{hari} $\rightarrow$ \textsc{Unit}. The defined labels are as follows:
\begin{itemize}[leftmargin=*,topsep=2pt]
\setlength{\itemsep}{2pt}
\setlength{\parskip}{0pt}
\setlength{\parsep}{0pt}
\item \textsc{[Int:Numeral]}, e.g. \textit{satu} [one], \textit{puluh} [(times) ten]
\item \textsc{[Int:Digit]}, e.g. \textit{12}, \textit{1,5}, \textit{XVII} [17]
\item \textsc{[Int:Ordinal]}, e.g. \textit{ke-2} [2nd], \textit{ketiga} [third], \textit{ke XVII} [17th]
\item \textsc{[Unit]}, e.g. \textit{hari} [day], \textit{musim} [season]
\item \textsc{[Unit:Duration]}, e.g. \textit{setahun} [a year], \textit{berjam-jam} [hours]
\item \textsc{[Field:Year]}, e.g. \textit{'86}, \textit{2015}
\item \textsc{[Field:Decade]}, e.g. \textit{70-an} [70's], \textit{limapuluhan} [fifties]
\item \textsc{[Field:Time]}, e.g. \textit{08.30}, \textit{WIB}
\item \textsc{[Field:Date]}, e.g. \textit{10/01/2015}
\item \textsc{[Field:PartOfDay]}, e.g. \textit{pagi} [morning]
\item \textsc{[Field:DayOfWeek]}, e.g. \textit{Selasa} [Tuesday]
\item \textsc{[Field:MonthOfYear]}, e.g. \textit{Januari} [January]
\item \textsc{[Field:SeasonOfYear]}, e.g. \textit{kemarau} [dry], \textit{gugur} [autumn]
\item \textsc{[Field:NamedDay]}, e.g. \textit{Natal} [Christmas]
\end{itemize}

In expressing \textsc{Time} and \textsc{Date}, some tokens are commonly used before the temporal expression, which by themselves cannot be considered a temporal expression (e.g. \textit{\underline{\smash{pukul}} 08.30} [08:30], \textit{\underline{\smash{tanggal}} 10 Januari} [January 10]). Hence, we define labels for those tokens as follows:
\begin{itemize}[leftmargin=*,topsep=2pt]
\setlength{\itemsep}{2pt}
\setlength{\parskip}{0pt}
\setlength{\parsep}{0pt}
\item \textsc{[Pre:Time]}, i.e. \textit{pukul}
\item \textsc{[Pre:Date]}, i.e. \textit{tanggal}
\end{itemize}

Apart from \textsc{[Int]}, \textsc{[Unit]} and \textsc{[Field]}, some tokens can be considered a single temporal expression. Moreover, some tokens preceding or following \textsc{[Int]}, \textsc{[Unit]} and \textsc{[Field]} can be included in the temporal expression extent to further define the expression. Such tokens are labelled as follows:
\begin{itemize}[leftmargin=*,topsep=2pt]
\setlength{\itemsep}{2pt}
\setlength{\parskip}{0pt}
\setlength{\parsep}{0pt}
\item \textsc{[Date:Solo]}, e.g. \textit{dulu} [in the past], \textit{kini} [now]
\item \textsc{[Date:Begin]}, e.g. \textit{masa} [period], \textit{zaman} [times] (they are usually combined with other tokens, e.g. \textit{masa lalu} [the past], \textit{zaman sekarang} [nowadays])
\item \textsc{[Duration:Solo]}, e.g. \textit{sebentar} [for a while]
\item \textsc{[Quantifier]}, e.g. \textit{beberapa} [a few]
\item \textsc{[Modifier]}, e.g. \textit{sekitar} [around], \textit{penghujung} [the end of]
\item \textsc{[Current]}, e.g. \textit{ini} [this], \textit{sekarang} [now]
\item \textsc{[Earlier]}, e.g. \textit{kemarin} [yesterday], \textit{lalu} [last]
\item \textsc{[Later]}, e.g. \textit{besok} [tomorrow], \textit{mendatang} [next]
\item \textsc{[Set]}, e.g. \textit{setiap} [each], \textit{sehari-hari} [daily]
\end{itemize}

\begin{figure}[t]
\center
\begin{tikzpicture}
\tikzset{vertex/.style = {shape=circle,draw,minimum size=1em}}
\tikzset{edge/.style = {->,> = latex'}}

\node[initial,state] (0) at (0,0) {$q_0$};
\node[state] (1) at (3,0) {$q_1$};
\node[accepting,state] (2) at (6,0) {$q_2$};
\node[state] (3) at (9,0) {$q_3$};
\node[accepting,state] (4) at (12,0) {$q_4$};

\path[->] (0) edge [bend left] node [pos=0.5, above] {\scriptsize \textsc{[Int:Numeral]}:\textsc{date}} (1)
	(1) edge [bend left] node [pos=0.5, above] {\scriptsize \textsc{[Unit]}:\textsc{duration}} (2)
	(2) edge [bend left] node [pos=0.5, above] {\scriptsize yang:\textsc{date}} (3)
	(3) edge [bend left] node [pos=0.5, above] {\scriptsize \textsc{[Earlier]}:\textsc{date}} (4);
\path[every edge/.style={gray,draw=gray}]	
	(0) edge [bend right] node [pos=0.5, below] {\scriptsize \textsc{[Int:Digit]}:\textsc{date}} (1)
	(3) edge [loop above] node [pos=0.5, above] {\scriptsize akan:\textsc{date}} (3)
	(3) edge [bend right] node [pos=0.5, above] {\scriptsize \textsc{[Later]}:\textsc{date}} (4)
	(2) edge [bend right] node [pos=0.5, above] {\scriptsize \textsc{[Later]}:\textsc{date}} (4)
	(1) edge [bend right] node [pos=0.5, below] {\scriptsize \textsc{[Field:MonthOfYear]}:\textsc{date}} (4)
	(4) edge [loop below] node [pos=0.5, below] {\scriptsize \textsc{[Field:Year]}:\textsc{date}} (4);

\end{tikzpicture}
\caption{Part of the FST's transition diagram used to recognize \textit{tiga hari yang lalu} [three days ago] as a temporal expression of \textsc{date} type.}
\label{fst}
\end{figure}
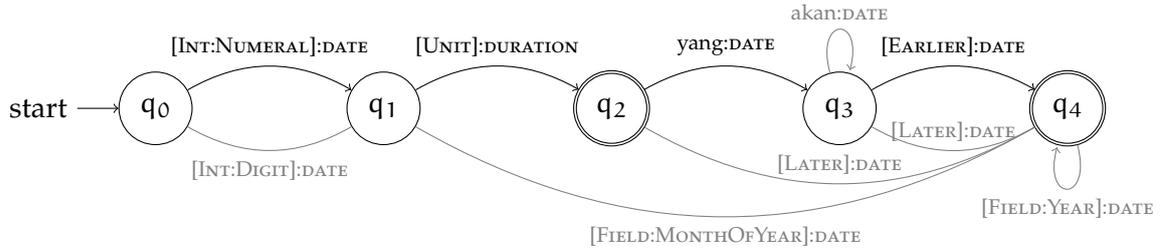

We then build a deterministic finite state transducer (FST) to recognize a temporal expression and to label it with one of the TIMEX3 types, i.e. \textsc{date}, \textsc{duration}, \textsc{time} and \textsc{set}. We define the FST $T = (Q,\Sigma,\Gamma,\delta,\omega,q_0,F)$ such that:
\begin{itemize}
\item $Q$ is a finite set of states;
\item $\Sigma$ as the \textit{input alphabet} is a finite set of previously defined token labels \{\textsc{[Int:Numeral]}, \textsc{[Int:Digit]}, ..., \textsc{[Later]}, \textsc{[Set]}\}  $\cup$ \{\textit{yang}, \textit{ke}, \textit{akan}, \textit{dan}\} (i.e. function words that are often used in temporal expressions);
\item $\Gamma$ as the \textit{output alphabet} is a finite set of temporal expression types \{\textsc{date}, \textsc{duration}, \textsc{time}, \textsc{set}\};
\item $\delta:Q\times\Sigma\rightarrow Q$ is the transition function;
\item $\omega:Q\times\Sigma\rightarrow \Gamma$ is the output function;
\item $q_0 \in Q$, is the start state;
\item $F \subseteq Q$, is the set of final states;
\end{itemize}

Figure~\ref{fst} shows a small part of the developed FST's transition diagram that is used to recognize the temporal expression \textit{tiga hari yang lalu} [three days ago] as of \textsc{date} type, with \textit{tiga} is initially labelled as \textsc{[Int:Numeral]}, \textit{hari} as \textsc{[Unit]} and \textit{lalu} as \textsc{[Earlier]} using the regular expression rules.

We used \textit{OpenFST}\footnote{\url{http://www.openfst.org}} to minimize the built FST, resulting in a deterministic FST with 26 states (of which 8 are final states) and 177 arcs. The complete FST is specified in a text file using the AT\&T FSM format\footnote{\url{http://github.com/paramitamirza/IndoTimex/blob/master/lib/fst/timex.fst}}, and visualized as a transition diagram using OpenFST\footnote{\url{http://github.com/paramitamirza/IndoTimex/blob/master/lib/fst/timex.pdf}}.

\subsection{Temporal Tagging}

Given a document in the TimeML annotation format, we first parse the document creation time (DCT) inside the DCT tag and the document content inside the \texttt{TEXT} tag. The content is further tokenized following a simple splitting rule with white-spaces and punctuations as delimiters, except for tokens containing digits (e.g. \textit{08.30}, \textit{'86}, \textit{ke-2}, \textit{70-an}, \textit{10/01/2015}).

Given a list of tokens and the document creation time, the tagging algorithm goes as described in Appendix \ref{app:temporal-tagging-algo}.

\subsection{Evaluation}

\paragraph{Dataset} The dataset comprises 75 news articles taken from www.kompas.com, and is made available for download\footnote{\url{http://github.com/paramitamirza/IndoTimex/tree/master/dataset}}. The preparation of the dataset includes cleaning the HTML files and converting the text into the TimeML document format. As shown in Table~\ref{corpora}, during the development phase only 50 news articles are used to develop the time grammar and the transducer. The rest 25 articles are manually annotated with temporal expressions and used for the evaluation phase.

\begin{table}[th!]
\centering
\begin{tabular}{lcc}
\hline
\textbf{Corpus} & \textbf{\# of docs} & \textbf{\# of tokens}\\
\hline
development & 50 & 17,026\\
evaluation & 25 & 9,549\\
\hline
\end{tabular}
\caption{Corpora used in development and evaluation phases.}
\label{corpora}
\end{table}

\paragraph{Evaluation Results} Table~\ref{result} shows the performance results of each task in temporal expression tagging, including temporal expression recognition and normalization. 

\begin{table}[t]
\centering
\begin{tabular}{lccc|c}
\hline
\textbf{Task} & \textbf{P} & \textbf{R} & \textbf{F1} & \textbf{Acc}\\
\hline
Timex recognition & 89.57\% & 96.43\% & 92.87\% & -\\
Timex normalization & 88.04\% & 82.65\% & 85.26\% & 85.71\%\\
\hline
\end{tabular}
\caption{Performance results on temporal expression recognition and normalization tasks on Indonesian documents, in terms of precision (P), recall(R), F1-score (F1) and accuracy (Acc).}
\label{result}
\end{table}

There are 211 temporal expressions identified by our method in the evaluation data. With 189 correctly identified entities, 22 false positives and 7 false negatives, the system yields 89.57\% precision, 96.43\% recall and 92.87\% F1-score.

Among the false positives, 11 entities which are actually flight numbers (e.g. \textit{8501}) are tagged as \textsc{date}, while 5 entities which are part of a geographic coordinate (e.g. \textit{08 derajat \underline{\smash{50 menit 43 detik}} selatan} [08 degrees 50 minutes 43 seconds south] or \textit{\underline{\smash{03.22.46}} Lintang Selatan} [3$^{\circ}$ 22' 46'' South]) are tagged as \textsc{duration} or \textsc{time}. There are 6 entities identified incorrectly due to the ambiguous nature of \textit{dulu/dahulu}, which could mean `in the past' or `first' (as in "John wants to say goodbye \textit{first} before leaving") depending on the context.

Introducing a threshold for the reasonable maximum year number that could appear in a text (e.g. year 3000) will decrease the number of falsely extracted flight numbers (e.g. 8501) because it is in the same format as a year. It might also help to include temporal signals such as \textit{pada} [on/at] or \textit{selama} [during] in the transducer, to ensure that the following tokens are indeed temporal expressions. 

We could also include in the transducer the expressions that can rule out the following tokens to be part of temporal expressions. For example, if we find \textit{derajat} [degree], we can make sure that even though the following tokens are usually part of temporal expressions (i.e. \textit{menit} [minutes] and \textit{detik} [seconds]), the transducer will end up in a non-final state. The same strategy could be applied if the following tokens denote geographical directions such as \textit{Lintang Selatan} [South] or \textit{Bujur Timur} [East].

The false negatives include \textit{esok hari} [tomorrow] and \textit{setengah hari} [half a day], which are due to the incomplete transducer. Another cases are \textit{jauh-jauh hari sebelumnya} [many days before] and \textit{2-3 menit} [2-3 minutes], which are due to the incomplete regular expressions to recognize indefinite quantifiers for expressing durations (i.e. \textit{jauh-jauh} [many] and 2-3). 

In determining the temporal expression types (i.e. \textsc{date}, \textsc{time}, \textsc{duration} and \textsc{set}), the system achieves a perfect accuracy. Meanwhile, for normalizing the correctly identified temporal expressions, the system achieves 85.71\% accuracy, resulting in 85.26\% F1-score. 

Most incorrect cases in the normalization task are because of the wrong anchor time, since we always use the document creation time as the anchor time in resolving the values. For example, in the documents, the expressions \textit{saat itu} [that moment] mostly refer to the previously mentioned temporal expressions of \textsc{time} type. 

There are 3 temporal expressions of which TimeNorm failed to normalize, including \textit{saat yang sama} [the same moment], \textit{tanggal 24 kemarin} [24th yesterday] and \textit{pukul 13.25 kemudian} [13:25 later].

As a future improvement, we consider including temporal signals (e.g. \textit{pada} [on/at], \textit{selama} [during]) in the transducer to make sure that the following tokens are indeed part of temporal expressions, as well as including expressions that rule out the following or preceding tokens to be part of temporal expressions (e.g. \textit{derajat} [degree], \textit{Lintang Selatan} [South]). This strategy might be useful to reduce the number of false positives.

\section{Conclusions}


\paragraph{Timex Recognition} In the EVENTI task, for recognizing temporal expressions in Italian texts, our statistical approach performed best in strict matching and in classifying the timex types, with 82.7\% F1-score and 80\% F1-score respectively. Furthermore, our system performed best in the pilot task, i.e. on historical texts, showing that our approach is robust with respect to domain changes.

We have also developed a rule-based system for recognizing temporal expressions in Indonesian documents. Even though the system could still be improved, particularly by completing the regular expression rules and the finite state transducer, the system achieves good results of 92.87\% F1-score. The built framework can be easily extended to accommodate other low resource languages, requiring only modifications of the regular expression rules and the finite state transducer. 

\paragraph{Timex Normalization} For timex normalization, we have extended an existing tool for English, TimeNorm, for both Italian and Indonesian languages. The adaptation is quite straightforward, because how time is expressed is more or less the same in all languages, involving time units (e.g. \timex{week}, \timex{hour} and relative time functions (e.g. \textit{two days ago}, \textit{now})). There are subtle differences depending on local conventions that need to be addressed such as punctuations used, 
day-month order, 
how to tell the time (e.g. \timex{half past eight} vs \timex{otto e mezzo} [eight and a half] for Italian vs \timex{setengah sembilan} [half (to) nine] for Indonesian), etc. Few modifications of the TimeNorm grammar and code are required in order to deal with the characteristics of Italian and Indonesian temporal expressions.

For Italian, the performance for timex normalization is still considerably lower than the state-of-the-art system for English, UWTime~\parencite{lee-EtAl:2014:P14-1}, i.e. 66.5\% vs 82.4\% F1-scores. This suggests that the TimeNorm adaptation for Italian can still be improved, for example by including \textit{semester} or \textit{half-year} as a unit. For Indonesian, the system can achieve 85.26\% F1-score in normalizing temporal expressions. However, some improvements are required to cope with superfluous temporal expressions such as \textit{tanggal 24 kemarin} [(on) 24th \textit{yesterday}] and \textit{pukul 13.25 kemudian} [(at) 13:25 \textit{later}].

Furthermore, for both Italian and Indonesian, we could implement different strategies to select the correct anchor time for some temporal expressions, instead of always using the document creation time. The best approximation would be to use the preceding temporal expressions of the same type (if any) as the anchor time.

\paragraph{Event and Temporal Relation Extraction} In EVENTI, the individual performances of our Italian event extraction system and temporal relation extraction system were quite good, with 86.7\% F1-score and 73.3\% F1-score resp. However, for the complete end-to-end temporal information processing the temporal awareness score was only 34.1\%. This result is quite similar to the TempEval-3 results for English, most probably related to the sparse annotation of temporal relations in the dataset. Without any specific adaptation to historical text, our system yields comparable results.
In a close future, our system for temporal information processing of Italian texts will be included in the TextPro tools suite.

\paragraph{} In general, the work confirms that statistical approaches for temporal information processing are robust across languages, given the availability of annotated texts and natural language processing tools (e.g. PoS tagger, dependency parser) for the language of interest. However, some tasks are still best solved with rule-based methods, e.g. timex normalization. Furthermore, for low-resource languages such as Indonesian, statistical approaches are more time consuming to implement, since we should first develop annotated data and basic NLP tools.

Cross-lingual annotation or cross-lingual model transfer approaches are often proposed to solve NLP tasks for low-resource languages. If the necessary resources are already available for a closely related language, they can be utilized to facilitate the construction of a model or annotation for the target language. For example, \textcite{nakov-ng:2009:EMNLP} utilized Malay language to improve statistical machine translation for Indonesian $\rightarrow$ English, considering that more resources are available for Malay, and that Malay and Indonesian are closely related.

\chapter{Conclusions and Outlook}\label{ch:conclusion}

\section{Conclusions}

The goal of temporal information processing is to construct structured information about events and temporal-causal relations between them, given the fact that news and narrative texts often describe dynamic information of events that occur in a particular temporal order or causal structure. In this thesis, building an integrated system for extracting such temporal-causal information from text has been our main focus. Furthermore, since temporal and causal relations are closely related, given the presumed constraint of event precedence in causality, we explored ways to exploit this presumption to improve the performance of our integrated temporal and causal relation extraction system.

In Chapter \ref{ch:background}, besides some natural language processing foundations, we have provided background information about machine learning approaches that are used in this thesis. Chapter \ref{ch:auto-event-extraction} introduced the task of temporal information processing, and outlined the state-of-the-art methods for extracting temporal information from text. Based on results reported on an evaluation campaign related to temporal information processing, i.e. TempEval-3, we highlighted the fact that the overall performance of end-to-end temporal information processing systems from raw text suffers due to the lacking temporal relation extraction systems, with 36\% F1-score. This was the main reason underlying our choice to focus our attention on the extraction of relations between events.

\paragraph{Temporal Relations} In Chapter \ref{ch:temp-rel-type} we have described our approach to build a hybrid temporal relation extraction system, \textit{TempRelPro}, which combines rule-based and machine learning modules in a sieve-based architecture inspired by CAEVO \parencite{chambers-etal:2014:TACL}. However, our architecture is arguably simpler and more efficient than CAEVO since (i) the temporal closure inference module is run only once and (ii) we use less classifiers in general. We have evaluated TempRelPro in three different evaluation settings, i.e. \textit{TempEval-3}, \textit{TimeBank-Dense} and \textit{QA-TempEval}, in which TempRelPro is shown to achieve state-of-the-art performances. However, TempRelPro still performs poorly in labelling the temporal relation types of event-event pairs, compared to its performance for pairs of temporal expressions (timex-timex) and event-timex. 

\paragraph{Causal Relations} One direction to address this issue is to build a causal relation extraction system, considering the temporal constraint of event precedence in causality. Apart from being an effort to improve the temporal relation extraction system, the extraction of causal chains of events in a story can also benefit question answering and decision support systems, among others. 

Looking at the existing resources for causality annotation, we could not find one that provides a comprehensive account of how causality can be expressed in a text. Therefore, we have presented in Chapter \ref{ch:annotating-causality} our guidelines for annotating explicit causality between events, inheriting the concept of events, event relations and signals in TimeML, without limiting our effort to specific connectives. Our annotation effort on \textit{TimeBank}---a freely available TimeML corpus that already contains temporal entity and temporal relation annotation---resulted in \textit{Causal-TimeBank}. Causal-TimeBank contains 318 causal links, much less than the 2,519 temporal links between events found in the corpus. This shows that causal relations, particularly explicit ones, appear relatively rarely in texts. Our analysis on the corpus statistics sheds light on the behaviour of causal markers in texts. For instance, there are several ambiguous causal signals and causative verbs, which occur abundantly but do not always carry a causation sense. 

Our next step was to exploit this corpus and the obtained corpus statistics for building (and evaluating) a causal relation extraction system. Chapter \ref{ch:caus-rel-recognition} provided details on our hybrid approach for building a system for identifying causal links between events, \textit{CauseRelPro}, making use of the previously mentioned Causal-TimeBank. Again, we adopted a sieve-based architecture for combining the rule-based and machine-learned modules, which is proven to benefit temporal relation extraction. An evaluation of CauseRelPro using the Causal-TimeBank corpus in stratified 10-fold cross-validation resulted in 40.95\% F1-score, much better than our previous data-driven system for causal relations reported in \textcite{mirza-tonelli:2014:Coling} with 33.88\% F1-score.

\paragraph{Integrated Temporal and Causal Relation Extraction System} In Chapter \ref{ch:integrated-system}, following the analysis of the interaction between temporal and (explicit) causal relations in texts, we presented our approach for integrating our temporal and causal relation extraction systems. The integrated system, \textit{CATENA}---CAusal and Temporal relation Extraction from NAtural language texts---, is a combination of TempRelPro and CauseRelPro, exploiting the presumption about event precedence when two events are connected by causality. The interaction between TempRelPro and CauseRelPro in the integrated architecture is realized by (i) using the output of TempRelPro (temporal link labels) as features for CauseRelPro, and (ii) using the output of CauseRelPro as a post-editing method for correcting the mislabelled output of TempRelPro.

Confirming the finding of several previous works \parencite{bethard-martin:2008:ACLShort, rink:2010, mirza-tonelli:2014:Coling}, using temporal information as features boosted the performance of our causal relation extraction system. Through an ablation test, we found that without temporal link labels as features, the F1-score drops from 62\% to 57\%, with a significant recall drop from 54\% to 46\%. We also found that the post-editing rules would improve the output of temporal relation labelling, even though this phenomenon is not captured statistically in the TempEval-3 evaluation due to the sparse annotation of the evaluation corpus. Nevertheless, explicit causality found in a text is very infrequent, and hence, cannot contribute much in improving the performance of the temporal relation extraction system.

\paragraph{Word Embeddings} While morpho-syntactic, context and time-value information features are sufficient for determining the temporal order of timex-timex and event-timex pairs, the lack of lexical-semantic information about event words may contribute to TempRelPro's low performance on event-event pairs. Chapter \ref{ch:deep-learning} discusses our preliminary investigation into the potentiality of exploiting word embeddings for alleviating this issue, specifically in using word embeddings as lexical-semantic features for the supervised temporal relation type classifier included in TempRelPro. 

We have compared two pre-trained word vectors from GloVe and Word2Vec, and found that Word2Vec embeddings yield better results. We also found that concatenating the two head word vectors of event pairs is the best combination method, although subtraction may also bring advantages for some relation types such as \texttt{IDENTITY} or \texttt{BEGINS}/\texttt{BEGUN\_BY}. 

In a 10-fold cross-validation setting, we found that combining word embeddings and traditional features results in significant improvement. However, using the same feature vector evaluated on the TempEval-3 evaluation corpus, the classifier's performance does not improve in general, despite of performance gains in identifying several relation types, i.e. \texttt{IDENTITY}, \texttt{SIMULTANEOUS} and \texttt{IS\_INCLUDED}.

\paragraph{Training Data Expansion} In Chapter \ref{ch:training-data-expansion} we have presented our investigation into the effect of training data expansion for temporal and causal relation extraction. In particular, we investigated the impact of (i) \textit{temporal reasoning on demand} for temporal relation type classification and (ii) \textit{self-training} for causal relation extraction.

In (i), our objective is to improve not only the quantity but also the quality of training data for temporal relation type classification. We made use of a temporal reasoner module that checks the temporal graph consistency and infers new temporal links, based on temporal closure inference on the initial set of annotated temporal links in a document. However, the temporal reasoner is only run when it is estimated to be effective, hence the term `temporal reasoning on demand'. According to our experiments, deduction may be beneficial when the estimated number of deducible temporal links falls below a certain threshold, which is experimentally assessed. With this setting, removing inconsistent documents prior to deduction can also have a positive impact on classification performances.

In (ii), we employed self-training to bootstrap the training data, along with a causal-link propagation method. The propagation method relies on an assumption that news texts often describe the same set of events by rewording the underlying story. Thus, if we found a causal relation between two events in a news, the same relation holds every time the two events are mentioned in similar news, in which the causality may be expressed differently than in the original news. Our experiments show that self-training and causal-link propagation can boost the performance of a causal relation extraction system, albeit our simplified implementation of self-training (only one iteration is performed), and the size of the unlabelled dataset being not significantly larger than the original training set. Despite the limited number of newly acquired training examples (324 through self-training and 32 through causal link propagation), they still have a significant impact on the classifier performance, since the original training corpus contains only 318 causal links.

\paragraph{Multilinguality} Finally, Chapter \ref{ch:multilinguality} summarizes our efforts in the adaptation of temporal information processing for texts in languages other than English, i.e. Italian and Indonesian. In general, the work confirms that statistical approaches for temporal information processing are robust across languages, given the availability of annotated texts and natural language processing tools for the language of interest. However, some tasks are still best solved with rule-based methods, e.g. timex normalization. Furthermore, for low-resource languages such as Indonesian, statistical approaches require more efforts since we should first construct annotated data and basic NLP tools. 

\section{Ongoing and Future Work}

\paragraph{Implicit Temporal and Causal Relations} The identification of temporal and causal relations between two events is relatively straightforward given an explicit marker (e.g. \textit{before}, \textit{because}) connecting the two events, tense-aspect-modality information embedded in the event words or specific syntactic construction involving the two events. It becomes more challenging when such an overt indicator is lacking, which is often the case when two events take place in different sentences. In the TempEval-3 evaluation corpus, 32.76\% of the event pairs do not occur in the same sentences. Furthermore, our Causal-TimeBank corpus only contains 318 causal links; more could be found if we do not limit our annotation to overtly expressed causal links (via causal signals and causal verbs) and also consider the implicit ones.

Our preliminary work with word embeddings is motivated by this issue, since most research on implicit relations incorporate word-based information in the form of word pair features. The results of our experiments in Chapter \ref{ch:deep-learning} shed some light on how word embeddings can potentially improve a classifier performance for temporal ordering of events. We have seen different advantages brought by different ways of combining word vectors, i.e., concatenation works well for identifying relations such as \texttt{BEFORE}/\texttt{AFTER} and \texttt{INCLUDES}/\texttt{IS\_INCLUDED}, whereas subtraction may benefit relations such as \texttt{IDENTITY} and \texttt{BEGINS}/\texttt{BEGUN\_BY}. We would like to take advantage of \textit{ensemble learning}, particularly \textit{stacking}, to learn a super-classifier that decides the best label for an event pair, given different predictions by other classifiers. In this case, the other classifiers could be classification models trained on traditional feature vector, concatenated word vectors and/or subtracted word vectors. 

We would also like to apply the same method to extract implicit causality between events. However, given the limited amount of training data for causal relations in Causal-TimeBank, it may be difficult to obtain a robust classification model based on word pair features. There are several resources that we can use to expand our training data, for instance, causality annotated between nominals \parencite{girju-EtAl:2007:SemEval-2007}, the parallel temporal-causal corpus by \textcite{BETHARD08.229}, and causality annotated between verbal event pairs \parencite{do-chan-roth:2011:EMNLP,riaz-girju:2013:SIGDIAL}. Another way to expand the training data would be to run the rule-based module in CauseRelPro, considering its high precision, on boundless unlabelled data to retrieve significant amount of event pairs connected by causal links for training the classification models with word embeddings as features.

Furthermore, instead of using general-purpose word embeddings, several works presented methods for building task-specific word embeddings \parencite{hashimoto-EtAl:2015:CoNLL,boros-EtAl:2014:EMNLP2014,nguyen-grishman:2014:P14-2,tang-EtAl:2014:SemEval}, which may also be beneficial for temporal ordering and causality extraction task. 



\appendix
\cleardoublepage
\part{Appendix}
\chapter{Appendix}

\section{Temporal Signal Lists}
\label{app:temporal-signals}

\begin{scriptsize}
\begin{longtable}{p{3.5cm}p{2.5cm}|p{3.5cm}p{2.5cm}}
\hline
\multicolumn{2}{c|}{\textbf{Event-related Signals}} & \multicolumn{2}{c}{\textbf{Timex-related Signals}} \\
\textbf{Text} & \textbf{Cluster} & \textbf{Text} & \textbf{Cluster} \\
\hline
\endfirsthead
\multicolumn{4}{c}%
{\tablename\ \thetable\ -- \textit{Continued from previous page}} \\
\hline
\textbf{Text} & \textbf{Cluster} & \textbf{Text} & \textbf{Cluster} \\
\hline
\endhead
\hline \multicolumn{4}{r}{\textit{Continued on next page}} \\
\endfoot
\hline
\endlastfoot
just as soon as  &  as soon as & at  &  at\\
just as long as  &  as long as & by  &  by\\
at the very moment  &  at the same time & in  &  in\\
at the same time  &  at the same time & on  &  on\\
so far as  &  as long as & for  &  for\\
read out by  &  followed by & by  &  by\\
prior to making  &  prior to & from  &  from\\
in advance of  &  prior to & to  &  to\\
immediately followed by  &  followed by & during  &  during\\
being pursued by  &  followed by & between  &  between\\
before proceeding with  &  prior to & after  &  after\\
before proceeding to  &  prior to & before  &  before\\
at one time  &  at the same time & up to a maximum of  &  up to\\
as swiftly as  &  as soon as & to a maximum of  &  up to\\
as speedily as  &  as soon as & up to  &  up to\\
as soon as  &  as soon as & up till  &  up to\\
as rapidly as  &  as soon as & within  &  within\\
as quickly as  &  as soon as & upon  &  after\\
as quick as  &  as soon as & until  &  until\\
as promptly as  &  as soon as & under  &  within\\
as much as  &  as long as & till  &  until\\
as long as  &  as long as & since  &  since\\
as fast as  &  as soon as & still  &  still\\
as far as  &  as long as & throughout  &  during\\
as expeditiously as  &  as soon as & through  &  during\\
as early as  &  as soon as & recently  &  recently\\
read by  &  followed by & previously  &  formerly\\
pursued by  &  followed by & previous  &  former\\
prior to  &  prior to & preliminary  &  early\\
monitoring of  &  followed by & preceding  &  former\\
monitored by  &  followed by & over  &  over\\
in parallel  &  at the same time & next  &  next\\
in conjunction  &  at the same time & latterly  &  recently\\
followed by  &  followed by & later  &  later\\
follow-up on  &  followed by & lately  &  lately\\
follow-up of  &  followed by & just  &  immediately\\
first of  &  prior to & initial  &  early\\
attended by  &  followed by & further  &  later\\
applied by  &  followed by & formerly  &  formerly\\
ahead of  &  prior to & former  &  former\\
yet  &  still & following  &  next\\
with  &  with & first  &  early\\
within  &  during & early  &  early\\
whilst  &  while & earlier  &  earlier\\
while  &  while & beyond  &  after\\
whereas  &  while & beforehand  &  formerly\\
whenever  &  when & already  &  formerly\\
when  &  when & ago  &  ago\\
urgently  &  immediately & afterwards  &  later\\
upon  &  after & afterward  &  later\\
until  &  until &  & \\
unless  &  if &  & \\
ultimately  &  eventually &  & \\
till  &  until &  & \\
thus  &  then &  & \\
throughout  &  during &  & \\
through  &  during &  & \\
thirdly  &  finally &  & \\
therefore  &  afterwards &  & \\
thereafter  &  afterwards &  & \\
then  &  afterwards &  & \\
tentatively  &  initially &  & \\
subsequently  &  eventually &  & \\
subsequent  &  next &  & \\
still  &  still &  & \\
soon  &  immediately &  & \\
someday  &  once &  & \\
since  &  since &  & \\
simultaneously  &  simultaneously &  & \\
secondly  &  then &  & \\
readily  &  immediately &  & \\
quickly  &  immediately &  & \\
promptly  &  immediately &  & \\
previously  &  formerly &  & \\
previous  &  former &  & \\
preceding  &  former &  & \\
potentially  &  eventually &  & \\
possibly  &  eventually &  & \\
over  &  during &  & \\
originally  &  formerly &  & \\
once  &  once &  & \\
nonetheless  &  still &  & \\
next  &  next &  & \\
nevertheless  &  still &  & \\
moreover  &  meanwhile &  & \\
meanwhile  &  meanwhile &  & \\
meantime  &  meanwhile &  & \\
long-standing  &  former &  & \\
later  &  later &  & \\
lastly  &  finally &  & \\
jointly  &  simultaneously &  & \\
into  &  into &  & \\
instantly  &  immediately &  & \\
initially  &  initially &  & \\
increasingly  &  still &  & \\
incoming  &  next &  & \\
impending  &  next &  & \\
immediately  &  immediately &  & \\
if  &  if &  & \\
furthermore  &  meanwhile &  & \\
further  &  later &  & \\
forthcoming  &  next &  & \\
formerly  &  formerly &  & \\
former  &  former &  & \\
foreseeable  &  next &  & \\
following  &  next &  & \\
follows  &  follow &  & \\
followed  &  follow &  & \\
follow  &  follow &  & \\
firstly  &  initially &  & \\
finally  &  eventually &  & \\
ex-  &  former &  & \\
eventually  &  eventually &  & \\
even  &  still &  & \\
earlier  &  earlier &  & \\
during  &  during &  & \\
directly  &  immediately &  & \\
despite  &  despite &  & \\
definitively  &  finally &  & \\
contemporaneously  &  simultaneously &  & \\
consistently  &  still &  & \\
consecutively  &  simultaneously &  & \\
continue  &  follow &  & \\
concurrently  &  simultaneously &  & \\
concomitantly  &  simultaneously &  & \\
beyond  &  after &  & \\
beforehand  &  previously &  & \\
before  &  before &  & \\
as  &  as &  & \\
anyway  &  still &  & \\
ancient  &  former &  & \\
always  &  still &  & \\
also  &  still &  & \\
already  &  already &  & \\
again  &  still &  & \\
afterwards  &  afterwards &  & \\
afterward  &  afterwards &  & \\
after  &  after &  & \\
\hline
\end{longtable}
\end{scriptsize}

\section{Causal Verb \& Signal Lists}
\label{app:causal-verbs-signals}

\begin{scriptsize}
\begin{longtable}{p{2cm}p{2.5cm}|p{6.5cm}p{2.5cm}}
\hline
\multicolumn{2}{c|}{\textbf{Causal Verbs}} & \multicolumn{2}{c}{\textbf{Causal Signals}} \\
\textbf{Text} & \textbf{Cluster} & \textbf{Text} & \textbf{Cluster} \\
\hline
\endfirsthead
\multicolumn{4}{c}%
{\tablename\ \thetable\ -- \textit{Continued from previous page}} \\
\hline
\textbf{Text} & \textbf{Cluster} & \textbf{Text} & \textbf{Cluster} \\
\hline
\endhead
\hline \multicolumn{4}{r}{\textit{Continued on next page}} \\
\endfoot
\hline
\endlastfoot
bribe & \texttt{CAUSE} & \textbf{Pattern} & \\
cause & \texttt{CAUSE} & \texttt{on ([athe]+\textbackslash \textbackslash s)?([a-z]+\textbackslash \textbackslash s)?grounds?} of & because of\\
compel & \texttt{CAUSE} & \texttt{on ([athe]+\textbackslash \textbackslash s)?([a-z]+\textbackslash \textbackslash s)?basis of} & because of\\
convince & \texttt{CAUSE} & \texttt{in ([a-z]+\textbackslash \textbackslash s)?light of} & because of\\
drive & \texttt{CAUSE} & \texttt{in ([a-z]+\textbackslash \textbackslash s)?pursuance of} & because of\\
impel & \texttt{CAUSE} & \texttt{on ([athe]+\textbackslash \textbackslash s)?([a-z]+\textbackslash \textbackslash s)?background of} & because of\\
incite & \texttt{CAUSE} & \texttt{on ([a-z]+\textbackslash \textbackslash s)?account of} & because of\\
induce & \texttt{CAUSE} & \texttt{for ([athe]+\textbackslash \textbackslash s)?sake of} & because of\\
influence & \texttt{CAUSE} & \texttt{by ([a-z]+\textbackslash \textbackslash s)?virtue of} & because of\\
inspire & \texttt{CAUSE} & \texttt{by ([a-z]+\textbackslash \textbackslash s)?reason of} & because of\\
persuade & \texttt{CAUSE} & \texttt{by ([a-z]+\textbackslash \textbackslash s)?cause of} & because of\\
prompt & \texttt{CAUSE} & \texttt{because of the ([a-z]+\textbackslash \textbackslash s)?need to} & due to\\
push & \texttt{CAUSE} & \texttt{due to the ([a-z]+\textbackslash \textbackslash s)?need to} & due to\\
force & \texttt{CAUSE} & \texttt{due ([a-z]+\textbackslash \textbackslash s)?to} & due to\\
enforce & \texttt{CAUSE} & \texttt{owing ([a-z]+\textbackslash \textbackslash s)?to} & due to\\
rouse & \texttt{CAUSE} & \texttt{thanks ([in]+\textbackslash \textbackslash s)?([a-z]+\textbackslash \textbackslash s)?to} & due to\\
set & \texttt{CAUSE} & \texttt{thanks ([a-z]+\textbackslash \textbackslash s)?to} & due to\\
spur & \texttt{CAUSE} & \texttt{under the ([a-z]+\textbackslash \textbackslash s)?influence of} & in consequence of\\
start & \texttt{CAUSE} & \texttt{in ([athe]+\textbackslash \textbackslash s)?([a-z]+\textbackslash \textbackslash s)?wake of} & in consequence of\\
stimulate & \texttt{CAUSE} & \texttt{in ([anthe]+\textbackslash \textbackslash s)?([a-z]+\textbackslash \textbackslash s)?aftermath of} & in consequence of\\
entail & \texttt{CAUSE} & \texttt{in ([athe]+\textbackslash \textbackslash s)?([a-z]+\textbackslash \textbackslash s)?wake} & in consequence\\
generate & \texttt{CAUSE} & \texttt{in ([athe]+\textbackslash \textbackslash s)?([a-z]+\textbackslash \textbackslash s)?aftermath} & in consequence\\
trigger & \texttt{CAUSE} & \texttt{in ([a-z]+\textbackslash \textbackslash s)?answer to} & in response to\\
spark & \texttt{CAUSE} & \texttt{in ([a-z]+\textbackslash \textbackslash s)?response to} & in response to\\
fuel & \texttt{CAUSE} & \texttt{in ([a-z]+\textbackslash \textbackslash s)?responding to} & in response to\\
ignite & \texttt{CAUSE} & \texttt{in ([a-z]+\textbackslash \textbackslash s)?replying to} & in response to\\
reignite & \texttt{CAUSE} & \texttt{in ([a-z]+\textbackslash \textbackslash s)?reaction to} & in response to\\
inflict & \texttt{CAUSE} & \texttt{in ([a-z]+\textbackslash \textbackslash s)?retaliation for} & in exchange for\\
provoke & \texttt{CAUSE} & \texttt{in ([a-z]+\textbackslash \textbackslash s)?exchange for} & in exchange for\\
have & \texttt{CAUSE-AMBIGUOUS} & \texttt{in ([a-z]+\textbackslash \textbackslash s)?order to} & in order to\\
move & \texttt{CAUSE-AMBIGUOUS} & \texttt{as ([athe]+\textbackslash \textbackslash s)?([a-z]+\textbackslash \textbackslash s)?result of} & as a result of\\
get & \texttt{CAUSE-AMBIGUOUS} & \texttt{as ([athe]+\textbackslash \textbackslash s)?([a-z]+\textbackslash \textbackslash s)?reaction to} & as a result of\\
make & \texttt{CAUSE-AMBIGUOUS} & \texttt{as ([anthe]+\textbackslash \textbackslash s)?([a-z]+\textbackslash \textbackslash s)?outcome of} & as a result of\\
send & \texttt{CAUSE-AMBIGUOUS} & \texttt{as ([athe]+\textbackslash \textbackslash s)?([a-z]+\textbackslash \textbackslash s)?follow-up to} & as a result of\\
aid & \texttt{ENABLE} & \texttt{as ([anthe]+\textbackslash \textbackslash s)?([a-z]+\textbackslash \textbackslash s)?effect of} & as a result of\\
allow & \texttt{ENABLE} & \texttt{as ([athe]+\textbackslash \textbackslash s)?([a-z]+\textbackslash \textbackslash s)?consequence of} & as a result of\\
authorize & \texttt{ENABLE} & \texttt{as ([athe]+\textbackslash \textbackslash s)?([a-z]+\textbackslash \textbackslash s)?result} & as a result\\
authorise & \texttt{ENABLE} & \texttt{as ([athe]+\textbackslash \textbackslash s)?([a-z]+\textbackslash \textbackslash s)?reaction} & as a result\\
empower & \texttt{ENABLE} & \texttt{as ([anthe]+\textbackslash \textbackslash s)?([a-z]+\textbackslash \textbackslash s)?outcome} & as a result\\
enable & \texttt{ENABLE} & \texttt{as ([athe]+\textbackslash \textbackslash s)?([a-z]+\textbackslash \textbackslash s)?follow-up} & as a result\\
ensure & \texttt{ENABLE} & \texttt{as ([anthe]+\textbackslash \textbackslash s)?([a-z]+\textbackslash \textbackslash s)?effect} & as a result\\
facilitate & \texttt{ENABLE} & \texttt{as ([athe]+\textbackslash \textbackslash s)?([a-z]+\textbackslash \textbackslash s)?consequence} & as a result\\
guarantee & \texttt{ENABLE} & \texttt{for th[eioa][st]e* ([a-z]+\textbackslash \textbackslash s)?reasons?} & for reason\\
permit & \texttt{ENABLE} & \texttt{it [i']s ([a-z]+\textbackslash \textbackslash s)*why} & is why\\
provide & \texttt{ENABLE} & \texttt{th[ai][st] (, ([a-z]+\textbackslash \textbackslash s)+, )*[i']s ([a-z]+\textbackslash \textbackslash s)*why} & is why\\
activate & \texttt{ENABLE-AMBIGUOUS} & in such a way that & so that\\
afford & \texttt{ENABLE-AMBIGUOUS} & for the reason that & so that\\
help & \texttt{ENABLE-AMBIGUOUS} & to ensure that & so that\\
leave & \texttt{ENABLE-AMBIGUOUS} & because of & because of\\
let & \texttt{ENABLE-AMBIGUOUS} & attributable to & due to\\
bar & \texttt{PREVENT} & in return & in response\\
block & \texttt{PREVENT} & in response & in response\\
constrain & \texttt{PREVENT} & in such a way as to & in order to\\
deter & \texttt{PREVENT} & such that & so that\\
discourage & \texttt{PREVENT} & so that & so that\\
dissuade & \texttt{PREVENT} & thus & therefore\\
hamper & \texttt{PREVENT} & therefore & therefore\\
hinder & \texttt{PREVENT} & thereby & therefore\\
hold & \texttt{PREVENT} & hence & therefore\\
impede & \texttt{PREVENT} & consequently & therefore\\
prevent & \texttt{PREVENT} & because & because\\
protect & \texttt{PREVENT} & since & since\\
restrain & \texttt{PREVENT} & as & as\\
restrict & \texttt{PREVENT} & so & so\\
deny & \texttt{PREVENT} & by & by\\
obstruct & \texttt{PREVENT} & from & from\\
inhibit & \texttt{PREVENT} &  & \\
prohibit & \texttt{PREVENT} &  & \\
forestall & \texttt{PREVENT} &  & \\
impede & \texttt{PREVENT} &  & \\
avert & \texttt{PREVENT} &  & \\
avoid & \texttt{PREVENT} &  & \\
preclude & \texttt{PREVENT} &  & \\
keep & \texttt{PREVENT-AMBIGUOUS} &  & \\
save & \texttt{PREVENT-AMBIGUOUS} &  & \\
stop & \texttt{PREVENT-AMBIGUOUS} &  & \\
affect & \texttt{AFFECT} &  & \\
influence & \texttt{AFFECT} &  & \\
determine & \texttt{AFFECT} &  & \\
change & \texttt{AFFECT} &  & \\
impact & \texttt{AFFECT} &  & \\
afflict & \texttt{AFFECT} &  & \\
undermine & \texttt{AFFECT} &  & \\
alter & \texttt{AFFECT} &  & \\
interfere & \texttt{AFFECT} &  & \\
link-to & \texttt{LINK-R} &  & \\
link-with & \texttt{LINK-R} &  & \\
relate-to & \texttt{LINK-R} &  & \\
connect-with & \texttt{LINK-R} &  & \\
associate-with & \texttt{LINK-R} &  & \\
lead-to & \texttt{LINK} &  & \\
stem-from & \texttt{LINK-R} &  & \\
depend-on & \texttt{LINK-R} &  & \\
rely-on & \texttt{LINK-R} &  & \\
result-in & \texttt{LINK} &  & \\
result-from & \texttt{LINK-R} &  & \\
\hline
\end{longtable}
\end{scriptsize}

\section{Temporal Tagging Algorithm}
\label{app:temporal-tagging-algo}

\begin{algorithmic}[1]
 \renewcommand{\algorithmicrequire}{\textbf{Input:}}
 \renewcommand{\algorithmicensure}{\textbf{Output:}}
 \REQUIRE list of tokens $tok$, DCT $dct$, FST $T$
 \ENSURE string $out$ of text annotated with TIMEX3 tags
  \\ \textit{Recognizing temporal expressions}
  \STATE $starts \leftarrow$ empty dictionary
  \STATE $ends \leftarrow$ empty list
  \STATE $timex \leftarrow$ empty dictionary 
  \STATE $start \leftarrow -1$
  \STATE $end \leftarrow -1$
  \STATE $tmx\_type \leftarrow O$
  \STATE $i \leftarrow 0$  
  \WHILE {$i <$ length of $tok$}
  \STATE $tlabel \leftarrow $ token label of $tok[i]$ based on regex
  	\IF {$start$ is $-1$} 
    	\IF {$tlabel$ is in input labels of $T.initial$}
      		\STATE $start \leftarrow i$
      		\STATE $(q,type) \leftarrow$ $T.transition(q_0, tlabel)$
      		\IF {$q$ is in $T.final$}
      			\STATE $end \leftarrow i$
      			\STATE $tmx\_type \leftarrow type$
      		\ENDIF
      	\ENDIF
    \ELSE
    	\IF {$T.transition(q, tlabel)$ is not null}
    		\STATE $(q,type) \leftarrow$ $T.transition(q, tlabel)$
    		\IF {$q$ is in $T.final$}
      			\STATE $end \leftarrow i$
      			\STATE $tmx\_type \leftarrow type$
      		\ENDIF
    	\ELSE
      		\IF {$start > -1$ and $end > -1$}
      			\STATE $starts[start] \leftarrow$ $tmx\_type$
        		\STATE Add $end$ to $ends$
        		\STATE $timex[start] \leftarrow tok[start...end]$
        	\ENDIF
        	\STATE $start \leftarrow -1$
  			\STATE $end \leftarrow -1$
  			\STATE $tmx\_type \leftarrow O$
  		\ENDIF
    \ENDIF
    \STATE $i \leftarrow i+1$
  \ENDWHILE
  \\ \textit{Normalizing temporal expressions}  
  \STATE $timex\_norm \leftarrow$ empty dictionary
  \FOR {$key$ in $timex$}
    \STATE $timex\_norm[key] \leftarrow normalize(timex[key],dct)$
  \ENDFOR
  \\ \textit{TIMEX3 tagging}
  \STATE $out \leftarrow$ empty string
  \STATE $tid \leftarrow 1$
  \FOR {$i=0$ to length of $tok$}  
    \IF {$i$ in keys of $starts$}
      \STATE $timex\_id \leftarrow tid$    
      \STATE $timex\_type \leftarrow starts[i]$
      \STATE $timex\_value \leftarrow timex\_norm[i]$                  
      \STATE $out \leftarrow out$ $+$ TIMEX3 opening tag (with $timex\_id$, $timex\_type$ and $timex\_value$) $+$ $tok[i]$ $+$ space 
      \STATE $tid \leftarrow tid+1$
    \ELSIF {$i$ in $ends$}
      \STATE $out \leftarrow out$ $+$ $tok[i]$ $+$ TIMEX3 closing tag $+$ space
    \ELSE
      \STATE $out \leftarrow out$ $+$ $tok[i]$ $+$ space
    \ENDIF
  \ENDFOR
 \RETURN $out$ 
 \end{algorithmic}

\cleardoublepage
\manualmark
\markboth{\spacedlowsmallcaps{\bibname}}{\spacedlowsmallcaps{\bibname}} 
\refstepcounter{dummy}
\addtocontents{toc}{\protect\vspace{\beforebibskip}} 
\addcontentsline{toc}{chapter}{\tocEntry{\bibname}}
\label{app:bibliography}
\printbibliography

\end{document}